\documentclass{article}

\usepackage[main, final]{neurips_2026}
\usepackage{arydshln}
\usepackage[utf8]{inputenc}
\usepackage[T1]{fontenc}
\usepackage{hyperref}
\hypersetup{pdftitle={WATERFALL: Workflow for Adaptive Training with Evolutionary Reward Formulation and Automated Learning Loops}, pdfauthor={Eleftherios Triantafyllidis, Filippos Christianos, Zhibin Li, Bernd Bickel}}
\usepackage{etoc}
\usepackage{url}
\usepackage{booktabs}
\usepackage{amsfonts}
\usepackage{nicefrac}
\usepackage{microtype}
\usepackage[table]{xcolor}
\usepackage{csquotes}
\usepackage{amsmath}
\usepackage{algorithm}
\usepackage{algpseudocode}
\usepackage{listings}
\usepackage{caption}
\usepackage{calc}
\usepackage{tcolorbox}
\usepackage{xcolor} \usepackage{contour} \contourlength{0.35pt} \newcommand{\outlinedwhite}[1]{\scalebox{1.2}{\contour{black}{\textcolor{white}{\bfseries\scriptsize #1}}}}
\newcommand{\outlinedcolor}[2]{\scalebox{1.2}{\contour{black}{\textcolor{#1}{\bfseries\scriptsize #2}}}}
\usepackage{multirow}
\usepackage{makecell}
\usepackage{array}
\usepackage{graphicx} 
\usepackage{subcaption}
\usepackage{rotating}
\usepackage{wrapfig}
\definecolor{successgreen}{RGB}{40,165,70}
\definecolor{failred}{RGB}{190,30,30}
\title{\textbf{WATERFALL} \\ \textbf{W}orkflow for \textbf{A}daptive \textbf{T}raining with \textbf{E}volutionary \textbf{R}eward \textbf{F}ormulation and \textbf{A}utomated \textbf{L}earning \textbf{L}oops}

\author{%
  \normalsize Eleftherios Triantafyllidis\thanks{\scriptsize \textbf{Correspondence}: \texttt{eleftherios.triantafyllidis@gmail.com}. \textbf{Code}: \url{https://github.com/etriantafyllidis/waterfall}. \textbf{Video}: \url{https://youtu.be/_epI2r3bQ1U}. \textbf{OpenReview}: \url{https://openreview.net/forum?id=TI1lG64bTY}. \normalsize} \\
  \small ETH Zurich \\
  \small Zurich, Switzerland \\
  \footnotesize \texttt{etriantafyll@ethz.ch} \\
  \And
  \normalsize Filippos Christianos \\
  \small Independent Researcher \\
  \small London, UK \\
  \footnotesize \texttt{f.christianos@gmail.com} \\
  \And
  \normalsize Zhibin Li \\
  \small UCL \\
  \small London, UK \\
  \footnotesize \texttt{alex.li@ucl.ac.uk} \\
  \And
  \normalsize Bernd Bickel \\
  \small ETH Zurich \\
  \small Zurich, Switzerland \\
  \footnotesize \texttt{bickelb@ethz.ch} \\
  \normalsize
}

\begin{document}
{\setlength{\tabcolsep}{5pt}\maketitle}
\vspace{-0.9cm}
\begin{abstract}
Reward design remains a central bottleneck in reinforcement learning (RL), particularly in sparse-reward, long-horizon and partially observable settings requiring dependency chaining, memory, and deceptive affordance disambiguation. While recent foundation-model approaches reduce manual reward engineering, most assume goal-conditioning, task descriptions, expert demonstrations, or ground-truth metrics. These assumptions are particularly problematic in partially observable environments, where task-relevant objectives, affordances, and dynamics must be discovered through interaction rather than disclosed \emph{a priori}. We formalise this problem setting by introducing \textbf{WATERFALL}, an iterative population-based workflow for automated reward discovery under the strict observability constraints inherent in POMDP environments, without relying on semantic task descriptors, pre-disclosed environment dynamics, goal conditioning, or ground-truth metrics. WATERFALL achieves this by first (i) synthesising diverse programmatic reward candidates via persona-conditioned generation, thereafter (ii) evaluating candidate behaviour from raw visual rollouts utilising a Swiss-system evolutionary tournament judged by vision-language models to identify elites, and finally (iii) leveraging longitudinal assessment histories to iteratively mutate, escalate, or prune candidates. We evaluate WATERFALL across fully observable and reveal-gated partially observable environments ranging from discrete to continuous, comparing against sparse-reward RL, intrinsic-motivation and foundation-model reward-synthesis baselines. Our results indicate that WATERFALL discovers and refines reward programs without access to privileged information, outperforming baselines that retain their native information advantages, most clearly in the POMDP regime. Extensive ablations also substantiate that visual evidence, tournament-based selection, and iterative refinement are jointly important. Enforcing this contract is what separates autonomous reward discovery from benchmark-privileged reward shaping.
\end{abstract}

\vspace{-0.5cm}
\etocdepthtag.toc{mtmain} 
\section{Introduction}
Designing reward functions remains one of the most persistent bottlenecks in Reinforcement Learning (RL). For tasks with sparse or delayed supervision, the reward is often the difference between efficient learning and complete failure, yet hand-engineering a robust dense reward requires substantial domain expertise and repeated trial-and-error, and notably limits cross-task generalisation \citep{sutton2018reinforcement, ng1999policy, Triantafyllidis2023}. Intrinsic-motivation methods, such as Count-Based Exploration (CBE), the Intrinsic Curiosity Module (ICM) and Random Network Distillation (RND), ease this sparsity by rewarding novelty or prediction error \citep{tang2017exploration,pmlr-v70-pathak17a,burda2019rnd}, yet cannot infer the valuable task semantics that challenging RL problems require.

To overcome this limitation, Large Language Models (LLMs) and Vision-Language Models (VLMs) automate reward specification from language, images, demonstrations, or code, as in the zero-shot reward models including \textbf{RoboCLIP}, \textbf{Code as Reward}, \textbf{Text2Reward (T2R)}, \textbf{Language to Rewards (L2R)}, \textbf{RL-VLM-F} \citep{rocamonde2024vlmrm,sontakke2023roboclip,pmlr-v235-venuto24a,xie2024text2reward,pmlr-v229-yu23a,pmlr-v235-wang24bn}. Recent work extended this with iterative evolutionary refinement, including \textbf{Eureka}, \textbf{REvolve}, \textbf{LaRes}, \textbf{RF-Agent}, \textbf{R*}, \textbf{PRM4RL}, \textbf{RE-GoT}, and \textbf{GRACE} \citep{ma2024eureka, hazra2025revolve,li2025lares,gao2025rfagent,li2025rstar,zhang2025prm4rl,yao2025regot,sapora2026grace}. Together, these methods show that foundation models can act as semantic planners, reward programmers, critics, and refinement engines.

Nevertheless, collectively, these approaches operate under privileged assumptions. Task goals are named explicitly in language, ontologies are known in advance, expert trajectories are available or scalar fitness signals such as success rates are accessed \citep{xie2024text2reward,pmlr-v229-yu23a,ma2024eureka,hazra2025revolve,li2025lares,gao2025rfagent,li2025rstar,sapora2026grace,zhang2025prm4rl}. These assumptions become \textbf{problematic in partially observable environments (POMDPs)}, where relevant goals, objects, affordances, and dependencies are revealed only through exploration. Disclosing these \emph{a priori} becomes privileged information, therefore solving a simpler problem than the policy faces \citep{kaelbling1998planning, chevalier2019babyai, NEURIPS2023_e8916198}.

This is the setting we target, \emph{\textbf{reward discovery under strict observability parity}}, where the reward designer operates on the policy's observable interface and reveal-gated knowledge acquired through interaction. Consider robotic exploration before a downstream task is assigned: A drawer conceals an object whose existence, and any interaction it affords, becomes known only once the drawer is opened. This exposes a fallacy: How can an external reward interface specify reward logic over structure whose existence is established after the interaction that reveals it? That is our \textbf{\emph{research question}}. Supporting learning without repeatedly supplying task-specific knowledge is valuable in real-world exploration, where privileged supervision may be unavailable. Parity tests reward construction without asymmetric task-information channels. It is deliberately stricter than deployability, but allows a shared high-level goal. Our goal-free experiments additionally withhold task specifications, testing how far interaction and pretrained priors can ground executable rewards for behaviours discovered through exploration. What matters is whether reward discovery proceeds from the agent's evidence or \enquote{\emph{hand-holding}} by an external, artificial source. Both are useful, but only the former applies where task structure is initially unknown and must be discovered through interaction.

\begin{figure}[ht]
    \centering
    \includegraphics[
    width=\linewidth,
    trim={5 0 3 0},
    clip]{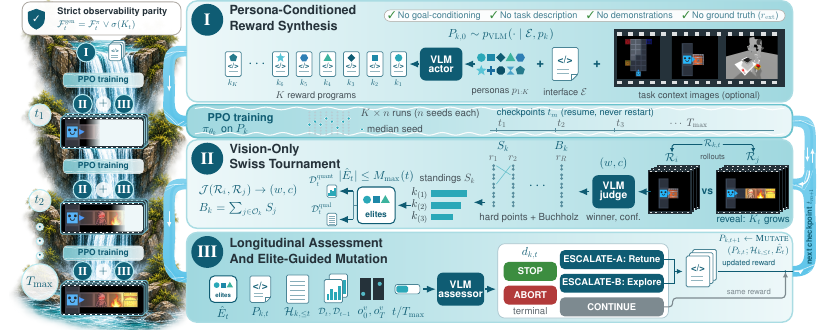}
    \vspace{-0.6cm}
    \caption{\textbf{WATERFALL:} A population-based workflow for reward discovery under strict observability parity that (i) synthesises diverse reward candidates, (ii) trains independent PPO policies, (iii) evaluates behaviour through vision pairwise judgements aggregated by Swiss-system tournaments, and (iv) utilises assessment histories to escalate, prune, or refine candidates via elite-guided mutation.}
    \vspace{-0.5cm}
    \label{fig:WATERFALL_Overview}
\end{figure}

To address this, we introduce the \textbf{W}orkflow for \textbf{A}daptive \textbf{T}raining with \textbf{E}volutionary \textbf{R}eward \textbf{F}ormulation and \textbf{A}utomated \textbf{L}earning \textbf{L}oops \textsc{\textbf{(WATERFALL)}}, depicted in \autoref{fig:WATERFALL_Overview}, which restricts reward generation and evaluation to non-privileged interfaces and observable evidence. We contribute (i) the strict observability-parity setting for reward discovery, (ii) a workflow integrating persona-conditioned synthesis, vision-judge Swiss tournaments, and longitudinal elite-guided mutation under this contract and (iii) an evaluation on discrete and continuous, fully and partially observable environments against RL, intrinsic-motivation and reward-synthesis baselines, (iv) extensive ablations and sensitivity analyses isolating iterative refinement, visual evidence, and information privilege.

\section{Related Work}
\label{sec:related_work}
\vspace{-0.25cm}
In reward design, potential-based shaping preserves optimal policies, yet requires a potential function \citep{ng1999policy}, difficult to specify in POMDPs. Many approaches instead learn rewards from expert behaviour \citep{ng2000irl,abbeel2004apprenticeship,ziebart2008maximum}, adversarial learning \citep{ho2016gail,fu2018airl}, or trajectory preferences \citep{christiano2017preferences,wirth2017preference_survey}. Intrinsic motivation also aid exploration under sparse signals \citep{tang2017exploration, 10611483, bellemare2016unifying}. Foundation models enable reward discovery without prescribed objectives or demonstrations, raising the question of what task information they require.

\subsection{Foundation Models for Reward Synthesis}
Amongst pretrained-VLM approaches, \textbf{VLM-RMs} evaluate rewards from text prompts \citep{rocamonde2024vlmrm}, \textbf{RoboCLIP} derives rewards from a demonstration or description \citep{sontakke2023roboclip}, \textbf{RL-VLM-F} elicits goal-conditioned visual preferences \citep{pmlr-v235-wang24bn}, and \textbf{Code as Reward (CAR-VLM)} distils VLM feedback into executable reward code \citep{pmlr-v235-venuto24a}. Among LLM approaches, \textbf{T2R} generates reward code from a language goal in a known environment \citep{xie2024text2reward}, \textbf{L2R} maps instructions to reward parameters \citep{pmlr-v229-yu23a}, and \textbf{EUREKA} iteratively optimises rewards given task descriptions and code \citep{ma2024eureka}. Reward-search methods further explore reward programs via population evolution, tree search, or graph decomposition \citep{li2025lares, gao2025rfagent, li2025rstar, yao2025regot, sapora2026grace}. Collectively, these methods synthesise effective rewards, but rely on task semantics supplied through goals, descriptions, or human data. These assumptions do not hold in reveal-gated POMDPs, where dynamics and ontologies may not be known beforehand and must instead be grounded through iterative exploration.

\subsection{Positioning of WATERFALL and Evaluated Baselines}
We discuss the following reward-search methods as \emph{closely related work} rather than empirical baselines. Their \emph{privileged information} assumptions help position our discovery setting. \textbf{LaRes} searches over reward populations using explicit task descriptions, environment variables, and success-rate feedback \citep{li2025lares}. \textbf{RF-Agent} drives its MCTS-based reward design with explicit task information and training feedback \citep{gao2025rfagent}. Likewise, \textbf{R*} injects a task description and goal, combining reward-structure evolution with preference-based parameter alignment \citep{li2025rstar}. \textbf{GRACE} avoids scalar success rates by recovering rewards from expert demonstrations, which themselves supply privileged task supervision \citep{sapora2026grace}. \textbf{RE-GoT} evaluates videos with a VLM, but also receives an environment abstraction and task description to construct a text-attributed graph \citep{yao2025regot}. These methods use supplied task specifications or demonstrations, whereas \textbf{WATERFALL tests discovery \emph{without} them}, particularly important in reveal-gated POMDPs.

We evaluate \textbf{WATERFALL} against an extensive \emph{coverage set} spanning reward-specification interfaces, while preserving each baseline's native information interface. \textbf{CAR-VLM} \citep{pmlr-v235-venuto24a} represents vision-conditioned programmatic reward synthesis. \textbf{T2R} \citep{xie2024text2reward} represents one-shot language-to-reward-code generation. \textbf{L2R} \citep{pmlr-v229-yu23a} represents language-conditioned reward parameterisation. \textbf{EUREKA} \citep{ma2024eureka} represents iterative LLM-based evolutionary reward-code optimisation driven by downstream training feedback. Process Reward Models \textbf{(PRMs)} \citep{zhang2025prm4rl} provide a diagnostic comparison of visual progress scoring. Together, these baselines span the principal axes of \emph{modality}, \emph{reward representation}, and \emph{refinement style}. We also include \textbf{RL} and \textbf{RL wCBE/ICM/RND} as non-foundation-model comparisons for sparse reward and exploration \citep{tang2017exploration,pmlr-v70-pathak17a,burda2019rnd}. Remaining baselines are excluded since they are variants of these same specification axes, which the coverage set samples.

Beyond coverage, WATERFALL differs informationally and algorithmically. It selects elites from \emph{\textbf{purely visual proxy judgements}}, without any access to privileged information or undiscovered task structure, inherent in POMDPs \citep{kaelbling1998planning,chevalier2019babyai,NEURIPS2023_e8916198}. Related comparison-based approaches include Bradley--Terry models \citep{bradley1952bt} and duelling bandits \citep{yue2012dueling}. WATERFALL uses Swiss-system tournaments \citep{fide2023tiebreak} to aggregate repeated pairwise evidence under noisy, seed-sensitive evaluation \citep{agarwal2021precipice} before longitudinal histories drive elite-guided mutation, reducing champion brittleness and isolated VLM misclassification. We claim \emph{minimal}, not \emph{zero}, semantics, since executable reward generation cannot proceed without a code interface. Ours exposes the same state and action vectors available to the policy, while the sparse reward term, episode length, and environment dynamics remain hidden.

\section{Methods}
\label{sec:methods}
\vspace{-0.15cm}
\paragraph{Problem setting and strict observability parity.}
We consider fully observable and reveal-gated partially observable environments. In the fully observable regime, the environment is an MDP
\(\mathcal{M}_{\mathrm{FO}}=(\mathcal{S},\mathcal{A},P,r_{\mathrm{ext}},\gamma)\);
in the partially observable regime, it is a POMDP
\(\mathcal{M}_{\mathrm{PO}}=(\mathcal{S},\mathcal{A},P,\Omega,O,r_{\mathrm{ext}},\gamma)\),
where observations \(o_t\sim O(\cdot\mid s_t)\). Let
\(\mathcal{F}^{\pi}_t=\sigma(o_{0:t},a_{0:t-1})\)
denote the sigma-algebra generated by the policy-visible observation--action history.
The external task reward \(r_{\mathrm{ext}}\) is sparse and terminal.
We study reward discovery under \emph{strict observability parity}: the reward generator may use only policy-visible history plus legitimately revealed knowledge,
\[
\mathcal{F}^{\mathrm{gen}}_t
=
\mathcal{F}^{\pi}_t \vee \sigma(K_t),
\]
where \(K_t\) is monotonically non-decreasing and contains only semantics made observable by the environment. FO variants expose relevant variables from \(t=0\) (\(K_t\equiv K\)); PO variants retain masking and reveal-gating. Providing full semantic schemas for as-yet unobserved entities would therefore violate the protocol. Semantic task information is not intrinsically undesirable, as parity restricts \emph{asymmetric} disclosure, so a task specification shared by the policy and reward generator remains compatible. It constrains the information used to construct a reward, whereas deployability concerns the signals needed at execution. Goal-free experiments test a harder discovery setting, where visual judgements still rely on pretrained priors and the same behaviour can fit multiple objectives.

\paragraph{Training and evidence collection.}
For each candidate \(k\), WATERFALL trains an independent PPO policy \(\pi_{\theta_k}(a_t\mid \mathcal{F}^{\pi}_t)\) using its generated reward program under a common optimiser and budget (Algorithm~\ref{alg:waterfall}). Rollout interiors are sampled independently of environment events under \(\pi_{\mathrm{sample}}(t)\), with boundary frames always included, \(\{o^v_0,o^v_T\}\subseteq\tau_{\mathrm{raw}}\), so the judge infers success from non-curated evidence rather than event-triggered captures. Candidates are assessed at $t\in\{1,10,20,\ldots,100\}$M steps for Env.~I--III and $t\in\{0.5,1,2,\ldots,10\}$M for Env.~IV, where the first checkpoint is a fail-fast screen and the last is $T_{\max}$; training progress $p=t/T_{\max}$ contextualises early behaviour. PPO details, sampling schedules, hyperparameters, and the full Algorithms~\ref{alg:waterfall-part1}--\ref{alg:waterfall-part2} are in Appendices~\ref{app:full_methods} and~\ref{app:evaluation_protocol}.

\begin{figure}[h]
\vspace{-0.5cm}
\begin{algorithm}[H]
\caption{\textbf{WATERFALL}: Condensed Algorithmic Pseudocode}
\label{alg:waterfall}
\small
\begin{algorithmic}[1]
\Require environment context \(\mathcal{E}\), candidates \(K\), checkpoints \(\mathcal{T}=\{t_m\}\), budget \(T_{\max}\)
\State \textbf{Stage I:} assign personas \(\{p_k\}_{k=1}^K\); generate \(\{P_{k,0}\}\); initialise policies \(\{\pi_{\theta_k}\}\)
\State \(\mathcal{K}\leftarrow\{1,\dots,K\}\)
\For{\(t\in\mathcal{T}\) \textbf{with} \(t\le T_{\max}\) \textbf{and} \(\mathcal{K}\neq\emptyset\)}
  \State \textbf{RL:} train each \(k\in\mathcal{K}\) under \(P_k\) until \(t\)
  \State \textbf{Inference:} collect rollout evidence \(\mathcal{R}_{k,t}\) including \(\{o^v_0,o^v_T\}\)
  \State \textbf{Stage II:} run Swiss pairwise judging; update standings and histories \(\mathcal{H}_{k,\le t}\)
  \State \textbf{Stage III:} assessment actor outputs \(d_{k,t}\in\{\texttt{CONTINUE},\texttt{ABORT},\texttt{STOP},\texttt{ESCALATE-A/B}\}\)
  \If{there exists \(k\) with \(d_{k,t}=\texttt{STOP}\)}
    \State optionally terminate the run
  \EndIf
  \ForAll{\(k\in\mathcal{K}\)}
    \If{\(d_{k,t}\in\{\texttt{ESCALATE-A},\texttt{ESCALATE-B}\}\)}
      \State update \(P_k\) using candidate history and elite priors
    \EndIf
  \EndFor
  \State \(\mathcal{K}\leftarrow\{k\in\mathcal{K}: d_{k,t}\notin\{\texttt{ABORT},\texttt{STOP}\}\}\)
\EndFor
\end{algorithmic}
\end{algorithm}
\vspace{-0.75cm}
\end{figure}

\subsection{Architecture Stage I: Persona-Conditioned Reward Synthesis}
\begin{wrapfigure}[24]{r}{0.4\linewidth}
    \centering
    \vspace{-0.5cm}
\includegraphics[
    width=\linewidth,
    trim={0.15cm 0.25cm 2.5cm 0.10cm},
    clip
]{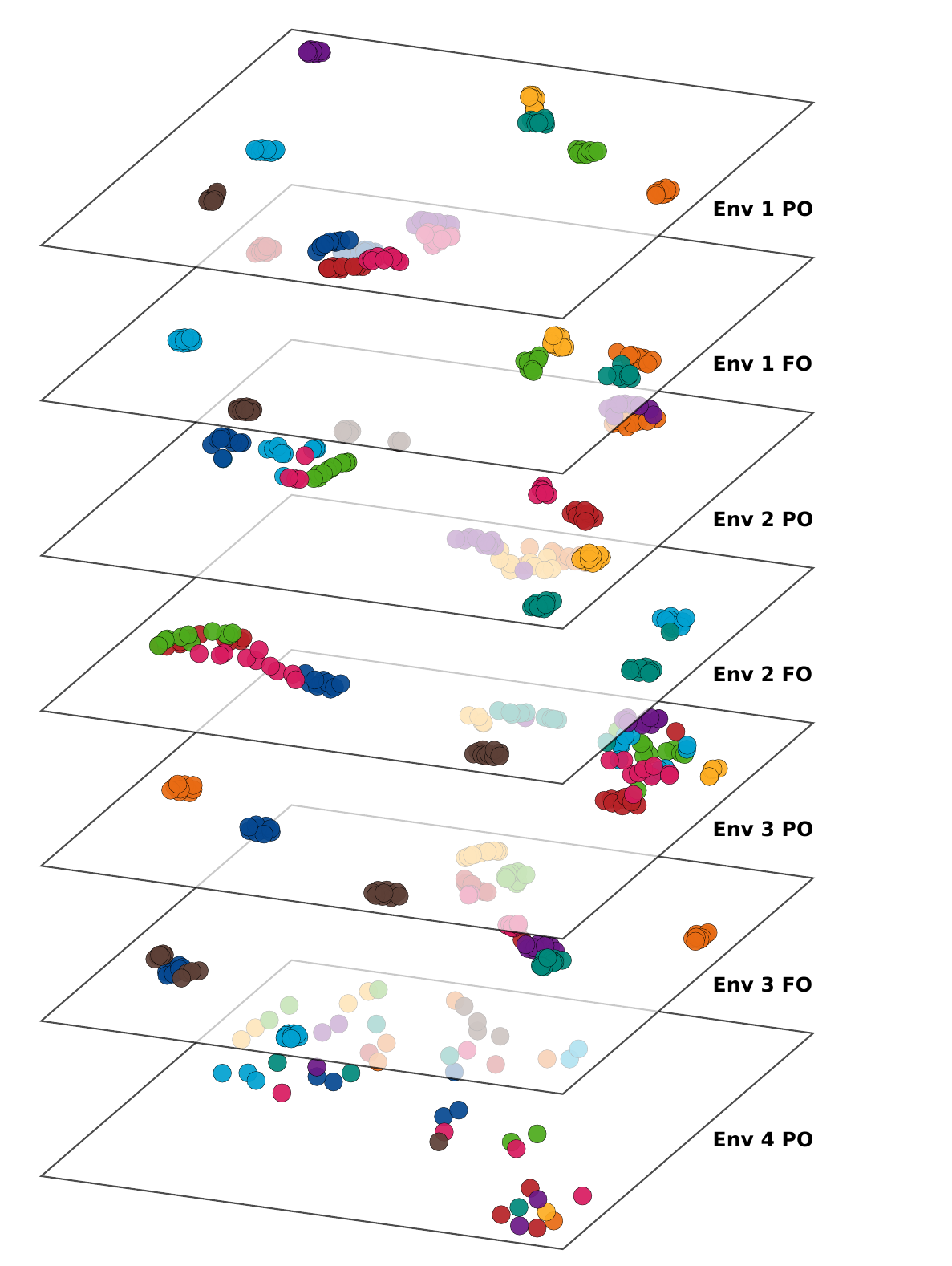}
    \includegraphics[width=\linewidth]{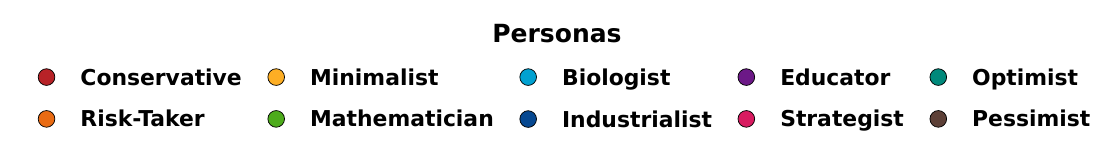}
    \vspace{-0.5cm}
    \caption{\textbf{Persona-Conditioned Behavioural Diversity t-SNE.}}
    \label{fig:stage_i_persona_tsne}
\end{wrapfigure}

To diversify initial reward hypotheses, WATERFALL conditions VLM code generation on a fixed persona set \(\mathcal{P}=\{p^{(1)},\dots,p^{(P)}\}\) (we use \(P=10\)). Candidate \(k\) is assigned a fixed persona \(p_k\in\mathcal{P}\), and its initial reward program is sampled as
\begin{equation}
P_{k,0} \sim p_{\mathrm{VLM}}\!\left(\,\cdot \mid \mathcal{E}, p_k \right),
\end{equation}
where \(\mathcal{E}\) contains the permitted observation-action coding interface, excluding task specifications, ground-truth reward code, and unrevealed dynamics. Keeping \(p_k\) fixed across the candidate lifetime preserves a coherent inductive bias, allowing tournament selection and later mutations to improve the population. Persona inventories and prompts are in Appendix~\ref{app:prompts}. A qualitative t-SNE diagnostic (\autoref{fig:stage_i_persona_tsne}) shows persona-conditioned candidates occupying heterogeneous regions of behaviour space rather than one reward-design mode, suggesting distinct behavioural priors. Details in Appendix~\ref{app:persona_diversity}.

\subsection{Architecture Stage II: Vision-Judge Swiss Tournament Evaluation}
Stage~II bounds multimodal prompt load through pairwise comparisons of rollout evidence \(\mathcal{R}_{k,t}=\{\rho^{(e)}_{k,t}\}_{e=1}^{E}\), where \(\rho^{(e)}_{k,t}\) is the sampled frame sequence of inference episode \(e\) for candidate \(k\) in the active set \(\mathcal{K}_t\), so each judge call carries \(2E\) rollouts regardless of \(|\mathcal{K}_t|\).

\paragraph{Vision-only pairwise judge.}
We use a vision-only judge \(\mathcal{J}(\mathcal{R}_{i,t},\mathcal{R}_{j,t}) \rightarrow (w_{i,j,t}, c_{i,j,t})\), whose parsed outcome comprises \(w_{i,j,t}\in\{i,j,\bot\}\) and confidence \(c_{i,j,t}\in[0,1]\) from rollout evidence alone, with \(\bot\) denoting no valid winner. No returns, success rates, learning curves, or ground-truth reward signals are exposed. The judge also emits per-episode \texttt{SUCCESS/FAIL/UNCERTAIN} labels and qualitative progress signals, with outcome counts used only as supporting evidence. Pairwise comparison is a natural fit for ranking under relative feedback \citep{bradley1952bt, yue2012dueling}.

\paragraph{Swiss tournament and elite pool.}
We rank candidates with a Swiss-system tournament of up to \(R\) rounds (configured \(R=5\)), which keeps the comparison budget bounded while preferentially matching high-ranked candidates and avoiding rematches. Tournament-style ranking is also a standard evolutionary selection mechanism \citep{goldberg1991tournament,blickle1996tournament}. We rank by Swiss hard points, breaking ties by Buchholz strength-of-schedule \citep{fide2023tiebreak}, with confidence-weighted soft scores as auxiliary evidence. At each checkpoint, the standings induce a small tie-aware elite pool \(\hat{E}\) of leading or near-tied candidates. Scoring rules, tie-break computation, and elite-pool caps are in Appendix~\ref{app:full_methods}. Quantitative and qualitative tournament digests are appended to lightweight per-candidate history files for Stage~III. The qualitative digests help the assessor \textbf{\emph{guard against hallucinated success}} by preserving contradictions and uncertain terminal interpretations that raw \texttt{SUCCESS} counts would conceal (see Appendix~\ref{app:robustness}).

\subsection{Architecture Stage III: Longitudinal Assessment and Elite-guided Mutation}
\paragraph{Assessment actor.}
At each checkpoint, the per-candidate assessment actor receives the reward program \(P_{k,t}\), history \(\mathcal{H}_{k,\le t}\), current tournament digest, boundary evidence \(\{o^v_0,o^v_T\}\), training progress \(p=t/T_{\max}\), and elite reward programs as priors. Its decision \(d_{k,t}\) is \texttt{CONTINUE}, \texttt{ABORT}, \texttt{STOP}, \texttt{ESCALATE-A}, or \texttt{ESCALATE-B}. Either escalation produces an updated program \(P_{k,t+1}\).

\paragraph{Longitudinal evidence.}
The actor is non-Markovian at the meta-level: it conditions on append-only lineage logs and current plus immediately prior tournament digests to detect whether the last mutation improved behaviour, while keeping ranking and elite selection tied to the current checkpoint.

\paragraph{Decisions.}
Checkpoint decisions combine judge-reported outcome labels, qualitative progress evidence, tournament standings, and longitudinal history, but never ground-truth task-success metrics. \texttt{STOP} records judged task completion, \texttt{ABORT} persistent failure, \texttt{CONTINUE} plausible or inconclusive progress, and escalation suspected reward misspecification or insufficient exploration. Results~II highlight WATERFALL's robustness to \texttt{STOP} calls, identifying successful reward candidates.

\paragraph{Elite-guided mutation.}
Under \texttt{ESCALATE-A}, the actor debugs or retunes reward logic, e.g., rescaling terms, adjusting termination conditions, or restructuring shaping components, without modifying environment dynamics or the action interface. Under \texttt{ESCALATE-B}, it adds exploration-driving mechanisms such as intrinsic bonuses or entropy-regularised incentives. Mutations use candidate-isolated prompts to avoid cross-contamination, optionally adapting compatible elite terms into \(P_{k,t}\) via guided mutation rather than crossover. Prompt contracts and mutation templates are in Appendix~\ref{app:prompts}.

\vspace{-0.3cm}
\section{Experimental Setup and Evaluation}
\vspace{-0.15cm}
\label{sec:evaluation}
\paragraph*{Experimental Overview} To achieve the observability constraints inherent in reveal-gated POMDPs, vision evaluations use fog-masked rollouts for Env.~I--III POs, and physically occluded for Env.~IV, matching each policy's observability regime. FOs instead use unmasked rollouts. WATERFALL receives precisely \autoref{fig:results_overview} (top row), preserving policy--architecture visual parity. Text-only baselines, including WATERFALL's no-images ablation, receive deterministic serialisations of the same observation stream with gated reveals (Appendix~\ref{app:textual_parity}). Zero-shot baselines receive the full POMDP ontology at generation time, which favours them but avoids conflating missing iterative discovery with reward-code quality. We preserve each method's native information interface so that comparisons evaluate reward quality rather than whether it can operate outside its design assumptions (\autoref{tab:implemented_interfaces}).

\paragraph*{Randomness, Seeds, Sampling and Evaluation Protocol}
All reward-synthesis methods receive \(K=10\) initial candidates, and each valid candidate is trained with \(n=10\) independent RL seeds for Env.~I--III and \(n=5\) for Env.~IV. Each baseline's native information interface and refinement protocol is preserved, even if advantageous. Curves average task returns across training seeds, conditional on the initial candidate portfolio. VLM rollouts use a median-ranked seed within each candidate to avoid best-seed cherry-picking and limit sensitivity to atypical training outcomes (Appendix~\ref{app:representative_inference_sampling}).

\subsection{Environments}
\label{results_environments}
We evaluate procedurally randomised discrete and continuous-control environments under \textbf{partial observability with strict parity} and \textbf{minimal semantic side-information}. Layout randomisation and departures from the original environments reduce LLM/VLM reliance on memorised task priors. Env.~I--III include FO and PO variants, whereby FO exposes task-relevant observations from $t=0$, while PO reveals them only after exploration. Env.~IV has no FO variant, deliberately, since a concealed button sits behind a solid cover, and no camera, depth or tactile sensor in reality can recover an object through an opaque barrier. Exposing it would mean reading privileged states rather than sensing, so a reward designer using that channel would work only in simulation and fail at deployment, which is exactly the asymmetry parity excludes. The environments span sparse-reward exploration, memory, deception, tool-use, and sequential chaining. Full details, including hyperparameters, observation, action, and reward specifications are in Appendix~\ref{app:environment_details}.

\paragraph{Environment I: \enquote{MiniGrid's Locked Room Variant} \cite{NEURIPS2023_e8916198}} Inspired by MiniGrid's \texttt{LockedRoomEnv}, this six-room, \(13{\times}13\), gridworld omits the mission string to avoid goal-conditioned hints. The agent must discover the key\(\rightarrow\)door\(\rightarrow\)goal-room dependency, whereby the matching key lies outside the randomly selected locked room containing the goal chest. Other doors are distractors. PO observations are local and egocentric, while FO exposes global task-relevant state from episode start. Success occurs by unlocking the room containing the goal chest.

\paragraph{Environment II: \enquote{Dark Corridor} \cite{NEURIPS2023_e8916198}} Inspired by MiniGrid's \texttt{Memory}, this randomised, \(9{\times}9\), cross-shaped corridor requires discovering tool\(\rightarrow\)blocker\(\rightarrow\)goal dependencies: a locked door, fire, and enemy block the goal arm in random order, while the matching key, bucket, and sword occupy the remaining arms. Success requires clearing all three blockers with the appropriate tools. PO observations are local and occlusion-limited, whereas FO additionally exposes global entity positions and states from episode start.

\paragraph{Environment III: \enquote{Medusa's Revenge}} This randomised two-chamber, \(19{\times}19\), gridworld requires the agent to acquire a key, unlock the second chamber, and obtain an artefact to defeat a rotating guardian protecting the true goal. The guardian emits a lethal ray with warning and active phases, while a deceptive chest near the start terminates the episode. In PO, the second chamber and its associated observations remain hidden until the door is unlocked, whereas FO exposes it from start.

\paragraph{Environment IV: \enquote{Reveal-Gated Robotic Button Task}} Inspired by Metaworld's button press \cite{pmlr-v100-yu20a}, this continuous-control POMDP uses a simulated Franka arm to discover a button$\rightarrow$cover$\rightarrow$button dependency. Pressing the visible button opens a cover that physically occludes a second button, whose coordinates remain masked until opening finishes and pressing it completes the task. Three continuous Cartesian force actions control an end-effector IK target.

\section{Results}
\label{sec:results}
\subsection*{Results I: WATERFALL vs Baselines in POMDP and FOMDP Settings}
\label{subsec:results_I}
\paragraph*{POMDP Results.} We first compare WATERFALL with the baselines under the evaluation protocol in \autoref{sec:evaluation}. From \autoref{fig:results_overview}.(A.1-A.3 and A.7), we initially observe that WATERFALL outperforms baselines in the intended reveal-gated POs. We also observe that in Env.~I (A.1) several baselines learn meaningful policies, showing that the task is not constructed to be solvable only by WATERFALL. WATERFALL attains higher returns and greater sample efficiency (except L2R in A.1), even though baselines retain their native information advantages. Consult \autoref{fig:results_overview} (top row), \enquote{Privileged Inputs}. Zero-shot baselines L2R, T2R and CAR-VLM receive the full reveal-gated ontology at generation time ($t=0$), an additional information advantage. In contrast, EUREKA's inherent iterative refinement architecture is not given this additional advantage and must discover the reveal-gated ontologies through exploration, as intended in POMDPs, which WATERFALL also shares. Even still, WATERFALL notably outperforms EUREKA despite EUREKA accessing ground-truth metrics and goal descriptors, which our implementation never receives. This indicates WATERFALL's advantage stems not merely from iterative refinement, but from \textbf{\emph{iteration grounded in non-privileged behavioural evidence}}, \textbf{\emph{vision-judge tournament selection}}, and \textbf{\emph{longitudinal reward mutation}}. 

\vspace{-0.15cm}
\paragraph*{FOMDP Results.} In the FO variants (\autoref{fig:results_overview}.A.4--A.6), where task-relevant state is visible from \(t{=}0\), several baselines improve, indicating that privileged observability reduces the reward-discovery burden. This is explicitly why we test FO and PO. Notably, FO does not necessarily make the problem easier for sparse-reward RL; it helps the reward designers as the task is visible to them.

\vspace{-0.15cm}
\paragraph*{PRM Cost and Fidelity Diagnostic.} We evaluate visual PRMs as a complementary progress-scoring diagnostic on random-policy and frozen $1$M/$10$M Env.~III PO rollouts (costing $\approx\$24.61$) under (i) \textbf{vision-only}, (ii) \textbf{memory-augmented}, and (iii) \textbf{privileged-rubric} settings. Agreement with the instrumented progress reference is limited (best MCC $0.32$; \autoref{fig:results_overview}.D). Most importantly, applying PRMs online during PPO would require repeated foundation-model calls, scaling poorly with environment steps with substantial costs for this configuration, with costs extrapolated to $\approx\$81.7$k for $10$M steps, at stride $4$ which is a projection. For full details consult the Appendix~\ref{app:prm_diagnostic}.

\subsection*{Results II: WATERFALL's VLM Judge Audit of STOP Calls}
\label{subsec:results_II}
\begin{figure}[htbp]
\centering
\vspace{-1cm}
\begin{subfigure}[t]{1.0\textwidth}
  \centering
  \includegraphics[width=\linewidth, trim=15 15 15 15,
    clip]{figures/waterfall_environment_overview.pdf}
\end{subfigure}

\begin{subfigure}[t]{0.3295\textwidth}
  \centering
  \includegraphics[width=\linewidth, trim=3 3 3 3,
    clip]{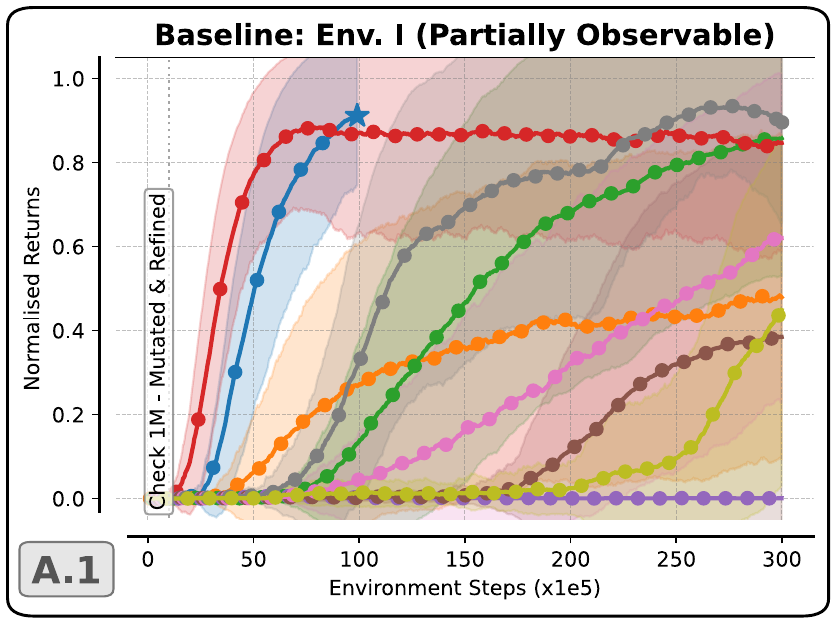}
\end{subfigure}
\begin{subfigure}[t]{0.3295\textwidth}
  \centering
  \includegraphics[width=\linewidth, trim=3 3 3 3,
    clip]{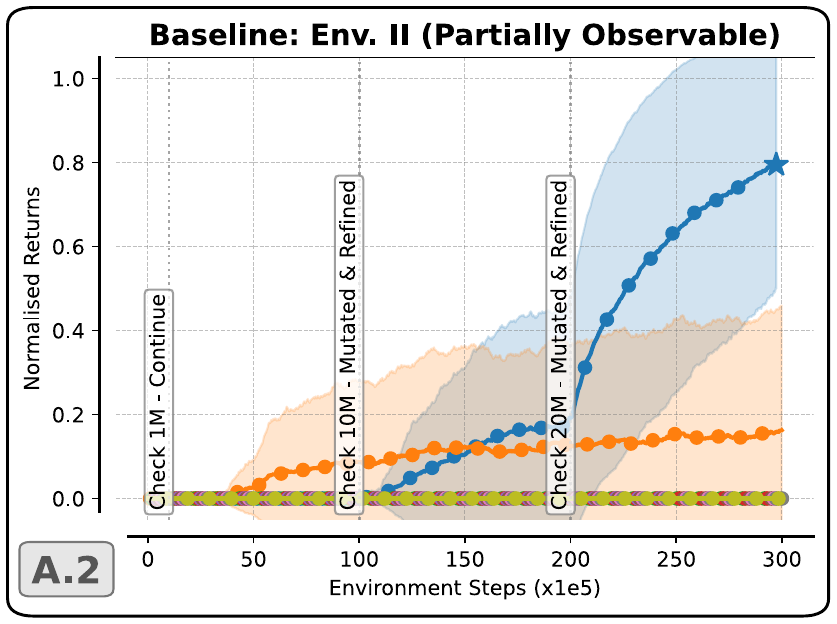}
\end{subfigure}
\begin{subfigure}[t]{0.3295\textwidth}
  \centering
  \includegraphics[width=\linewidth, trim=3 3 3 3,
    clip]{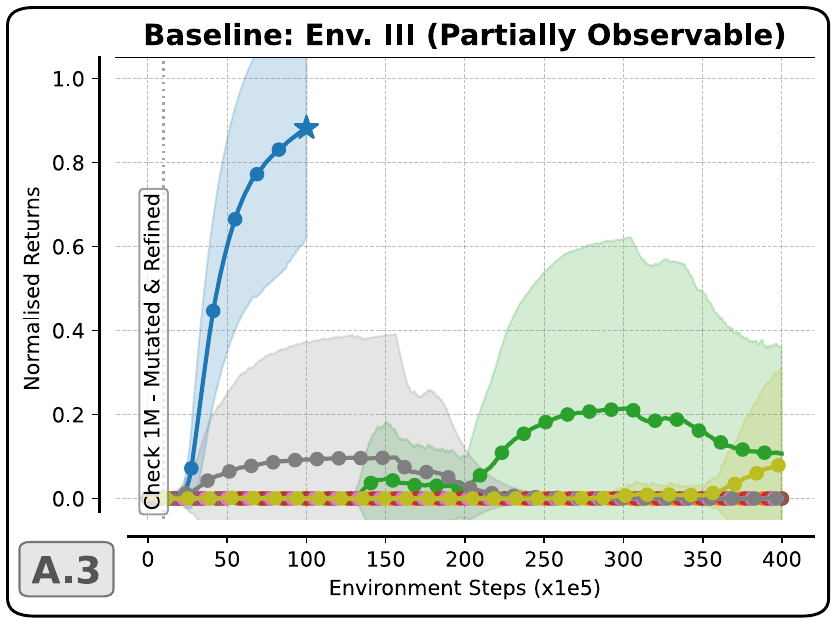}
\end{subfigure}

\begin{subfigure}[t]{0.3295\textwidth}
  \centering
  \includegraphics[width=\linewidth, trim=3 3 3 3,
    clip]{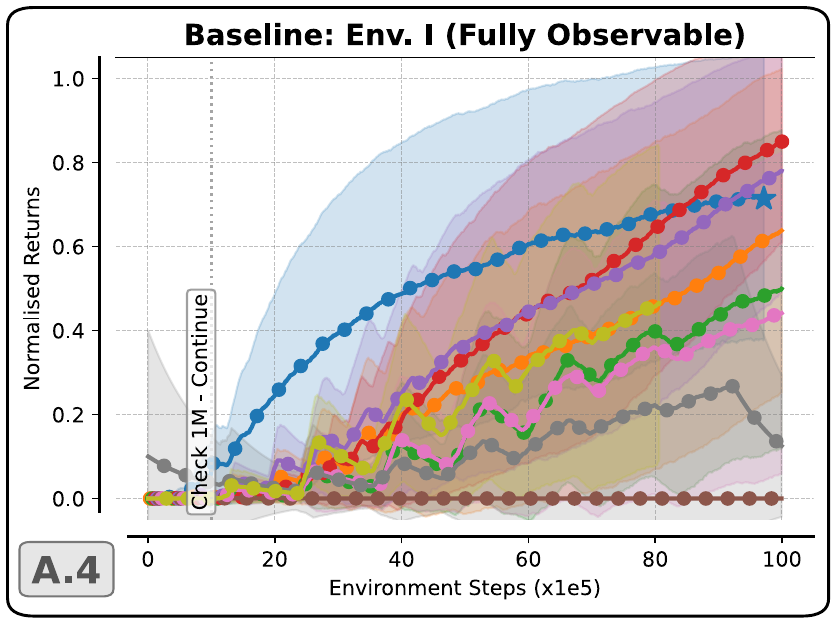}
\end{subfigure}
\begin{subfigure}[t]{0.3295\textwidth}
  \centering
  \includegraphics[width=\linewidth, trim=3 3 3 3,
    clip]{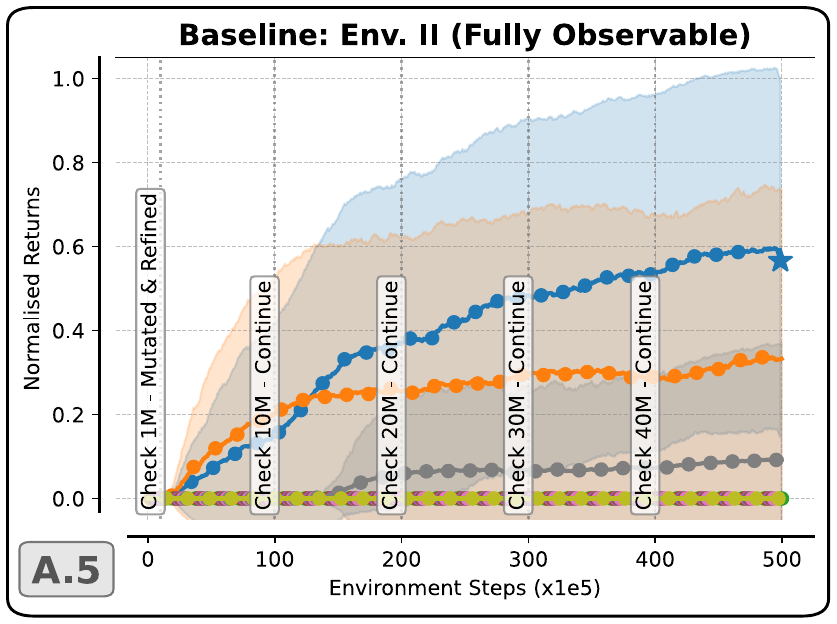}
\end{subfigure}
\begin{subfigure}[t]{0.3295\textwidth}
  \centering
  \includegraphics[width=\linewidth, trim=3 3 3 3,
    clip]{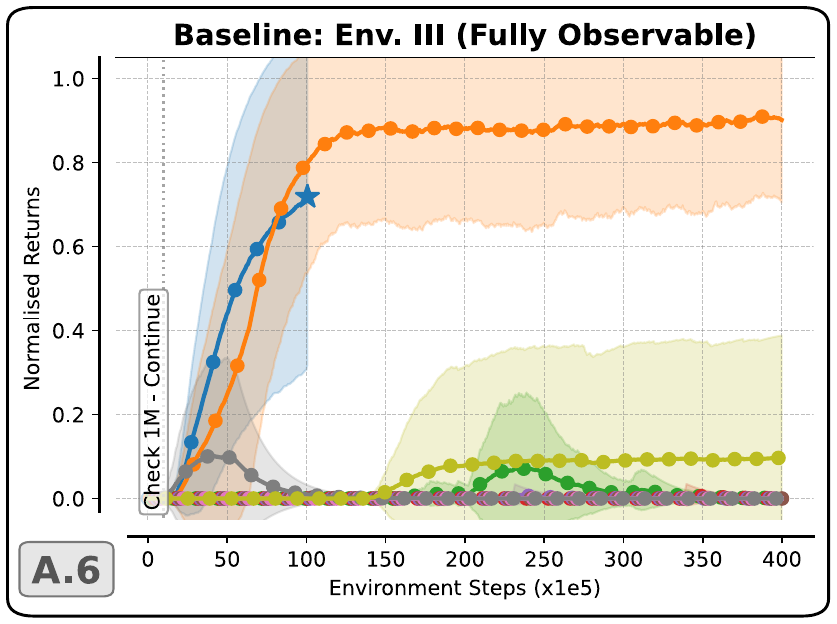}
\end{subfigure}

\begin{subfigure}[t]{0.3295\textwidth}
  \centering
  \includegraphics[width=\linewidth, trim=3 3 3 3,
    clip]{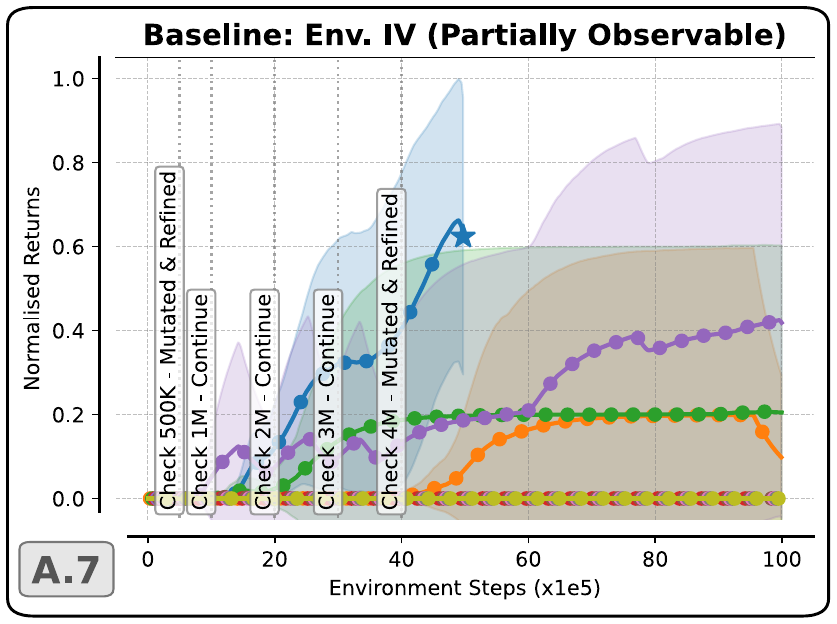}
\end{subfigure}
\begin{subfigure}[t]{0.3295\textwidth}
  \centering
  \includegraphics[width=\linewidth, trim=3 3 3 3,
    clip]{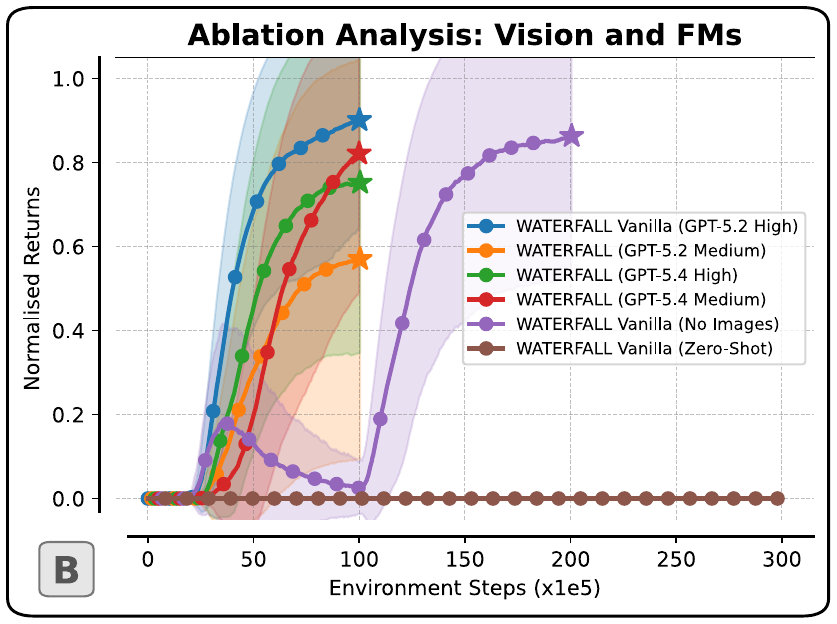}
\end{subfigure}
\begin{subfigure}[t]{0.3295\textwidth}
  \centering
  \includegraphics[width=\linewidth, trim=3 3 3 3,
    clip]{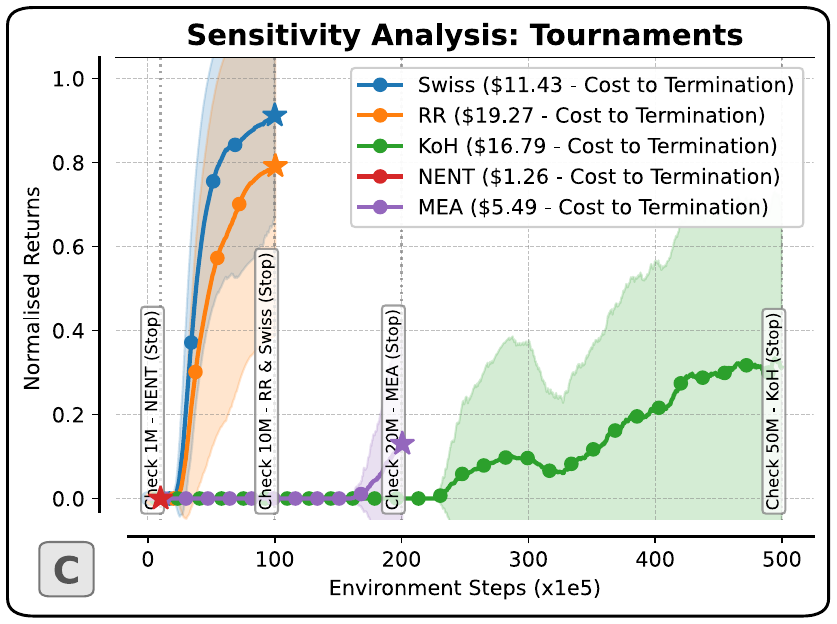}
\end{subfigure}%

\par\vspace{-0.35cm}
\noindent
\begin{subfigure}[t]{0.33\textwidth}
  \vspace{0pt}
  \centering
  \includegraphics[width=\linewidth, trim=3 3 3 3]{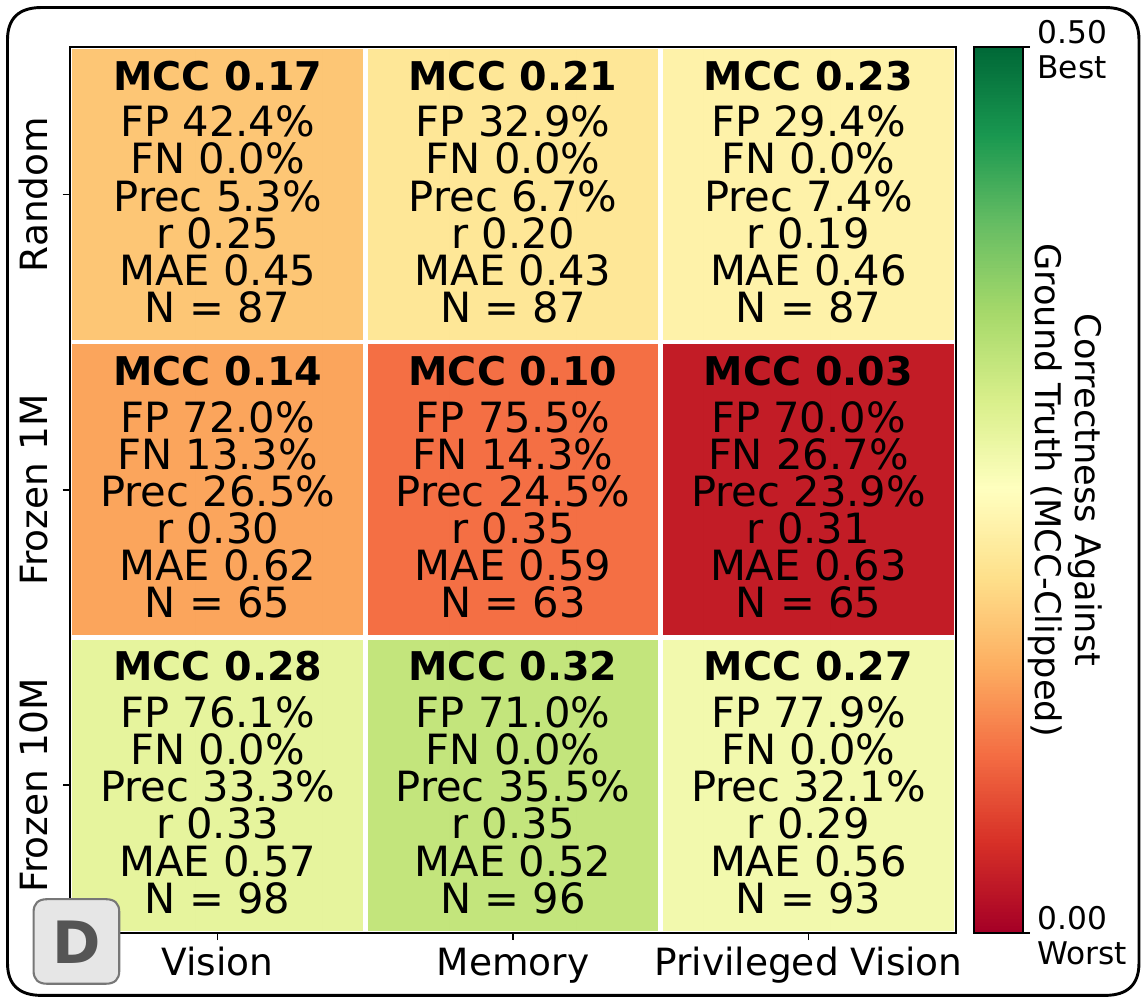}\\[2pt]   \parbox{\linewidth}{\scriptsize\raggedright Rows are rollout conditions, columns PRM regimes. Cell colour shows clipped MCC (higher is better); reporting MCC, FP/FN rates, precision, Pearson $r$, MAE, and sample count.}
\end{subfigure}\hfill%
\begin{subfigure}[t]{0.665\textwidth}
  \vspace{0pt}
  \centering
  \includegraphics[
    width=\linewidth,
    trim=8 8 8 8,
    clip
  ]{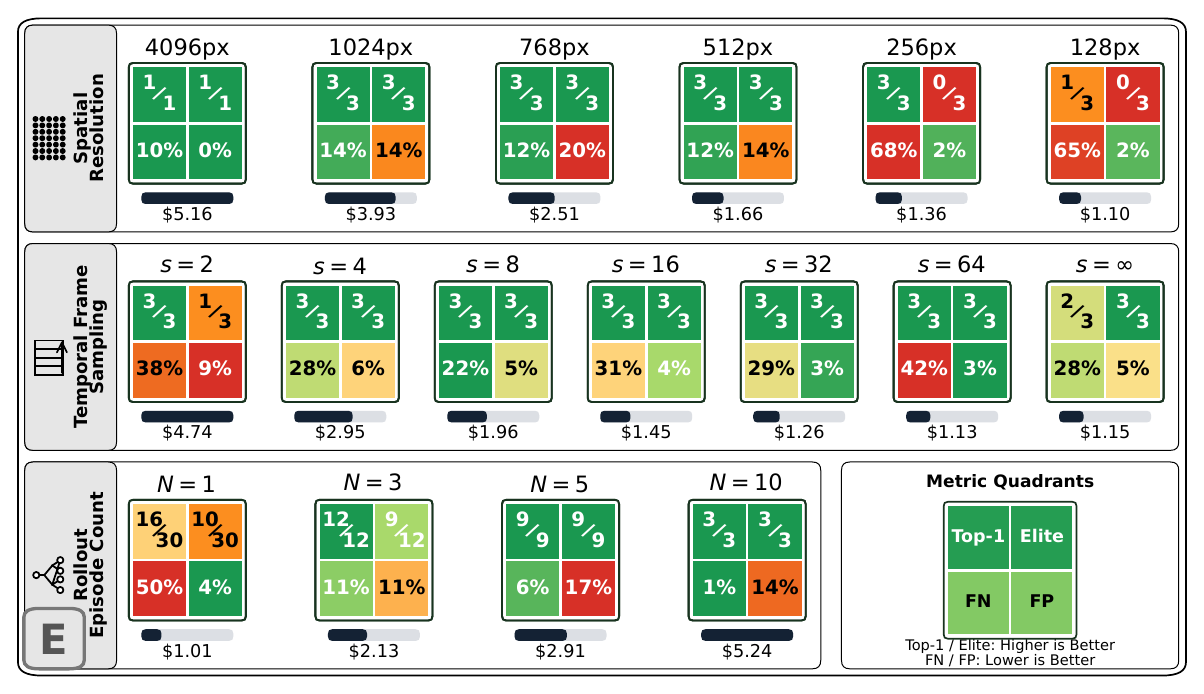}
\end{subfigure}
\caption{\textbf{Results Overview.} \small \textbf{Top Row:} Environments and baselines, with the inputs they consume. \textbf{A.1-7:} WATERFALL and baselines. \textbf{B:} Vision ablation and foundation-model sensitivity. \textbf{C:} Downstream tournament topologies. \textbf{D:} PRM ground-truth accuracy. \textbf{E:} Vision sensitivity at the frozen $10$M checkpoint for Env. III. \(E^\star=\{k5,k8\}\); three independent judge runs per condition. Quadrants show top-1 solved, \(E^\star\subseteq\hat{E}\), FN, and FP. Top-1/Elite colours are absolute, FN/FP block-normalised, with raw rates printed. Episode-count denominators are subsets \(\times\) judge runs. Bars show mean USD/run. \textbf{Learning curves:} Smoothed mean $\pm$ one standard deviation of normalised ground-truth returns over $n=10$ training seeds ($n=5$ in A.7). Stars mark judge \texttt{STOP} calls. Overlapping zero-return baseline markers are staggered for visibility. Reward-synthesis baselines show the best of $K$ candidates by ground-truth success, while WATERFALL uses VLM selection.}
\label{fig:results_overview}
\vspace{-0.5cm}
\end{figure}

We first analyse the variation of the plotted mean returns (see \autoref{fig:results_overview}.A) which is the result of the median-seed rollout assessed by the judge exhibiting consistent success, whereas the curve averages all seeds including weaker non-median runs. Stage~III's \texttt{STOP} decision reflects behaviour judged from a representative training seed in Stage~II's vision judge. Success in the judged rollouts therefore need not coincide with a mean return near one. We audit all primary \texttt{STOP} calls (eleven) across Env.~I--IV.

A candidate passes when at least $8/10$ seeds in Env.~I--III, or $4/5$ in Env.~IV, achieve $\geq50\%$ completion over their last $1{,}000$ logged training episodes. Ten of eleven candidates pass, with $94/105$ seed runs ($89.5\%$) meeting the threshold (\autoref{tab:stop_audit}) and exhibiting overall robustness. The exception is Env.~II FO, where \texttt{STOP} at $50$M steps per policy leaves only $6/10$ seeds meeting the threshold. Env.~III FO illustrates the remaining variation: eight seeds perform well ($85.3$--$92.1\%$) while two perform poorly ($11.1$--$12.1\%$). Completion rates and checkpoints are in the Appendix~\ref{app:stop_audit}. These findings motivate the vision ablation in Results~III and the visual sensitivity tests in Results~VI.

\subsection*{Results III: Ablation Evaluation of Vision and Iterative Refinement}
From \autoref{fig:results_overview}, WATERFALL's full visual iterative refinement is important: zero-shot remains flat at zero, while removing images notably weakens convergence. Visual rollout assessment and checkpointed refinement therefore jointly drive sample-efficient reward discovery in the POMDP.

\subsection*{Results IV: Sensitivity Evaluation of Foundation Model}
\label{results_foundation_model_sensitivity}
Next, we test sensitivity to the foundation model used for Stage~II judging and Stage~III assessment, mutation, and escalation in Env.~III, in two settings. \textbf{Results IV.A: Frozen Evaluator Sensitivity} With the \(10\)M checkpoint frozen and only the judge varied across \(16\) foundation-model configurations (Appendix, \autoref{fig:foundation_model_sensitivity}), the cost-effective \textit{gpt-5-mini} and \textit{gpt-5.4-mini (high)} recover the solved elite pair \(E^\star=\{k5,k8\}\) as reliably as the costlier \textit{gpt-5.2/5.3} and \textit{gpt-4.1}, whereas nano-scale models are substantially less reliable. \textbf{Results IV.B: End-to-End Downstream Branching} We rerun WATERFALL end-to-end from Stage~I with \textit{gpt-5.2} (high reasoning, our default, and medium) and \textit{gpt-5.4} (high and medium). \autoref{fig:results_overview}.B shows that WATERFALL is not dependent on the latest frontier model, although convergence speed, selection stability, and final return remain model-sensitive.

\subsection*{Results V: Balancing Tournament Topology}
\label{results_tournaments_ablation}
We compare \textbf{KoH}, \textbf{Round Robin (RR)}, and \textbf{Swiss (ours)} on frozen checkpoints and in downstream branches. KoH is cheap (\(|\mathcal{K}_t|-1\) matches) yet order-sensitive, RR is exhaustive (\(|\mathcal{K}_t|(|\mathcal{K}_t|-1)/2\)) yet expensive, and Swiss (\(\approx R\lfloor|\mathcal{K}_t|/2\rfloor\)) provides a tunable middle ground. In Results V.B, we add two non-tournament controls: \textbf{No Evolution, No Tournaments (NENT)}, with only independent Stage~III self-refinement, and \textbf{Monolithic Evolutionary Assessment (MEA)}, which keeps evolutionary refinement but replaces pairwise tournaments with absolute all-candidate judging. See Appendix~\ref{app:topology_ablation} for selection regret ($\Delta_{\text{sel}}(t)$), top-3 ranking fidelity against RR ($\mathrm{IoU}_3(t)$), and judge-call cost.

\paragraph*{Results V.A: Frozen Topology --- Swiss Matches RR} 
Freezing \(\mathcal{K}_t\) and \(\{\mathcal{R}_{k,t}\}\) at \(t=1\)M and \(t=10\)M isolates judge topology (\autoref{tab:frozen_topology_ablation_summary}). At \(1\)M, RR and Swiss select \(k7\), while KoH selects \(k1\). At \(10\)M, Swiss exactly matches RR on top-\(3\), \(T^{\mathrm{Swiss}}_3=T^{\mathrm{RR}}_3\), with \(\mathrm{IoU}^{\mathrm{Swiss}}_3=1.0\), whereas KoH remains topology-sensitive.  Costs are $\$2.97$, $\$8.48$ and $\$15.40$ for KoH, Swiss and RR respectively.

\paragraph*{Results V.B - Downstream Branch Extension --- Swiss Outperforms RR}
Under topology-specific selection (\autoref{fig:results_overview}.C) Swiss performs best, RR remains competitive yet costlier, and KoH remains brittle and produces weaker policies. NENT and MEA prematurely return \texttt{STOP} at $1$M and $20$M, respectively, despite failing to solve the task. MEA achieves non-zero returns thanks to its evolutionary refinement, yet remains brittle to VLM judge misclassification without tournaments. These results show that Swiss best balances cost and selection quality over RR and KoH, while NENT/MEA confirm that both evolutionary refinement and tournament-based calibration are necessary. Swiss's higher downstream return reflects its selection, mutation, pruning, and training path, without establishing superior ranking accuracy, since Swiss and RR produce the same top-three ranking and elite pair in Results V.A. Although KoH is cheaper per checkpoint, it requires six assessments to reach \texttt{STOP} at $50$M, compared with two for Swiss at $10$M, and still yields a weaker policy at greater total API cost.

\subsection*{Results VI: Sensitivity Evaluation of Visual Evidence}
\label{results_vision_evidence_sensitivity}
Since vision is central to WATERFALL, we test how reliably the VLM judge selects elites when (i) \textbf{image resolution}, (ii) \textbf{frame density}, or (iii) \textbf{rollout episode-count} is reduced. All sensitivity analyses use the frozen 10M checkpoint in Env. III, where the solved elite pair is identifiable as \(E^\star=\{k5,k8\}\). We omit 1M because early-training entropy renders candidate rankings too noisy for controlled evidence ablations. Across conditions, we report top-1 solved recovery, \(\hat{k}_1\in E^\star\), meaning that the highest-ranked candidate belongs to the high-success reference pair \(E^\star\). Elite-pair retention means that both remain in the selected elite pool \(\hat{E}\). Each trial is a complete Swiss tournament on fixed rollout evidence, repeated three times per subset. Episode-count evaluations (Results VI.C) also vary the selected episodes, with subset counts and trial totals reported in Appendix~\ref{results:sensitivity_vision_granularity}.

\paragraph{Direct Judge Accuracy and Confidence.}
Beyond elite recovery, we compare the judge's preferences with independently measured task success. With three episodes per candidate at $512$px and stride $s=8$, the judge selects the better-performing candidate in $169/180$ non-tied comparisons ($93.9\%$; Spearman $\rho=0.846$; Brier score $0.081$). Reference success is measured on the seven episodes not shown to the judge. Confidence is less well calibrated (five-bin ECE $0.157$; \autoref{tab:judge_correctness}). Overall, agreement declines when images become too coarse or rollout coverage too sparse.

\paragraph*{Results VI.A: Spatial Resolution Sensitivity}
\label{results_resolution_sensitivity}
We evaluate progressively downsampled rollout images, while keeping the candidate pool, judge, and temporal sampling policy fixed. From \autoref{fig:results_overview}.E, the solved elite pair is preserved from the \(4096\)px reference down to \(512\)px, with both top-1 solved recovery and \(E^\star\subseteq\hat{E}\) holding across all repeated judge runs. Below this threshold, \(256\)px still ranks a solved candidate top-1 yet loses solved-pair elite retention, while \(128\)px becomes unstable. Overall, the sharp increase in FN indicates that coarse visual evidence impairs structural discrimination.

\paragraph*{Results VI.B: Temporal Frame-Sampling Sensitivity}
We next vary the recording stride \(s\in\{2,4,8,16,32,64,\infty\}\) to test temporal granularity, FP/FN calibration, and cost. We used the \(s=2\) dataset as the highest-fidelity one to realise coarser finite strides (\(s\geq4\)), thereby reducing trajectory-realisation noise from policy stochasticity. \(s=\infty\) retains only start and terminal frames as an intentionally hard boundary-only condition. We fix the resolution at \(512{\times}512\)px to control API costs, while VI.A shows that this resolution preserves VLM judge reliability. From \autoref{fig:results_overview}.E, finite temporal subsampling of \(s\in\{4,8,16,32,64\}\) is largely robust, both top-1 solved recovery and \(E^\star\subseteq\hat{E}\) hold in \(3/3\) judge runs as costs fall. The densest (\(s=2\)) still selects a solved top-1 candidate but weakens elite-pair retention (\(1/3\)), while boundary-only (\(s=\infty\)) retains \(E^\star\subseteq\hat{E}\) in \(3/3\) runs but recovers top-1 solved in only \(2/3\). Qualitative digests support this, as intermediate strides appear to improve the VLM judge's temporal context (Stage~II) to identify key trajectories. Overall, moderate subsampling preserves selection at lower costs, while more frames do not always improve judging.

\vspace{-0.25cm}
\paragraph*{Results VI.C: Rollout Episode-Count Sensitivity}
Using a fixed ten-episode master pool, we test \(N_{\mathrm{ep}}\in\{1,3,5,10\}\) episodes per candidate at $512{\times}512$px and stride $s=8$. From \autoref{fig:results_overview}.E, \(N_{\mathrm{ep}}=1\) is brittle, top-1 solved recovery occurs in only \(16/30\) subset--judge trials and \(E^\star\subseteq\hat{E}\) in only \(10/30\), with a high FN rate. \(N_{\mathrm{ep}}=3\) yields higher reliability, top-1 solved (\(12/12\)) and elite pair \(9/12\). With \(N_{\mathrm{ep}}\geq5\), \(E^\star\subseteq\hat{E}\) holds in all trials, making \(N_{\mathrm{ep}}=5\) the practical cost--stability compromise and \(N_{\mathrm{ep}}=10\) the highest-confidence setting. Evaluation costs scale linearly with \(N_{\mathrm{ep}}\).

\vspace{-0.2cm}
\section{Discussion}
\vspace{-0.2cm}
\paragraph{Main takeaways.}
Against sparse-reward RL, intrinsic-motivation, and reward-synthesis baselines, WATERFALL gains are largest in reveal-gated POMDPs, where task structure must be grounded through interaction \citep{kaelbling1998planning}, while fully observable variants reduce this discovery burden and several baselines recover. Extensive ablations support contributions from \emph{iterative refinement}, \emph{pairwise visual comparison}, and \emph{longitudinal mutation}, while Swiss tournaments balance selection fidelity and evaluation cost. Together, these findings support reward discovery from non-privileged observable evidence, with pretrained priors guiding the interpretation of useful behaviour.

\vspace{-0.25cm}
\paragraph{Limitations and future work.}
Several future work improvements remain. \textbf{First}, while WATERFALL is stress-tested on discrete FO/POMDP settings and a continuous POMDP, broader evaluations on real robots would be beneficial. \textbf{Second}, WATERFALL introduces several design hyperparameters, including candidate count, checkpoint spacing, tournament rounds, rollout count, recording stride, and image resolution. These choices can create a plausible cost--stability trade-off, consistent with the fragility of RL evaluation under noise \citep{agarwal2021precipice}. Our targeted ablations and sensitivity tests narrow the plausible operating regions and quantify these trade-offs extensively. \textbf{Third}, VLM judging is costly, yet aborting or early-stopping candidates reduces consumed training budgets under the evaluated protocol (details in Appendix~\ref{app:compute_accounting}) and costs are also far lower than PRMs, whose online process supervision would be prohibitively expensive, so we ran a diagnostic implementation instead. \textbf{Fourth}, vision judgement can become fallible, though we measure strong pairwise agreement with task success ($93.9\%$), while noting that reported confidence should not be read as a calibrated probability (ECE $0.157$). Likewise, auditing every primary \texttt{STOP} call finds $94/105$ seed runs above threshold, with ten out of eleven candidates passing; held-out seeds and calibrated termination remain open questions. \textbf{Finally}, performance can depend on foundation-model capability, reasoning settings, and API drift. Cheaper or open-weight evaluators, adaptive evidence allocation, selective process supervision, and sparse human feedback under the same information contract merit further study.

\vspace{-0.25cm}
\section{Conclusion}
\vspace{-0.25cm}
We introduced \textbf{WATERFALL}, a workflow for automated reward discovery under \emph{strict observability parity}. Rather than assuming goals, task descriptors, demonstrations, or scalar success signals, WATERFALL generates reward candidates from non-privileged interfaces, selects them through vision-only Swiss tournaments, and refines them via longitudinal assessment and elite-guided mutation. Across the evaluated settings, WATERFALL discovers effective reward programs and outperforms baselines in reveal-gated POMDPs despite their native information advantages, while several baselines recover in FO variants. These findings support learning from newly revealed interactions without requiring their reward logic to be specified in advance by a privileged source.

\vspace{-0.25cm}
\begin{ack}
\vspace{-0.3cm}
Supported by the Computational Design Lab at ETH Zurich. We declare no competing interests.
\end{ack}

\bibliographystyle{unsrtnat}
\bibliography{bibliography}

\newpage
\appendix

\begin{center}
    {\Large\bfseries WATERFALL: Technical Appendix\par}
    \vspace{0.15cm}
    {\small
    \textbf{W}orkflow for \textbf{A}daptive \textbf{T}raining with
    \textbf{E}volutionary \textbf{R}eward \textbf{F}ormulation and
    \textbf{A}utomated \textbf{L}earning \textbf{L}oops\par}
    {\footnotesize\itshape Advances in Neural Information Processing Systems (NeurIPS) 2026\par}
    {\small
    \textbf{Eleftherios Triantafyllidis},
    \textbf{Filippos Christianos},
    \textbf{Zhibin Li},
    and \textbf{Bernd Bickel}\par}
    \rule{\linewidth}{0.4pt}
\end{center}

\vspace{-0.5cm}

\section*{Technical Appendix}
\label{app:statements}
\vspace{-0.2cm}
This Technical Appendix complements the main paper by specifying the exact interfaces, protocols, and implementation details required to reproduce \textbf{WATERFALL} and to audit its strict information constraints. The main text focuses on the conceptual pipeline and empirical findings; here we provide the operational definitions and low-level procedures that were omitted due to space. In particular, we document: (i) the stage-conditioned input--output contracts and attached evidence artefacts for the Actor/Judge modules; (ii) the precise observability-parity mechanism used to support text-only baselines without privileged side-information; (iii) environment specifications for all FO/POMDP variants, including observation schemas, masking, and reveal-gating rules; (iv) the inference sampling policy and boundary-capture requirements used to eliminate privileged temporal priors; (v) the Swiss-system tournament details (pairings, rematch avoidance, tie-breaks) and how tournament outputs are persisted into longitudinal per-candidate histories; and (vi) training and evaluation hyperparameters, compute budgets, checkpoint schedules, and failure-handling rules (gate-and-repair).

\subsection*{Author Contributions}
\textbf{Conceptualisation:} Eleftherios Triantafyllidis and Filippos Christianos. \\
\textbf{Methods:} Eleftherios Triantafyllidis and Filippos Christianos. \\
\textbf{Implementation and Evaluations:} Eleftherios Triantafyllidis. \\
\textbf{Infrastructure Support and Computing Resources:} Eleftherios Triantafyllidis and Bernd Bickel. \\
\textbf{Funding:}  Eleftherios Triantafyllidis, Filippos Christianos and Bernd Bickel. \\
\textbf{Paper Writing:} All authors. \\
\textbf{Additional Funding Notes:} A small fraction of API costs, following additional ablation studies and sensitivity evaluations, was borne by the first and second authors' personal funds. \\

\vspace{-0.2cm}
\subsection*{Broader Impacts} 
WATERFALL may reduce manual reward engineering for RL research, especially in POMDPs where task semantics must be discovered through interaction. Positive impacts include improved stress tests for autonomous reward discovery and reduced dependence on hand-crafted task-specific reward functions. Potential risks include over-reliance on noisy foundation-model judgements, reward misalignment in more open-ended domains, and increased compute/API costs. We mitigate these risks by evaluating only in synthetic environments, preserving audit logs for evolutionary decisions, and treating VLM judgements as noisy proxy evidence rather than authoritative ground truth.

Nevertheless, RL policies trained with foundation-model-generated rewards call for caution when deployed in the real world. Strong performance in simulations does not guarantee safe behaviour in real life. Therefore, such policies should be thoroughly validated before deployment. Used responsibly, these models could improve several aspects of society by automating complex, repetitive, strenuous and dangerous tasks, off-loading them from humans and thereby minimising risk. Ultimately, how these models are used remains the responsibility of the engineers and researchers who use them.

\subsection*{Existing Assets and Licences}
\label{app:existing_assets}
WATERFALL and the evaluated baselines use standard third-party software and APIs, including Unity 2022.3 LTS for custom environment implementation (Unity Terms of Service), Unity ML-Agents 0.30.0 for the RL interface and PPO training (Apache 2.0), PyTorch for neural-network training (BSD-style licences), NumPy/SciPy for numerical computation (BSD-style licences), Matplotlib for plotting (Matplotlib License), and OpenAI API models for VLM/LLM-based reward synthesis under the OpenAI API terms. MiniGrid-inspired tasks informed environment design through cited prior work; no corresponding source code, third-party datasets, or human-subject data are redistributed.

\etocdepthtag.toc{mtappendix}
\etocsettagdepth{mtmain}{none}
\etocsettagdepth{mtappendix}{subsection}
\renewcommand{\contentsname}{Technical Appendix Contents}
\tableofcontents
\newpage

\section{Expanded Methodological Details}
\label{app:full_methods}
This section expands on the method in the main text. We first formalise the fully and partially observable settings and the observability-parity constraint, then detail WATERFALL's three stages: \textbf{Stage~I} synthesises persona-conditioned reward candidates, \textbf{Stage~II} ranks the behaviour they induce through VLM-judged Swiss tournaments over visual rollouts, and \textbf{Stage~III} uses longitudinal assessment histories to continue, abort, escalate or stop each candidate. We close with the full algorithm and its state machine. The prompts implementing each stage are listed in Appendix~\ref{app:prompts}, and the code inputs in Appendix~\ref{app:code_input}.

\paragraph{Problem setting.}
We consider both a fully observable Markov decision process (MDP) and a partially observable MDP (POMDP). In the fully observable regime, the environment is an MDP
\(
\mathcal{M}_{\mathrm{FO}}=(\mathcal{S},\mathcal{A},P,r,\gamma)
\),
with state \(s_t\in\mathcal{S}\), action \(a_t\in\mathcal{A}\), transition kernel \(P(s_{t+1}\!\mid s_t,a_t)\), reward \(r(s_t,a_t,s_{t+1})\), and discount \(\gamma\in(0,1)\). In the partially observable regime, the environment is a POMDP
\(
\mathcal{M}_{\mathrm{PO}}=(\mathcal{S},\mathcal{A},P,\Omega,O,r,\gamma)
\),
where the agent receives observations \(o_t\in\Omega\) (e.g., RGB frames and/or structured channels) drawn from the observation model \(O(o_t\mid s_t)\) and acts with a policy \(\pi_\theta(a_t\mid \mathcal{I}^{\pi}_t)\) conditioned on its information history \(\mathcal{I}^{\pi}_t=\sigma(o_{0:t},a_{0:t-1})\). The external task signal $r_{\mathrm{ext}}$ is sparse and terminal and is used to evaluate task performance. WATERFALL policies optimise their generated reward programs.

\paragraph{World partial observability and progressive semantic grounding.}
We evaluate reward synthesis under \emph{world partial observability} and \emph{incomplete semantic grounding}, where task-relevant semantics are discovered only through interaction rather than pre-disclosed by the experimenter. Concretely, in reveal-gated PO environments, the generative reward designer is constrained to the same above-boundary information available to the policy, augmented only by a time-varying knowledge set \(K_t\) that becomes available \emph{iff} the corresponding evidence is made observable by environment dynamics (e.g., unlocking a door or first encountering an object). Formally, we restrict the generator to
\(
\mathcal{I}^{\text{gen}}_t=\mathcal{I}^{\pi}_t \vee \sigma(K_t),
\)
with \(K_t\) monotonically non-decreasing and no privileged access to unrevealed latent variables. Providing a complete semantic schema \emph{a priori} for entities that have not yet been observed constitutes oracle side-information, as it injects structure not available to an embodied agent before exploration. A fully observable (FO) regime is obtained by changing the observation function so that relevant variables are observable from \(t=0\) (hence \(K_t\equiv K\)), whereas a PO regime retains masking and reveal-gating (so \(K_t\) is acquired through interaction). Accordingly, one-shot reward synthesis baselines are not expected to solve reveal-gated POMDPs \emph{without violating} this information protocol. FO results provide an informative comparison, while PO results evaluate discovery under masking and reveal-gating. Neither regime establishes a universal performance bound for a reward-synthesis method. Nevertheless, we support that the distinction between FO and PO matters, when claims about ontological uncertainty and the method's independence to relying on external channels providing task-priors (goals, task-information, expert trajectories etc) are made.  

\paragraph{Parity, deployability and goal-free discovery.} Deployability constrains what a learned reward reads at execution, whereas parity constrains what the reward designer receives while constructing it. A designer with complete task knowledge can write a fully deployable reward; however, if the generator alone receives the task specification, hidden ontology or ground-truth performance, successful behaviour no longer distinguishes discovery of reward structure from competent implementation of disclosed information. In this case parity is therefore an attributional control, analogous to blinding, not a claim that deployment requires withholding goals. Goal-free discovery is a separate, stronger regime, motivated by capability and affordance discovery before an operator has authored a task ontology (agentic learning), staged instruction or reward library, as in Env.~IV, where the second button enters neither the image nor the observation vector until the first press opens the cover. We do not claim that reward design should generally be goal-free, nor that observations identify a unique intended reward, since both may require human intervention or following specific protocols that would be violated by fully autonomous agentic AI.

\paragraph{Inference Sampling and Avoidance of Privileged Temporal Priors:} To eliminate dependencies on hand-engineered event triggers and event-tailoring (e.g., $\text{OnUnlock} \rightarrow \text{Capture Rollout(-s)}$) and demonstrate robustness against sampling bias, we decouple vision observation recording from environment dynamics, adhering strictly to an arbitrary sampling policy $\pi_{\text{sample}}(t)$ (e.g., log-spaced or fixed-interval). This validates WATERFALL's and ultimately the VLM Evolutionary Judge's capacity to infer goal attainment $\hat{y}$ from a raw, uncurated experience stream $\tau_{\text{raw}}$, thereby providing rigorous evidence of generalisation without privileged temporal priors. We additionally enforce task-agnostic \emph{boundary captures} at episode start and termination, i.e., $\{o^{v}_{0},\,o^{v}_{T}\}\subseteq \tau_{\text{raw}}$, since WATERFALL’s evaluation of $\hat{y}$ is \emph{anchored} by this minimal evidence to ensure identifiability and avoid degenerate sampling regimes where $\pi_{\text{sample}}(t)$ entirely misses the success/failure transition. Lastly, this sparsity serves as a stress test against the necessity for temporal recurrence mechanisms in the evaluation loop, confirming that the VLM Judge bridges observational gaps via abductive reasoning over persistent state changes (e.g., unlocked doors, inventory updates) rather than requiring dense, contiguous event streams. In simple terms, we deliberately ignore intermediate semantic events and strictly adhere to the interval environment step recordings and episode boundaries hence avoiding task-specific engineering to sample human-informed events which would notably limit generalisability. 

\paragraph{Inference sampling.}
For Env.~I--III, rollout interiors follow the task-agnostic sampling policy $\pi_{\mathrm{sample}}(t)$ independently of environment events, with mandatory episode boundaries $\{o^v_0,o^v_T\}\subseteq\tau_{\mathrm{raw}}$. This tests whether the judge can interpret task progress from temporally sparse visual evidence without event-triggered interior captures.

\paragraph{Assessment Schedule and Training.}
To optimise the evolutionary compute budget, we apply \textit{fail-fast} initial screening checkpoints at $1$M steps for Env.~I--III, and $0.5$M for Env.~IV. These initial checkpoints allow pruning degenerate candidates such as hard execution failures, while later checkpoints evaluate whether negligible return persists. Subsequent assessments occur at $\{10,20,\ldots\}$M for Env.~I--III and $\{1,2,\ldots,10\}$M for Env.~IV, up to $T_{\max}$. The judge receives training progress $p=t/T_{\max}$ to contextualise early behaviour. Screening, ranking, and longitudinal decisions follow the same information restrictions at each checkpoint. 

\paragraph{Policy optimisation (RL Per Candidate).}
For each reward candidate $k$, we train an RL policy $\pi_{\theta_k}(a_t \mid \mathcal{I}^{\pi}_t)$ using its generated reward program and PPO. Concretely, at each update we maximise the clipped surrogate objective
\begin{equation}
\begin{aligned}
\mathcal{L}^{\mathrm{PPO}}(\theta)
&= \mathbb{E}_t\!\left[\min\!\big(r_t(\theta)\hat{A}_t,\ \mathrm{clip}(r_t(\theta),1-\epsilon,1+\epsilon)\hat{A}_t\big)\right]\\
&\quad - \alpha\,\mathbb{E}_t\!\left[\big(V_\psi(\mathcal{I}^{\pi}_t)-\hat{V}_t\big)^2\right]\\
&\quad + \beta\,\mathbb{E}_t\!\left[\mathcal{H}\!\left(\pi_{\theta}(\cdot\mid\mathcal{I}^{\pi}_t)\right)\right],
\end{aligned}
\end{equation}
where $r_t(\theta)=\pi_\theta(a_t\mid\mathcal{I}^{\pi}_t)/\pi_{\theta_{\text{old}}}(a_t\mid\mathcal{I}^{\pi}_t)$ and $\hat{A}_t$ is an advantage estimate (e.g., GAE).
At assessment checkpoints, we freeze $\theta_k$ for evaluation rollouts; thereafter, training proceeds under the updated reward program according to the checkpoint schedule until the budget $T_{\max}$ (or early abortion) is reached. Hyperparameters are reported in Appendix~\ref{app:hyperparameters}. Hereinafter, we detail the stages of WATERFALL. The algorithmic overview of these stages is outlined in condensed form in Algorithm~\ref{alg:waterfall}, while the full end-to-end architecture of WATERFALL is provided in Algorithms~\ref{alg:waterfall-part1} and~\ref{alg:waterfall-part2}.

\begin{algorithm}[t]
\caption{\textbf{WATERFALL} (condensed): staged reward generation, tournament selection, and assessment-driven evolution}
\label{alg:waterfall_appendix}
\begin{algorithmic}[1]
\Require Env.\ context $\mathcal{E}$, candidates $K$, checkpoints $\mathcal{T}=\{t_m\}$, budget $T_{\max}$.
\State \textbf{Stage I (Init):} assign personas $\{p_k\}_{k=1}^K$; generate programs $\{P_{k,0}\}$; init policies $\{\pi_{\theta_k}\}$.
\State Active set $\mathcal{K}\leftarrow\{1,\dots,K\}$.
\For{$t \in \mathcal{T}$ \textbf{with} $t\le T_{\max}$ \textbf{and} $\mathcal{K}\neq\emptyset$}
  \State \textbf{RL:} train each $k\in\mathcal{K}$ under $P_{k}$ until $t$.
  \State \textbf{Inference:} collect rollout evidence $\mathcal{R}_{k,t}$ incl.\ boundary frames $\{o^{v}_{0},o^{v}_{T}\}$.
  \State \textbf{Stage II (Tournament):} Swiss judge $\Rightarrow$ standings $(S_k,B_k,\tilde{S}_k,\bar{c}_k)$; append history $\mathcal{H}_{k,\le t}$.
  \State \textbf{Stage III (Assessment):} actor outputs $d_{k,t}\in\{\texttt{CONTINUE},\texttt{ABORT},\texttt{STOP},\texttt{ESCALATE-A/B}\}$.
  \If{exists $k$ with $d_{k,t}=\texttt{STOP}$}
    \State \textbf{optional:} terminate run (solved).
  \EndIf
  \ForAll{$k\in\mathcal{K}$}
    \If{$d_{k,t}\in\{\texttt{ESCALATE-A},\texttt{ESCALATE-B}\}$}
      \State $P_k \leftarrow \textsc{Mutate}(P_k,\mathcal{H}_{k,\le t},\text{elite priors})$
    \EndIf
  \EndFor
  \State $\mathcal{K}\leftarrow\{k\in\mathcal{K}: d_{k,t}\notin\{\texttt{ABORT},\texttt{STOP}\}\}$ \Comment{prune inactive}
\EndFor
\end{algorithmic}
\end{algorithm}

\subsection{Architecture Level I: Reward Synthesis with Candidate Persona Injections}
To increase behavioural and code-level diversity across the $K$ reward programs (and reduce mode collapse towards near-identical shaping terms), WATERFALL conditions VLM code generation on a small set of \emph{personas}---acting as behavioural priors with high-level inductive biases that steer reward design preferences (e.g., conservative dense shaping vs.\ exploration-seeking sparse incentives) without changing the environment API or privileged information access.
Formally, let $\mathcal{P}=\{p^{(1)},\dots,p^{(P)}\}$ denote a fixed persona set (we use $P{=}10$).
Each candidate $k\in\{1,\dots,K\}$ is assigned a persona $p_k\in\mathcal{P}$, and its initial reward program is sampled from a persona-conditioned program distribution
\begin{equation}
P_{k,0} \sim p_{\text{VLM}}\!\left(\,\cdot \mid \mathcal{E}, p_k \right),
\end{equation}
where $\mathcal{E}$ is the environment code context; no external goal specification is provided to the generator.
Crucially, we keep $p_k$ \emph{fixed for the lifetime of candidate $k$} so that mutations remain consistent with the candidate's inductive bias, while tournament selection and subsequent per-candidate updates still allow the population to improve over checkpoints.
This persona-conditioning is complementary to sampling stochasticity (e.g., temperature): it encourages qualitatively distinct reward hypotheses even when candidates share the same environment context and prompt template, improving portfolio coverage under a limited context window budget. The exact persona definitions are provided in Appendix~\ref{app:personas}.

\subsection{Architecture Level II: Evolutionary Tournaments via Iterative Visual Rollout Assessments}
A practical limitation of VLM-supervised RL pipelines is \emph{context window saturation}: jointly supplying large payloads of rollout imagery \emph{and} multiple candidate programs in a single prompt can exceed or degrade the model's effective use of context, especially when salient evidence is distributed throughout long inputs \citep{liu2024lost}. To preserve both scalability in the number of candidates ($K$) and reliability of qualitative judgements, WATERFALL performs \emph{piecewise} (pairwise) comparisons during assessment checkpoints rather than a single monolithic multi-way prompt.

\paragraph{Vision-only pairwise judge.}
At checkpoint $t$, each active candidate $k \in \mathcal{K}_t$ is evaluated in inference for $E$ episodes, producing a set of rollout images $\mathcal{R}_{k,t}=\{\rho^{(e)}_{k,t}\}_{e=1}^{E}$.
We define a \emph{vision-only} pairwise judge
$
\mathcal{J}(\mathcal{R}_{i,t},\mathcal{R}_{j,t}) \rightarrow (w_{i,j,t}, c_{i,j,t}),
$
whose parsed outcome comprises $w_{i,j,t}\in\{i,j,\bot\}$ and confidence $c_{i,j,t}\in[0,1]$, where $\bot$ denotes no valid winner. Judgements are \emph{based solely on rollout evidence}; no reward curves, success rates, or ground-truth signals are exposed. The judge is instructed to provide both (i) per-episode \texttt{SUCCESS/FAIL/UNCERTAIN} labels and (ii) qualitative “progress signals” (e.g., reduced dithering/backtracking, sustained proximity to goal-relevant objects, correct interaction order emerging), using the outcome counts only as supporting evidence.

\paragraph{Swiss-System Tournament.} We adopt a Swiss-system tournament with at most $R$ rounds (configured $R{=}5$). Within each round, candidates are ordered by hard points with seeded random tie-breaking, then greedily paired with the highest-ranked remaining opponent they have not previously faced.
Let $P_r \subseteq \mathcal{K}_t \times \mathcal{K}_t$ be the set of round-$r$ pairings (disjoint pairs). For each played match $(i,j)\in P_r$, the judge returns $(w_{i,j,t}^{(r)},c_{i,j,t}^{(r)})$. Unmatched candidates receive byes only in rounds containing at least one comparison; otherwise, the tournament ends. We assign \textbf{Swiss hard points} as:
\begin{equation}
s_k^{(r)} \;=\;
\begin{cases}
1, & \text{if $k$ wins its round-$r$ match or receives a bye}\\
\frac{1}{2}, & \text{if its match has no valid winner}\\
0, & \text{otherwise}
\end{cases}
\qquad\Rightarrow\qquad
S_k \;=\; \sum_{r=1}^{R} s_k^{(r)}.
\end{equation}
The displayed leaderboard is sorted by descending hard points $S_k$ and then descending \textbf{Buchholz} $B_k$; remaining ties preserve the initial seeded candidate order. Let $\mathcal{O}_k$ be the set of opponents faced by candidate $k$ across all rounds (excluding byes). The Buchholz score is
\begin{equation}
B_k \;=\; \sum_{j\in\mathcal{O}_k} S_j,
\end{equation}
i.e., the sum of the final hard scores of $k$'s opponents (higher is better). This discourages “easy paths” and makes ties less sensitive to bracket idiosyncrasies in noisy proxy judgements. Thereafter, aside from the aforementioned hard points, we also utilise an auxiliary signal as an \textbf{uncertainty-aware soft score.} Because $c_{i,j,t}^{(r)}$ is the judge’s self-reported confidence and may be imperfectly calibrated, we do \emph{not} threshold it (e.g., we avoid rules like “ignore wins below $0.6$”). Instead, we use it as a continuous, uncertainty-aware auxiliary metric:
for a match between $i$ and $j$ with confidence $c\in[0,1]$ assigned to the reported winner,
\begin{equation}
\tilde{S}_i \mathrel{+}= 
\begin{cases}
c, & \text{if $w=i$}\\
1-c, & \text{if $w=j$}
\end{cases}
\qquad
\tilde{S}_j \mathrel{+}= 
\begin{cases}
1-c, & \text{if $w=i$}\\
c, & \text{if $w=j$}
\end{cases}
\end{equation}
For matches without a valid winner, each candidate receives $\frac{1}{2}$ soft point; byes contribute no soft points or match-confidence observations. Summing over played matches gives $\tilde{S}_k$. We additionally report the per-candidate average confidence $\bar{c}_k$ across matches as a diagnostic of proxy certainty. Soft scores $\tilde{S}_k$ and average confidence $\bar{c}_k$ are retained in the tournament digest as diagnostic context for Stage~III; neither enters deterministic elite selection. The comparison in \hyperref[results_tournaments_ablation]{Results~V} evaluates the selection fidelity and API cost of Swiss tournaments against alternative ranking protocols.

\paragraph{Tie-aware elite pool selection (adaptive, capped).}
At each checkpoint $t$, we convert Swiss standings into a small \emph{elite pool} used as prior context for Stage~III mutation decisions.
Let $S_k$ denote the Swiss hard score and $B_k$ the Buchholz tie-break for candidate $k\in\mathcal{K}_t$, and define $S^*=\max_{k\in\mathcal{K}_t} S_k$.
We first form a tie-aware preliminary set of near-leaders
\begin{equation}
\mathcal{E}'_t \;=\;\Big\{k\in\mathcal{K}_t \;:\; S_k \ge S^{*}-\Delta\Big\},
\qquad \Delta=1,
\end{equation}
which retains candidates within one Swiss win of the leader to mitigate proxy-judge noise and bracket idiosyncrasies.
We then sort $\mathcal{E}'_t$ by the lexicographic key $(S_k,B_k,k)$ in descending order, using higher candidate ID as the final deterministic tie-break, and select the final elite pool as
\begin{equation}
M_{\max}(t)=\min\!\Big(3,\;\big\lceil 0.3\,|\mathcal{K}_t|\big\rceil\Big),
\qquad
\mathcal{E}_t \;=\; \text{Top}_{M_{\max}(t)}\!\left(\mathcal{E}'_t\right).
\end{equation}
The pool contains $\min(M_{\max}(t),|\mathcal{E}'_t|)$ candidates, retaining up to three reward hypotheses within one hard point of the leader. A unique leader therefore does not necessarily produce a singleton pool.

\paragraph{Sequential mutation for context isolation and diversity.}
We implement mutation in a \emph{per-candidate sequential} manner.
Batching multiple distinct codebases into a single prompt increases the risk of cross-contamination (e.g., hallucinated shared variables or blended logic), whereas candidate-isolated prompts reduce bleed-through and allow prompt-level stochasticity (e.g., temperature/seed variations) to maintain diversity across mutations.
Formally, for each candidate $k$, we sample an independent mutation stochasticity source $\xi_{k,t}$ and produce
$
P_{k,t+1} \leftarrow \textsc{Mutate}\!\left(P_{k,t};\, \mathcal{H}_{k,\le t},\, P_{\hat{k}_t,t},\, \xi_{k,t}\right).
$

\paragraph{Temporal evidence and chronic failure.}
Tournament outputs are persisted across checkpoints as a lightweight textual history $\mathcal{H}_{k,\le t}$.
This enables temporal reasoning over learning dynamics: repeated failures across multiple checkpoints (``chronic failure'') provide substantially stronger evidence for abortion than a single snapshot, while early-stage instability at low training progress can be treated as inconclusive/continue rather than premature termination.
Concretely, decisions such as \texttt{ABORT} and aggressive inheritance are calibrated by training progress (percentage of the total budget elapsed) and longitudinal tournament history, not solely by a single checkpoint outcome.

\subsection{Architecture Level III: Iterative Candidate Assessments and Escalations}

\paragraph{Elite-Guided Mutation (EGM)}
Following tournament selection, WATERFALL performs \emph{elite-guided mutation} to update candidate reward programs.
Let $P_{k,t}$ denote the reward program (code) for candidate $k$ at checkpoint $t$.
The assessment actor receives: (i) the candidate code $P_{k,t}$, (ii) the accumulated tournament summaries $\mathcal{H}_{k, \le t}$ (via the digest), and (iii) (optionally) the elite code $P_{\hat{k}_t,t}$ as a \emph{prior}.
If escalation is triggered, the actor proposes an updated program $P_{k,t+1}$ by copying and adapting selected parameters/terms from $P_{\hat{k}_t,t}$ (elitist inheritance) and applying localised edits (mutation) to $P_{k,t}$.
We avoid the term \emph{crossover} unless two parent programs are explicitly recombined; in our implementation, the operator is guided mutation with an elite reference.

\paragraph{Assessment actor and evolutionary decisions.}
At each assessment checkpoint $t$, WATERFALL invokes an \emph{assessment actor} that serves as a per-candidate meta-controller over the evolutionary loop. For each active candidate $k\in\mathcal{K}_t$, the actor is conditioned on: (i) the current reward program $P_{k,t}$, (ii) a persistent longitudinal per-candidate assessment history file in the form of a digest $\mathcal{H}_{k,\le t}$ aggregating all prior checkpoint outcomes, (iii) the current tournament context at $t$ (qualitative judge summaries and quantitative standings, e.g., $(S_k,B_k,\tilde{S}_k,\bar{c}_k)$), (iv) minimal rollout boundary evidence $\{o^{v}_{0},o^{v}_{T}\}$, and (v) training progress $p=t/T_{\max}$. The actor outputs a discrete decision
$
d_{k,t}\in\{\texttt{CONTINUE},\texttt{ABORT},\texttt{STOP},\texttt{ESCALATE-A},\texttt{ESCALATE-B}\},
$
where \texttt{CONTINUE/ABORT/STOP} return no code, and \texttt{ESCALATE-A/B} returns an updated reward program.

\paragraph{Longitudinal evolutionary momentum and assessment history.}
The policy may use within-episode memory, while the Stage~III assessment actor conditions on cross-checkpoint histories to avoid brittle snapshot decisions under noisy proxy judging.
For each candidate $k$, we maintain an append-only lineage log $\mathcal{H}_{k,\le t}$ that records prior checkpoint outcomes and decisions, and we re-attach it at every checkpoint so the actor can reason over trajectories (e.g., distinguish early volatility from chronic failure).
In addition, we attach the current tournament digests $\mathcal{D}_{t}=\{\mathcal{D}^{\mathrm{quant}}_{t},\mathcal{D}^{\mathrm{qual}}_{t}\}$ together with the immediately prior digests $\mathcal{D}_{t-1}$ (explicitly labelled as \texttt{HISTORY FILE — PRIOR CHECKPOINT ONLY (t-1)}), providing a first-order temporal difference that acts as an evolutionary ``momentum'' signal.
Concretely, the actor conditions on
\[
\big(P_{k,t},\,\mathcal{H}_{k,\le t},\,\mathcal{D}_{t},\,\mathcal{D}_{t-1},\,p=t/T_{\max}\big),
\]
and is instructed to use $\mathcal{D}_{t-1}$ for regression/improvement detection only (never for ranking or elite selection), enabling decisions to reflect \emph{direction of change} induced by the last mutation (e.g., improved success evidence from $t\!-\!1$ to $t$ suggests the update helped).

\paragraph{Decision criteria.}
The assessment actor interprets judge-reported \texttt{SUCCESS/FAIL/UNCERTAIN} labels, qualitative progress signals, current tournament standings (quantitative rankings), prior assessment history, and training progress. It returns \texttt{STOP} for judged task completion, \texttt{ABORT} for persistent failure, \texttt{CONTINUE} for plausible or inconclusive progress, and \texttt{ESCALATE-A/B} for suspected reward misspecification or insufficient exploration. These are model-generated decisions under the supplied prompt, rather than deterministic thresholds on task success or a fixed numerical improvement window. The deterministic gate validates the decision format and any required code; the separate ground-truth audit does not trigger the decisions.

\paragraph{Escalations and elite-guided mutation.}
Under \texttt{ESCALATE-A}, the actor debugs and/or retunes the reward logic (e.g., rescaling terms, correcting termination conditions via \texttt{RequestEndEpisode()}, or restructuring shaping components) without modifying environment dynamics or the action interface. Under \texttt{ESCALATE-B}, the actor introduces algorithmic exploration drivers (e.g., intrinsic bonuses or entropy-regularised incentives) to improve state-space coverage. When beneficial, the actor may consult a small set of top-performing candidates (an elite pool) as priors and inherit compatible parameters/terms, yielding an updated program $P_{k,t+1}$ consistent with the information constraints of the current checkpoint. WATERFALL's prompts are listed in Appendix~\ref{app:prompts}.

\subsection{Algorithmic Overview and Technical Details of WATERFALL's State Machine}
\paragraph{Stage-conditioned model inputs (I/O contract).}
WATERFALL assembles prompts from stage-specific evidence folders to preserve interface parity while avoiding privileged signals. In \textbf{Training} (Stage~1), the Actor receives the environment template $\mathcal{E}$ (API/context) and (optionally) task context images, and emits $K$ candidate reward programs $\{P_{k,0}\}_{k=1}^K$; a lightweight Judge reviews the generated code for protocol compliance and viability (no learning curves). The RL template is generically defined as \enquote{RLEnvironment} to limit semantic leakage of underlying environment dynamics from pre-training priors, rather than, for example, \enquote{MiniGrid\_UnlockMultiRoom}. The contents of the RL template are categorised into two key code parts (i) above boundary and (ii) below boundary. The above boundary part is fed to the VLM directly, alongside the images of the environment. This part of the code is limited to the state and action vector of the agent, and do NOT expose at any point the sparse reward term (nor inject it to allow the foundation model to purely infer it in its own) nor the length the episodes which also must be determined by the foundation model (though is terminated after a max step hidden). The below boundary part contains the full ground truth dynamics of the environment which are entirely hidden from the VLM to avoid any semantic leakage or goal-conditioning. In the \textbf{Tournament} stage (Stage~2), the Judge receives rollout imagery $\mathcal{R}_{k,t}$ and produces pairwise outcomes for ranking. In \textbf{Assessment} (Stage~3), the Actor makes discrete decisions $d_{k,t}$ (continue/abort/stop/escalate) from longitudinal summaries $\mathcal{H}_{k,\le t}$, active candidate codes, best performing candidates from evolutionary rankings - including evolutionary digests both quantitative and qualitative (all of which stem from Stage~2) and training progress $p=t/T_{\max}$; only escalations require emitting updated code $P_{k,t+1}$.

\paragraph{Deterministic checks, model review, and repair.}
The deterministic gate $G$ checks the required candidate IDs and basic source-code structure before a separate model-based Judge $J$ reviews code-contract compliance. Gate failures return the pending batch for correction; model rejections trigger repair of the rejected candidates while accepted candidates are retained. Stage~I uses the Training Actor for repair, while Stage~III uses the Assessment Fixer. Revised outputs pass through the gate and model review again. During Stage~III repair, a \texttt{CONTINUE} decision can retain the previous program instead of introducing a mutation. 

\paragraph{Assessment actor decisions (meta-controller).}
At each checkpoint $t$, the assessment actor outputs a per-candidate decision
\(
d_{k,t}\in\{\texttt{CONTINUE},\texttt{ABORT},\texttt{STOP},\texttt{ESCALATE\mbox{-}A},\texttt{ESCALATE\mbox{-}B}\},
\)
parsed from a strict, deterministic decision block. \texttt{ABORT} removes $k$ from the active set $\mathcal{K}_{t+1}$, \texttt{STOP} terminates training of $k$ following judged task completion (optionally ending the run), and \texttt{CONTINUE} leaves both policy training and the reward program unchanged. Escalations trigger program updates (see below): Option~A focuses on reward debugging/retuning, while Option~B introduces exploration-driving terms; non-escalation decisions emit no code.

\paragraph{Active-set persistence and checkpoint archival.}
We maintain a persistent active-set mirror of candidate programs on disk and archive snapshots at each checkpoint $t$ for reproducibility. Let $\mathcal{K}_t$ denote active candidate IDs at checkpoint $t$; after parsing $d_{k,t}$ we update $\mathcal{K}_{t+1}=\{k\in\mathcal{K}_t:\ d_{k,t}\notin\{\texttt{ABORT},\texttt{STOP}\}\}$ and write an archived code snapshot tagged by the checkpoint index (and unit, episodes vs.\ environment steps), enabling deterministic restarts and post-hoc audits.

\paragraph{Checkpoint units.}
Checkpoint indices $t$ are reported in environment steps by default and trained as such (and may be equivalently configured in episodes in our implementation); we always expose the unit explicitly to the Judge/Actor via the prompt to prevent ambiguity.

\begin{algorithm}
\caption{WATERFALL \\ 
\small \textbf{W}orkflow for \textbf{A}daptive \textbf{T}raining with \textbf{E}volutionary \textbf{R}eward \textbf{F}ormulation and \textbf{A}utomated \textbf{L}earning \textbf{L}oops \\
\textbf{Part I: Evolutionary Subroutines and Helper Functions}}
\label{alg:waterfall-part1}
\footnotesize
\begin{algorithmic}[1]
\Require Environment context $\mathcal{E}$ (permitted observation/action interface and visual context);
\Statex \hspace{1.25em} candidate count $K$, persona set $\mathcal{P}$, Swiss rounds $R$, inference episodes $E$;
\Statex \hspace{1.25em} checkpoint schedule $\mathcal{T}=\{1,10,20,\ldots,100\}$M for Env.~I--III or $\mathcal{T}=\{0.5,1,2,\ldots,10\}$M for Env.~IV, with the corresponding budget $T_{\max}$;
\Statex \hspace{1.25em} sampling policy $\pi_{\text{sample}}(t)$ and boundary capture rule $\{o^{v}_0,o^{v}_T\}\subseteq \tau_{\text{raw}}$.
\Ensure Best discovered reward program(s) and associated policies, or failure after exhausting $T_{\max}$.

\Function{GenerateInitialPrograms}{$\mathcal{E}, K, \mathcal{P}$}
  \For{$k=1$ to $K$}
    \State Assign persona $p_k\in\mathcal{P}$ (fixed for candidate lifetime)
    \State Query Actor to sample program $P_{k,0}\sim p_{\text{VLM}}(\cdot \mid \mathcal{E}, p_k)$
  \EndFor
  \State \Return $\{P_{k,0}\}_{k=1}^K$
\EndFunction

\Function{GateAndRepair}{$\{P_{k,t}\}$}
  \State $\mathcal{U}\leftarrow$ IDs of submitted programs awaiting acceptance
  \While{$\mathcal{U}\neq\emptyset$}
    \State Apply deterministic format gate $G$ to the pending batch
    \If{$G=\texttt{pass}$}
      \State Query stage-specific model Judge $J$; remove accepted IDs from $\mathcal{U}$
    \EndIf
    \State Repair remaining IDs in $\mathcal{U}$ via Actor/Fixer; keep accepted programs fixed
    \State For Stage~III \texttt{CONTINUE}, retain prior code and remove that ID from $\mathcal{U}$
  \EndWhile
  \State \Return accepted or retained programs $\{P_{k,t}\}$
\EndFunction

\Function{TrainRLToCheckpoint}{$k, P_{k,t}, t_m$}
  \State Train policy $\pi_{\theta_k}$ under reward program $P_{k,t}$ until environment steps reach $t_m$
  \State \Return trained $\theta_k$
\EndFunction

\Function{InferenceRollouts}{$k,\theta_k,E,\pi_{\text{sample}}$}
  \State Run inference for $E$ episodes to collect raw stream $\tau_{\text{raw}}$ under $\pi_{\text{sample}}(t)$
  \State Enforce boundary captures: include $\{o^{v}_0,o^{v}_T\}$ for each episode
  \State \Return rollout evidence $\mathcal{R}_{k,t}=\{\rho^{(e)}_{k,t}\}_{e=1}^{E}$
\EndFunction

\Function{SwissTournament}{$\{\mathcal{R}_{k,t}\}_{k\in\mathcal{K}_t}, R$}
  \State Initialise Swiss hard scores $S_k\leftarrow 0$ and (optional) soft scores $\tilde{S}_k\leftarrow 0$
  \For{$r=1$ to $R$}
    \State Compute $P_r$ by hard-point order with seeded random ties and greedy pairing without rematches     \If{$P_r=\emptyset$}       \State \textbf{break}     \EndIf     \State Award $+1$ hard point to each unmatched candidate (bye)
    \ForAll{$(i,j)\in P_r$}
      \State Judge returns $(w_{i,j,t}^{(r)}, c_{i,j,t}^{(r)})\leftarrow \mathcal{J}(\mathcal{R}_{i,t},\mathcal{R}_{j,t})$
      \State Update hard points: winner $+1$, loser $+0$; if no valid winner, each receives $+\frac{1}{2}$
      \State Update diagnostic soft points: winner $+c_{i,j,t}^{(r)}$, loser $+(1-c_{i,j,t}^{(r)})$; if no valid winner, each receives $+\frac{1}{2}$
      \State Record qualitative summary snippet for both candidates (progress/failure mode)
    \EndFor
  \EndFor
  \State Compute Buchholz tie-break $B_k=\sum_{j\in\mathcal{O}_k}S_j$ (excluding byes)
    \State \Return standings $\{(S_k,B_k,\tilde{S}_k,\bar{c}_k)\}$ and qualitative summaries $\{\text{sum}_{k,t}\}$
\EndFunction
\algstore{waterfall_break_a}
\end{algorithmic}
\end{algorithm}

\begin{algorithm}[t]
\caption*{\textbf{WATERFALL, Part I (continued):} Evolutionary Subroutines and Helper Functions}
\footnotesize
\begin{algorithmic}[1]
\algrestore{waterfall_break_a}

\Function{UpdateHistory}{$k,t,\text{sum}_{k,t},(S_k,B_k,\tilde{S}_k,\bar{c}_k)$}
  \State Append an entry to the per-candidate history file $\mathcal{H}_{k,\le t}$
  \State \Return updated $\mathcal{H}_{k,\le t}$
\EndFunction

\Function{AssessmentActorDecision}{$k,t,P_{k,t},\mathcal{H}_{k,\le t},\{o^{v}_0,o^{v}_T\},p$}
  \State Provide Actor with: code $P_{k,t}$; longitudinal history $\mathcal{H}_{k,\le t}$;
  \Statex \hspace{1.75em} tournament context at $t$ (qualitative + standings); boundary frames $\{o^{v}_0,o^{v}_T\}$; progress $p=t/T_{\max}$
  \State Actor outputs decision $d_{k,t}\in\{\texttt{CONTINUE},\texttt{ABORT},\texttt{STOP},\texttt{ESCALATE-A},\texttt{ESCALATE-B}\}$
  \State \Return $d_{k,t}$
\EndFunction

\Function{Mutate}{$k,t,d_{k,t},P_{k,t},\mathcal{H}_{k,\le t},\mathcal{E}_{t}$}
  \State Optionally attach elite prior program(s) $\mathcal{E}_{t}=\{P_{j,t}: j\in \text{ElitePool}(t)\}$
  \If{$d_{k,t}=\texttt{ESCALATE-A}$}
    \State Produce $P_{k,t+1}$ via reward debugging/retuning (rescaling, reshaping, termination logic fixes)
  \ElsIf{$d_{k,t}=\texttt{ESCALATE-B}$}
    \State Produce $P_{k,t+1}$ via exploration-driving terms (intrinsic/entropy-like incentives; still within API)
  \EndIf
  \State \Return updated program $P_{k,t+1}$
\EndFunction
\algstore{waterfall_break} 
\end{algorithmic}
\end{algorithm}

\begin{algorithm}[tp]
\caption{WATERFALL \\ 
\small \textbf{W}orkflow for \textbf{A}daptive \textbf{T}raining with \textbf{E}volutionary \textbf{R}eward \textbf{F}ormulation and \textbf{A}utomated \textbf{L}earning \textbf{L}oops \\
\textbf{Part II: Main Execution Procedure}}
\label{alg:waterfall-part2}
\footnotesize
\begin{algorithmic}[1]
\algrestore{waterfall_break} 

\Procedure{WATERFALL}{$\mathcal{E},K,\mathcal{T},T_{\max}$}
  \State \textbf{Stage 1: Initialisation.} $\{P_{k,0}\}\leftarrow$ \Call{GenerateInitialPrograms}{$\mathcal{E},K,\mathcal{P}$}
  \State $\{P_{k,0}\}\leftarrow$ \Call{GateAndRepair}{$\{P_{k,0}\}$}
  \State Initialise policies $\{\pi_{\theta_k}\}$ and active set $\mathcal{K}_{0}\leftarrow\{1,\dots,K\}$
  \For{$m=1$ to $|\mathcal{T}|$ with $t_m\le T_{\max}$ and $\mathcal{K}_{m-1}\neq\emptyset$}
    \State $p\leftarrow t_m/T_{\max}$
    \State \textbf{RL training to checkpoint.}
    \ForAll{$k\in\mathcal{K}_{m-1}$}
      \State $\theta_k\leftarrow$ \Call{TrainRLToCheckpoint}{$k,P_{k,m-1},t_m$}
    \EndFor

    \State \textbf{Inference evidence collection.}
    \ForAll{$k\in\mathcal{K}_{m-1}$}
      \State $\mathcal{R}_{k,t_m}\leftarrow$ \Call{InferenceRollouts}{$k,\theta_k,E,\pi_{\text{sample}}$}
    \EndFor

    \State \textbf{Stage 2: Swiss tournament and history update.}
    \State $\{(S_k,B_k,\tilde{S}_k,\bar{c}_k)\},\{\text{sum}_{k,t_m}\}\leftarrow$ \Call{SwissTournament}{$\{\mathcal{R}_{k,t_m}\}_{k\in\mathcal{K}_{m-1}},R$}
    \ForAll{$k\in\mathcal{K}_{m-1}$}
      \State $\mathcal{H}_{k,\le t_m}\leftarrow$ \Call{UpdateHistory}{$k,t_m,\text{sum}_{k,t_m},(S_k,B_k,\tilde{S}_k,\bar{c}_k)$}
    \EndFor

    \State \textbf{Stage 3: Assessment decisions and evolutionary updates.}
    \State Determine the capped elite pool $\mathcal{E}_{t_m}$ using descending $(S_k,B_k,k)$ among candidates within one hard point of the leader.
    \ForAll{$k\in\mathcal{K}_{m-1}$}
      \State Extract boundary frames $\{o^{v}_0,o^{v}_T\}$ from $\mathcal{R}_{k,t_m}$
      \State $d_{k,t_m}\leftarrow$ \Call{AssessmentActorDecision}{$k,t_m,P_{k,m-1},\mathcal{H}_{k,\le t_m},\{o^{v}_0,o^{v}_T\},p$}
      \If{$d_{k,t_m}=\texttt{STOP}$}
        \State Record judged completion for $k$ (optionally terminate entire run)
      \ElsIf{$d_{k,t_m}=\texttt{ABORT}$}
        \State Remove $k$ from the active set
      \ElsIf{$d_{k,t_m}\in\{\texttt{ESCALATE-A},\texttt{ESCALATE-B}\}$}
        \State $P_{k,m}\leftarrow$ \Call{Mutate}{$k,t_m,d_{k,t_m},P_{k,m-1},\mathcal{H}_{k,\le t_m},\mathcal{E}_{t_m}$}
      \Else
        \State $P_{k,m}\leftarrow P_{k,m-1}$ \Comment{\texttt{CONTINUE}}
      \EndIf
    \EndFor
    \State Update active set $\mathcal{K}_{m}\leftarrow\{k\in\mathcal{K}_{m-1}: d_{k,t_m}\notin\{\texttt{ABORT},\texttt{STOP}\}\}$
    \State $\mathcal{U}_m\leftarrow\{k\in\mathcal{K}_{m}:d_{k,t_m}\in\{\texttt{ESCALATE-A},\texttt{ESCALATE-B}\}\}$
    \State $\{P_{k,m}\}_{k\in\mathcal{U}_m}\leftarrow$ \Call{GateAndRepair}{$\{P_{k,m}\}_{k\in\mathcal{U}_m}$}
  \EndFor
\EndProcedure
\end{algorithmic}
\end{algorithm}

\section{Expanded Evaluation Protocol}
\label{app:evaluation_protocol}
This section specifies how WATERFALL and the baselines were evaluated. We first describe the four procedurally randomised environments, their FO and PO variants and the POMDP specifications (see Appendix~\ref{app:environment_details}). We then summarise the state of the art in reward synthesis, the information interface each implemented baseline receives (refer to Appendix~\ref{app:baseline_information_interfaces}), and how textual parity and gated entity reveals are enforced for baselines and ablations. Finally, we give the seed and evaluation protocol, the representative inference sampling that selects the rollouts shown to the judge (see Appendix~\ref{app:representative_inference_sampling}), and the reproducibility details, including compute accounting, visual capture, hyperparameters and the checkpoint schedule (consult Appendix~\ref{app:reproducibility}).

\subsection{Expansion on Evaluated Environments}
\label{app:environment_details}

\begin{figure}[t]
    \centering
    \includegraphics[width=\linewidth]{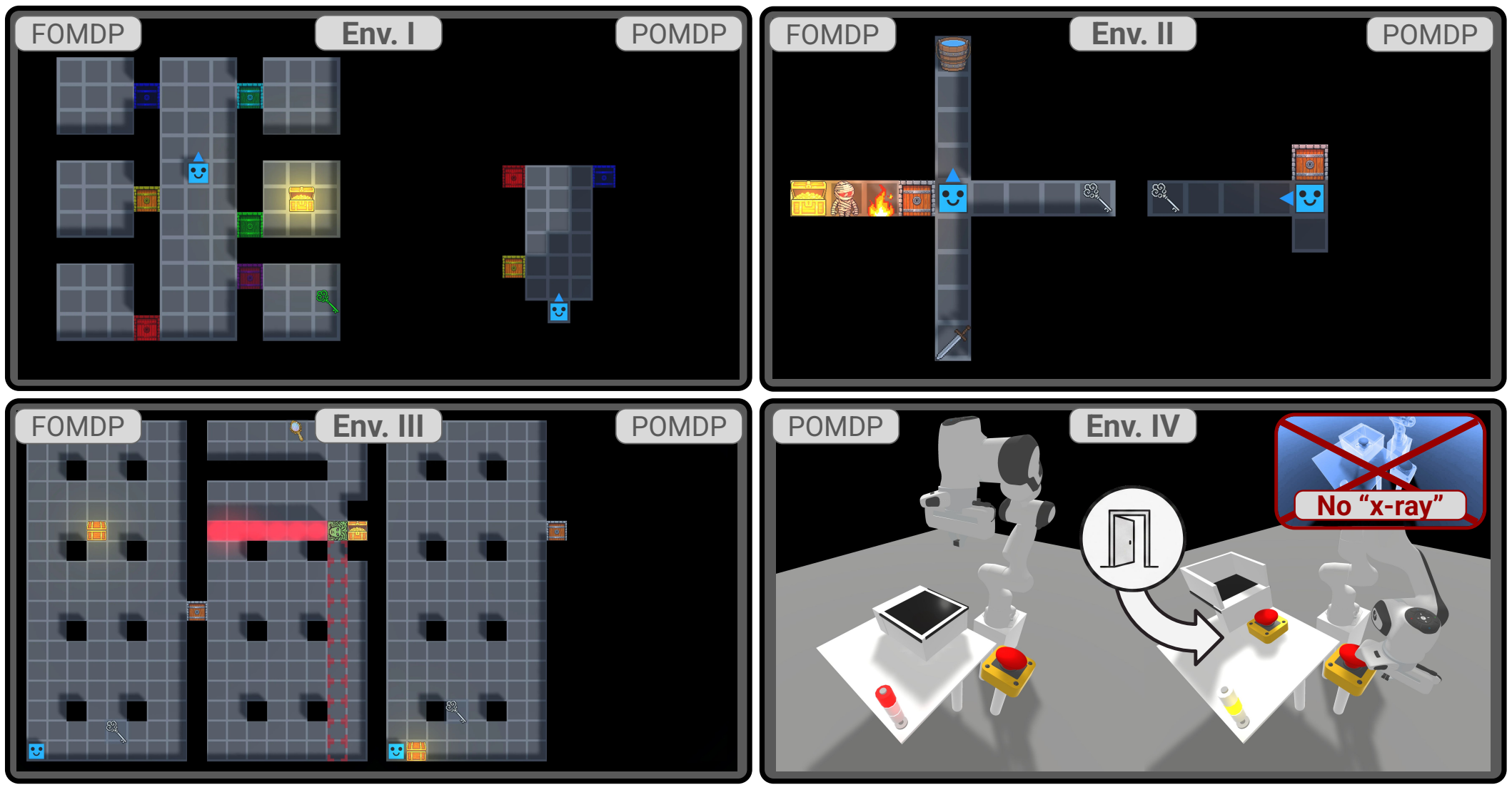}
    \caption{The evaluated procedurally randomised environments entailing discrete and continuous control with their FO and PO variants where applicable.}
    \label{fig:results_overview_appendix}
\end{figure}

\paragraph{Environment I: \enquote{Locked Room Variant} \cite{NEURIPS2023_e8916198}} Heavily inspired by MiniGrid's \texttt{LockedRoomEnv}, this environment retains the canonical six-room, single-locked-room layout on a randomised $13\times13$ grid, yet removes the mission string and associated semantic hint. The agent must instead infer the key$\rightarrow$door$\rightarrow$goal dependency through interaction: the goal is placed inside the locked room, the matching coloured key is spawned anywhere outside it, and the remaining doors are closed but unlocked distractors. \textbf{States, Actions and Rewards:} Let $s_t=(G_t,p_t,\theta_t,i_t,d_t)$ denote grid layout, agent position, orientation, inventory, and door states. The action space is $\mathcal{A}=\{\textsc{TurnLeft},\textsc{TurnRight},\textsc{Forward},\textsc{Pickup},\textsc{Drop},\textsc{Toggle},\textsc{Stay}\}$. PO observations are local and egocentric, $o_t=(R_t,\theta_t,i_t)$, where $R_t\in\mathbb{R}^{7\times 7\times 3}$ is a cone-shaped ray tensor over anonymised $(\mathrm{type},\mathrm{color},\mathrm{state})$ channels. Success yields $r_t=\max(0,\,1-0.9\,t/H)$ with $H=320$, and $0$ otherwise. \textbf{FO Variant:} The 181-dimensional observation comprises agent position and orientation (3), inventory occupancy and item colour (2), open-chest position (2), key position and colour (3), position, colour and open/closed state for each of the six doors ($6\times4$), and the same $7\times7\times3$ ray values as PO (147). These global entity fields are available from episode start, while the seven-action interface is unchanged. \textbf{Challenges:} Sparse reward, partial observability, delayed dependency chaining, and distractor-door disambiguation. \textbf{Semantic Availability:} Minimal.

\paragraph{Environment II: \enquote{Dark Corridor} \cite{NEURIPS2023_e8916198}} Inspired by MiniGrid's \texttt{Memory}, this environment instantiates a randomised $9\times 9$ cross-shaped corridor in which the agent starts at the centre and must recover a latent tool$\rightarrow$blocker dependency through interaction alone. One corridor arm contains the true goal (open chest) at depth $4$, while the three intermediate cells on that same arm are occupied by a random permutation of blockers---a locked door, fire, and an enemy. The remaining three arms each contain exactly one tool at depth $4$---a key, a bucket, and a sword---which must be collected and used via \textsc{Toggle} to clear the corresponding blocker and progressively open access to the goal. The action space is $\mathcal{A}=\{\textsc{TurnLeft},\textsc{TurnRight},\textsc{Forward},\textsc{Pickup},\textsc{Drop},\textsc{Toggle},\textsc{Stay}\}$, and PO observations are local, egocentric, and partial, $o_t=(R_t,\theta_t,i_t)$, where $R_t$ is a T-shaped ray observation comprising one left ray, one right ray, and four forward rays over anonymised $(\mathrm{type},\mathrm{color},\mathrm{state})$ channels, together with normalised orientation $\theta_t$ and inventory state $i_t$; equivalently, $R_t\in\mathbb{R}^{6\times 3}$ and $o_t\in\mathbb{R}^{21}$. Visibility is additionally occluded by walls and closed doors along the forward ray, reinforcing the memory burden. Success is achieved by clearing all three blockers with the appropriate tools and accessing the goal; failures occur on direct contact with fire or the enemy, or by horizon truncation with $H=192$. \textbf{FO Variant:} The 43-dimensional observation comprises agent position and orientation (3), inventory occupancy and item type (2), open-chest position (2), position and active-state flags for the key, bucket, sword, door, fire and enemy ($6\times3$), and the same T-shaped ray values as PO (18). These global entity fields are available from episode start, while the seven-action interface is unchanged. \textbf{Challenges:} Sparse-reward exploration, tool-use sequencing, delayed dependency chaining, and memory under partial observability. \textbf{Semantic Availability:} Minimal.

\paragraph{Environment III: \enquote{Medusa's Revenge}} This environment, developed for this work, uses a $19\times19$ grid comprising a $17\times17$ interior and a one-cell boundary wall. A central wall separates two chambers. The door, key, decoy chest, mirror, and vertical placement of the guardian--goal assembly are randomised, while the agent starts at grid coordinate $(1,17)$ in the left chamber. The guardian occupies the only approach to the true chest, whose other adjacent cells are walls.  \textbf{States and Actions:} Both variants use a 25-dimensional observation: normalised agent position (2), eight immediate-neighbour occupancy/hazard values (8), key possession (1), normalised door, key and decoy coordinates (6), and a reveal-gated buffer (8). The buffer contains mirror possession, normalised guardian direction, and the coordinates of the guardian, mirror and true chest. Coordinates are divided by the grid size; coordinates of a collected mirror or defeated guardian become $-1$. In PO, all eight buffer entries remain $-1$ and the right chamber is visually covered until the door is unlocked. FO exposes the buffer from episode start and omits this visual cover. Actions $0$--$4$ apply grid displacements $(0,1)$, $(0,-1)$, $(-1,0)$, $(1,0)$ and $(0,0)$, respectively. Entering an item cell collects it; attempting to enter the locked door opens it when the key is held. \textbf{Dynamics and Termination:} The guardian rotates through four cardinal directions every eight action applications, emitting an active lethal ray and a non-lethal warning along its next direction. Walls and locked doors block both rays. Waiting and blocked moves still advance the guardian's cycle. The mirror protects against the active ray and enables defeating the guardian by direct contact. Failures arise from unprotected guardian or active-ray contact, entering the decoy, or reaching the configured horizon of 256 action applications; the horizon check precedes processing that action. \textbf{Rewards and Evaluation:} Native sparse feedback is $1$ for success and this reward assignment is excluded from \texttt{WATERFALL}. WATERFALL policies only optimise their own generated rewards. Episode logging records the binary task-completion flag separately from training reward. \textbf{Challenges:} Sparse feedback, dependency chaining, deceptive affordances, temporal reasoning, and discovery of hidden task structure. \textbf{Semantic Availability:} The reward-generation interface excludes the hidden dynamics and success predicates; PO buffer semantics become available through the door-unlock reveal.

\paragraph{Environment IV: \enquote{Reveal-Gated Robotic Button Task} \cite{pmlr-v100-yu20a}}
\label{app:env4_implementation}
Inspired by MetaWorld's button-press task, this continuous-control POMDP uses a simulated Franka arm and requires discovery of a button$\rightarrow$cover$\rightarrow$button dependency. The first button is initially visible, whereas a second button is physically occluded by an opaque cover. Pressing the first button changes a traffic light from red to yellow and rotates the cover by $-180^\circ$ over $3$ simulated seconds. Only once this motion completes does the second button become visually observable and its position become available to the policy; pressing it changes the light to green and completes the task. \textbf{States and Actions:} The 22-dimensional observation comprises end-effector linear velocity (3), angular velocity (3), position (3), quaternion orientation (4), the visible-button position (3), three traffic-light indicators, and a three-coordinate reveal-gated buffer. Positions, velocities, and actions use the \texttt{originReference} coordinate frame. Before reveal, the hidden buffer is fixed to $(-1111,-1111,-1111)$; after the cover finishes opening, it contains the revealed button's position. Button-position observations use Gaussian positional noise with configured parameter $0.01$. The action space is three continuous Cartesian force components. \textbf{Dynamics and Termination:} The visible and revealed buttons register presses when their \texttt{originReference} vertical positions cross $0.395$ and $0.335$, respectively. Episodes terminate on successful activation of the revealed button, end-effector--IK-target drift exceeding $0.5$, persistent IK non-convergence, or $10{,}240$ action applications. The executed \texttt{DecisionRequester} configuration repeats each selected policy action for ten action applications, yielding a maximum of $1{,}024$ policy decisions per episode; ML-Agents' built-in \texttt{MaxStep} termination is disabled. \textbf{Rewards and Evaluation:} Ground-truth task completion is logged independently of training reward. The hidden runtime assigns its sparse success reward only when none of the reward-synthesis compile flags are active; WATERFALL-generated candidates define \texttt{WATERFALL}, thereby compile-time excluding this ground-truth reward during WATERFALL training. \textbf{Challenges:} Continuous contact control, reveal-gated partial observability, sequential dependency discovery, and robust control under IK dynamics. \textbf{Semantic Availability:} Minimal; the second-button position remains masked until legitimate interaction reveals it, while the hidden task dynamics and ground-truth success predicate are excluded from the reward-generation interface.

\subsection{POMDP Environment Specifications}
\label{app:pomdp_environment_specs}
\begin{table}[p]
\centering
\caption{\textbf{POMDP environment specifications for Environments I--IV.}
Implemented observation spaces, action spaces, horizons, reveal gates, and terminal conditions for the four PO environments.}
\vspace{0.5cm}
\label{tab:pomdp_environment_specs}
\begin{footnotesize}
\setlength{\tabcolsep}{4pt}
\renewcommand{\arraystretch}{1.05}
\begin{tabular}{>{\raggedright\arraybackslash}p{2.1cm}*{4}{>{\raggedright\arraybackslash}p{2.6cm}}}
\toprule
\textbf{Specification} & \textbf{Env. I} & \textbf{Env. II} & \textbf{Env. III} & \textbf{Env. IV} \\
\midrule
Environment name
& MiniGrid Locked Room
& Dark Corridor
& Medusa's Revenge
& Reveal-Gated Robotic Button Task \\
Grid size
& \(13{\times}13\)
& \(9{\times}9\)
& \(19{\times}19\)
& Continuous physics \\
Episode horizon
& 320
& 192
& 256
& 10,240 action applications; 1,024 policy decisions \\
\hline

\multicolumn{5}{c}{\textbf{Observation Space}} \\
\hline
Observation vector size
& 150
& 21
& 25
& 22 \\
Local perception
& \(7{\times}7{\times}3\) egocentric rays
& \(6{\times}3\) T-shaped rays
& 8-neighbour ray values
& Proprioception and three traffic-light indicators \\
Agent state
& Orientation
& Orientation
& Agent position
& End-effector position, quaternion, linear and angular velocity \\
Inventory state
& Key held; key colour
& Tool held; tool type
& Key possession
& None \\
Known coordinates
& None
& None
& Door, key, decoy chest
& Visible-button position in the \texttt{originReference} frame \\
Reveal-gated state
& Room/object semantics when observed
& Tool/blocker semantics when observed
& 8 hidden values after door unlock
& Second-button position after the cover fully opens \\
\hline

\multicolumn{5}{c}{\textbf{Action Space}} \\
\hline
Action vector size
& 1 discrete branch
& 1 discrete branch
& 1 discrete branch
& 3 continuous components \\
Number of actions
& 7
& 7
& 5
& Continuous \\
Actions
& Turn left/right, forward, pickup, drop, toggle, stay
& Turn left/right, forward, pickup, drop, toggle, stay
& Move north/south/west/east, stay
& Cartesian force commands to the IK target \\
\hline

\multicolumn{5}{c}{\textbf{Task Structure and Termination}} \\
\hline
Core dependency
& Key \(\rightarrow\) locked door \(\rightarrow\) chest room
& Tools \(\rightarrow\) blockers \(\rightarrow\) goal arm
& Key \(\rightarrow\) door \(\rightarrow\) mirror \(\rightarrow\) Medusa/goal
& Visible button \(\rightarrow\) cover opens \(\rightarrow\) revealed button \\
Success terminal
& Unlock the room containing the goal chest
& Clear all three blockers with the appropriate tools
& Obtain the artefact and defeat the guardian protecting the true goal
& Activate the revealed second button \\
Failure terminal
& Timeout
& Fire/enemy contact or timeout
& Decoy chest, unprotected Medusa/ray contact, or timeout
& End-effector--target drift, persistent IK non-convergence, or timeout \\
Primary challenge
& Sparse reward, memory, locked-room discovery
& Tool-use sequencing, occlusion, memory
& Long-horizon chaining, deception, hidden chamber
& Continuous contact control, physical occlusion, sequential discovery \\
\hline
\end{tabular}
\end{footnotesize}
\end{table}

The POMDP variants intentionally restrict the policy and reward-discovery interface to locally observable evidence. Environments I and II use egocentric ray observations with inventory state, whereas Environment III includes a reveal-gated buffer that stays masked until the locked door is opened, while Environment IV physically occludes the second button, masking its coordinates in both the observation vector and the rendered view until the cover fully opens after the robot presses the first button. Thus, any semantic information about unrevealed regions or entities is unavailable until produced by legitimate interaction with the environment.

\subsection{State of the Art Reward Synthesis}
We summarise, to the best of our knowledge, the state of the art on utilising LLMs and VLMs for reward synthesis in \autoref{tab:baseline_taxonomy}.

\begin{table}[t]
\centering
\caption{\textbf{Reward-Synthesis Baselines. }Overview of architectures for automated reward synthesis with LLMs and VLMs, categorised by information privilege, modality, refinement style, evolutionary selection, and required human involvement.}
\vspace{0.25cm}
\small
\setlength{\tabcolsep}{3.6pt}
\renewcommand{\arraystretch}{1.12}

\newsavebox{\waterfallTaxonomyBody}
\sbox{\waterfallTaxonomyBody}{%
\begin{tabular}{lcccccc}
\toprule
Method &
\multicolumn{2}{c}{Information Privilege} &
Modality &
Refinement &
Evolutionary &
Required Human In-Loop? \\
\cmidrule(lr){2-3}
& Goal-Conditioned & Ground Truth\textsuperscript{11} & & & & \\
\midrule

WATERFALL (ours) &
No &
No &
Vision+Text &
Iterative &
\shortstack{Yes (Swiss-Tourn.)} &
No \\

EUREKA \citep{ma2024eureka} &
Yes\textsuperscript{1} &
Yes &
Text &
Iterative &
\shortstack{Yes (Elite/Greedy)\textsuperscript{10}} &
No\textsuperscript{14}\\

L2R \citep{pmlr-v229-yu23a} &
Yes\textsuperscript{2} &
Yes\textsuperscript{2} &
Text &
Iterative &
No &
Yes\textsuperscript{2} \\

CAR-VLM / VLM-CaR \citep{pmlr-v235-venuto24a} &
Yes\textsuperscript{3} &
Yes\textsuperscript{3} &
Vision+Text &
One-shot\textsuperscript{12} &
No &
No\textsuperscript{3} \\

Text2Reward \citep{xie2024text2reward} &
Yes\textsuperscript{2} &
Yes &
Text &
One-shot/Iter.\textsuperscript{4} &
No &
No\textsuperscript{4} \\

REvolve \citep{hazra2025revolve} &
Yes\textsuperscript{1} &
No\textsuperscript{5} &
Text &
Iterative &
\shortstack{Yes (Elo)\textsuperscript{6}} &
Yes\textsuperscript{5} \\

LaRes \citep{li2025lares} &
Yes &
Yes\textsuperscript{7} &
Text &
Iterative &
\shortstack{Yes (Evo)} &
No \\

RF-Agent \citep{gao2025rfagent} &
Yes &
Yes\textsuperscript{7} &
Text &
Iterative &
No &
No \\

RE-GoT \citep{yao2025regot} &
Yes\textsuperscript{15} &
No\textsuperscript{8} &
Vision+Text &
Iterative &
No &
No \\

R* \citep{li2025rstar} &
Yes &
Yes\textsuperscript{7} &
Text &
Iterative &
\shortstack{Yes (Evo)} &
No \\

FORGE (withdrawn)\textsuperscript{9} \citep{fan2025forge_withdrawn} &
Yes &
Yes &
Text &
Iterative &
\shortstack{Yes (Evo)} &
No \\

GRACE \citep{sapora2026grace} &
Yes (Demos)\textsuperscript{13} &
No\textsuperscript{13} &
Text &
Iterative &
\shortstack{Yes (Evo)} &
Yes \\

\end{tabular}%
}
\resizebox{\ifdim\width>\textwidth\textwidth\else\width\fi}{!}{%
\begin{tabular}{@{}l@{}}
\usebox{\waterfallTaxonomyBody}\\
\midrule
\multicolumn{1}{p{\dimexpr\wd\waterfallTaxonomyBody-2\tabcolsep\relax}}{\footnotesize
\textsuperscript{1} Goal specified via task/env context (text/code), not goal images.\;
\textsuperscript{2} “Interactive feedback/optim” shorthand (language goal with an interactive feedback or optimisation loop).\;
\textsuperscript{3} Uses goal image and/or trajectory-based verification; no interactive human feedback loop is queried during training.\;
\textsuperscript{4} Table entry refers to configurations where iterative refinement can use human feedback (optionally).\;
\textsuperscript{5} Fitness depends on (minimal) human preference feedback.\;
\textsuperscript{6} Elo-style ranking from pairwise comparisons.\;
\textsuperscript{7} Search/controller consumes scalar evaluation/training metrics (e.g., success/return).\;
\textsuperscript{8} Uses VLM rollout evaluation rather than scalar GT metrics.\;
\textsuperscript{9} Withdrawn ICLR 2026 submission (OpenReview).\;
\textsuperscript{10} “Elite/Greedy” denotes elitist selection (keep best-so-far reward/program each iteration).
\textsuperscript{11} “Ground Truth” denotes exposing ground truth rewards either by explicitly providing reward plots of those rewards or success rates and/or metrics stemming from ground truths. This is particularly the case for methods explicitly relying on collecting such ground truth metrics when performing iterative reward synthesis.\;
\textsuperscript{12} One-shot inclusive of a verifier loop.
\textsuperscript{13} Requires privileged supervision in the form of expert trajectories from an expert policy or human demo. Those demos encode the task objective implicitly (goal/intent).
\textsuperscript{14} Not inherently required and is optional for RLHF.
\textsuperscript{15} Explicitly requires environment codes and a text description of the task goal to map out its initial graph-based reasoning paths.
\par\medskip\noindent\textbf{Core Baselines Included:} RL (sparse $r_{\text{ext}}$), RL wCBE \citep{tang2017exploration}, RL wICM \citep{pmlr-v70-pathak17a} and RL wRND \citep{burda2019rnd}, while not part of the aforementioned architectures leveraging reward synthesis, these algorithms were included as part of the core experimental exploration-based baselines leveraging classical intrinsic motivation.
} \\
\bottomrule
\end{tabular}%
}
\label{tab:baseline_taxonomy}
\end{table}

\subsection{Implemented baseline information interfaces}
\label{app:baseline_information_interfaces}
The main comparisons preserve each method's native information interface. \autoref{tab:implemented_interfaces} records the PO configurations used in these experiments; the architectural taxonomy in \autoref{tab:baseline_taxonomy} describes the broader methods. The no-images ablation changes WATERFALL's evidence modality while retaining its restricted interface. Ground-truth task-success statistics used in the STOP and frozen-judge audits are excluded from WATERFALL's model prompts and decisions.

\begin{table}[htbp]
\centering
\small
\setlength{\tabcolsep}{3pt}
\renewcommand{\arraystretch}{1.12}
\caption{\textbf{Inputs used in the PO comparisons across all methods.} Descriptions concern the evaluated implementations. Code/API is restricted for WATERFALL and compact for the listed synthesis baselines. PRMs are a separate diagnostic, not a trained PPO baseline.}
\vspace{0.25cm}
\label{tab:implemented_interfaces}
\begin{tabular}{@{}>{\raggedright\arraybackslash}p{0.16\linewidth}>{\raggedright\arraybackslash}p{0.24\linewidth}>{\raggedright\arraybackslash}p{0.29\linewidth}>{\raggedright\arraybackslash}p{0.24\linewidth}@{}}
\toprule
Method & Supplied task information & Behavioural evidence & Feedback/refinement \\
\midrule
WATERFALL & No prescribed goal; restricted code/API & PO visual rollouts; reveal-gated interface & No GT scalar; tournament/history \\
No-images & Same restricted interface & Text digests with gated reveals & No GT scalar; text tournament/history \\
T2R & Task instruction and pre-disclosed schema & No policy rollouts & Success criterion in prompt; one-shot \\
L2R & Instruction, motion descriptor, pre-disclosed schema & No policy rollouts & Success criterion in prompt; two-call generation \\
CAR-VLM & Start/goal images; pre-disclosed compact interface & Goal disclosed visually; no policy rollouts & No scalar fitness; one-shot \\
EUREKA & Goal description and schema & No policy rollouts & Success/length/reward statistics; reflection \\
Sparse PPO & No task prompt & Policy observations & Sparse task reward \\
PPO + ICM/RND & No task prompt & Policy observations & Sparse plus observation-based intrinsic reward \\
PPO + CBE & No task prompt & Policy observations plus hidden visitation key & Sparse plus count bonus \\
\bottomrule
\end{tabular}
\end{table}

\subsection{Textual Parity and Gated Entity Reveals for Baseline Comparisons, Ablation and POMDPs} 
\label{app:textual_parity}
\paragraph{Deterministic parity artefacts (DBD/GER/VPER).}
Let $o_t$ denote the policy observation at time $t$ under the experimental regime (FO or PO). For WATERFALL's no-images ablation, we define a deterministic serialisation (Dense Behaviour Digest) as a mapping
\[
D:\ \mathcal{O}\rightarrow \Sigma^\star,\qquad \tilde{o}_t = D(o_t),
\]
where $\Sigma$ is a finite alphabet of tokens and $D$ is \emph{round-trip exact} under a fixed parser/formatter (we serialise each scalar with a round-trip safe float format so that parsing recovers the identical value). To prevent semantic oracle information, entities are initially anonymised by observable identifiers $u\in\mathcal{U}$ and may only be assigned persistent semantic tokens once legitimately discovered. We formalise gated entity reveals as an event-log
\[
\mathrm{GER}_t \;=\; \{(u,\phi(u))\mid u\in\mathcal{U},\ g_t(u)=1\},
\]
where $g_t(u)\in\{0,1\}$ is an environment-defined reveal gate that can activate only when the corresponding evidence becomes observable under the regime; once activated, $g_{t'}(u)=1$ for all $t'\ge t$. Finally, to approximate purely visual ``appearance-change'' cues without adding hidden state, we log visual parity event reveals as a deterministic transition signal
\[
e_t \;=\; V(x_{t-1},x_t),
\]
where $x_t$ denotes the rendered frame (or equivalent public visual state) and $V$ emits a small discrete event label (or \texttt{None}) for publicly observable changes (e.g., a door opening). Together, $(\tilde{o}_t,\mathrm{GER}_{\le t},e_{\le t})$ provide a text-based parity interface that is informationally contained in the public observation stream, ensuring downstream reasoning operates without privileged access.

To support text-only LLM baselines and WATERFALL's no-images ablation while preserving strict observability parity, we convert the environment observation stream into a deterministic behaviour digest (DBD) $\tilde{o}_t = D(o_t)$ that serialises the full observation schema into a compact, lossless textual form; when context budgets require, we use a bounded variant $D_{\text{lite}}$ over a recent window. Because many identifiers in $o_t$ are intentionally anonymised (to prevent semantic leakage and encourage ontological discovery), we provide gated entity reveals (GER) as event-driven mapping emissions $\mathrm{GER}_t=\{(u,\phi(u))\mid u\in\mathcal{U}_t,\ g_t(u)=1\}$, where $g_t$ activates only upon legitimate in-environment discovery, after which $\phi(u)$ may be referenced consistently in subsequent steps. Finally, to approximate purely visual “appearance-change” cues without adding hidden state, we log visual parity event reveals (VPER) as short deterministic signals $e_t = V(x_{t-1},x_t)$ (e.g., door opens, glow appears), enabling language-only methods to condition on the same publicly observable events available to vision-based variants. To ground visual evidence $x_t$ (pixels) in the code-level symbols required by programmatic rewards, we introduce a reveal-gated matching $\phi_t$ from anonymised observable ids $u\in\mathcal{U}$ to semantic tokens $s\in\mathcal{S}$, disclosed only when observable, i.e., $s=\phi_t(u)$ is emitted iff $g_t(u)=1$, ensuring that downstream reasoning operates on aligned semantics rather than unconstrained guessing. 

\subsection{Seeds and Evaluation Protocol} 
Primary runs used one fixed configuration per environment; sensitivity analyses varying the foundation model are reported separately.

\paragraph{Randomness, seeds, and evaluation protocol.}
We distinguish (i) a \emph{generation seed} $s_{\mathrm{gen}}$, which controls the stochasticity of \textsc{WATERFALL}'s zero-shot reward synthesis, and (ii) \emph{training seeds} $\{s_{\mathrm{train}}^{(s)}\}_{s=1}^{n}$, which control RL optimisation noise and environment randomisation during policy learning.
We deterministically fix the initial candidate portfolio:
$\mathcal{K}(s_{\mathrm{gen}})=\{k_1,\dots,k_K\}, \qquad K=10$. For each candidate $k\in\mathcal{K}(s_{\mathrm{gen}})$, we train and evaluate policies under $n$ independent training seeds $\{s_{\mathrm{train}}^{(s)}\}_{s=1}^{n}$, with $n=10$ for Env.~I--III and $n=5$ for Env.~IV.
Learning curves aggregate over these $n$ training runs per candidate, with the initial candidate set fixed by $s_{\mathrm{gen}}$. Frozen-evaluator tables instead report repeated judge calls on fixed rollout evidence, as specified for each study.
Baselines are evaluated under the same environment/training-seed protocol $\{s_{\mathrm{train}}^{(s)}\}_{s=1}^{n}$ for comparability.
The same foundation model, part of OpenAI's GPT family, was held fixed across WATERFALL and all baselines and held consistent within each experiment. Due to provider-side changes that caused \texttt{gpt-5.2} to reject our visual input payloads as too large, Env.~I, II and IV used \texttt{gpt-5.4}, while Env.~III used \texttt{gpt-5.2}. These foundation models remained consistent within each experiment and across all reward-synthesis baselines therein. Persistent conversation was disabled, i.e., each call conditions only on the current prompt and explicitly attached artefacts (no cross-call memory).

\subsection{Representative inference sampling protocol}
\label{app:representative_inference_sampling}
Deep RL training is stochastic, and individual seeds can produce atypically strong or weak behaviours. To prevent the VLM evaluator from overfitting to such aleatoric variation, we use a representative inference protocol when generating visual rollouts for Stage~II and Stage~III assessment. For each valid reward candidate, we train $n=10$ independent seeds in Env.~I--III and $n=5$ in Env.~IV. At each assessment checkpoint, we run deterministic inference for every seed and estimate its mean episode return \(\bar{R}_i\) over \(N_{\mathrm{eval}}\) episodes. We then select the representative seed \(S_{\mathrm{med}}=\arg\min_i |\bar{R}_i-\mathrm{median}(\{\bar{R}_1,\dots,\bar{R}_n\})|\). The visual rollouts shown to the VLM are generated from $S_{\mathrm{med}}$, providing a central observed training outcome rather than selecting the best-performing seed.

Unless environment throughput made this infeasible, we used \(N_{\mathrm{eval}}=1000\) episodes per seed. For binary success-rate estimates, this gives a conservative upper bound on the standard error of \(SE=\sqrt{p(1-p)/N_{\mathrm{eval}}}\leq \sqrt{0.5(1-0.5)/1000}\approx1.6\%\). This balances estimator precision with the computational cost of repeated inference. Hyperparameters were generated by WATERFALL for each environment and then held fixed across all baselines within that environment.

\subsection{Reproducibility, Randomness, and Artefact Layout}
\label{app:reproducibility}
Appendix~\ref{app:full_methods} specifies the algorithm and Appendix~\ref{app:prompts} the prompt contracts. Appendix~\ref{app:environment_details} describes the environments, Appendix~\ref{app:hyperparameters} reports training settings, and Appendix~\ref{app:representative_inference_sampling} details inference sampling. The \hyperref[app:compute_accounting]{compute-accounting section} separates recorded and estimated costs, while \autoref{tab:stop_audit} reports the seed-level termination audit. The public code release is described under \hyperref[app:artefact_availability]{artefact availability}.

Table~\ref{tab:reproducibility_summary} summarises the experimental protocol used for the main WATERFALL runs and baselines. Learning curves aggregate over $n=10$ training seeds for Env.~I--III and $n=5$ for Env.~IV, conditional on one realised initial candidate portfolio. For WATERFALL, the candidate count is \(K=10\), matching the \(P=10\) persona-conditioned Stage-I generation setup. Where applicable, reward-synthesis baselines are allocated the same initial candidate budget \(K=10\), while each baseline's native information interface and refinement protocol are preserved.

\begin{table}[t]
\centering
\footnotesize
\renewcommand{\arraystretch}{1.25}
\newcommand{\reproCatWidth}{0.24\linewidth}
\newcommand{\reproSettingWidth}{0.65\linewidth}
\caption{\textbf{Reproducibility summary for WATERFALL and baselines.}}
\vspace{0.25cm}
\begin{tabular}{
>{\raggedright\arraybackslash\bfseries}p{\reproCatWidth}
>{\raggedright\arraybackslash}p{\reproSettingWidth}
}
\specialrule{\heavyrulewidth}{\abovetopsep}{0pt}
\textbf{Category}\rule[-0.9ex]{0pt}{3.2ex} & \textbf{Setting} \\
\specialrule{\lightrulewidth}{0pt}{0pt}
Hardware & AMD Ryzen 9 5950X; 128GB RAM; NVIDIA RTX 3090 24GB \\
Operating System & Ubuntu 24.04 LTS \\
Simulator & Unity3D 2022.3 LTS \\
RL Toolkit & Unity ML-Agents 0.30.0 with PyTorch backend \\
Core RL optimiser & PPO for WATERFALL candidates and RL baselines \\
Additional RL Modules & LSTM, ICM, RND, and CBE where specified \\
WATERFALL Candidates & \(K=10\) reward candidates \\
Persona Count & \(P=10\) Stage-I personas \\
Generation Seeds & One zero-shot generation seed for the main candidate portfolio \\
Training Seeds & \(n=10\) for Env.~I--III; \(n=5\) for Env.~IV \\
Checkpoint Schedule & Env.~I--III: \(\{1,10,20,\ldots,100\}\)M; Env.~IV: \(\{0.5,1,2,\ldots,10\}\)M steps \\
Policy Evaluation & Policy parameters frozen at each checkpoint before inference rollouts \\
Evaluation Episodes & \(N_{\mathrm{eval}}=1000\) episodes per seed where feasible \\
Representative Rollout & Median-seed rollout selected by checkpoint mean return \\
Tournament Topology & Swiss-system tournament, default \(R=5\) rounds \\
Elite Pool & Tie-aware near-leader pool capped at \(\min(3,\lceil0.3|\mathcal{K}_t|\rceil)\) \\
Default VLM/LLM & \texttt{gpt-5.4} (Env.~I, II, IV) and \texttt{gpt-5.2} (Env.~III), high reasoning effort \\ 
Conversation Memory & Disabled; each prompt call is stateless \\
Main Baselines & RL, RL+ICM, RL+RND, RL+CBE, T2R, L2R, CAR-VLM, EUREKA \\
PRM Usage & Diagnostic only, not online PPO supervision \\
Visual Capture & RGBA32 (Lossless WebP) \\
Raw Capture Resolution & \(4096\times4096\) px before deterministic downsampling where used \\
Resolution Sensitivity & \(4096,1024,768,512,256,128\) px frozen-tournament tests; lower resolutions repeated across three fixed-evidence judge calls \\
Temporal Sampling Sensitivity & Strides \(s\in\{2,4,8,16,32,64,\infty\}\), repeated across three fixed-evidence judge calls \\
Episode-Count Sensitivity & \(N_{\mathrm{ep}}\in\{1,3,5,10\}\) over deterministic subsets from a ten-episode master pool, repeated across three fixed-evidence judge calls \\
Error Bands & Learning curves: processed mean \(\pm1\) population standard deviation across seeds; see \hyperref[app:learning_curve_processing]{curve processing} \\
Artefact Logging & Reward snapshots, rollout evidence, tournament digests, histories, repair traces \\
Code availability & Publicly available at \url{https://github.com/etriantafyllidis/waterfall} \\
\bottomrule
\end{tabular}
\label{tab:reproducibility_summary}
\end{table}

\subsection*{Experimental Setup.}
All experiments were run on a single workstation, running the Ubuntu 24.04 LTS version OS, with an AMD Ryzen 9 5950X processor, 128 GB RAM (3600 MHz), and an NVIDIA GeForce RTX 3090 GPU with 24 GB VRAM. The PyTorch ML-Agents 0.30.0 toolkit was used alongside the 2022.3 LTS version Unity3D. The core learning algorithms that were used: PPO, LSTM, ICM, RND and CBE depending on the experiments tested.

\paragraph{Learning-curve processing and statistical reporting.}
\label{app:learning_curve_processing}
Learning curves use the logged binary task-completion indicator, $y_{s,e}\in\{0,1\}$, for episode $e$ of training seed $s$. They do not plot the accumulated generated reward; hence, the normalised returns in \autoref{fig:results_overview} are task-completion rates. Within each seed's retained plotting range, the plotting routine applies
\[
z_{s,e}=\frac{y_{s,e}-m_s}{M_s-m_s},
\qquad m_s=\min_e y_{s,e},\quad M_s=\max_e y_{s,e},
\]
when $M_s>m_s$; constant all-zero and all-one sequences remain zero and one, respectively. This transformation leaves the binary completion indicators unchanged. When enabled, a centred moving average is applied within each seed, with edge windows normalised by their available observations. Seed traces are then linearly interpolated onto a common cumulative-step grid. At each grid point, the mean and population standard deviation are computed over seeds with observations covering that point. Optional exponential smoothing applies separately to both sequences, using $u_j=\alpha u_{j-1}+(1-\alpha)v_j$. Shading shows the resulting mean plus or minus the resulting standard deviation. Curves use $n=10$ training seeds for Env.~I--III and $n=5$ for Env.~IV, conditional on the realised candidate portfolio.

\paragraph{Wall-clock and API-cost accounting.}
\label{app:compute_accounting}
We distinguish aggregate environment steps, active PPO wall-clock time, PPO stream-hours, API latency, and API charges. PPO elapsed time is reconstructed from the union of recorded active training intervals; stream-hours sum the durations of individual candidate--seed streams. API time sums recorded response latencies.

Policies resume from existing checkpoints, including optimiser and normaliser state, rather than restarting at each assessment. Training cost scales with consumed candidate--seed budgets,
\[
C_{\mathrm{RL}}=\sum_{k=1}^{K}\sum_{s=1}^{n}T_{k,s},
\]
while the pairwise comparison budget at checkpoint $t$ is at most
\[
C_{\mathrm{pairs}}(t)\leq R\left\lfloor |\mathcal{K}_t|/2\right\rfloor.
\]
Summary, generation, assessment, and repair requests are counted separately from pairwise matches. Env.~II FO accounting through its $50$M \texttt{STOP} checkpoint includes $162$ API requests, and Env.~IV PO accounting through its $5$M \texttt{STOP} checkpoint includes $171$ requests.

\begin{table}[htbp]
\centering
\caption{\textbf{Env.~III PO accounting through the 10M STOP checkpoint.} PPO consumes $1.00$B aggregate environment steps and $1{,}601$ stream-hours. PPO time measures active elapsed training time; model-stage times sum API latencies. Rollout capture and orchestration are not included.}
\vspace{0.25cm}
\label{tab:waterfall_stage_cost}
\small
\renewcommand{\arraystretch}{1.15}
\begin{tabular}{lrrrr}
\specialrule{\heavyrulewidth}{\abovetopsep}{0pt}
\textbf{Component} & \textbf{Time (h)} & \textbf{API calls} & \textbf{Tokens (M)} & \textbf{API cost (USD)} \\
\specialrule{\lightrulewidth}{0pt}{0pt}
Stage I: generation & 0.59 & 11 & 0.12 & 1.12 \\
PPO training & 18.00 & -- & -- & -- \\
Stage II: tournaments & 0.72 & 51 & 4.06 & 8.48 \\
Stage III: assessment/mutation & 0.64 & 23 & 0.71 & 2.95 \\
\midrule
\rowcolor{white}
Accounted total & 19.95 & 85 & 4.89 & 12.55 \\
\bottomrule
\end{tabular}
\end{table}

\begin{table}[htbp]
\centering
\caption{\textbf{Primary WATERFALL compute accounting.} \textbf{Env-steps:} environment steps summed across all candidate--seed training runs. Invalid candidates were not trained and add no compute. \textbf{PPO wall-clock:} elapsed hours of PPO training with streams running in parallel. \textbf{PPO stream-h:} summed run time of every individual training stream, equivalent to compute-hours. \textbf{VLM wall-clock:} summed foundation-model request latency. \textbf{VLM cost:} summed API charges, including initial reward generation. \textbf{VLM tokens:} total tokens across all foundation-model requests.}
\vspace{0.25cm}
\label{tab:waterfall_compute}
\renewcommand{\arraystretch}{1.15}
\resizebox{\ifdim\width>\linewidth\linewidth\else\width\fi}{!}{%
\begin{tabular}{lrrrrrr}
\specialrule{\heavyrulewidth}{\abovetopsep}{0pt}
\rule[-0.9ex]{0pt}{3.2ex}%
\textbf{Setting} & \textbf{Env-steps (B)} & \textbf{PPO wall-clock (h)} & \textbf{PPO stream-h} & 	\textbf{VLM wall-clock (h)} & \textbf{VLM cost (USD)} & \textbf{VLM tokens (M)}\\
\specialrule{\lightrulewidth}{0pt}{0pt}
I FO & 0.71 & 10.27 & 721 & 2.30 & 20.82 & 5.68\\
I PO & 1.00 & 19.04 & 1,668 & 3.44 & 25.11 & 6.28\\
II FO & 3.40 & 47.72 & 3,193 & 6.15 & 54.12 & 14.56\\
II PO & 2.60 & 64.20 & 4,233 & 6.61 & 52.05 & 13.25\\
III FO & 1.00 & 15.79 & 1,541 & 2.53 & 15.89 & 6.23\\
III PO & 1.00 & 18.00 & 1,601 & 1.95 & 12.55 & 4.89\\
IV PO  & 0.175 & -- & 7,622 & 2.45 & 46.98 & 16.56\\
\specialrule{\heavyrulewidth}{0pt}{\belowbottomsep}
\end{tabular}%
}
\end{table}

\paragraph{Baseline compute coverage.}
Baseline computes including WATERFALL's are seen in \autoref{tab:baseline_compute}.
\begin{table}[htbp]
\centering
\caption{\textbf{Total Training Compute} Complete tally across all environments.}
\vspace{0.25cm}
\label{tab:baseline_compute}
\small
\renewcommand{\arraystretch}{1.15}
\begin{tabular}{lrrr}
\specialrule{\heavyrulewidth}{\abovetopsep}{0pt}
\rule[-0.9ex]{0pt}{3.2ex}%
\textbf{Method} & \textbf{Steps (B)} & \textbf{PPO stream-h} & \textbf{API cost (USD)} \\
\specialrule{\lightrulewidth}{0pt}{0pt}
WATERFALL & 9.89 & 20,579 & 227.51 \\
CAR-VLM & 45.64 & 92,220 & 19.20 \\
EUREKA & 44.69 & 90,589 & 95.60 \\
L2R & 42.82 & 86,997 & 17.38 \\
T2R & 27.84 & 63,866 & 20.39 \\
Sparse PPO & 6.18 & 10,219 & -- \\
PPO + ICM & 6.15 & 10,488 & -- \\
PPO + CBE & 6.19 & 10,122 & -- \\
PPO + RND & 6.11 & 10,268 & -- \\
\specialrule{\heavyrulewidth}{0pt}{\belowbottomsep}
\end{tabular}
\end{table}

Attributed API costs for the vision-evidence sensitivity studies total $\$178.63$: $\$36.81$ for resolution, $\$43.89$ for temporal sampling, and $\$97.94$ for episode count. Requests reused as frozen references are not additional expenditure and must not be counted twice when forming overall totals.

\paragraph{Artefact availability.}
\phantomsection\label{app:artefact_availability}
The appendix documents the algorithm, prompt contracts, observability-parity rules, environment specifications, training settings, inference sampling, compute accounting, and representative evaluation artefacts. WATERFALL's executable code, simulator source files, environment wrappers and prompts are publicly available at \url{https://github.com/etriantafyllidis/waterfall}.

We distinguish three sources of stochasticity: (i) RL optimisation noise (network initialisation, minibatch sampling, and environment stochasticity), indexed by training seed $s\in\{1,\dots,n\}$; (ii) generative synthesis noise, indexed by a per-request randomness source $\xi$ (e.g., model temperature/seed) used in Stage I (initial programs) and Stage III (mutations); and (iii) tournament sampling noise induced by the inference rollout randomisations and the sampling policy $\pi_{\text{sample}}(t)$. WATERFALL treats these sources as independent by construction. Each reward candidate $k$ is trained over $n$ independent RL seeds, and each prompt call is executed in a fresh, stateless context (no persistent conversation). This ensures that diversity arises from $\xi$ and not from compounding hidden context.

\paragraph{Deterministic logging.}
All artefacts are written to disk in a deterministic, file-system-grounded structure to enable post-hoc audits. At minimum, each checkpoint $t$ produces: (i) the active reward program snapshot $P_{k,t}$ for each candidate; (ii) inference evidence $\mathcal{R}_{k,t}$ (images and/or parity text) generated under $\pi_{\text{sample}}(t)$ with enforced boundary frames; (iii) tournament match summaries and the aggregated tournament digest; (iv) the per-candidate longitudinal history file $\mathcal{H}_{k,\le t}$; and (v) any gate-and-repair rejection traces from Stages I/III. This design ensures that every evolutionary decision is traceable to immutable evidence artefacts.

\paragraph{Runtime heterogeneity and artefact volume.}
Although all candidates were trained under the same optimiser and checkpoint schedule, we observed non-trivial \emph{wall-clock heterogeneity} across candidates. This did not stem from differences in PPO hyperparameters, but from \emph{candidate-induced environment throughput}: different synthesised reward programs can alter the empirical episode-length distribution, the frequency of resets, and the number of terminal events encountered before a fixed environment-step budget is reached. Let $N$ denote a fixed training budget in environment steps and let $\bar{L}_k$ be the mean realised episode length for candidate $k$. Then the number of completed episodes before a checkpoint scales as
\[
E_k \;\approx\; \frac{N}{\bar{L}_k}.
\]
Consequently, candidates with systematically shorter realised episodes induce more frequent reset and bookkeeping cycles, which can reduce wall-clock throughput despite identical RL hyperparameters. Under deterministic per-episode/per-checkpoint artefact writing, this also increases the volume of logged artefacts approximately proportionally to the number of completed episodes, i.e.,
\[
|\mathcal{A}_k| \;\propto\; E_k \;\approx\; \frac{N}{\bar{L}_k},
\]
where $|\mathcal{A}_k|$ denotes the artefact footprint for candidate $k$ up to a fixed checkpoint. In practice, this manifested as some candidates requiring substantially longer real time to reach the same environment-step budget, while simultaneously producing markedly larger logging files. We treat this as an expected systems-level consequence of \textsc{WATERFALL}'s intentional candidate diversity: persona-conditioned synthesis can generate reward programs with different termination strictness, shaping density, and behavioural priors, thereby inducing different episode-length regimes even when the downstream trainer configuration is held fixed. This observation further motivates our fail-fast check-pointing and longitudinal assessment design, since pathological or chronically inefficient candidates can be identified and pruned without conflating optimiser settings with candidate-specific runtime effects.

\subsection{Visual Observation Capture and Preprocessing}
\paragraph*{Recording Protocol and Data Compression} The initial visual evidence is acquired directly within the simulation engine via a dedicated parallel recording module. At each designated environment step $t$, the agent's view is rendered into an off-screen \texttt{RenderTexture} and read into an RGBA32 \texttt{Texture2D} buffer. Because the underlying graphics API utilises a bottom-up $y$-axis coordinate system, the raw byte array undergoes a deterministic vertical inversion prior to encoding. Finally, the uncompressed tensor is passed via a native C-interop layer directly to \texttt{libwebp} (specifically invoking \texttt{WebPEncodeLosslessRGBA}). This pipeline serialises the observation $o_{t}^{v}$ into a lightweight WebP byte stream in-memory, ensuring minimal blocking latency on the environment's concurrent stepping throughput.

\paragraph*{Downsampling for Sensitivity Evaluation} To mitigate the computational overhead of high-resolution visual rollouts during VLM inference, we apply a deterministic downsampling pipeline to the raw $4096 \times 4096$ pixel captures. Given the discrete, hard-edged nature of the gridworld environments, standard anti-aliasing filters (e.g., Lanczos or Bicubic) inherently introduce gradient artefacts that exponentially bloat file sizes under compression. Consequently, we strictly utilise nearest-neighbour interpolation, defined as $f(x, y) = I(\lfloor x \cdot s \rceil, \lfloor y \cdot s \rceil)$ where $s$ is the scaling factor, ensuring semantic and colour boundaries remain perfectly sharp. The resulting downscaled tensors are encoded using lossless WebP compression, maximising token-efficiency for the foundation models while preserving strict visual fidelity and minimising HTTP payload footprints.

\subsection{Checkpoint schedule, model saving, and evaluation phases} Training proceeds through checkpoints $\mathcal{T}=\{1,10,20,\ldots\}$M steps for Env.~I--III and $\mathcal{T}=\{0.5,1,2,\ldots,10\}$M steps for Env.~IV, up to the environment-specific budget $T_{\max}$. At each checkpoint $t_m$, we (i) freeze policy parameters $\theta_k$ for evaluation rollouts, (ii) run $N_{\mathrm{eval}}$ inference episodes per seed to estimate performance and select the median seed, whose $E$ recorded episodes form the rollout evidence $\mathcal{R}_{k,t_m}$, (iii) conduct the Swiss tournament using a vision-only judge $\mathcal{J}$, and (iv) update the longitudinal per-candidate history $\mathcal{H}_{k,\le t_m}$ used by the Stage III assessment actor.  

\subsection{Hyperparameters}
\label{app:hyperparameters}
All WATERFALL candidates, reward synthesis and RL-based baselines were trained with the same PPO configuration within each environment and observability setting to ensure consistency. Consequently, within a given environment, methods differ only in the reward source or auxiliary exploration signal, not in the downstream optimiser, training budget, network architecture, seed protocol, or checkpoint schedule. For reward-synthesis baselines, the same downstream PPO settings were used after reward generation. For intrinsic-motivation baselines, the PPO settings in \autoref{tab:ppo_hyperparameters_pomdp} were kept fixed and the auxiliary module in \autoref{tab:intrinsic_hyperparameters} was added. Reward-synthesis methods receive an initial allocation of $K=10$ candidate slots. Env.~I--III use $n=10$ training seeds and $T_{\max}=100$M steps per policy; Env.~IV uses $n=5$ and $T_{\max}=10$M.

In PPO, we optimise the standard clipped actor--critic objective with clipping parameter \(\epsilon\), entropy coefficient \(\beta\), discount factor \(\gamma\), and GAE parameter \(\lambda\). The rollout buffer contains the environment transitions collected before each update phase, while the horizon is the trajectory truncation length used during rollout collection.

\begin{table}[t]
\centering
\caption{\textbf{PPO hyperparameters for the PO variants of Environments I--IV.}
The same PPO settings were used for WATERFALL candidates and all RL-based baselines within each environment. Env.~I--III use a maximum training budget of \(T_{\max}=100\mathrm{M}\) environment steps per policy, while Env.~IV uses \(T_{\max}=10\mathrm{M}\).}
\vspace{0.25cm}
\label{tab:ppo_hyperparameters_pomdp}
\begin{small}
\begin{tabular}{lcccc}
\toprule
\multicolumn{5}{c}{\underline{\textbf{PPO Hyperparameter Settings}}} \\ \addlinespace

\textbf{Hyperparameter} & \textbf{Env. I} & \textbf{Env. II} & \textbf{Env. III} & \textbf{Env. IV} \\
\midrule
Minibatch size & 2048 & 2048 & 2048 & 4096 \\
Rollout buffer size & 20480 & 20480 & 20480 & 40960 \\
Learning rate & \(3.0{\times}10^{-4}\) & \(3.0{\times}10^{-4}\) & \(3.0{\times}10^{-4}\) & \(3.0{\times}10^{-4}\) \\
Learning-rate schedule & Linear & Linear & Linear & Linear \\
Number of epochs & 3 & 3 & 3 & 3 \\
Clipping parameter \(\epsilon\) & 0.2 & 0.2 & 0.2 & 0.2 \\
GAE parameter \(\lambda\) & 0.95 & 0.95 & 0.95 & 0.95 \\
Entropy coefficient \(\beta\) & \(5.0{\times}10^{-3}\) & \(3.0{\times}10^{-3}\) & \(1.0{\times}10^{-2}\) & \(5.0{\times}10^{-3}\) \\
Discount factor \(\gamma\) & 0.999 & 0.999 & 0.995 & 0.995 \\
Extrinsic reward strength & 1.0 & 1.0 & 1.0 & 1.0 \\
Trajectory horizon & 256 & 128 & 256 & 512 \\
Maximum environment steps \(T_{\max}\) & \(100\mathrm{M}\) & \(100\mathrm{M}\) & \(100\mathrm{M}\) & \(10\mathrm{M}\) \\
\midrule

\multicolumn{5}{c}{\textbf{Policy and Value Network Settings}} \\
\midrule
Observation normalisation & Yes & Yes & No & Yes \\
Hidden layers & 2 & 2 & 3 & 3 \\
Hidden units per layer & 256 & 128 & 256 & 512 \\
Recurrent memory & Yes & Yes & No & No \\
Memory sequence length & 64 & 64 & N/A & N/A \\
Memory size & 128 & 128 & N/A & N/A \\
\hline
\end{tabular}
\end{small}
\end{table}

\paragraph{Env.~II FO configuration.}
For Env.~II FO, WATERFALL uses observation normalisation and two hidden layers of 128 units, with recurrent memory disabled. This differs from the recurrent Env.~II PO configuration in \autoref{tab:ppo_hyperparameters_pomdp}.

\begin{table}[t]
\centering
\caption{\textbf{Auxiliary intrinsic-motivation hyperparameters.}
For RL+ICM and RL+RND, the PPO configuration in \autoref{tab:ppo_hyperparameters_pomdp} is unchanged and the auxiliary reward signal below is added.}
\vspace{0.25cm}
\label{tab:intrinsic_hyperparameters}
\begin{small}
\begin{tabular}{lcc}
\toprule
\multicolumn{3}{c}{\underline{\textbf{Intrinsic-Motivation Settings}}} \\

\textbf{Hyperparameter} & \textbf{RL+ICM} & \textbf{RL+RND} \\
\midrule
Auxiliary reward strength & 0.1 & 0.1 \\
Auxiliary discount factor & 0.999 & 0.999 \\
Auxiliary learning rate & \(3.0{\times}10^{-4}\) & \(3.0{\times}10^{-4}\) \\
Auxiliary hidden layers & 2 & 2 \\
Auxiliary hidden units per layer & 128 & 128 \\
Encoding size & 128 & N/A \\
\bottomrule
\end{tabular}
\end{small}
\end{table}

For the count-based exploration baseline, the auxiliary reward is computed from episodic hidden-state visitation counts as
\[
r^{\mathrm{CBE}}_t
=
\eta \frac{1}{\sqrt{N(x_t)}} ,
\]
where \(N(x_t)\) is the visit count of the discrete environment-specific state key and \(\eta=10^{-3}\) is the exploration strength used in the environment implementations. This auxiliary bonus is added on top of the same PPO configuration reported in \autoref{tab:ppo_hyperparameters_pomdp}.

\section{Expanded Experimental Results}
\label{app:expanded_results}
This section expands on the six results in the main text, in the same order for readability. \textbf{Expanded Results~I} details the comparison against sparse-reward RL, intrinsic-motivation and reward-synthesis baselines, together with the PRM progress-scoring diagnostic (Appendix~\ref{app:prm_diagnostic}). \textbf{Expanded Results~II} audits all eleven primary \texttt{STOP} calls against withheld ground-truth task completion. \textbf{Expanded Results~III} covers the vision and refinement ablations. \textbf{Expanded Results~IV--VI} finally report the sensitivity analyses, i.e. the foundation-model evaluator, the tournament topology, and the visual evidence given to the judge (image resolution, recording stride and episode count).

\subsection{Expanded Results I: WATERFALL vs Baselines}
\label{app:expanded_results_i}
As per the main manuscript, following methods were evaluated, ranging from standard sparse reward mechanisms as per the extrinsic mechanisms of RL, to intrinsic motivated learning up to the most relevant ones - SOTA reward synthesis baselines. Due to the popularity and value of PRMs we chose to include those as well in our analysis.
\begin{itemize}
    \item \textbf{WATERFALL}
    \item \textbf{RL}
    \item \textbf{RL with Count-Based Exploration (CBE)}
    \item \textbf{RL with Intrinsic Curiosity Module (ICM)}
    \item \textbf{RL with Random Network Distillation (RND)}
    \item \textbf{Text2Reward (T2R)}
    \item \textbf{Language to Reward (L2R)}
    \item \textbf{Code As Reward (CAR)-VLM}
    \item \textbf{EUREKA}
        \item \textbf{Process Reward Models (PRMs)}
\end{itemize}

\paragraph{PRM progress-scoring diagnostic.}
\label{app:prm_diagnostic}
We score local visual transitions at stride $s=4$ from random-policy and frozen $1$M/$10$M Env.~III PO rollouts. Vision PRM uses current-episode visual evidence; Memory PRM additionally uses prior episode history; Privileged PRM receives a task rubric. Predictions are matched to instrumented sidecar records by episode and frame identity.

For the binary diagnostic, a positive prediction has PRM score $>0$, and a positive reference has \texttt{gt\_score\_heuristic}$>0$. This reference represents instrumented progress events rather than a uniquely specified dense task reward. Classification statistics use completed queries with matched reference records.

\begin{table}[htbp]
\centering
\small
\caption{\textbf{PRM progress-scoring diagnostic in Env.~III PO.} FPR is $\mathrm{FP}/(\mathrm{FP}+\mathrm{TN})$ and precision is $\mathrm{TP}/(\mathrm{TP}+\mathrm{FP})$. These classification metrics differ from the count-based diagnostics used in the frozen VLM tournament studies.}
\vspace{0.25cm}
\begin{tabular}{llrrrr}
\toprule
\textbf{Policy} & \textbf{PRM regime} & \textbf{Scored/attempted} & \textbf{MCC} & \textbf{FPR} & \textbf{Precision} \\
\midrule
Random & Memory & 87/87 & 0.211 & 32.9\% & 6.7\% \\
Random & Privileged & 87/87 & 0.229 & 29.4\% & 7.4\% \\
Random & Vision & 87/87 & 0.174 & 42.4\% & 5.3\% \\
1M & Memory & 63/65 & 0.102 & 75.5\% & 24.5\% \\
1M & Privileged & 65/65 & 0.031 & 70.0\% & 23.9\% \\
1M & Vision & 65/65 & 0.143 & 72.0\% & 26.5\% \\
10M & Memory & 96/99 & 0.321 & 71.0\% & 35.5\% \\
10M & Privileged & 93/99 & 0.266 & 77.9\% & 32.1\% \\
10M & Vision & 98/99 & 0.283 & 76.1\% & 33.3\% \\
\bottomrule
\end{tabular}
\label{tab:prm_diagnostic}
\end{table}

The diagnostic contains $753$ attempted queries, of which $741$ completed, with total logged cost $\$24.61$. Using the measured mean cost per attempted query, $c$, extrapolated online cost is $C(M,s)=cM/s$: approximately $\$81.7$k for $M=10$M and $s=4$. This projection assumes one comparable query every four transitions and is not a measured online PPO experiment. Its query schedule differs from WATERFALL's checkpointed assessment.

\subsection{Expanded Results II: STOP Audit}
\label{app:expanded_results_ii}
\phantomsection\label{app:stop_audit}
We audited eleven primary \texttt{STOP} calls across all seven evaluated settings, using all ten existing training seeds per candidate in Env.~I--III and all five in Env.~IV. Ground-truth task-success metrics were withheld from the decision process and used only for this audit.

\begin{table}[htbp]
\centering
\small
\setlength{\tabcolsep}{4pt}
\renewcommand{\arraystretch}{1.12}
\caption{\textbf{Audit of primary \texttt{STOP} calls across training seeds.} A seed is classified as solved at $\geq 50\%$ task completion over its last $1{,}000$ logged training episodes; a candidate passes when at least $80\%$ of its seeds qualify ($\geq 8/10$ in Env.~I--III; $\geq 4/5$ in Env.~IV). Ranges summarise per-seed completion rates within each group; --- denotes no below-threshold seeds. STOP checkpoints are nominal environment-step budgets per policy.}
\vspace{0.25cm}
\label{tab:stop_audit}
\setlength{\aboverulesep}{0pt}
\setlength{\belowrulesep}{0pt}
\begin{tabular}{lllcccc}
\toprule
\rule[-1.0ex]{0pt}{5.6ex}%
\textbf{Call No.} & \textbf{Environment} & \textbf{Candidate} & \shortstack{\textbf{STOP}\\\textbf{(M steps)}} & \shortstack{\textbf{Seeds}\\\textbf{solved}} & \multicolumn{2}{c}{\textbf{Completion range (\%)}} \\
\cmidrule(lr){6-7}
\rule[-0.9ex]{0pt}{3.2ex}%
 & & & & & Solved seeds & Below threshold \\
\midrule
No. \#1 & Env I FO   & \texttt{k1} & 10 & 10/10 & 77.7--86.8 & --- \\
No. \#2 & Env I FO   & \texttt{k7} & 10 & 10/10 & 78.8--84.6 & --- \\
No. \#3 & Env I PO   & \texttt{k1} & 10 & 10/10 & 94.7--99.6 & --- \\
No. \#4 & Env I PO   & \texttt{k6} & 10 & 10/10 & 66.1--72.9 & --- \\
No. \#5 & Env I PO   & \texttt{k8} & 10 & 9/10  & 66.8--92.4 & 23.0 \\
No. \#6 & Env II FO  & \texttt{k9} & 50 & 6/10  & 66.8--95.0 & 31.3--42.3 \\
No. \#7 & Env II PO  & \texttt{k6} & 30 & 10/10 & 62.7--98.6 & --- \\
No. \#8 & Env III FO & \texttt{k1} & 10 & 8/10  & 85.3--92.1 & 11.1--12.1 \\
No. \#9 & Env III PO & \texttt{k5} & 10 & 9/10  & 91.4--95.9 & 41.6 \\
No. \#10 & Env III PO & \texttt{k8} & 10 & 8/10  & 91.2--95.7 & 6.3--23.5 \\
No. \#11 & Env IV PO  & \texttt{k7} & 5  & 4/5   & 87.6--98.2 & 17.7 \\
\midrule
\addlinespace[2pt]
\rowcolor{white}
\multicolumn{3}{@{}l}{\textbf{All eleven candidates}} & & \textbf{94/105} & \textbf{62.7--99.6} & \textbf{6.3--42.3} \\
\addlinespace[2pt]
\bottomrule
\end{tabular}
\end{table}

Ten of eleven audited candidates pass the stated criterion (\autoref{tab:stop_audit}), with $94/105$ seed runs meeting the seed-level threshold ($89.5\%$). Env.~II FO is the exception: at its $50$M-step \texttt{STOP} checkpoint, mean task completion is $66.48\%$ across ten seeds, but only $6/10$ meet the threshold. Env.~IV PO passes at $5$M steps, with mean completion of $79.22\%$ across five seeds and $4/5$ meeting the threshold. These means use each seed's last $1{,}000$ logged training episodes. In Env.~III FO, eight seeds achieve $85.3$--$92.1\%$ completion, while two achieve $11.1$--$12.1\%$, explaining how successful representative-seed behaviour can coexist with a lower mean curve. All seven audited PO candidates pass. 

\subsection{Expanded Results III: Vision and Refinement Ablations}
\label{app:expanded_results_iii}
\autoref{fig:results_overview}.B compares three Env.~III PO variants. \textbf{Zero-shot} trains the Stage~I candidates of the full run without Stages~II and III, and every candidate stays at zero completion. \textbf{No-images} reruns the full loop, including Stage~I, but its judge reads deterministic text digests (Appendix~\ref{app:textual_parity}) instead of rollout frames. Overall, we observe that zero-shot with downstream training fails the task, no-images does converge but in a less sample-efficient manner, while the default use of WATERFALL's architecture remains beneficial. From a cost perspective, zero-shot remains substantially cheaper but yields complete failure, no-images remains a cost-effective option (API cost \$5.83 vs \$12.55 for the full run, with 73\% fewer tokens) at the price of poorer sample efficiency, and WATERFALL's default vision-based assessment is an effective middle ground.

\subsection{Expanded Results IV: Sensitivity Evaluation of Foundation Model}
\label{app:expanded_results_iv}
\begin{figure}[t]
    \centering
    \includegraphics[width=\textwidth]{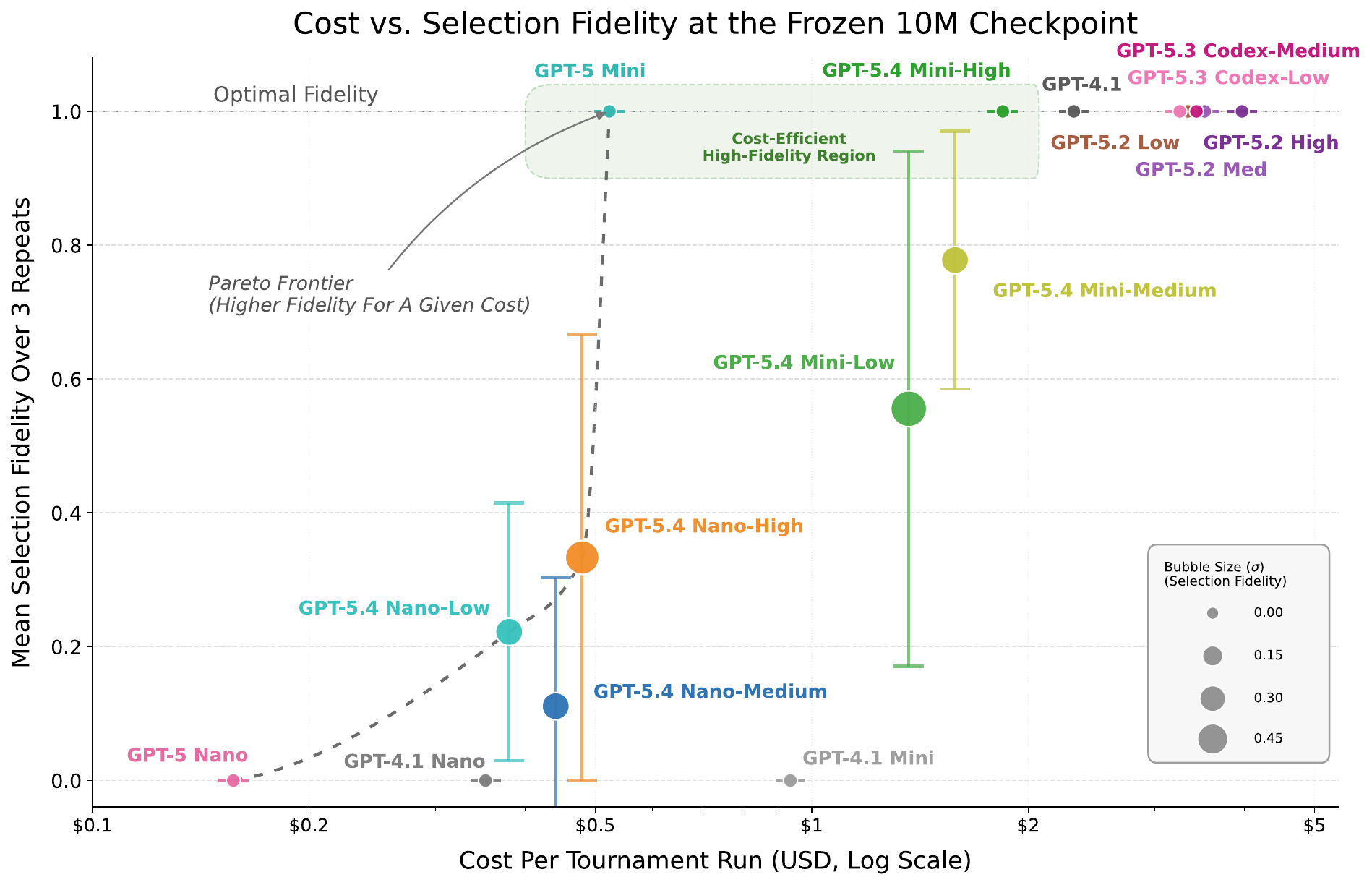}
    \caption{\textbf{Foundation-model sensitivity.}
    Frozen \(10\mathrm{M}\)-step evaluator sweep: points show mean selection fidelity over three Swiss tournaments, bars show \(\pm1\sigma\), and bubble size encodes \(\sigma\).}
    \label{fig:foundation_model_sensitivity}
\end{figure}

This experiment evaluates WATERFALL's sensitivity to the foundation model used in Stage~II tournaments and Stage~III assessment, mutation, and escalation. We consider two settings. \textbf{Experiment I:} \textit{\textbf{end-to-end downstream branching.}} We rerun WATERFALL from candidate generation through full training using (i) \textit{\textbf{gpt-5.2 (high; default)}}, (ii) \textit{\textbf{gpt-5.2 (medium)}}, (iii) \textit{\textbf{gpt-5.4 (high)}}, and (iv) \textit{\textbf{gpt-5.4 (medium)}}. Prompts, stages, candidate interfaces, and the initial environment-context image are held fixed to limit confounding. From \autoref{fig:results_overview}.B, we infer that WATERFALL is not dependent on the strongest available frontier model, supporting accessibility, robustness, and lower-cost deployment, although convergence speed, selection stability, and final return remain model-sensitive. \textbf{Experiment II:} \textit{\textbf{frozen evaluator sensitivity.}} We freeze WATERFALL policy checkpoints at \(t=10\mathrm{M}\) environment steps and vary only the visual tournament judge. For each evaluator \(m\), rollout evidence \(\mathcal{R}_{k,t}\), candidate pool \(\mathcal{K}_t\), tournament topology, prompt template, and image preprocessing are fixed, and selection fidelity is measured against the high-success ground-truth elite pair \(E^\star=\{k5,k8\}\). We report \(\mathbb{I}[\hat{k}_1\in E^\star]\), \(\mathbb{I}[E^\star\subseteq T_3]\), and \(\mathbb{I}[E^\star\subseteq\hat{E}]\), denoting correct top-1 selection, solved-elite recovery within the top-3, and solved-elite inclusion in the selected elite pool. Images are downsampled to \(1024\)px, which preserves visual granularity according to \hyperref[results_vision_evidence_sensitivity]{Results~VI} while reducing token pressure and cost. This isolates evaluator-side variation: changes in \(\hat{k}_1\), \(T_3\), or \(\hat{E}\) arise from the judge model rather than policy training, mutation, or trajectory stochasticity. From \autoref{fig:foundation_model_sensitivity}, we infer that cost-effective evaluators remain viable: \textit{\textbf{gpt-5-mini}} and \textit{\textbf{gpt-5.4-mini (high)}} recover the correct elite pair perfectly, matching costlier \textit{\textbf{gpt-5.2/5.3}} variants and \textit{\textbf{gpt-4.1}}.

\paragraph{Cost-sensitive foundation-model evaluator sensitivity.}
We evaluate whether WATERFALL's frozen Stage-II visual judge requires frontier-scale evaluators. Holding the checkpoint, rollout evidence, image preprocessing, Swiss tournament protocol, prompts, and ground-truth elite set fixed, we vary only the foundation-model evaluator. The ground-truth high-success elite pair is \(E^\star=\{k5,k8\}\). For each evaluator, we run three repeated frozen Swiss tournaments and compute the composite selection fidelity
\[
F=\frac{1}{3}\left(
\mathbb{I}[\hat{k}_1\in E^\star]
+\mathbb{I}[E^\star\subseteq T_3]
+\mathbb{I}[E^\star\subseteq\hat{E}]
\right),
\]
where \(\hat{k}_1\) is the top-ranked candidate, \(T_3\) is the top-three ranking, and \(\hat{E}\) is the selected elite pool. Results show that evaluator cost is not monotonically predictive of selection quality. GPT-5 Mini achieves perfect selection fidelity at low cost (\(\$0.52\)/run), matching GPT-5.4 Mini-H and the more expensive GPT-5.2/5.3 variants. In contrast, GPT-5.4 Nano variants are unstable, and GPT-4.1 Mini/Nano fail despite moderate cost. At this frozen checkpoint, some lower-cost evaluators therefore preserve elite selection across all three repetitions, although this result does not hold for every low-cost model. \autoref{fig:foundation_model_sensitivity} summarises these results and \autoref{tab:foundation_model_sensitivity_appendix} provides the detailed breakdown.

\begin{table}[p]
\centering
\caption{\textbf{Frozen foundation-model sensitivity at the 10M checkpoint in Env.~III.} All rows use identical frozen rollout evidence and Swiss tournament settings, varying only the VLM evaluator. The high-success ground-truth elite pair is $E^\star=\{k5,k8\}$. The composite selection fidelity is $F=\frac{1}{3}\left(\mathbb{I}[\hat{k}_1\in E^\star]+\mathbb{I}[E^\star\subseteq T_3]+\mathbb{I}[E^\star\subseteq\hat{E}]\right)$, reported as mean $\pm$ sample standard deviation over three repeated tournaments. Cost/run is mean $\pm$ sample standard deviation in USD, and Tok./run is the mean total input-plus-output token count; both include all API calls recorded for each tournament. Count FP/Miss/U reports the mean normalised success-count excess, success-count deficit, and uncertainty fraction defined in \eqref{eq:fm_count_diagnostics}. EvMiss counts matches flagged \texttt{EvidenceMissing=True} across the three quantitative digests.}
\vspace{0.25cm}
\footnotesize
\setlength{\tabcolsep}{3pt}
\renewcommand{\arraystretch}{1.25}
\resizebox{\ifdim\width>\textwidth\textwidth\else\width\fi}{!}{%
\begin{tabular}{lccccccccc}
\toprule
\textbf{Evaluator} & \textbf{Cost/run} & \textbf{Tok./run} & $\bar{F}$ & $\hat{k}_1{\in}E^\star$ & $E^\star{\subseteq}T_3$ & $E^\star{\subseteq}\hat{E}$ & $\hat{E}{=}E^\star$ & \textbf{Count FP/Miss/U} & \textbf{EvMiss} \\
\midrule
GPT-5 Nano & $0.157\pm0.003$ & 2.28M & $0.00\pm0.00$ & 0/3 & 0/3 & 0/3 & 0/3 & 0.04/0.22/0.86 & 0 \\
GPT-4.1 Nano & $0.352\pm0.005$ & 3.50M & $0.00\pm0.00$ & 0/3 & 0/3 & 0/3 & 0/3 & 0.19/0.16/0.40 & 14 \\
GPT-5 Mini & $0.523\pm0.021$ & 1.71M & $1.00\pm0.00$ & 3/3 & 3/3 & 3/3 & 3/3 & 0.23/0.02/0.11 & 0 \\
GPT-4.1 Mini & $0.934\pm0.028$ & 2.29M & $0.00\pm0.00$ & 0/3 & 0/3 & 0/3 & 0/3 & 0.41/0.08/0.06 & 2 \\
GPT-4.1 & $2.314\pm0.080$ & 1.10M & $1.00\pm0.00$ & 3/3 & 3/3 & 3/3 & 2/3 & 0.09/0.00/0.00 & 0 \\
GPT-5.4 Nano-L & $0.380\pm0.006$ & 1.76M & $0.22\pm0.19$ & 2/3 & 0/3 & 0/3 & 0/3 & 0.00/0.25/0.04 & 0 \\
GPT-5.4 Nano-M & $0.441\pm0.008$ & 1.81M & $0.11\pm0.19$ & 0/3 & 1/3 & 0/3 & 0/3 & 0.04/0.12/0.04 & 1 \\
GPT-5.4 Nano-H & $0.480\pm0.033$ & 1.77M & $0.33\pm0.33$ & 1/3 & 2/3 & 0/3 & 0/3 & 0.07/0.08/0.04 & 0 \\
GPT-5.4 Mini-L & $1.364\pm0.030$ & 1.71M & $0.56\pm0.38$ & 3/3 & 1/3 & 1/3 & 0/3 & 0.19/0.01/0.21 & 0 \\
GPT-5.4 Mini-M & $1.582\pm0.040$ & 1.74M & $0.78\pm0.19$ & 3/3 & 3/3 & 1/3 & 1/3 & 0.11/0.11/0.45 & 0 \\
GPT-5.4 Mini-H & $1.844\pm0.105$ & 1.77M & $1.00\pm0.00$ & 3/3 & 3/3 & 3/3 & 3/3 & 0.13/0.11/0.41 & 1 \\
GPT-5.2 Low & $3.333\pm0.112$ & 1.74M & $1.00\pm0.00$ & 3/3 & 3/3 & 3/3 & 3/3 & 0.05/0.07/0.12 & 0 \\
GPT-5.2 Med & $3.518\pm0.102$ & 1.72M & $1.00\pm0.00$ & 3/3 & 3/3 & 3/3 & 3/3 & 0.10/0.03/0.08 & 0 \\
GPT-5.2 High & $3.965\pm0.204$ & 1.78M & $1.00\pm0.00$ & 3/3 & 3/3 & 3/3 & 3/3 & 0.10/0.02/0.08 & 0 \\
GPT-5.3 Codex-L & $3.248\pm0.163$ & 1.68M & $1.00\pm0.00$ & 3/3 & 3/3 & 3/3 & 3/3 & 0.21/0.00/0.02 & 0 \\
GPT-5.3 Codex-M & $3.426\pm0.151$ & 1.69M & $1.00\pm0.00$ & 3/3 & 3/3 & 3/3 & 3/3 & 0.15/0.01/0.07 & 0 \\
\bottomrule
\end{tabular}%
}

\vspace{0.4cm}

\setlength{\tabcolsep}{2pt}
\begin{tabular}{llll}
\toprule
\textbf{Evaluator} & \textbf{Top-1 runs} & \textbf{Top-3 runs} & \textbf{Elite-pool runs} \\
\midrule
GPT-5 Nano & \makecell[l]{k10/k4/k7} & \makecell[l]{\{k10,k3,k4\}; \{k4,k7,k5\}; \{k7,k5,k3\}} & \makecell[l]{\{k10,k3,k4\}; \{k4,k7\}; \{k7,k5\}} \\
GPT-4.1 Nano & \makecell[l]{k7/k7/k7} & \makecell[l]{\{k7,k8,k3\}; \{k7,k4,k8\}; \{k7,k10,k3\}} & \makecell[l]{\{k7,k8\}; \{k7,k4,k8\}; \{k7,k10\}} \\
GPT-5 Mini & \makecell[l]{k8/k8/k8} & \makecell[l]{\{k8,k5,k1\}; \{k8,k5,k2\}; \{k8,k5,k3\}} & \makecell[l]{\{k8,k5\}; \{k8,k5\}; \{k8,k5\}} \\
GPT-4.1 Mini & \makecell[l]{k10/k10/k10} & \makecell[l]{\{k10,k1,k3\}; \{k10,k7,k4\}; \{k10,k3,k5\}} & \makecell[l]{\{k10,k1,k3\}; \{k10,k7\}; \{k10,k3\}} \\
GPT-4.1 & \makecell[l]{k8/k5/k5} & \makecell[l]{\{k8,k5,k3\}; \{k5,k8,k3\}; \{k5,k1,k8\}} & \makecell[l]{\{k8,k5\}; \{k5,k8\}; \{k5,k1,k8\}} \\
GPT-5.4 Nano-L & \makecell[l]{k8/k5/k1} & \makecell[l]{\{k8,k1,k3\}; \{k5,k1,k3\}; \{k1,k3,k4\}} & \makecell[l]{\{k8,k1\}; \{k5,k1,k3\}; \{k1,k3,k4\}} \\
GPT-5.4 Nano-M & \makecell[l]{k1/k10/k1} & \makecell[l]{\{k1,k8,k5\}; \{k10,k1,k8\}; \{k1,k10,k3\}} & \makecell[l]{\{k1,k8\}; \{k10,k1,k8\}; \{k1\}} \\
GPT-5.4 Nano-H & \makecell[l]{k1/k3/k8} & \makecell[l]{\{k1,k8,k5\}; \{k3,k10,k6\}; \{k8,k1,k5\}} & \makecell[l]{\{k1,k8\}; \{k3,k10,k6\}; \{k8,k1\}} \\
GPT-5.4 Mini-L & \makecell[l]{k8/k5/k8} & \makecell[l]{\{k8,k1,k4\}; \{k5,k8,k1\}; \{k8,k6,k9\}} & \makecell[l]{\{k8\}; \{k8,k5,k1\}; \{k8\}} \\
GPT-5.4 Mini-M & \makecell[l]{k8/k8/k8} & \makecell[l]{\{k8,k7,k5\}; \{k8,k5,k1\}; \{k8,k6,k5\}} & \makecell[l]{\{k8\}; \{k8,k5\}; \{k8,k6\}} \\
GPT-5.4 Mini-H & \makecell[l]{k8/k8/k8} & \makecell[l]{\{k8,k5,k9\}; \{k8,k5,k2\}; \{k8,k5,k10\}} & \makecell[l]{\{k8,k5\}; \{k8,k5\}; \{k8,k5\}} \\
GPT-5.2 Low & \makecell[l]{k8/k8/k8} & \makecell[l]{\{k8,k5,k3\}; \{k8,k5,k2\}; \{k8,k5,k4\}} & \makecell[l]{\{k8,k5\}; \{k8,k5\}; \{k8,k5\}} \\
GPT-5.2 Med & \makecell[l]{k8/k8/k8} & \makecell[l]{\{k8,k5,k3\}; \{k8,k5,k10\}; \{k8,k5,k3\}} & \makecell[l]{\{k8,k5\}; \{k8,k5\}; \{k8,k5\}} \\
GPT-5.2 High & \makecell[l]{k8/k8/k8} & \makecell[l]{\{k8,k5,k3\}; \{k8,k5,k3\}; \{k8,k5,k10\}} & \makecell[l]{\{k8,k5\}; \{k8,k5\}; \{k8,k5\}} \\
GPT-5.3 Codex-L & \makecell[l]{k8/k8/k8} & \makecell[l]{\{k8,k5,k10\}; \{k8,k5,k1\}; \{k8,k5,k1\}} & \makecell[l]{\{k8,k5\}; \{k8,k5\}; \{k8,k5\}} \\
GPT-5.3 Codex-M & \makecell[l]{k8/k8/k8} & \makecell[l]{\{k8,k5,k1\}; \{k8,k5,k1\}; \{k8,k5,k10\}} & \makecell[l]{\{k8,k5\}; \{k8,k5\}; \{k8,k5\}} \\
\bottomrule
\end{tabular}
\label{tab:foundation_model_sensitivity_appendix}
\end{table}

\paragraph{Success-count diagnostics.}
For candidate $k$ in tournament $r$, let $S_{kr}$, $F_{kr}$, and $U_{kr}$ denote the total \texttt{SUCCESS}, \texttt{FAIL}, and \texttt{UNCERTAIN} counts recorded in the quantitative digest, aggregated across that candidate's repeated appearances. Define $n_{kr}=S_{kr}+F_{kr}+U_{kr}$ and $N_r=\sum_{k=1}^{10}n_{kr}$. Each appearance with valid parsed evidence contributes three episode assessments; repeated judgements of the same frozen episodes count separately. Comparisons flagged \texttt{EvidenceMissing=True} contain zero recorded counts and contribute neither to the diagnostic numerators nor to $N_r$. These flags denote missing parsed counts, which can arise even when the model response contains descriptive evidence.

Let $g_k$ be the ground-truth success fraction over the three frozen episodes: $g_1=g_3=1/3$, $g_5=g_8=1$, and $g_k=0$ otherwise. With $[x]_+=\max(x,0)$, the tournament-level diagnostics are
\begin{equation}
\label{eq:fm_count_diagnostics}
\begin{aligned}
\mathrm{FP}^{\mathrm{cnt}}_r
&=\frac{\sum_{k=1}^{10}[S_{kr}-n_{kr}g_k]_+}{N_r},\\
\mathrm{Miss}^{\mathrm{cnt}}_r
&=\frac{\sum_{k=1}^{10}[n_{kr}g_k-S_{kr}]_+}{N_r},\\
\mathrm{U}_r
&=\frac{\sum_{k=1}^{10}U_{kr}}{N_r}.
\end{aligned}
\end{equation}
Positive parts are taken after aggregating each candidate's counts within a tournament. The table reports the arithmetic mean of each diagnostic over the three tournaments, rather than pooling their observations. \texttt{UNCERTAIN} counts remain in the denominator and contribute to $\mathrm{U}_r$, but do not count as declared successes; they can therefore contribute to a success-count deficit. FP and Miss quantify aggregate count discrepancies, within which opposing episode-level errors may cancel. They are not conventional false-positive/false-negative rates or probability-calibration measures.

\paragraph{Notes on Closed-model Reproducibility.}
A limitation of evaluating foundation-model-driven RL systems is that closed API endpoints provide only partial visibility into the evaluator: model weights, multimodal tokenisation, backend routing, and request-admissibility constraints are not fully exposed to researchers. In our experiments, identical visual payloads could be accepted by some endpoints but rejected by others, despite nominally similar context-window specifications. We therefore report model/version, reasoning effort, image resolution, sampling stride, and input-feasibility failures explicitly, and treat such failures as systems-level evaluator constraints rather than policy-learning failures. This motivates future evaluation on open-weight or locally hosted multimodal models where the full inference stack can be fixed and audited. One example of our sensitivity analysis on foundation models was that despite identical visual payloads for all models, GPT-5.1 failed due to endpoint-specific multimodal input limits and was therefore omitted from the analysis.

\subsection{Expanded Results V: Balancing Tournament Topology}
\label{app:topology_ablation}

\paragraph{Frozen tournament-topology ablation protocol.}
To isolate the effect of tournament topology from downstream training and mutation confounds, we performed a \emph{frozen} topology ablation at two checkpoints: $t=1$M and $t=10$M environment steps. For each checkpoint, we fixed the candidate pool $\mathcal{K}_t$ and reused the same rollout evidence $\{\mathcal{R}_{k,t}\}_{k\in\mathcal{K}_t}$ while varying only the tournament topology used by the vision judge. Concretely, we compared: (i) \textbf{KoH}, which performs a sequential bracket over a randomised candidate order and yields a single champion after $|\mathcal{K}_t|-1$ matches; (ii) \textbf{RR}, which exhaustively compares all unordered pairs and therefore requires $|\mathcal{K}_t|(|\mathcal{K}_t|-1)/2$ matches; and (iii) \textbf{Swiss}, which performs $R$ rounds of score-aware pairings and yields an approximately linear-in-$R$ comparison budget.

\paragraph{Metrics.}
We report four complementary quantities. First, \emph{top-$1$ agreement}, i.e., whether a topology selects the same champion as RR. Second, \emph{ranking fidelity}, measured by the top-$3$ overlap
\[
\mathrm{IoU}_3(t)=\frac{|T^{\mathrm{method}}_3(t)\cap T^{\mathrm{RR}}_3(t)|}{|T^{\mathrm{method}}_3(t)\cup T^{\mathrm{RR}}_3(t)|}.
\]
Third, \emph{selection regret},
\[
\Delta_{\mathrm{sel}}(t)=g_{\hat{k}_{\mathrm{opt}},t}-g_{\hat{k}_{\mathrm{method}},t},
\]
where $g_{k,t}$ denotes the ground-truth performance statistic at checkpoint $t$ (here instantiated as the number of successful evaluation episodes out of $3$; if a binary solved/not-solved criterion is preferred, this can be substituted without changing the topology comparison logic). Fourth, \emph{evaluation cost}, summarised by match count, total requests, token usage, and dollar cost.

\paragraph{Checkpoint $t=1$M.}
At the early fail-fast checkpoint, all candidates are ground-truth failures, so $\Delta_{\mathrm{sel}}(1\mathrm{M})=0$ for all three topologies under the success-count definition above. The informative signal at this checkpoint is therefore \emph{relative ranking structure}, not scalar regret. RR and Swiss both select $k7$ as the top-$1$ candidate, whereas KoH selects $k1$. The RR top-$3$ is $\{k7,k5,k9\}$, the Swiss top-$3$ is $\{k7,k1,k5\}$, and the KoH top-$3$ is $\{k1,k7,k9\}$, yielding
\[
\mathrm{IoU}_3^{\mathrm{Swiss}}(1\mathrm{M})=0.5, 
\qquad
\mathrm{IoU}_3^{\mathrm{KoH}}(1\mathrm{M})=0.5.
\]
Thus, at $1$M the main distinction is that Swiss preserves the exhaustive RR winner while KoH does not, even though all candidates are still objectively poor.

\paragraph{Checkpoint $t=10$M.}
At the later checkpoint, the candidate pool is substantially more separated. Under the provided ground truth, $k5$ and $k8$ are the only fully solved candidates ($3/3$ successes), while $k1$ and $k3$ are partial candidates ($1/3$ successes each). Here, RR ranks $\{k8,k5,k1\}$ as the top-$3$ and selects elite set $\{k8,k5\}$. Swiss matches this ranking exactly and yields the same elite set:
\[
T^{\mathrm{Swiss}}_3(10\mathrm{M}) = T^{\mathrm{RR}}_3(10\mathrm{M}) = \{k8,k5,k1\},
\qquad
\mathrm{IoU}_3^{\mathrm{Swiss}}(10\mathrm{M})=1.0.
\]
By contrast, KoH selects $k5$ as champion and returns only $\{k5\}$ as elite, with top-$3$ $\{k5,k8,k7\}$, yielding
\[
\mathrm{IoU}_3^{\mathrm{KoH}}(10\mathrm{M})=0.5.
\]
Notably, all three topologies still achieve zero top-$1$ selection regret at $10$M because KoH selects $k5$ and RR/Swiss select $k8$, and both are fully solved under the ground truth. This highlights an important point: \emph{for evolutionary pipelines, topology quality is not captured by top-$1$ scalar regret alone}. The more informative distinction is whether the topology preserves the \emph{broader elite structure} that governs inheritance and mutation in subsequent stages.

\paragraph{Cost.}
Across the two frozen checkpoints, KoH required $18$ matches and $20$ total requests, RR required $90$ matches and $92$ total requests, and Swiss required $49$ matches and $51$ total requests. The corresponding total evaluation costs were \$2.97 (KoH), \$15.40 (RR), and \$8.48 (Swiss). Hence, RR is the most informative but also the most expensive, KoH is the cheapest but the most topology-sensitive, and Swiss provides the intended compromise: it recovers the exhaustive RR structure at the later checkpoint while requiring substantially fewer comparisons.  

\paragraph{Consolidated summary.}
\autoref{tab:frozen_topology_ablation_summary} consolidates the frozen-topology ablation: per-checkpoint top-\(1\)/top-\(3\) selections, elite sets, and ranking fidelity against the exhaustive RR reference, together with the aggregate evaluation cost. It makes explicit that Swiss preserves the RR winner at \(1\)M and exactly recovers the RR top-\(3\) and elite set at \(10\)M using roughly half the comparison budget, whereas KoH's cheap sequential bracket diverges at both checkpoints.

\begin{table}[htbp]
\centering
\caption{\textbf{Frozen tournament-topology ablation (Environment III, Medusa's Revenge, POMDP).} Top-\(1\)/top-\(3\) selections (\(T_3\)), elite sets (\(\hat{E}\)), and ranking fidelity against the exhaustive RR reference (\(\mathrm{IoU}_3\); RR entries are marked ``--'' because RR is the reference itself) at the two frozen checkpoints, alongside matches, judge requests, and logged API cost aggregated across both checkpoints. Top-\(1\) selection regret is zero for all topologies at both checkpoints: all candidates are ground-truth failures at \(1\)M, whereas \(k5\) and \(k8\) are both fully solved at \(10\)M. Swiss exactly recovers the exhaustive RR top-\(3\) and elite set at \(10\)M (\(\mathrm{IoU}_3=1.0\)) using \(49/90\approx54\%\) of RR's matches and approximately \(55\%\) of its cost, whereas KoH remains topology-sensitive at both checkpoints.}
\vspace{0.25cm}
\footnotesize
\setlength{\tabcolsep}{3pt}
\renewcommand{\arraystretch}{1.05}
\resizebox{\ifdim\width>\textwidth\textwidth\else\width\fi}{!}{%
\begin{tabular}{lcccccccccc}
\toprule
& \multicolumn{3}{c}{\textbf{Checkpoint \(t=1\)M}}
& \multicolumn{4}{c}{\textbf{Checkpoint \(t=10\)M}}
& \multicolumn{3}{c}{\textbf{Total (both checkpoints)}} \\
\cmidrule(lr){2-4}
\cmidrule(lr){5-8}
\cmidrule(lr){9-11}
\textbf{Topology}
& \textbf{Top-1}
& \(\mathbf{T_3}\)
& \(\mathbf{IoU_3}\)
& \textbf{Top-1}
& \(\mathbf{T_3}\)
& \(\mathbf{\hat{E}}\)
& \(\mathbf{IoU_3}\)
& \textbf{Matches}
& \textbf{Requests}
& \textbf{Cost} \\
\midrule
KoH
& \(k1\)
& \(\{k1,k7,k9\}\)
& \(0.5\)
& \(k5\)
& \(\{k5,k8,k7\}\)
& \(\{k5\}\)
& \(0.5\)
& \(18\)
& \(20\)
& \(\$2.97\) \\
RR
& \(k7\)
& \(\{k7,k5,k9\}\)
& --
& \(k8\)
& \(\{k8,k5,k1\}\)
& \(\{k8,k5\}\)
& --
& \(90\)
& \(92\)
& \(\$15.40\) \\
Swiss
& \(k7\)
& \(\{k7,k1,k5\}\)
& \(0.5\)
& \(k8\)
& \(\{k8,k5,k1\}\)
& \(\{k8,k5\}\)
& \(1.0\)
& \(49\)
& \(51\)
& \(\$8.48\) \\
\bottomrule
\end{tabular}%
}
\label{tab:frozen_topology_ablation_summary}
\end{table}

\paragraph{Interpretation.}
The frozen ablation supports the role of Swiss as a \emph{cost-aware robustness layer} for noisy visual comparison. At $1$M, Swiss already tracks the exhaustive winner more closely than KoH; at $10$M, Swiss exactly matches the exhaustive RR top-$3$ and elite set, whereas KoH collapses to a narrower champion path. This is precisely the regime relevant to \textsc{WATERFALL}: the downstream mutation stage does not require a perfect total ordering, but it does require a topology that preserves a reliable set of strong candidates under a limited comparison budget.

\paragraph{Downstream topology branches and cost.} Under topology-specific downstream selection (\autoref{fig:results_overview}.C), Swiss achieves the highest return, RR remains competitive at greater API cost, and KoH produces weaker policies. Swiss reaches \texttt{STOP} at $10$M after two assessments, whereas KoH reaches $50$M after six assessments and still underperforms. Thus, lower cost per checkpoint does not imply lower cost to termination. The return difference reflects the full selection, mutation, pruning, and training path: it does not establish superior frozen ranking accuracy, since Swiss and RR recover the same top-three ranking and elite pair at $10$M. The downstream API costs in \autoref{fig:results_overview}.C exclude the shared initial-generation cost of $\$1.117$.

We additionally evaluate two non-tournament monolithic controls only in the downstream branch-extension setting. \textbf{No Evolution, No Tournaments (NENT)} removes Stage~II entirely by removing tournaments, elite exchange, and cross-candidate selection, while retaining only independent Stage~III self-refinement. Under this \emph{Monolithic Non-Evolutionary Assessment} setting, the Stage~III assessor---conditioned only on boundary rollout evidence (start/end frames)---prematurely returned \texttt{STOP} at the $1$M checkpoint despite zero ground-truth goal attainments. This indicates that tournament-guided relative comparison and elite-mediated evolutionary structure are important for \emph{suppressing early false positives}. We posit that, in the absence of pairwise calibration against competing candidates, the assessor over-interprets weak local progress cues as evidence of task completion.

\paragraph{Monolithic evolutionary assessment.}
\textbf{Monolithic Evolutionary Assessment (MEA)} ablates tournament structure while retaining evolutionary refinement by replacing pairwise tournaments with a single monolithic all-candidate assessment prompt. We matched rollout evidence across conditions where feasible. MEA receives all candidates in one monolithic prompt, exposing the same rollout set as the pairwise judges yet without structured pairwise decomposition. To accommodate the monolithic visual payload, images were downsampled to $768$px, which as supported by \autoref{tab:resolution_sensitivity_appendix_summary}, does not materially affect the judge's reasoning ability. Thus, failures reflect the limits of absolute K-way visual assessment rather than reduced evidence. For MEA, Stage~II receives full rollout evidence for absolute $K$-way elite selection, while Stage~III remains boundary-frame based per candidate. This setting tests monolithic assessment while retaining evolutionary selection and mutation, with the visual-payload adaptation stated above. MEA passes the $1$M and $10$M checks and continues training with improving but still sub-solution returns, before incorrectly returning \texttt{STOP} at $20$M. Thus, monolithic evolutionary assessment delays, but does not eliminate, false-positive termination.

\paragraph{Vision payload for the KoH downstream branch (Results V.B).} At later \textbf{KoH} branch checkpoints beyond $20$M environment steps, the image payload exceeded the input-context budget of the evaluation model (gpt5.2 at 400k input), so we \emph{uniformly downsampled all rollout frames} while keeping the judge, prompt template, candidate pool, and temporal sampling policy $\pi_{\mathrm{sample}}(t)$ fixed. We chose uniform downsampling over alternatives such as switching to a different model (e.g. gpt5.4) or altering the recording stride (e.g. from 12 to any higher interval) because it preserves the same episode identities, ordering, and boundary-anchored evidence $\{o^{v}_{0},o^{v}_{T}\}\subseteq\tau_{\mathrm{raw}}$, thereby introducing a notably smaller and more controlled confound than changing either the evaluator itself or the temporal information content; importantly, KoH was already unsuccessful prior to this downsampling intervention (at the $20$M checkpoint), so the negative downstream result does not rely on the downsampled regime either way. Furthermore, we support that downsampling from 4096px to 1024px has minimal to negligible effects as substantiated by our results on resolution sensitivity in \autoref{tab:resolution_sensitivity_appendix_summary}.

\subsection{Expanded Results VI: Vision Evidence Sensitivity}
\phantomsection
\label{results:sensitivity_vision_granularity}
\label{resolution_sampling}
\label{appendix_resolution_sampling}
\paragraph{Protocol and metrics.}
Because WATERFALL selects reward candidates through visual Swiss tournaments, we evaluate whether candidate selection remains stable when visual evidence is systematically reduced. All experiments in this appendix use the frozen \(10\)M checkpoint of Medusa's Revenge (POMDP), where the ground-truth high-success elite pair is identifiable as
\[
E^\star=\{k5,k8\}.
\]
The checkpoint, candidate pool, rollout-generation protocol, Swiss topology, prompt template, and evaluator are held fixed unless explicitly varied. Repeated runs vary only the VLM judge call on fixed rollout evidence, isolating evaluator-side decision stability from trajectory-realisation noise. We report four selection diagnostics: whether the top-ranked candidate belongs to \(E^\star\), whether both candidates in \(E^\star\) appear in the ordered top-three \(T_3\), whether both candidates in \(E^\star\) are included in the selected elite pool \(\hat{E}\), and whether the selected elite pool exactly equals \(E^\star\). Exact equality is stricter than needed for WATERFALL mutation because including an additional near-tied candidate can still preserve both solved elites for downstream inheritance. We use the $10$M checkpoint because candidate performance is sufficiently separated to identify the high-success reference pair.

\paragraph{Direct judge correctness and confidence.}
For the episode-count study, reference success rates are computed from the non-shown complement of the ten-episode pool when $N_{\mathrm{ep}}<10$. At $N_{\mathrm{ep}}=10$, no complement remains and the reference uses the full pool. Pairwise accuracy measures agreement with the ordering of unequal reference success rates. Brier scores compare reported winner confidence with whether that winner is correct; ECE uses five equal-width confidence bins.

\begin{table}[htbp]
\centering
\caption{\textbf{Direct VLM-judge correctness and confidence diagnostics at the frozen Env.~III PO 10M checkpoint.} Pairwise accuracy excludes comparisons tied in reference task success. Spearman \(\rho\) compares tournament ranking with reference task success, and Brier evaluates reported winner confidence against correctness. ``Both elites'' denotes \(E^\star=\{k5,k8\}\subseteq\hat{E}\). These frozen-evidence diagnostics assess evaluator correctness separately from elite-recovery outcomes in \autoref{fig:results_overview}.}
\vspace{0.25cm}
\footnotesize
\setlength{\tabcolsep}{7.5pt}
\renewcommand{\arraystretch}{1.05}
\begin{tabular}{llrrrrrc}
\toprule
\textbf{Study} & \textbf{Setting} & \textbf{Runs} & \textbf{Non-tied} & \textbf{Accuracy} & \(\boldsymbol{\rho}\) & \textbf{Brier} & \textbf{Both elites} \\
\midrule
\multirow{4}{*}{\textbf{Episode count}}
& \(N_{\mathrm{ep}}=1\)  & 30 & 462 & 83.5\% & 0.738 & 0.142 & 10/30 \\
& \(N_{\mathrm{ep}}=3\)  & 12 & 180 & 93.9\% & 0.846 & 0.081 & 9/12 \\
& \(N_{\mathrm{ep}}=5\)  & 9  & 129 & 89.9\% & 0.797 & 0.083 & 9/9 \\
& \(N_{\mathrm{ep}}=10\) & 3  & 46  & 97.8\% & 0.869 & 0.044 & 3/3 \\
\midrule
\multirow{7}{*}{\textbf{Temporal stride}}
& \(s=2\)        & 3 & 49 & 93.9\%  & 0.765 & 0.085 & 1/3 \\
& \(s=4\)        & 3 & 48 & 100.0\% & 0.865 & 0.057 & 3/3 \\
& \(s=8\)        & 3 & 48 & 100.0\% & 0.874 & 0.061 & 3/3 \\
& \(s=16\)       & 3 & 48 & 100.0\% & 0.856 & 0.059 & 3/3 \\
& \(s=32\)       & 3 & 47 & 93.6\%  & 0.797 & 0.087 & 3/3 \\
& \(s=64\)       & 3 & 48 & 100.0\% & 0.865 & 0.065 & 3/3 \\
& \(s=\infty\)   & 3 & 48 & 95.8\%  & 0.851 & 0.080 & 3/3 \\
\midrule
\multirow{5}{*}{\textbf{Spatial resolution}}
& \(1024\)px & 3 & 48 & 85.4\% & 0.772 & 0.120 & 3/3 \\
& \(768\)px  & 3 & 47 & 83.0\% & 0.751 & 0.114 & 3/3 \\
& \(512\)px  & 3 & 45 & 93.3\% & 0.874 & 0.074 & 3/3 \\
& \(256\)px  & 3 & 50 & 80.0\% & 0.674 & 0.171 & 0/3 \\
& \(128\)px  & 3 & 50 & 66.0\% & 0.455 & 0.226 & 0/3 \\
\bottomrule
\end{tabular}
\label{tab:judge_correctness}
\end{table}

At three episodes, five-bin ECE is $0.157$. Restricting to comparisons with a reference success-rate margin of at least $0.20$ gives $151/162$ correct decisions ($93.2\%$). These diagnostics share frozen policies and recorded episodes, and some subsets overlap; they are not independent replications of the complete reward-discovery procedure. The high-success reference pair $E^\star=\{k5,k8\}$ is distinct from the candidate-level criterion used in the primary STOP audit. 

\paragraph{Spatial-resolution sensitivity.}
We first test whether Swiss tournament outcomes depend critically on high-fidelity imagery by uniformly downsampling rollout images. We evaluate
\[
r \in \{4096,1024,768,512,256,128\}\ \mathrm{px},
\]
using \(4096\)px as the high-fidelity reference condition. Let \(\hat{k}^{(j)}_r\) and \(\hat{E}^{(j)}_r\) denote the top-ranked candidate and elite pool at resolution \(r\) under judge repetition \(j\). We report top-1 agreement with the reference tournament,
\[
A_1(r)=\frac{1}{J_r}\sum_{j=1}^{J_r}\mathbb{I}\!\left[\hat{k}^{(j)}_r=\hat{k}_{4096}\right],
\]
and exact elite-set recovery,
\[
A_E(r)=\frac{1}{J_r}\sum_{j=1}^{J_r}\mathbb{I}\!\left[\hat{E}^{(j)}_r=E^\star\right].
\]
The \(4096\)px setting is a single reference due to input-limit and cost constraints; all lower-resolution settings are repeated across three independent VLM judge calls.

\begin{table}[htbp]
\centering
\caption{\textbf{Spatial-resolution sensitivity at the frozen \(10\)M checkpoint.} \(A_1(r)\) measures top-1 agreement with the \(4096\)px reference tournament, whose top-ranked candidate is \(k8\). \(A_E(r)\) measures exact elite-set recovery of \(E^\star=\{k5,k8\}\). Costs are logged API costs in USD; lower-resolution costs are reported as mean\(\pm\)std across three repeated judge calls.}
\vspace{0.25cm}
\footnotesize
\setlength{\tabcolsep}{3pt}
\renewcommand{\arraystretch}{1.05}
\resizebox{\ifdim\width>\textwidth\textwidth\else\width\fi}{!}{%
\begin{tabular}{lcccccccc}
\toprule
\textbf{Resolution} & \textbf{Runs} & \(\mathbf{A_1(r)}\) & \(\mathbf{Top1{\in}E^\star}\) & \(\mathbf{E^\star{\subseteq}T_3}\) & \(\mathbf{A_E(r)}\) & \textbf{Cost/run} & \textbf{Total cost} & \textbf{Pattern} \\
\midrule
4096px & 1 & \(1/1\) & \(1/1\) & \(1/1\) & \(1/1\) & \(\$5.155\) & \(\$5.16\) & Reference \\
1024px & 3 & \(3/3\) & \(3/3\) & \(3/3\) & \(3/3\) & \(\$3.926\pm0.124\) & \(\$11.78\) & Invariant \\
768px  & 3 & \(3/3\) & \(3/3\) & \(3/3\) & \(3/3\) & \(\$2.510\pm0.183\) & \(\$7.53\) & Invariant \\
512px  & 3 & \(3/3\) & \(3/3\) & \(3/3\) & \(3/3\) & \(\$1.658\pm0.020\) & \(\$4.97\) & Invariant \\
256px  & 3 & \(0/3\) & \(3/3\) & \(0/3\) & \(0/3\) & \(\$1.358\pm0.052\) & \(\$4.07\) & Retains \(k5\) only \\
128px  & 3 & \(1/3\) & \(1/3\) & \(2/3\) & \(0/3\) & \(\$1.099\pm0.098\) & \(\$3.30\) & Unstable \\
\bottomrule
\end{tabular}%
}
\label{tab:resolution_sensitivity_appendix_summary}
\end{table}

\begin{table}[htbp]
\centering
\caption{\textbf{Per-run spatial-resolution sensitivity.} Stable recovery holds from \(1024\)px through \(512\)px. At \(256\)px, the judge still ranks a solved candidate first, but loses the second solved elite \(k8\). At \(128\)px, selection becomes unstable and sometimes elevates non-solved candidates.}
\vspace{0.25cm}
\footnotesize
\setlength{\tabcolsep}{3pt}
\renewcommand{\arraystretch}{1.02}
\resizebox{\ifdim\width>\textwidth\textwidth\else\width\fi}{!}{%
\begin{tabular}{lclllcccc}
\toprule
\textbf{Res.} & \textbf{Run} & \textbf{Top-1} & \(\mathbf{T_3}\) & \(\mathbf{\hat{E}}\) & \(\mathbf{Top1{\in}E^\star}\) & \(\mathbf{E^\star{\subseteq}T_3}\) & \(\mathbf{\hat{E}=E^\star}\) & \textbf{Cost} \\
\midrule
4096 & 1 & \(k8\)  & \(\{k8,k5,k1\}\)   & \(\{k8,k5\}\)       & Y & Y & Y & \(\$5.155\) \\ \midrule
1024 & 1 & \(k8\)  & \(\{k8,k5,k10\}\)  & \(\{k8,k5\}\)       & Y & Y & Y & \(\$3.785\) \\
1024 & 2 & \(k8\)  & \(\{k8,k5,k2\}\)   & \(\{k8,k5\}\)       & Y & Y & Y & \(\$4.017\) \\
1024 & 3 & \(k8\)  & \(\{k8,k5,k1\}\)   & \(\{k8,k5\}\)       & Y & Y & Y & \(\$3.977\) \\
\midrule
768  & 1 & \(k8\)  & \(\{k8,k5,k2\}\)   & \(\{k8,k5\}\)       & Y & Y & Y & \(\$2.487\) \\
768  & 2 & \(k8\)  & \(\{k8,k5,k10\}\)  & \(\{k8,k5\}\)       & Y & Y & Y & \(\$2.703\) \\
768  & 3 & \(k8\)  & \(\{k8,k5,k2\}\)   & \(\{k8,k5\}\)       & Y & Y & Y & \(\$2.339\) \\
\midrule
512  & 1 & \(k8\)  & \(\{k8,k5,k3\}\)   & \(\{k8,k5\}\)       & Y & Y & Y & \(\$1.670\) \\
512  & 2 & \(k8\)  & \(\{k8,k5,k3\}\)   & \(\{k8,k5\}\)       & Y & Y & Y & \(\$1.669\) \\
512  & 3 & \(k8\)  & \(\{k8,k5,k3\}\)   & \(\{k8,k5\}\)       & Y & Y & Y & \(\$1.635\) \\
\midrule
256  & 1 & \(k5\)  & \(\{k5,k10,k1\}\)  & \(\{k5,k10\}\)      & Y & N & N & \(\$1.411\) \\
256  & 2 & \(k5\)  & \(\{k5,k10,k3\}\)  & \(\{k5,k10,k3\}\)   & Y & N & N & \(\$1.308\) \\
256  & 3 & \(k5\)  & \(\{k5,k1,k3\}\)   & \(\{k5,k1,k3\}\)    & Y & N & N & \(\$1.354\) \\
\midrule
128  & 1 & \(k10\) & \(\{k10,k5,k8\}\)  & \(\{k10,k5\}\)      & N & Y & N & \(\$1.178\) \\
128  & 2 & \(k8\)  & \(\{k8,k10,k4\}\)  & \(\{k8,k10,k4\}\)   & Y & N & N & \(\$0.989\) \\
128  & 3 & \(k10\) & \(\{k10,k5,k8\}\)  & \(\{k10,k5\}\)      & N & Y & N & \(\$1.130\) \\
\bottomrule
\end{tabular}%
}
\label{tab:resolution_sensitivity_appendix_full}
\end{table}

\paragraph{Temporal frame-sampling sensitivity.}
We next test sensitivity to temporal evidence density. We vary the recording stride as
\[
s\in\{2,4,8,16,32,64,\infty\},
\]
where \(s=\infty\) denotes an intentionally harsh boundary-only condition containing only the start and terminal frames of each episode. We omit \(s=1\) because it exceeded the input limits of the evaluator. To isolate temporal information content, the \(s=2\) recording is treated as the highest-fidelity feasible base visual dataset and coarser finite strides are generated by deterministic subsampling rather than by re-recording different trajectories. Using the same trajectories avoids confounding temporal evidence with policy stochasticity or environment randomisation. Images are fixed at $512{\times}512$px. Qualitative digests suggest that the dense $s=2$ condition creates ambiguity near apparent goal states, whereas boundary-only judging relies heavily on terminal frames and degrades qualitative judge assessment whereby candidates may be exploring trajectories and interactions that could be useful yet not clearly successful or not. Intermediate strides retain temporal context for distinguishing trajectories while reducing API cost. These are only qualitative interpretations of the observed selections and dedicated experiments on such rollouts would solidify this.

\begin{table}[htbp]
\centering
\caption{\textbf{Temporal frame-sampling sensitivity at the frozen \(10\)M checkpoint.} All finite-stride conditions use the same underlying rollout evidence, with coarser conditions generated by deterministic subsampling. The top-ranked candidate is solved for every finite stride. Both solved elites are consistently retained from \(s=4\) through \(s=64\), whereas \(s=2\) and the boundary-only setting show reduced stability in at least one metric.}
\vspace{0.25cm}
\footnotesize
\setlength{\tabcolsep}{3pt}
\renewcommand{\arraystretch}{1.05}
\resizebox{\ifdim\width>\textwidth\textwidth\else\width\fi}{!}{%
\begin{tabular}{lcccccccc}
\toprule
\textbf{Stride \(s\)} & \textbf{Runs} & \(\mathbf{Top1{\in}E^\star}\) & \(\mathbf{E^\star{\subseteq}T_3}\) & \(\mathbf{E^\star{\subseteq}\hat{E}}\) & \(\mathbf{\hat{E}=E^\star}\) & \textbf{Cost/run} & \textbf{Total cost} & \textbf{Top-1 runs} \\
\midrule
\(2\)        & 3 & \(3/3\) & \(1/3\) & \(1/3\) & \(0/3\) & \(\$4.737\pm0.054\) & \(\$14.21\) & \(k5,k5,k5\) \\
\(4\)        & 3 & \(3/3\) & \(3/3\) & \(3/3\) & \(0/3\) & \(\$2.948\pm0.153\) & \(\$8.84\)  & \(k5,k5,k5\) \\
\(8\)        & 3 & \(3/3\) & \(3/3\) & \(3/3\) & \(0/3\) & \(\$1.955\pm0.039\) & \(\$5.87\)  & \(k5,k5,k5\) \\
\(16\)       & 3 & \(3/3\) & \(3/3\) & \(3/3\) & \(0/3\) & \(\$1.451\pm0.037\) & \(\$4.35\)  & \(k5,k5,k5\) \\
\(32\)       & 3 & \(3/3\) & \(3/3\) & \(3/3\) & \(2/3\) & \(\$1.259\pm0.042\) & \(\$3.78\)  & \(k5,k5,k5\) \\
\(64\)       & 3 & \(3/3\) & \(3/3\) & \(3/3\) & \(0/3\) & \(\$1.131\pm0.019\) & \(\$3.39\)  & \(k5,k5,k5\) \\
\(\infty\)  & 3 & \(2/3\) & \(3/3\) & \(3/3\) & \(0/3\) & \(\$1.148\pm0.069\) & \(\$3.45\)  & \(k1,k5,k5\) \\
\bottomrule
\end{tabular}%
}
\label{tab:temporal_sampling_sensitivity_appendix_summary}
\end{table}

\begin{table}[t]
\centering
\caption{\textbf{Per-run temporal frame-sampling sensitivity.} Repeated judge calls vary only the VLM evaluation, not the underlying trajectories. Exact elite equality is stricter than elite preservation: several runs include both solved candidates plus a near-tied additional candidate, which is acceptable for downstream mutation but recorded here for completeness.}
\vspace{0.25cm}
\footnotesize
\setlength{\tabcolsep}{3pt}
\renewcommand{\arraystretch}{1.02}
\begin{tabular}{lclllccccc}
\toprule
\textbf{Stride} & \textbf{Run} & \textbf{Top-1} & \(\mathbf{T_3}\) & \(\mathbf{\hat{E}}\) & \(\mathbf{Top1{\in}E^\star}\) & \(\mathbf{E^\star{\subseteq}T_3}\) & \(\mathbf{E^\star{\subseteq}\hat{E}}\) & \(\mathbf{\hat{E}=E^\star}\) & \textbf{Cost} \\
\midrule
\(2\)       & 1 & \(k5\) & \(\{k5,k1,k6\}\) & \(\{k5,k1\}\)       & Y & N & N & N & \(\$4.798\) \\
\(2\)       & 2 & \(k5\) & \(\{k5,k3,k1\}\) & \(\{k5,k3,k1\}\)    & Y & N & N & N & \(\$4.718\) \\
\(2\)       & 3 & \(k5\) & \(\{k5,k8,k1\}\) & \(\{k5,k8,k1\}\)    & Y & Y & Y & N & \(\$4.694\) \\ \midrule
\(4\)       & 1 & \(k5\) & \(\{k5,k1,k8\}\) & \(\{k5,k1,k8\}\)    & Y & Y & Y & N & \(\$2.790\) \\
\(4\)       & 2 & \(k5\) & \(\{k5,k8,k1\}\) & \(\{k5,k8,k1\}\)    & Y & Y & Y & N & \(\$3.096\) \\
\(4\)       & 3 & \(k5\) & \(\{k5,k8,k1\}\) & \(\{k5,k8,k1\}\)    & Y & Y & Y & N & \(\$2.958\) \\ \midrule
\(8\)       & 1 & \(k5\) & \(\{k5,k8,k1\}\) & \(\{k5,k8,k1\}\)    & Y & Y & Y & N & \(\$1.971\) \\
\(8\)       & 2 & \(k5\) & \(\{k5,k8,k1\}\) & \(\{k5,k8,k1\}\)    & Y & Y & Y & N & \(\$1.984\) \\
\(8\)       & 3 & \(k5\) & \(\{k5,k8,k1\}\) & \(\{k5,k8,k1\}\)    & Y & Y & Y & N & \(\$1.911\) \\ \midrule
\(16\)      & 1 & \(k5\) & \(\{k5,k1,k8\}\) & \(\{k5,k1,k8\}\)    & Y & Y & Y & N & \(\$1.488\) \\
\(16\)      & 2 & \(k5\) & \(\{k5,k8,k1\}\) & \(\{k5,k8,k1\}\)    & Y & Y & Y & N & \(\$1.414\) \\
\(16\)      & 3 & \(k5\) & \(\{k5,k1,k8\}\) & \(\{k5,k1,k8\}\)    & Y & Y & Y & N & \(\$1.451\) \\ \midrule
\(32\)      & 1 & \(k5\) & \(\{k5,k8,k1\}\) & \(\{k5,k8,k1\}\)    & Y & Y & Y & N & \(\$1.290\) \\
\(32\)      & 2 & \(k5\) & \(\{k5,k8,k4\}\) & \(\{k5,k8\}\)       & Y & Y & Y & Y & \(\$1.211\) \\
\(32\)      & 3 & \(k5\) & \(\{k5,k8,k1\}\) & \(\{k5,k8\}\)       & Y & Y & Y & Y & \(\$1.275\) \\ \midrule
\(64\)      & 1 & \(k5\) & \(\{k5,k1,k8\}\) & \(\{k5,k1,k8\}\)    & Y & Y & Y & N & \(\$1.152\) \\
\(64\)      & 2 & \(k5\) & \(\{k5,k8,k1\}\) & \(\{k5,k8,k1\}\)    & Y & Y & Y & N & \(\$1.118\) \\
\(64\)      & 3 & \(k5\) & \(\{k5,k8,k1\}\) & \(\{k5,k8,k1\}\)    & Y & Y & Y & N & \(\$1.122\) \\ \midrule
\(\infty\) & 1 & \(k1\) & \(\{k1,k5,k8\}\) & \(\{k1,k5,k8\}\)    & N & Y & Y & N & \(\$1.069\) \\
\(\infty\) & 2 & \(k5\) & \(\{k5,k8,k1\}\) & \(\{k5,k8,k1\}\)    & Y & Y & Y & N & \(\$1.197\) \\
\(\infty\) & 3 & \(k5\) & \(\{k5,k8,k1\}\) & \(\{k5,k8,k1\}\)    & Y & Y & Y & N & \(\$1.179\) \\
\bottomrule
\end{tabular}
\label{tab:temporal_sampling_sensitivity_appendix_full}
\end{table}

\paragraph{Rollout episode-count sensitivity.}
Finally, we test how many rollout episodes per candidate are needed for stable visual tournament selection. We use a fixed ten-episode master pool, \(512{\times}512\) images, temporal stride \(s=8\), the same frozen \(10\)M checkpoint, and the same episode-index subset for every candidate. The subsets are:
\[
\begin{aligned}
N_{\mathrm{ep}}=1 &: \{1\},\{2\},\ldots,\{10\},\\
N_{\mathrm{ep}}=3 &: \{1,2,3\},\{4,5,6\},\{7,8,9\},\{2,5,10\},\\
N_{\mathrm{ep}}=5 &: \{1,2,3,4,5\},\{6,7,8,9,10\},\{1,3,5,7,9\},\\
N_{\mathrm{ep}}=10 &: \{1,\ldots,10\}.
\end{aligned}
\]
The recorded ground-truth success rates in this master pool are \(k5=100\%\), \(k8=90\%\), \(k3=70\%\), \(k1=40\%\), and all other candidates \(0\%\). We therefore treat \(E^\star=\{k5,k8\}\) as the high-success elite pair, while noting that \(k3\) is a partially successful distractor that often tests whether the judge can distinguish consistent solution behaviour from partial progress.

For label calibration, each Swiss digest aggregates \texttt{SUCCESS/FAIL/UNCERTAIN} labels over repeated pairwise appearances. Exact per-episode confusion matrices cannot always be reconstructed from the aggregate digest alone, so we report conservative count-level diagnostics:
\[
\mathrm{FP}^{\min}_{\mathrm{succ}}=\sum_k \max(0,\hat{S}_k-S^\star_k),
\qquad
\mathrm{Miss}_{\mathrm{succ}}=\sum_k \max(0,S^\star_k-\hat{S}_k),
\]
where \(\hat{S}_k\) is the judge-reported aggregate success count and \(S^\star_k\) is the minimum ground-truth success count implied by the same candidate's match appearances and sampled episode subset. These quantities are diagnostics of label calibration inside each tournament, not the primary selection criterion.

\begin{table}[htbp]
\centering
\caption{\textbf{Aggregate episode-count sensitivity at the frozen \(10\)M checkpoint.} Single-episode evaluation is brittle. Three episodes are enough to recover a solved top-ranked candidate in every judge call, although complete solved-elite preservation is not perfect for all three-episode subsets. Five and ten episodes provide the most stable recovery of both solved elites, while exact elite equality is occasionally violated by inclusion of an additional near-tied candidate.}
\vspace{0.25cm}
\footnotesize
\setlength{\tabcolsep}{3pt}
\renewcommand{\arraystretch}{1.05}
\resizebox{\ifdim\width>\textwidth\textwidth\else\width\fi}{!}{%
\begin{tabular}{cccccccc}
\toprule
\(\mathbf{N_{\mathrm{ep}}}\) & \textbf{Subsets} & \textbf{Judge calls} & \(\mathbf{Top1{\in}E^\star}\) & \(\mathbf{E^\star{\subseteq}T_3}\) & \(\mathbf{E^\star{\subseteq}\hat{E}}\) & \(\mathbf{\hat{E}=E^\star}\) & \textbf{Cost/run} \\
\midrule
\(1\)  & 10 & 30 & \(16/30\) & \(18/30\) & \(10/30\) & \(8/30\) & \(\$1.015\pm0.080\) \\
\(3\)  & 4  & 12 & \(12/12\) & \(11/12\) & \(9/12\)  & \(9/12\) & \(\$2.131\pm0.174\) \\
\(5\)  & 3  & 9  & \(9/9\)   & \(9/9\)   & \(9/9\)   & \(8/9\)  & \(\$2.911\pm0.086\) \\
\(10\) & 1  & 3  & \(3/3\)   & \(3/3\)   & \(3/3\)   & \(2/3\)  & \(\$5.242\pm0.186\) \\
\bottomrule
\end{tabular}%
}
\label{tab:episode_count_sensitivity_appendix_summary}
\end{table}

\begin{table}[htbp]
\centering
\caption{\textbf{Compact per-subset episode-count sensitivity.} Each row aggregates three repeated VLM judge calls on fixed rollout evidence. Single-episode judging is brittle: solved top-1 recovery occurs in \(16/30\) judge calls, and both solved elites are preserved in the elite pool in \(10/30\). With \(N_{\mathrm{ep}}=3\), a solved candidate is always selected top-1, but subset \(\{4,5,6\}\) remains less stable because the partially successful distractor \(k3\) is repeatedly elevated. With \(N_{\mathrm{ep}}\ge5\), both solved elites are preserved in all runs, although exact elite equality is occasionally broken by an additional near-tied candidate.}
\vspace{0.25cm}
\footnotesize
\setlength{\tabcolsep}{3.2pt}
\renewcommand{\arraystretch}{1.05}
\resizebox{\ifdim\width>\textwidth\textwidth\else\width\fi}{!}{%
\begin{tabular}{cclcccc}
\toprule
\(\mathbf{N_{\mathrm{ep}}}\) & \textbf{Subset} & \textbf{Episodes} & \textbf{Top-1 across 3 runs} & \(\mathbf{E^\star{\subseteq}T_3}\) & \(\mathbf{E^\star{\subseteq}\hat{E}}\) & \(\mathbf{\hat{E}=E^\star}\) \\
\midrule
1  & 1  & \(\{1\}\)              & \(k5,k5,k8\) & \(3/3\) & \(2/3\) & \(1/3\) \\
1  & 2  & \(\{2\}\)              & \(k3,k3,k3\) & \(1/3\) & \(1/3\) & \(0/3\) \\
1  & 3  & \(\{3\}\)              & \(k8,k8,k8\) & \(1/3\) & \(1/3\) & \(1/3\) \\
1  & 4  & \(\{4\}\)              & \(k3,k3,k3\) & \(2/3\) & \(0/3\) & \(0/3\) \\
1  & 5  & \(\{5\}\)              & \(k5,k5,k8\) & \(3/3\) & \(3/3\) & \(3/3\) \\
1  & 6  & \(\{6\}\)              & \(k1,k1,k1\) & \(0/3\) & \(0/3\) & \(0/3\) \\
1  & 7  & \(\{7\}\)              & \(k5,k5,k5\) & \(2/3\) & \(0/3\) & \(0/3\) \\
1  & 8  & \(\{8\}\)              & \(k3,k3,k3\) & \(2/3\) & \(0/3\) & \(0/3\) \\
1  & 9  & \(\{9\}\)              & \(k3,k3,k5\) & \(1/3\) & \(0/3\) & \(0/3\) \\
1  & 10 & \(\{10\}\)             & \(k5,k5,k5\) & \(3/3\) & \(3/3\) & \(3/3\) \\
\midrule
3  & 1  & \(\{1,2,3\}\)          & \(k5,k5,k5\) & \(3/3\) & \(3/3\) & \(3/3\) \\
3  & 2  & \(\{4,5,6\}\)          & \(k5,k5,k5\) & \(2/3\) & \(0/3\) & \(0/3\) \\
3  & 3  & \(\{7,8,9\}\)          & \(k5,k5,k5\) & \(3/3\) & \(3/3\) & \(3/3\) \\
3  & 4  & \(\{2,5,10\}\)         & \(k5,k5,k5\) & \(3/3\) & \(3/3\) & \(3/3\) \\
\midrule
5  & 1  & \(\{1,2,3,4,5\}\)      & \(k5,k5,k5\) & \(3/3\) & \(3/3\) & \(3/3\) \\
5  & 2  & \(\{6,7,8,9,10\}\)     & \(k5,k5,k5\) & \(3/3\) & \(3/3\) & \(3/3\) \\
5  & 3  & \(\{1,3,5,7,9\}\)      & \(k8,k5,k5\) & \(3/3\) & \(3/3\) & \(2/3\) \\
\midrule
10 & 1  & \(\{1,\ldots,10\}\)    & \(k5,k5,k5\) & \(3/3\) & \(3/3\) & \(2/3\) \\
\bottomrule
\end{tabular}%
}
\label{tab:episode_count_sensitivity_appendix_compact}
\end{table}

\paragraph{Cost accounting for Results VI.}
Attributed API costs sum to $\$178.63$ across the vision-evidence sensitivity conditions: $\$36.81$ for spatial resolution, $\$43.89$ for temporal frame-sampling, and $\$97.94$ for rollout episode-count sensitivity; reused primary-reference requests are not additional expenditure. These costs are reported as systems-level reproducibility metadata rather than as performance metrics. They also motivate the fixed-evidence protocol: rerunning full independent policies or re-recording fresh rollout pools for every evidence condition would conflate visual-evidence effects with PPO stochasticity, environment realisation, and trajectory variance while multiplying API cost.

\section{Expanded Qualitative Analyses}
\label{app:qualitative}
This section complements the quantitative results with qualitative views of WATERFALL's behaviour. We visualise the behavioural diversity of the persona-conditioned candidates (Appendix~\ref{app:persona_diversity}) and show how tournament aggregation handles hallucinated and ambiguous \texttt{SUCCESS} labels (Appendix~\ref{app:robustness}). These views are descriptive: they illustrate the mechanisms behind the quantitative results rather than adding new claims.

\subsection{Persona-Conditioned Behavioural Diversity: PCA and t-SNE}
\label{app:persona_diversity}
\paragraph{Purpose and data.}
\autoref{fig:stage_i_persona_tsne} visualises behavioural variation among candidate policies. Each point represents a candidate--seed pair, obtained by averaging numerical rollout descriptors over $1{,}000$ inference episodes in Env.~I--III and $100$ in Env.~IV. Descriptors include episode length, unique visible-state counts, and action fractions. With ten candidates, complete layers contain $100$ points in Env.~I--III and $50$ in Env.~IV. Colours identify the assigned Stage-I personas. Rollouts are collected in inference mode with policy parameters held fixed. Each environment and observability setting is embedded independently.

\paragraph{Preprocessing and PCA.}
Within each layer, missing or non-finite entries are imputed using feature medians. Features with variance at most $10^{-12}$ are removed, and the remaining features are standardised to zero mean and unit variance. PCA reduces the descriptor dimension before t-SNE. For $N$ points and $D$ retained features, the number of components is
\[
q=\min(q_{\max},D,N-1),
\]
with component caps of $q_{\max}=20$ for Env.~I--III and $q_{\max}=10$ for Env.~IV PO in the reference configuration.

\paragraph{t-SNE settings.}
Each PCA representation is embedded in two dimensions using Euclidean distances, PCA initialisation, an automatic learning rate, and a maximum of $1{,}000$ optimisation iterations. Nominal perplexity is $p_0=10$ for Env.~I--III and $p_0=5$ for Env.~IV PO, adjusted for smaller samples as
\[
p=\min\!\left(p_0,\max\!\left(2,\frac{N-1}{3}\right)\right).
\]
For complete layers, this gives perplexities of $10$ and $5$, respectively. The reference random state is $7+i$, where $i$ indexes layers from zero in bottom-to-top input order.

\paragraph{Visualisation and interpretation.}
Embedding coordinates are clipped at their second and ninety-eighth percentiles, rescaled within each layer, and geometrically stacked for display. The layers have no shared coordinate system, so their alignment and cross-layer distances carry no behavioural meaning. The figure provides a qualitative description of variation in the selected descriptors; cluster sizes and separation are not quantitative diversity measures. This visualisation does not isolate the causal contribution of persona conditioning.

\subsection{Robustness to Label Noise, Ambiguity, and Evaluation Hallucinations}
\label{app:robustness}
VLM-based rollout evaluation can exhibit stochasticity, especially at early checkpoints where policies are weak, behaviours are not yet converged, and terminal states may be visually ambiguous. A common failure mode is an isolated spurious \texttt{SUCCESS} label assigned to a visually ambiguous terminal state, such as an opened lock/chest cue that is not actually task completion, or a trap-like/mimic-like terminal state. WATERFALL is designed not to treat such labels as ground truth. Instead, it couples hard-point tournament aggregation with confidence diagnostics and qualitative semantic consistency checks over behavioural signatures.

We report the \(1\mathrm{M}\)-step checkpoint because it is the noisiest evaluation point: policies are still unstable, many candidates terminate prematurely, and visual terminal states are most ambiguous. Later checkpoints are typically cleaner because successful candidates converge to more stable terminal behaviours, and the judge labels become more consistent with synchronised ground-truth sidecars. The examples below therefore stress-test the robustness mechanism at the point where VLM misclassification is most likely. We use \emph{hallucinated \texttt{SUCCESS}} only when the VLM assigns \texttt{SUCCESS} to a rollout that the synchronised ground-truth inference sidecar marks as failure. We use \emph{ambiguity} for cases where the visual evidence is incomplete or uncertain, even if no false-positive \texttt{SUCCESS} is established.

\paragraph{Uncertainty-aware tournament scoring.}
Within each Swiss-system tournament, WATERFALL maintains an uncertainty-aware soft score \(S_{\mathrm{soft}}\) using the judge confidence \(c\in[0,1]\). For a pairwise match where the judge declares candidate \(i\) the winner over candidate \(j\), the soft-score update is
\begin{equation}
\Delta S_{\mathrm{soft}}^{(i)} = c,
\qquad
\Delta S_{\mathrm{soft}}^{(j)} = 1-c .
\end{equation}
Confidence values closer to $1/2$ produce more symmetric diagnostic soft-score increments, hard-point updates do not depend on confidence. Elite selection uses hard points, Buchholz, and candidate ID, while confidence and qualitative digests provide context for Stage~III.

\paragraph{Qualitative semantic consistency filter.}
In addition to quantitative standings, WATERFALL stores per-candidate qualitative digests. These digests are not used to prove success. Rather, they provide an audit layer that flags inconsistencies between episode-level \texttt{SUCCESS/FAIL/UNCERTAIN} labels and recurrent visual terminal patterns, such as repeatedly ending at an ambiguous opened-object state or red mimic-like chest. This helps prevent over-interpreting raw \texttt{SUCCESS} counts when the visual evidence itself is ambiguous.

\paragraph{Empirical stress cases at the \(1\mathrm{M}\)-step checkpoint.}
The following digests distinguish three related failure modes. First, \emph{hallucinated-\texttt{SUCCESS} robustness}: in \textsc{Environment~1} POMDP, candidates \texttt{k5} and \texttt{k9} receive more raw VLM \texttt{SUCCESS} labels than \texttt{k3}, but synchronised ground-truth sidecars mark only \texttt{k3} as truly successful; WATERFALL selects \texttt{k3} as the sole elite. Second, \emph{ambiguity robustness}: in \textsc{Environment~2} POMDP, all candidates fail under ground-truth sidecars at \(1\mathrm{M}\), but the tournament still preserves candidates with the strongest relative tool-to-obstacle progress while explicitly recording uncertain terminal interpretations. Third, \textsc{Environment~3} provides direct hallucinated-\texttt{SUCCESS} stress cases: in the POMDP variant, candidates with false-positive \texttt{SUCCESS} labels under red/open chest terminal states are excluded from the elite pool; in the FO variant, an ambiguous false-positive \texttt{SUCCESS} label is contextualised by the qualitative digest, while another candidate with zero raw successes is selected based on stronger behavioural evidence.

\begin{lstlisting}[
  basicstyle=\ttfamily\tiny,
  frame=single,
  columns=fullflexible,
  breaklines=true,
  keepspaces=true,
  showstringspaces=false,
  aboveskip=0.6em,
  belowskip=0.2em,
  escapeinside={(*@}{@*)},
  caption={\textbf{Environment~1 POMDP: quantitative and qualitative robustness digest at \(1\mathrm{M}\) steps.}
  Candidate \texttt{k3} is selected as the sole elite despite having fewer raw \texttt{SUCCESS} labels than \texttt{k5} and \texttt{k9}. This illustrates hallucinated-\texttt{SUCCESS} robustness: WATERFALL does not simply rank by raw VLM \texttt{SUCCESS} counts, but instead combines Swiss pairwise evidence with qualitative behavioural consistency. Highlighted lines mark the relevant evidence.},
  label={lst:env1_po_robustness_digest_1m},
  captionpos=b
]
(*@\textbf{\sffamily\footnotesize Environment 1 POMDP -- MiniGrid Locked Room: Quantitative Vision-Judge Digest}@*)
========================================================================================================
 (*@\textbf{\sffamily TOURNAMENT DIGEST:}@*) CHECKPOINT 0 (ENVIRONMENT STEPS: 1000000)
 (*@\textbf{\sffamily Mode}@*): SwissSystem | (*@\textbf{\sffamily Rounds}@*): 5 | (*@\textbf{\sffamily Candidates}@*): 10 | (*@\textbf{\sffamily Bracket Randomisation}@*) = True
========================================================================================================

(*@\textbf{\sffamily --- LEADERBOARD (HIGHER SCORE IS BETTER) ---}@*)
(Sorted by: Score Descending > Buchholz Descending)
Rank  | Candidate  | Score    | Soft Score(*)    | Avg. Confidence    | Buchholz   | Total S/F/U(**)
-------------------------------------------------------------------------------------------------------------------
(*@\textcolor{red}{\bfseries\ttfamily 1~~~~~|~k3~~~~~~~~~|~5.0~~~~~~|~4.03~~~~~~~~~~~~~|~0.81~~~~~~~~~~~~~~~|~15.0~~~~~~~|~5/6/4}@*)
2     | k4         | 3.0      | 2.44             | 0.72               | 15.0       | 0/13/2
(*@\textcolor{red}{\bfseries\ttfamily 3~~~~~|~k5~~~~~~~~~|~3.0~~~~~~|~2.95~~~~~~~~~~~~~|~0.70~~~~~~~~~~~~~~~|~15.0~~~~~~~|~6/0/9}@*)
(*@\textcolor{red}{\bfseries\ttfamily 4~~~~~|~k9~~~~~~~~~|~3.0~~~~~~|~2.95~~~~~~~~~~~~~|~0.81~~~~~~~~~~~~~~~|~14.0~~~~~~~|~6/4/5}@*)
5     | k7         | 3.0      | 2.69             | 0.77               | 14.0       | 0/14/1
6     | k2         | 3.0      | 1.97             | 0.83               | 12.0       | 1/7/4
7     | k8         | 2.0      | 2.42             | 0.74               | 12.0       | 0/12/3
8     | k1         | 2.0      | 1.61             | 0.81               | 10.0       | 0/12/0
9     | k10        | 2.0      | 0.40             | 0.87               | 6.0        | 0/9/0
10    | k6         | 1.0      | 1.54             | 0.82               | 13.0       | 1/10/4

(*) Soft Score: winner receives +confidence, loser receives +(1-confidence), summed over matches.
(**) S/F/U: SUCCESS/FAIL/UNCERTAIN labels aggregated from visual rollouts.

(*@\textbf{\sffamily --- ELITE POOL SELECTION (Top 30\% within Delta=1) ---}@*)
Qualified Candidates: {3}
Total Qualified: 1

(*@\textbf{\sffamily\footnotesize Environment 1 POMDP -- Hallucinated-SUCCESS Robustness}@*)
========================================================================================================
 (*@\textbf{\sffamily TOURNAMENT QUALITATIVE DIGEST: CHECKPOINT 0 (ENVIRONMENT STEPS: 1000000)}@*)
========================================================================================================

(*@\textbf{\sffamily Candidate: k3}@*)
Behavioral signature: (*@\textcolor{red}{Repeatedly executes a complete-looking key-lock sequence: it collects colored keys, traverses multi-lock maps, and often reaches a glowing open treasure chest or a goal-adjacent opened blue/teal lock state.}@*) Even its non-successes remain structured around relevant keys and locks rather than collapsing into local dithering.
Progress signals: (*@\textcolor{red}{Multiple matches show an unambiguous glowing treasure chest at terminal, and many other episodes still end after key pickup and opened blue-lock progress.}@*)
Failure mode hypothesis: Likely (A) policy behaviour / incomplete execution in unresolved runs, but uncertain because several terminals stop right after major progress and may simply omit the final goal cue; repeated environment constraints are not evidenced.
Evidence: R1-M2, R2-M2, R3-M2, R4-M1, R5-M1

...

(*@\textbf{\sffamily Candidate: k5}@*)
Behavioral signature: Repeatedly and efficiently homes in on the same blue target-like object/door/lock from varied starts, often with short episodes and very little visible dithering. (*@\textcolor{red}{The interpretation remains uncertain because some summaries treat these early terminals as success-like while others say the final cue is visually ambiguous.}@*)
Progress signals: Persistent rapid convergence to the blue target across all matches is a strong proximity-to-solve signal, and several summaries explicitly judge that pattern as successful.
Failure mode hypothesis: Uncertain--non-success judgments are repeatedly about ambiguous early termination beside the blue target and limited visible key/use evidence, not a clearly repeated (A) or (B) pattern.
Evidence: R1-M3, R2-M1, R3-M1, R4-M2, R5-M1

...

(*@\textbf{\sffamily Candidate: k9}@*)
Behavioral signature: Efficient, blue-target-oriented policy that often reaches and opens the blue lock/door/box quickly, while still showing occasional broader exploration or red-object opening. (*@\textcolor{red}{Its main inconsistency is occasional longer wandering that ends at red or other intermediate states instead of blue completion.}@*)
Progress signals: Repeated blue unlock terminals across several matches, including very fast blue reaches, plus additional opened red/blue states in uncertain runs.
Failure mode hypothesis: Likely (A) policy behaviour in its misses--wandering/backtracking or settling for red/intermediate interactions instead of finishing at blue; some uncertainty remains when the final cue is not explicit, but repeated environment constraints are not supported.
Evidence: R1-M2, R2-M5, R3-M3, R4-M2, R5-M4
\end{lstlisting}

\begin{lstlisting}[
  basicstyle=\ttfamily\tiny,
  frame=single,
  columns=fullflexible,
  breaklines=true,
  keepspaces=true,
  showstringspaces=false,
  aboveskip=0.6em,
  belowskip=0.2em,
  escapeinside={(*@}{@*)},
  caption={\textbf{Environment~2 POMDP: quantitative and qualitative robustness digest at \(1\mathrm{M}\) steps.}
  All candidates fail under ground-truth sidecars at this early checkpoint, so this listing is primarily evidence for robustness to label ambiguity rather than a clean hallucinated-success rejection. Candidate \texttt{k6} receives mixed \texttt{SUCCESS/FAIL/UNCERTAIN} labels and is retained by the tournament standings, its qualitative digest records plausible tool-to-obstacle sequencing. Candidate \texttt{k3} is also retained as an elite despite zero raw successes. Highlighted lines mark the relevant evidence.},
  label={lst:env2_po_robustness_digest_1m},
  captionpos=b
]
(*@\textbf{\sffamily\footnotesize Environment 2 POMDP -- Dark Corridor: Quantitative Vision-Judge Digest}@*)
========================================================================================================
 (*@\textbf{\sffamily TOURNAMENT DIGEST:}@*) CHECKPOINT 0 (ENVIRONMENT STEPS: 1000000)
 (*@\textbf{\sffamily Mode}@*): SwissSystem | (*@\textbf{\sffamily Rounds}@*): 5 | (*@\textbf{\sffamily Candidates}@*): 10 | (*@\textbf{\sffamily Bracket Randomisation}@*) = True
========================================================================================================

(*@\textbf{\sffamily --- LEADERBOARD (HIGHER SCORE IS BETTER) ---}@*)
(Sorted by: Score Descending > Buchholz Descending)
Rank  | Candidate  | Score    | Soft Score(*)    | Avg. Confidence    | Buchholz   | Total S/F/U(**)
-------------------------------------------------------------------------------------------------------------------
(*@\textcolor{red}{\bfseries\ttfamily 1~~~~~|~k6~~~~~~~~~|~5.0~~~~~~|~4.14~~~~~~~~~~~~~|~0.83~~~~~~~~~~~~~~~|~14.0~~~~~~~|~3/7/5}@*)
(*@\textcolor{red}{\bfseries\ttfamily 2~~~~~|~k3~~~~~~~~~|~4.0~~~~~~|~2.93~~~~~~~~~~~~~|~0.74~~~~~~~~~~~~~~~|~15.0~~~~~~~|~0/15/0}@*)
3     | k8         | 3.0      | 2.68             | 0.74               | 15.0       | 0/13/2
4     | k1         | 3.0      | 2.79             | 0.78               | 13.0       | 0/15/0
5     | k2         | 3.0      | 2.77             | 0.69               | 12.0       | 0/15/0
6     | k5         | 2.0      | 2.43             | 0.75               | 13.0       | 0/15/0
7     | k10        | 2.0      | 2.04             | 0.74               | 11.0       | 0/11/4
8     | k9         | 2.0      | 2.17             | 0.76               | 10.0       | 0/15/0
9     | k7         | 1.0      | 2.18             | 0.72               | 12.0       | 0/15/0
10    | k4         | 0.0      | 0.87             | 0.83               | 10.0       | 0/15/0

(*) Soft Score: winner receives +confidence, loser receives +(1-confidence), summed over matches.
(**) S/F/U: SUCCESS/FAIL/UNCERTAIN labels aggregated from visual rollouts.

(*@\textbf{\sffamily --- ELITE POOL SELECTION (Top 30\% within Delta=1) ---}@*)
Qualified Candidates: {6, 3}
Total Qualified: 2

(*@\textbf{\sffamily\footnotesize Environment 2 POMDP -- Qualitative Evidence of Ambiguous Labels and Relative Progress}@*)
========================================================================================================
 (*@\textbf{\sffamily TOURNAMENT QUALITATIVE DIGEST: CHECKPOINT 0 (ENVIRONMENT STEPS: 1000000)}@*)
========================================================================================================

(*@\textbf{\sffamily Candidate: k6}@*)
Behavioral signature: k6 repeatedly executes multi-step item-to-obstacle behaviour: quick key or bucket pickup followed by movement toward the fire/mummy/door route, often reaching the door tile, treasure corridor, or an opened chest. Its main weakness is occasional overcommitment into the hazardous mummy/fire bottleneck.
Progress signals: (*@\textcolor{red}{Strong progress signals include repeated plausible bucket-fire-door and key-door-treasure sequences, plus multiple summaries with a visible opened treasure chest or treasure corridor at terminal.}@*) This is strong evidence of proximity to solving and occasional visible success.
Failure mode hypothesis: Repeated evidence supports mostly (A) policy behaviour on failures, especially hazard contact after otherwise useful progress. (*@\textcolor{red}{Some door-terminal episodes remain genuinely ambiguous, so uncertainty remains around whether a few non-success rollouts were incomplete or simply not visually confirmed.}@*)
Evidence: R1-M4, R2-M1, R3-M1, R4-M1, R5-M1

...

(*@\textbf{\sffamily Candidate: k3}@*)
Behavioral signature: k3 repeatedly matches the visible tool to the local obstacle: bucket in fire scenes, sword in mummy scenes, and key near locked doors. After that good first step, it usually loops, wanders, or times out without returning to complete the interaction.
Progress signals: (*@\textcolor{red}{The strongest signal is consistent context-sensitive item choice across all three recurring layout types, often with very quick pickup.}@*) That suggests proximity to solving the first subgoal, even though no extinguish, enemy defeat, door opening, or final completion is shown.
Failure mode hypothesis: Repeated evidence supports (A) policy behaviour: incomplete multi-step execution after a correct initial pickup, with stalling or backtracking before obstacle resolution. No repeated (B) episode-level blockage is evident, though some uncertainty remains from partial observability.
Evidence: R1-M5, R2-M2, R3-M2, R4-M1, R5-M2

...

(*@\textbf{\sffamily Candidate: k10}@*)
Behavioral signature: k10 shows two recurring modes: in some Episode 1 rollouts it makes a short, direct move toward the visible key with little dithering, while in Episode 3 it often turns into the mummy/fire/door path and terminates quickly. Episode 2 is frequently already terminal at step 0 or too sparse to reveal meaningful behaviour.
Progress signals: Its main positive signal is the repeated direct approach to the visible key in Episode 1, sometimes ending adjacent to it while still blue. Outside that narrow pattern, no consistent progress toward obstacle resolution is shown.
Failure mode hypothesis: Repeated evidence supports mainly (A) policy behaviour, especially the recurrent Episode 3 hazard entry into the mummy/fire corridor. (*@\textcolor{red}{Repeated step-0 Episode 2 terminals also introduce some uncertainty about possible episode-level anomalies, but the overall picture is still mostly behaviour-limited.}@*)
Evidence: R1-M4, R2-M3, R3-M5, R4-M5, R5-M4
\end{lstlisting}

\begin{lstlisting}[
  basicstyle=\ttfamily\tiny,
  frame=single,
  columns=fullflexible,
  breaklines=true,
  keepspaces=true,
  showstringspaces=false,
  aboveskip=0.6em,
  belowskip=0.2em,
  escapeinside={(*@}{@*)},
  caption={\textbf{Environment~3 POMDP: quantitative and qualitative robustness digest at \(1\mathrm{M}\) steps.}
  Candidates \texttt{k2} and \texttt{k6} receive isolated raw \texttt{SUCCESS} labels under red/open chest terminal states, but they are not selected as elites. The qualitative digest flags these terminal states as ambiguous or brittle, illustrating how WATERFALL surfaces potential VLM misclassification (false-positive VLM \texttt{SUCCESS} labels) rather than selecting directly by raw \texttt{SUCCESS} counts. Highlighted lines mark the relevant evidence.},
  label={lst:env3_po_robustness_digest_1m},
  captionpos=b
]
(*@\textbf{\sffamily\footnotesize Environment 3 POMDP -- Medusa's Revenge: Quantitative Vision-Judge Digest}@*)
========================================================================================================
 (*@\textbf{\sffamily TOURNAMENT DIGEST:}@*) CHECKPOINT 0 (ENVIRONMENT STEPS: 1000000)
 (*@\textbf{\sffamily Mode}@*): SwissSystem | (*@\textbf{\sffamily Rounds}@*): 5 | (*@\textbf{\sffamily Candidates}@*): 10 | (*@\textbf{\sffamily Bracket Randomisation}@*) = True
========================================================================================================

(*@\textbf{\sffamily --- LEADERBOARD (HIGHER SCORE IS BETTER) ---}@*)
(Sorted by: Score Descending > Buchholz Descending)
Rank  | Candidate  | Score    | Soft Score(*)    | Avg. Confidence    | Buchholz   | Total S/F/U(**)
-------------------------------------------------------------------------------------------------------------------
1     | k7         | 5.0      | 3.30             | 0.66               | 13.0       | 0/15/0
2     | k1         | 4.0      | 2.97             | 0.66               | 15.0       | 0/15/0
3     | k5         | 3.0      | 2.70             | 0.68               | 15.0       | 0/15/0
4     | k9         | 3.0      | 2.79             | 0.68               | 13.0       | 0/15/0
5     | k4         | 3.0      | 2.78             | 0.67               | 12.0       | 0/15/0
(*@\textcolor{red}{\bfseries\ttfamily 6~~~~~|~k2~~~~~~~~~|~3.0~~~~~~|~2.01~~~~~~~~~~~~~|~0.65~~~~~~~~~~~~~~~|~5.0~~~~~~~~|~3/9/0}@*)
7     | k3         | 2.0      | 2.26             | 0.66               | 13.0       | 0/15/0
8     | k8         | 2.0      | 2.45             | 0.65               | 13.0       | 0/15/0
9     | k10        | 1.0      | 1.12             | 0.72               | 12.0       | 0/12/0
(*@\textcolor{red}{\bfseries\ttfamily 10~~~~|~k6~~~~~~~~~|~0.0~~~~~~|~1.62~~~~~~~~~~~~~|~0.68~~~~~~~~~~~~~~~|~15.0~~~~~~~|~3/12/0}@*)

(*) Soft Score: winner receives +confidence, loser receives +(1-confidence), summed over matches.
(**) S/F/U: SUCCESS/FAIL/UNCERTAIN labels aggregated from visual rollouts.

(*@\textbf{\sffamily --- ELITE POOL SELECTION (Top 30\% within Delta=1) ---}@*)
Qualified Candidates: {7, 1}
Total Qualified: 2

(*@\textbf{\sffamily\footnotesize Environment 3 POMDP -- Hallucinated-SUCCESS Robustness}@*)
========================================================================================================
 (*@\textbf{\sffamily TOURNAMENT QUALITATIVE DIGEST: CHECKPOINT 0 (ENVIRONMENT STEPS: 1000000)}@*)
========================================================================================================

(*@\textbf{\sffamily Candidate: k2}@*)
Behavioral signature: Mixed profile: in many matches it approaches key/chest areas and often terminates with a chest turning red (mimic-like), but it sometimes shows more structured navigation toward the door/right side. (*@\textcolor{red}{In one matchup it reliably reaches a red/open chest terminal state labeled as success.}@*)
Progress signals: Demonstrates door-side navigation and key-region engagement in some episodes; (*@\textcolor{red}{strongest signal is a full set of labeled SUCCESS episodes reaching a red/open chest state (task interpretation uncertain across matches).}@*)
Failure mode hypothesis: (A) Policy-driven--frequent early endings after chest interaction (red/mimic terminal), suggesting poor trap avoidance; environment-driven constraints not repeatedly evidenced.
Evidence: R1-M1, R2-M4, R3-M5, R4-M4

...

(*@\textbf{\sffamily Candidate: k6}@*)
Behavioral signature: Frequently converges on nearby chests and terminates quickly with a red "mimic/trap" chest visible; multiple matches show extremely short episodes (step ~4). (*@\textcolor{red}{However, it also shows one matchup where it consistently reaches a red/open chest terminal state labeled as success.}@*)
Progress signals: (*@\textcolor{red}{Only clear progress signal is the labeled 3/3 SUCCESS set reaching the red/open chest terminal state (generalization uncertain given many other mimic-like failures).}@*)
Failure mode hypothesis: (A) Policy-driven--dominant pattern is brittle, short-horizon behavior ending after chest-trigger events; no repeated evidence that the environment itself prevents progress.
Evidence: R1-M4, R2-M3, R3-M5, R4-M5, R5-M4

...

(*@\textbf{\sffamily Cross-candidate semantic consistency evidence}@*)
Candidate k7: Recurrent red-mimic chest terminal states suggest a repeated mistake in chest interaction/avoidance rather than environmental impossibility.
Candidate k8: Highly repeatable immediate failures tied to chest-to-red mimic transitions suggest trap-chest interaction sensitivity.
Candidate k10: Repeated early terminations coincident with chest turning red (mimic-like) suggest a persistent hazardous interaction.
\end{lstlisting}

\begin{lstlisting}[
  basicstyle=\ttfamily\tiny,
  frame=single,
  columns=fullflexible,
  breaklines=true,
  keepspaces=true,
  showstringspaces=false,
  aboveskip=0.6em,
  belowskip=0.2em,
  escapeinside={(*@}{@*)},
  caption={\textbf{Environment~3 FO: quantitative and qualitative robustness digest at \(1\mathrm{M}\) steps.}
  The fully observable variant provides a second example of hallucinated or ambiguous \texttt{SUCCESS} labels under synchronised ground-truth failure. Candidate \texttt{k9} receives a raw \texttt{SUCCESS} label for a left-chest interaction that the qualitative digest describes as inconsistently interpreted, yet it is ranked fifth and excluded from the elite pool. Candidate \texttt{k8} is selected as an elite despite zero raw successes, showing that WATERFALL uses relative behavioural evidence rather than raw \texttt{SUCCESS} counts alone. Highlighted lines mark the relevant evidence.},
  label={lst:env3_fo_robustness_digest_1m},
  captionpos=b
]
(*@\textbf{\sffamily\footnotesize Environment 3 FO -- Medusa's Revenge: Quantitative Vision-Judge Digest}@*)
========================================================================================================
 (*@\textbf{\sffamily TOURNAMENT DIGEST:}@*) CHECKPOINT 0 (ENVIRONMENT STEPS: 1000000)
 (*@\textbf{\sffamily Mode}@*): SwissSystem | (*@\textbf{\sffamily Rounds}@*): 5 | (*@\textbf{\sffamily Candidates}@*): 10 | (*@\textbf{\sffamily Bracket Randomisation}@*) = True
========================================================================================================

(*@\textbf{\sffamily --- LEADERBOARD (HIGHER SCORE IS BETTER) ---}@*)
(Sorted by: Score Descending > Buchholz Descending)
Rank  | Candidate  | Score    | Soft Score(*)    | Avg. Confidence    | Buchholz   | Total S/F/U(**)
-------------------------------------------------------------------------------------------------------------------
1     | k4         | 5.0      | 3.59             | 0.72               | 13.0       | 2/13/0
(*@\textcolor{red}{\bfseries\ttfamily 2~~~~~|~k8~~~~~~~~~|~4.0~~~~~~|~3.16~~~~~~~~~~~~~|~0.72~~~~~~~~~~~~~~~|~15.0~~~~~~~|~0/15/0}@*)
3     | k1         | 3.0      | 2.57             | 0.67               | 15.0       | 0/15/0
4     | k3         | 3.0      | 2.82             | 0.75               | 15.0       | 0/15/0
(*@\textcolor{red}{\bfseries\ttfamily 5~~~~~|~k9~~~~~~~~~|~2.0~~~~~~|~2.30~~~~~~~~~~~~~|~0.66~~~~~~~~~~~~~~~|~14.0~~~~~~~|~1/13/1}@*)
6     | k5         | 2.0      | 2.25             | 0.68               | 12.0       | 0/15/0
7     | k2         | 2.0      | 2.33             | 0.72               | 12.0       | 0/15/0
8     | k6         | 2.0      | 2.18             | 0.66               | 11.0       | 0/15/0
9     | k7         | 2.0      | 1.51             | 0.69               | 9.0        | 0/12/0
10    | k10        | 1.0      | 1.29             | 0.68               | 11.0       | 0/12/0

(*) Soft Score: winner receives +confidence, loser receives +(1-confidence), summed over matches.
(**) S/F/U: SUCCESS/FAIL/UNCERTAIN labels aggregated from visual rollouts.

(*@\textbf{\sffamily --- ELITE POOL SELECTION (Top 30\% within Delta=1) ---}@*)
Qualified Candidates: {4, 8}
Total Qualified: 2

(*@\textbf{\sffamily\footnotesize Environment 3 FO -- Hallucinated-SUCCESS Robustness}@*)
========================================================================================================
 (*@\textbf{\sffamily TOURNAMENT QUALITATIVE DIGEST: CHECKPOINT 0 (ENVIRONMENT STEPS: 1000000)}@*)
========================================================================================================

(*@\textbf{\sffamily Candidate: k4}@*)
Behavioral signature: Consistently among the most goal-directed: repeatedly reaches the central door and pushes into right-side corridors/laser zones near the guarded area. (*@\textcolor{red}{Also shows occasional immediate termination after a left-side chest "opens"/changes state (treated as SUCCESS in multiple matches).}@*)
Progress signals: Clear chest-opening successes plus repeated penetration into the right-side hazard/goal region (strongest proximity signals among candidates shown).
Failure mode hypothesis: (A) Policy behavior--recurrent laser hazard contact on right-side attempts, with occasional left-side mimic/trap interaction (uncertain).
Evidence: R1-M2, R2-M3, R3-M1, R4-M1, R5-M1

...

(*@\textbf{\sffamily Candidate: k8}@*)
Behavioral signature: Repeatedly navigates through the central door into the right-side hazard region and spends extended time in the upper-right/laser-adjacent areas.
Progress signals: (*@\textcolor{red}{Consistent right-side access and at least one apparent key pickup signal (key sprite disappears) indicate strong proximity to the core task sequence.}@*)
Failure mode hypothesis: (A) Policy behavior--hazardous pathing into active lasers and incomplete final objective interaction despite reaching the correct region (uncertain).
Evidence: R1-M1, R2-M2, R3-M1, R4-M3, R5-M2

...

(*@\textbf{\sffamily Candidate: k9}@*)
Behavioral signature: Generally stable but indecisive: often wanders in left/middle and near the central corridor until timeout, with limited right-side penetration. (*@\textcolor{red}{Occasionally triggers a left-chest interaction event (sometimes labeled SUCCESS/opened red chest; other times mimic-like terminal with disappearance).}@*)
Progress signals: Shows at least one clear chest-opening success and other chest interaction signals, suggesting some ability to manipulate task objects (uncertain link to main objective).
Failure mode hypothesis: (A) Policy behavior--repeated timeouts from insufficient right-side/goal-directed progression, with intermittent unsafe mimic/trap interaction.
Evidence: R1-M3, R2-M4, R3-M3, R4-M1, R5-M3

...

(*@\textbf{\sffamily Cross-candidate semantic consistency evidence}@*)
Candidate k6: Repeatedly ends episodes shortly after the left chest turns into a red "mimic/trap"-like state, often with very short horizons.
Candidate k7: Frequently beelines to or starts near the left chest and then terminates almost immediately with the chest turning red "mimic"-like and the agent disappearing.
Candidate k10: Consistently terminates very quickly, almost always coinciding with a left-side chest turning into a red "mimic"-like state and the agent disappearing.
\end{lstlisting}

Together, \autoref{lst:env1_po_robustness_digest_1m}, \autoref{lst:env2_po_robustness_digest_1m}, \autoref{lst:env3_po_robustness_digest_1m}, and \autoref{lst:env3_fo_robustness_digest_1m} show three distinct robustness behaviours. \emph{Hallucinated-\texttt{SUCCESS} robustness} is clearest in \textsc{Environment~1} and \textsc{Environment~3}: candidates receiving false-positive VLM \texttt{SUCCESS} labels are either not selected or are contextualised by qualitative evidence rather than accepted as solved. \emph{Ambiguity robustness} is clearest in \textsc{Environment~2}: all candidates are still ground-truth failures, but the tournament preserves candidates with the strongest relative progress and explicitly records uncertain terminal interpretations. Across all cases, raw episode-level VLM labels are treated as noisy supporting evidence rather than decisive supervision, supporting the role of Swiss aggregation and qualitative digests as robustness and audit mechanisms for visual evaluation.

\section{WATERFALL's Architectural Prompt System}
\label{app:prompts}
In this section, we delineate the complete prompt pipeline employed by WATERFALL, encompassing the persona injection dictionaries, evolutionary processes, and stage-specific instructions. We emphasise that these prompts are designed for \textbf{\emph{high generalisability}}, remaining agnostic to specific environment dynamics. Crucially, and unlike prior work such as Eureka, WATERFALL operates under strict information constraints; at no point is the system exposed to: (i) goal-conditioned prompting, (ii) externally supplied expert reward programs, or (iii) ground-truth reward signals or scalar metrics.

\subsection{Pre-System Initialisation: Persona Dictionary}
\label{app:personas}
To mitigate mode collapse in the initial population, we precede the evolutionary loop with a dictionary initialisation phase. We curate a set of distinct behavioural personas (see Listing~\ref{lst:persona_prompt}) that act as inductive biases, $p_k$. These priors are subsequently injected into the Stage I synthesis prompt to ensure that the initial pool of synthesised reward candidates (default $K=10$) exhibits sufficient strategic diversity before the first round of training.

\begin{lstlisting}[
  basicstyle=\ttfamily\scriptsize,
  frame=single,
  columns=fullflexible,
  breaklines=true,
  keepspaces=true,
  showstringspaces=false,
  aboveskip=0.6em,
  belowskip=0.2em,
  escapeinside={(*@}{@*)},
  caption={Pre-Stage I: Persona Injections Fed to the Training Actor},
  label={lst:persona_prompt},
  captionpos=b
]
(*@\textbf{\underline{\sffamily Pre-System Stage Dictionary of Persona Injections}}@*)

(*@\textbf{\sffamily Conservative Engineer}@*): Prefer dense, reliable shaping rewards that guide the agent incrementally. Avoid risky sparse rewards. Prioritize safety and penalty avoidance over speed.
(*@\textbf{\sffamily Risk-Taker}@*): Prefer sparse rewards and intrinsic curiosity. Ignore small penalties. Encourage the agent to explore the edges of the state space even if it means failing initially.
(*@\textbf{\sffamily Minimalist}@*): Use the smallest possible code footprint. Rely on a single, powerful terminal reward. Avoid complex shaping functions or auxiliary variables. Keep logic simple and lean.
(*@\textbf{\sffamily Mathematician}@*): Use potential-based reward shaping (difference in potential between states). Focus on geometric distance and continuous reward signals derived from strict formulas.
(*@\textbf{\sffamily Biologist}@*): Treat energy as a resource. Implement rewards that mimic homeostasis or survival instincts. Penalize "movement waste" and reward efficiency and resource conservation.
(*@\textbf{\sffamily Industrialist}@*): Focus purely on throughput and speed. heavily penalize every time step to force the fastest possible completion. Ignore aesthetic or exploration metrics; only results matter.
(*@\textbf{\sffamily Educator}@*): Implement a curriculum-style reward. Heavily reward the completion of sub-goals (like picking up a key) even more than the final goal initially, to "teach" the mechanics step-by-step.
(*@\textbf{\sffamily Strategist}@*): Prioritize long-term planning. Use rewards that encourage setting up future success (like positioning) rather than immediate gratification. Penalize short-sighted greedy actions.
(*@\textbf{\sffamily Optimist}@*): Use only positive reinforcement. Do not use penalties (negative rewards). If the agent makes a mistake, give 0 reward. If it succeeds, give massive rewards. Focus on "promotion" signals.
(*@\textbf{\sffamily Pessimist}@*): Focus on "prevention". Heavily penalize prohibited states (collisions, wrong turns). Define the task primarily by what the agent must NOT do, sculpting behavior via constraints.
\end{lstlisting}

\subsection{Stage I: Reward Synthesis with Candidate Persona Injections}
Following dictionary initialisation, Stage~I generates $K$ reward candidates. For candidate $k$, the \emph{Training Actor} (Listing~\ref{lst:training_prompt}) samples $P_{k,0}\sim p_{\mathrm{VLM}}(\cdot\mid\mathcal{E},p_k)$ using the permitted environment interface and assigned persona. A deterministic format gate $G$ checks candidate IDs, brace balance, required imports, and the expected base class. Candidates passing this gate undergo separate model-based review by the \emph{Training Judge} $J$ (Listing~\ref{lst:judge_prompt}), which assesses compliance with the code contract, including preservation of the observation/action interface and opaque environment calls. Rejected outputs return to the Actor for repair and are checked again.

\begin{lstlisting}[
  basicstyle=\ttfamily\scriptsize,
  frame=single,
  columns=fullflexible,
  breaklines=true,
  keepspaces=true,
  showstringspaces=false,
  aboveskip=0.6em,
  belowskip=0.2em,
  escapeinside={(*@}{@*)},
  caption={Stage I System Prompt: Training Actor Instructions},
  label={lst:training_prompt},
  captionpos=b
]
(*@\textbf{\underline{\sffamily Stage I: Actor's Training Prompt}}@*)

There is an environment attached as an image for context. You need to solve this training environment with Reinforcement Learning (RL).

(*@\textbf{\sffamily Your Persona:}@*)
(*@\textit{\{\{DESIGN\_PERSONA\}\}}@*). You must strictly adhere to this persona.

(*@\textbf{\sffamily Your Task:}@*)
1. Briefly explain what you think is happening.
2. Provide the full, I underline (*@\underline{FULL}@*), code with a dense reward function for RL, that is also tuned by default, which will allow the agent to reach the goal. Assume the hidden functions are there, so these should be called but not declared. Environment complexity may increase indefinitely. The code needs to be fully functional, compilable and solve the environment without the omission of any functions that may be necessary nor coding comments insinuating how to implement functionalities. For this run we need: 
(*@\textit{\{\{TRAINING\_CANDIDATE\_IDS\}\}}@*)
3. At the very end, summarise this entire conversation providing as much context as possible of what I asked you and what you did, all in less than 200 words as such: [StartSummary] TextHere [EndSummary]

[Start] Candidate Rejection History [Start]
(*@\textit{\{\{REJECTION\_HISTORY\_DIGEST\}\}}@*)
[End] Candidate Rejection History [End]

(*@\textbf{\sffamily Additional Notes:}@*)
We are using a PyTorch-based toolkit called ML-Agents 0.30.0. You are given a code block containing necessary variables, including states and actions. You are not allowed to change the states or action logic. There are hidden functions which are not revealed to you, you are to provide full code WITHOUT declaring those hidden functions (nor with any placeholders or stubs) but calling them in their right place as per the template, but do assume they exist. You are also not to change any existing environment resets. Heuristics are not allowed and you should add a termination logic based on your reasoning by only calling RequestEndEpisode(). You must define the following at the top of your code: #define WATERFALL.

[RL Template and Image of Environment for Reference Appended Below]
\end{lstlisting}

\begin{lstlisting}[
  basicstyle=\ttfamily\scriptsize,
  frame=single,
  columns=fullflexible,
  breaklines=true,
  keepspaces=true,
  showstringspaces=false,
  aboveskip=0.6em,
  belowskip=0.2em,
  escapeinside={(*@}{@*)},
  caption={Stage I System Prompt: Training Code Judge},
  label={lst:judge_prompt},
  captionpos=b
]
(*@\textbf{\underline{\sffamily Stage I: Judge's Training Prompt}}@*)

You are the JUDGE for an RL code-generation pipeline.

(*@\textbf{\sffamily Context:}@*)
- The ACTOR produced the following reward candidate(-s)
- The ACTOR's code should have full functionality and entail a dense reward to solve said environment
- The ACTOR's called but did not declare any of the hidden functions as its instructions were: 'Assume the hidden functions are there, so these should be called but not declared.'
- The states and actions logic of the code are fixed by the base environment and MUST NOT have been changed.
- The ACTOR must exclusively call the hidden function RequestEndEpisode() for termination logic

(*@\textbf{\sffamily Objectives}@*)
The code(-s) of the ACTOR are attached below. 
Your role is to advise whether any violations have taken place based on these objectives:
* Verify if the attached candidate(-s) are present
* No changes in the states or actions of the code
* Code must be fully functional and compilable (with the exception of declaring hidden functions which is why you are given the template to check)
* No omission of any core functions or variables necessary to run it
* No placeholders, no pseudocode nor any "{...}", but full functionality 
* No heuristics
* Code must function independently without any further libraries
* Termination logic must exclusively call the hidden function RequestEndEpisode()
* The top of the code file must include: #define WATERFALL.
* The ACTOR is only allowed to call hidden functions that are present in the RLEnvironment template

If you see this prompt, a deterministic check has deemed that the bare minimums have been fulfilled - now verify all objectives.

(*@\textbf{\sffamily Additional Specific Notes:}@*)
We are using the PyTorch-based toolkit called ML-Agents 0.30.0 in Unity. For the generated C# codes, both 'public class' and 'public partial class' are allowed.

(*@\textbf{\sffamily Your Task:}@*)
- For each candidate ID listed in the attached files below, respond with your decision on a new line in this exact format:
CANDIDATE_K = ACCEPT
CANDIDATE_K = DECLINE (Explain why in a brief, single-line comment)
- You MUST provide a decision for every candidate.

(*@\textit{\{\{REJECTION\_HISTORY\_NOTICE\_START\}\}}@*)
IMPORTANT: You previously rejected these candidates, which have now been updated by the actor as per your recommendation. The rejection history digest is attached. If these are now fixed and there are no other issues, ACCEPT the candidate(-s)
(*@\textit{\{\{REJECTION\_HISTORY\_NOTICE\_END\}\}}@*)

[RL Template of Environment for Reference Appended Below]
[K-Code Candidates from the ACTOR Appended Below]
\end{lstlisting}

\subsection{Stage II: Evolutionary Tournaments via Iterative Visual Rollout Assessments}
At every assessment checkpoint $t$, we freeze the policies $\pi_{\theta_k}$ and collect visual rollout evidence $\mathcal{R}_{k,t}$ for all active candidates. To bypass the scalability limits of context windows, we employ a \emph{Vision-Based Pairwise Judge} (see Listing~\ref{lst:tournament_judge}) within a Swiss-System tournament. This judge takes two sets of rollouts $(\mathcal{R}_{i,t}, \mathcal{R}_{j,t})$ and returns a winner $w_{i,j,t}$ based purely on visual progress towards the goal, without access to scalar reward curves. To facilitate longitudinal reasoning in the subsequent stage, we simultaneously invoke a \emph{Tournament Summariser} (see Listing~\ref{lst:tournament_summary}). This actor aggregates the pairwise outcomes and qualitative observations into a concise per-candidate text digest, capturing behavioural signatures and potential failure modes (e.g., ``dithering'' or ``premature termination'') observed across the tournament rounds.

\begin{lstlisting}[
  basicstyle=\ttfamily\scriptsize,
  frame=single,
  columns=fullflexible,
  breaklines=true,
  keepspaces=true,
  showstringspaces=false,
  aboveskip=0.6em,
  belowskip=0.2em,
  escapeinside={(*@}{@*)},
  caption={Stage II System Prompt: Evolutionary Tournament Judge},
  label={lst:tournament_judge},
  captionpos=b
]
(*@\textbf{\underline{\sffamily Stage II: Judge's Evolutionary Tournament Prompt}}@*)

You are an evolutionary judge evaluating a portfolio of RL agents.

(*@\textbf{\sffamily Your Task:}@*)
1. *Identify the Elite*: You are given rollout images from a total of (*@\textit{\{\{NUMBER\_EPISODE\_RECORDINGS\}\}}@*) episodes from the following agents: (*@\textit{\{\{ACTIVE\_CANDIDATE\_IDS\}\}}@*). Purely based on the visual information, identify which candidate agent is showing the most promising behaviour of having achieved or in the process of achieving the goal objective of the environment. You must choose one clear winner.
2. *Describe the Rollouts*: Describe concisely what you observe in the rollout images for every episode out of the total (*@\textit{\{\{NUMBER\_EPISODE\_RECORDINGS\}\}}@*) for each (*@\textit{\{\{ACTIVE\_CANDIDATE\_IDS\}\}}@*) candidate. Do not invent any details, everything must be strictly supported by the rollout images acting as evidence, if you are uncertain of the behaviour mention so.
3. *Quantify the Success*: From the rollouts images of every single episode out of the total (*@\textit{\{\{NUMBER\_EPISODE\_RECORDINGS\}\}}@*) episodes, for each of the agents in question ((*@\textit{\{\{ACTIVE\_CANDIDATE\_IDS\}\}}@*)), quantify how many successes there are. Successes should be inferred purely based on the visual information provided on whether the goal objective is attained. For each episode, label the outcome as: SUCCESS / FAIL / UNCERTAIN. Count SUCCESS only when the goal is unambiguously attained visually. Use UNCERTAIN only when the goal state cannot be confirmed from images; otherwise choose SUCCESS or FAIL.

(*@\textbf{\sffamily Decision Rule:}@*)
1. Do not choose the winner based on SUCCESS counts alone. Episode randomisations may differ in difficulty. Use SUCCESS/FAIL/UNCERTAIN counts as supporting evidence, however balance it with candidate behaviour showing stronger overall competence and robustness across episodes.
2. If evidence is mixed or close to a tie between (*@\textit{\{\{ACTIVE\_CANDIDATE\_IDS\}\}}@*), choose the candidate that shows stronger progress signals toward the goal objective (e.g., consistent approach, visually observable changes that expand previously occluded states, exploiting newly revealed affordances in the observable state-space of the environment, spends more frames near goal-relevant objectives, correct interaction sequence, reduces dithering/backtracking), and state these signals explicitly.
3. Do not infer intent from agent code or training configuration; judge from images only
4. Provide a confidence score of your decision

(*@\textbf{\sffamily Output Format Request (Must Follow):}@*)
1. Current checkpoint N = {n} (*@\textit{\{\{EPISODES\_OR\_STEPS\}\}}@*)
2. Briefly describe the environment you observe <concise description, less than 50w>
3. Compared candidates: (*@\textit{\{\{ACTIVE\_CANDIDATE\_IDS\}\}}@*)
4. Winner: CANDIDATE_<ID> | Confidence: <0.0-1.0>
   * You must follow this exact syntax: Winner: CANDIDATE_X | Confidence: Y
   * No other special characters are allowed for this particular response
5. Episode-by-episode rollout description (per candidate):
   * CANDIDATE_<ID>
   * Episode i: <SUCCESS|FAIL|UNCERTAIN> -- <concise description, less than 25w>
   * Episode i+1: <SUCCESS|FAIL|UNCERTAIN> -- <concise description, less than 25w>
   * ... up to Episode (*@\textit{\{\{NUMBER\_EPISODE\_RECORDINGS\}\}}@*)
6. Success counts:
   * CANDIDATE_<ID>: SUCCESS = X / (*@\textit{\{\{NUMBER\_EPISODE\_RECORDINGS\}\}}@*), FAIL = Y / (*@\textit{\{\{NUMBER\_EPISODE\_RECORDINGS\}\}}@*), UNCERTAIN = Z / (*@\textit{\{\{NUMBER\_EPISODE\_RECORDINGS\}\}}@*)
7. Short summary why the particular winner is chosen and why the other(-s) are not. Evidence of this is required by referencing either the success/failure counts or explicitly referencing specific episode rollouts regarding learning progress signals. For each candidate and their failure cases, briefly provide a short diagnosis (from the visuals) whether failures appear driven by agent behaviour or by environmental/episode-level constraints suggested by the rollouts that could prevent task completion, and state your uncertainty; only attribute constraints when supported by clear, repeated visual patterns across episodes.

(*@\textbf{\sffamily Additional Notes:}@*)
1. Your response should not exceed 500 words
2. You may not make any other decisions except for deciding the best performing candidate

(*@\textit{\{\{FURTHER\_EVOLUTIONARY\_CONTEXT\_APPEARS\_HERE\_IF\_ANY\}\}}@*)

[Rollouts as Images From a Total of (*@\textit{\{\{NUMBER\_EPISODE\_RECORDINGS\}\}}@*) Episodes for (*@\textit{\{\{ACTIVE\_CANDIDATE\_IDS\}\}}@*) Appended Below]
\end{lstlisting}

\begin{lstlisting}[
  basicstyle=\ttfamily\scriptsize,
  frame=single,
  columns=fullflexible,
  breaklines=true,
  keepspaces=true,
  showstringspaces=false,
  aboveskip=0.6em,
  belowskip=0.2em,
  escapeinside={(*@}{@*)},
  caption={Stage II System Prompt: Tournament Summary Actor},
  label={lst:tournament_summary},
  captionpos=b
]
(*@\textbf{\underline{\sffamily Stage II: Actor's Evolutionary Summariser Prompt}}@*)

You will receive multiple Tournament Summary text files for the current checkpoint only.
Each file describes one pairwise match (winner+confidence, per-episode SUCCESS/FAIL/UNCERTAIN, brief behaviour notes). Use ONLY what is in these files;
do not invent details.

(*@\textbf{\sffamily Task:}@*)
Produce a per-candidate qualitative digest by aggregating patterns across all provided match summaries.

(*@\textbf{\sffamily For each candidate k:}@*)
- Behavioral signature: 1--2 sentences describing typical behaviour across matches (mark uncertain if weak evidence).
- Progress signals: strongest indications of improvement or proximity to solving; if none, write 'No consistent progress signals (uncertain).'
- Failure mode hypothesis: attribute failures to (A) policy behaviour or (B) environment/episode constraints ONLY if repeatedly supported; include uncertainty.
- Evidence: cite 2--5 match IDs as R{round}-M{match}, extracted from filenames; only cite matches that include that candidate.

(*@\textbf{\sffamily Output (plain text, follow exactly as below):}@*)
====================================================
 TOURNAMENT QUALITATIVE DIGEST: CHECKPOINT (*@\textit{\{\{CHECKPOINT\_INDEX\}\}}@*) (ENVIRONMENT STEPS: (*@\textit{\{\{ENV\_STEPS\}\}}@*))
====================================================
--- PER-CANDIDATE QUALITATIVE ASSESSMENT ---

Candidate: k1
Behavioral signature: ...
Progress signals: ...
Failure mode hypothesis: ...
Evidence: R?-M?, R?-M?, ...

Candidate: k2
...

Rules:
- No leaderboard/scores/re-ranking.
- Summarise; no long quotes.
- If insufficient evidence: 'Insufficient evidence (uncertain).' Still list available Evidence matches.

--- CANDIDATE NOTATIONS ---
(*@\textit{\{\{CANDIDATE\_NOTATIONS\}\}}@*)

[Append ONLY files starting with tournament_summary_check_{n} below]
\end{lstlisting}

\subsection{Stage III: Iterative Candidate Assessments and Escalations}
In the final stage of the loop, the \emph{Assessment Actor} (see Listing~\ref{lst:assessment_actor}) functions as a meta-controller, evaluating the learning trajectory of each candidate to output a discrete decision $d_{k,t} \in \{\texttt{CONTINUE}, \texttt{ABORT}, \texttt{STOP}, \texttt{ESCALATE-A/B}\}$. This decision is conditioned on the tournament digests $\mathcal{H}_{k, \le t}$, the current training progress $p=t/T_{\max}$, and boundary visual frames. If an escalation (mutation) is triggered, the actor modifies the reward code $P_{k,t}$ to produce $P_{k,t+1}$, potentially incorporating logic from elite candidates identified in Stage II. To maintain system robustness, any mutated code is subject to a rigorous review by the \emph{Assessment Judge} (see Listing~\ref{lst:assessment_judge}). If violations are detected (e.g., heuristics or API breaches), a specialised \emph{Fixer} (see Listing~\ref{lst:assessment_fixer}) is invoked to apply minimal necessary repairs, ensuring the evolutionary cycle remains automated and self-correcting.

\begin{lstlisting}[
  basicstyle=\ttfamily\scriptsize,
  frame=single,
  columns=fullflexible,
  breaklines=true,
  keepspaces=true,
  showstringspaces=false,
  aboveskip=0.6em,
  belowskip=0.2em,
  escapeinside={(*@}{@*)},
  caption={Stage III System Prompt: Assessment Actor},
  label={lst:assessment_actor},
  captionpos=b
]
(*@\textbf{\underline{\sffamily Stage III: Actor's Iterative Assessment and Escalation Decision Prompt}}@*)

You are evaluating the behaviour of an RL agent. The agent has now been trained for a total of N = {n} (*@\textit{\{\{EPISODES\_OR\_STEPS\}\}}@*) with the learning progress being currently at (*@\textit{\{\{LEARNING\_PROCESS\_PERCENTAGE\}\}}@*)% and will terminate at N = {n_max} (*@\textit{\{\{EPISODES\_OR\_STEPS\}\}}@*). The inference results are summarised below as text-only tournament summaries (vision-only judge output).

(*@\textbf{\sffamily Your Task:}@*)
Assess whether the candidate agent (ID: (*@\textit{\{\{ACTIVE\_CANDIDATE\_IDS\}\}}@*)), has learned the task or not and briefly describe why. Critically distinguish between failures of policy convergence and failures of reward specification; also verify that the code logic and termination conditions align with the environment in question to rule out potential implementation bugs. Base your judgement on (i) the tournament summaries (vision judge output), (ii) start and terminal inference images of the agent's behaviour and (iii) the candidate code, all of which are appended below. Choose one of the following decisions for this candidate:
1. Continue Training [Has not yet learned - No code changes]
2. Abort Training [Has failed learning - Kill candidate]
3. Stop Training [Has successfully learned]
4. Escalation Strategy Option [Code implementation corrections with/without adjustments in learning]
   * (*@\textbf{\sffamily Option A:}@*) Debug Code or/and Tune Existing Reward (Adjust code logic, weights or/and terms in the current reward function)
   * (*@\textbf{\sffamily Option B:}@*) Augmented Exploration Strategy (Implement algorithmic mechanisms to drive broader state-space coverage, such as Intrinsic Motivation, Entropy Regularisation, Noise Injection, Diversity Search, Uncertainty Estimation etc)
(*@\textbf{\sffamily Important Point No.1:}@*) As part of a prior evolutionary assessment, an evolutionary recommendation is given to you as an appended file, study it carefully alongside the tournament quantitative/qualitative digests appended across checkpoints.

(*@\textbf{\sffamily Important Point No.2:}@*) Finally, you are given as visuals the first and very last (terminal) frame from the inference results of this candidate for each of the N = {n} (*@\textit{\{\{EPISODES\_OR\_STEPS\}\}}@*) number of episodes. This serves as a visual aid to observe the agent's ability to navigate the environment and expose potentially new environment dynamics that were potentially not apparent in earlier stages of the training process. Alongside the qualitative assessment given to you by the evolutionary digest and these visuals; you should proceed in making an informed decision on how to proceed.

(*@\textbf{\sffamily Important Point No.3:}@*) GER (Gated Entity Reveal) is the event-log that records previously hidden entities and states once these become visible and known as part of the environment dynamics (e.g., hidden environment states and events). GER files are appended as part of the materials. You must use GER to interpret what information becomes available and when in relation to achieving or converging to the goal objective of the environment in question.

(*@\textbf{\sffamily Your Output:}@*)
Decide an option from above. If the decision is CONTINUE/ABORT/STOP, do not output code. If ESCALATE-OPTION-A/B: output the full updated code in a single fenced code block. The code must contain (if applicable) new rewards that are appropriately tuned or/and new exploration strategies. The code must be fully functional as well as compilable in a standalone, trainer-agnostic way - however assume the hidden functions are there, so these should only be called but not declared nor with any placeholders or stubs of them either. You are not allowed to change the states or action logic. Heuristics are not allowed and you should add a termination logic based on your reasoning by only calling RequestEndEpisode(). You must define the following at the top of your code: #define WATERFALL.

The first line of your response must be exactly in this format with your decisions:
(*@\textit{\{\{ACTIVE\_CANDIDATE\_IDS\}\}}@*) = CONTINUE|ABORT|STOP|ESCALATE-OPTION-A|ESCALATE-OPTION-B

If you choose to ESCALATE by adopting the logic or rewards from the evolutionary winner ((*@\textit{\{\{FINAL\_WINNER\_CANDIDATE\_ID\}\}}@*)), you must append exactly // PARAMETERS_INHERITED_FROM_(*@\textit{\{\{FINAL\_WINNER\_CANDIDATE\_ID\}\}}@*) to the end of your decision line.

(*@\textbf{\sffamily Additional Notes:}@*)
We are using the PyTorch-based toolkit called ML-Agents 0.30.0. 
You must name the class as: (*@\textit{\{\{ACTIVE\_CANDIDATE\_IDS\}\}}@*) : Agent.
Max step at which all training will stop is set at N = {n_max} (*@\textit{\{\{EPISODES\_OR\_STEPS\}\}}@*).
Success is defined as a minimum, once a single candidate has learned the policy.

[Candidate Code, Starting and Terminal Visual Frames, Training History Digest (if any), Rejection History Digest (if any) and Tournament Summaries Appended Below]
\end{lstlisting}

\begin{lstlisting}[
  basicstyle=\ttfamily\scriptsize,
  frame=single,
  columns=fullflexible,
  breaklines=true,
  keepspaces=true,
  showstringspaces=false,
  aboveskip=0.6em,
  belowskip=0.2em,
  escapeinside={(*@}{@*)},
  caption={Stage III System Prompt: Assessment Judge},
  label={lst:assessment_judge},
  captionpos=b
]
(*@\textbf{\underline{\sffamily Stage III: Judge's Iterative Assessment and Escalation Decision Prompt}}@*)

You are the JUDGE for an RL code-generation pipeline.

(*@\textbf{\sffamily Context:}@*)
- The ACTOR produced initially the following reward candidate(-s)
- The ACTOR's code should have full functionality and entail a dense reward to solve said environment
- The ACTOR's called but did not declare any of the hidden functions as its instructions were: 'Assume the hidden functions are there, so these should be called but not declared.'
- The states and actions logic of the code are fixed by the base environment and MUST NOT have been changed
- The ACTOR must exclusively call the hidden function RequestEndEpisode() for termination logic
- The ACTOR is evaluating at set training checkpoints the inference of the model and may have made decisions on new reward strategies

(*@\textbf{\sffamily Objectives}@*)
The code(-s) of the ACTOR are attached below. 
Your role is to advise whether any violations have taken place based on these objectives:
* Verify if the attached candidate(-s) are present. The specific candidate IDs to review are attached.
* No changes in the states or actions of the code
* Strictly verify that the methods and functions from the RLTemplate are not renamed and retained in the candidates code
* Code must be fully functional and compilable
* No omission of any core functions or variables necessary to run it
* No placeholders, no pseudocode nor any "{...}", but full functionality 
* No heuristics
* Code must function independently without any further libraries
* Termination logic must exclusively call the hidden function RequestEndEpisode()
* The top of the code file must include: #define WATERFALL.
* The ACTOR is only allowed to call hidden functions that are present in the RLEnvironment template

If you see this prompt, a deterministic check has deemed that the bare minimums have been fulfilled - now verify all objectives.

(*@\textbf{\sffamily Additional Specific Notes:}@*)
We are using the PyTorch-based toolkit called ML-Agents 0.30.0 in Unity. For the generated C# codes, both 'public class' and 'public partial class' are allowed.

(*@\textbf{\sffamily Your Task:}@*)
- For each candidate ID listed in the attached files below, respond with your decision on a new line in this exact format:
CANDIDATE_K = ACCEPT
CANDIDATE_K = DECLINE (Explain why in a brief, single-line comment)

- You MUST provide a decision for every candidate.

(*@\textit{\{\{REJECTION\_HISTORY\_NOTICE\_START\}\}}@*)
IMPORTANT: You previously rejected these candidates, which have now been updated by the actor as per your recommendation. The rejection history digest is attached. If these are now fixed and there are no other issues, ACCEPT the candidate(-s)
(*@\textit{\{\{REJECTION\_HISTORY\_NOTICE\_END\}\}}@*)

[RL Template of Environment for Reference Appended Below]
[K-Code Candidates from the ACTOR Appended Below]
\end{lstlisting}

\begin{lstlisting}[
  basicstyle=\ttfamily\scriptsize,
  frame=single,
  columns=fullflexible,
  breaklines=true,
  keepspaces=true,
  showstringspaces=false,
  aboveskip=0.6em,
  belowskip=0.2em,
  escapeinside={(*@}{@*)},
  caption={Stage III System Prompt: Assessment Fixer},
  label={lst:assessment_fixer},
  captionpos=b
]
(*@\textbf{\underline{\sffamily Stage III: Fixer's Prompt}}@*)

Your task is to repair the candidate code. Fix ONLY what the JUDGE flagged (plus any errors directly caused by your fix). Keep everything else unchanged.

(*@\textbf{\sffamily Context:}@*)
- The JUDGE previously DECLINED the code for the candidate(s) below due to specific violations listed in the rejection history (.txt).
- Apply the MINIMUM necessary edits to satisfy the JUDGE while preserving the intended reward strategy and behaviour.
- The code must remain otherwise IDENTICAL (including comments, structure, variable names, and formatting). Do not refactor.

(*@\textbf{\sffamily You are given:}@*)
(i) RL template (reference contract),
(ii) The latest rejected candidate code,
(iii) The judge's decline reasons + rejection history.

(*@\textbf{\sffamily Rules:}@*)
- Do NOT re-assess the agent. Do NOT change decisions. Do NOT provide what you would consider additional fixes or improvements.
- Do NOT change states/actions/observations/termination semantics unless the rejection history explicitly instructs you to do so.
- Output ONLY the full corrected code (no explanations).

(*@\textbf{\sffamily Closing Notes:}@*)
We are using the PyTorch-based toolkit called ML-Agents 0.30.0.

[RL Template, Rejected Candidate Code, Rejection History (Reason) Appended Below]
\end{lstlisting}

\subsection{Generation of Evolutionary Summaries and Longitudinal Historical Aggregations}
Following the vision-based Swiss tournament in Stage II, WATERFALL aggregates the pairwise outcomes into a structured \emph{Tournament Digest} (Listing~\ref{lst:tournament_digest}). This document provides quantitative tournament evidence to the Assessment Actor at Stage III, recording hard scores ($S_k$), Buchholz scores ($B_k$), diagnostic soft scores ($\tilde{S}_k$), and the deterministically selected elite pool. Crucially, we append a specific \emph{Evolutionary Recommendation} block (Listing~\ref{lst:evo_recommendation}) to this digest to prompt the Assessment Actor to interpret standings as a direct instruction set for evolutionary operators—guiding decisions on whether to inherit parameters from the ``Elite'' (top-ranking candidates) or abort underperformers based on the training budget. To support longitudinal reasoning beyond a single snapshot, we maintain a persistent \emph{Assessment History Digest} (Listing~\ref{lst:assessment_history}), which logs the decision trajectory (e.g., \texttt{ESCALATE} vs.\ \texttt{CONTINUE}) for each candidate across all prior checkpoints. Finally, an aggregated \emph{Rejection History Digest} (Listing~\ref{lst:rejection_history}) preserves a trail of judge-issued declines from Stages I and III, ensuring that the repair loop acts on accumulated feedback rather than isolated errors.

\begin{lstlisting}[
  basicstyle=\ttfamily\tiny,
  frame=single,
  columns=fullflexible,
  breaklines=true,
  keepspaces=true,
  showstringspaces=false,
  aboveskip=0.6em,
  belowskip=0.2em,
  escapeinside={(*@}{@*)},
  caption={Generated Artefact: Quantitative Tournament Digest (Template)},
  label={lst:tournament_digest},
  captionpos=b
]
========================================================================================================
 (*@\textbf{\sffamily TOURNAMENT DIGEST:}@*) CHECKPOINT (*@\textit{\{\{CHECKPOINT\_ID\}\}}@*) (ENVIRONMENT STEPS: (*@\textit{\{\{STEP\_COUNT\}\}}@*))
 (*@\textbf{\sffamily Mode}@*): SwissSystem | (*@\textbf{\sffamily Rounds}@*): (*@\textit{\{\{R\}\}}@*) | (*@\textbf{\sffamily Candidates}@*): (*@\textit{\{\{K\}\}}@*) | (*@\textbf{\sffamily Bracket Randomisation}@*) = True
========================================================================================================

(*@\textbf{\sffamily --- LEADERBOARD (HIGHER SCORE IS BETTER) ---}@*)
(Sorted by: Score Descending > Buchholz Descending)
Rank  | Candidate      | Score          | Soft Score(*)    | Avg. Confidence    | Buchholz       | Total S/F/U(**)
-------------------------------------------------------------------------------------------------------------------
1     | (*@\textit{\{k\_elite\}}@*)         | (*@\textit{\{S\_max\}}@*)         | (*@\textit{\{Soft\_S\}}@*)       | (*@\textit{\{Avg\_C\}}@*)          | (*@\textit{\{B\_score\}}@*)     | (*@\textit{\{S/F/U\}}@*)
2     | (*@\textit{\{k\_2\}}@*)             | (*@\textit{\{S\_k2\}}@*)          | (*@\textit{\{Soft\_S\}}@*)       | (*@\textit{\{Avg\_C\}}@*)          | (*@\textit{\{B\_score\}}@*)     | (*@\textit{\{S/F/U\}}@*)
3     | (*@\textit{\{k\_3\}}@*)             | (*@\textit{\{S\_k3\}}@*)          | (*@\textit{\{Soft\_S\}}@*)       | (*@\textit{\{Avg\_C\}}@*)          | (*@\textit{\{B\_score\}}@*)     | (*@\textit{\{S/F/U\}}@*)
...
(*@\textit{\{K\}}@*)     | (*@\textit{\{k\_last\}}@*)          | (*@\textit{\{S\_min\}}@*)         | (*@\textit{\{Soft\_S\}}@*)       | (*@\textit{\{Avg\_C\}}@*)          | (*@\textit{\{B\_score\}}@*)     | (*@\textit{\{S/F/U\}}@*)

(*) Notation on Soft Score: Uncertainty-aware metric. Winner gets +Confidence, Loser gets +(1-Confidence). Summed over matches.
(**) Notation on S/F/U: S=Success, F=Failure, U=Uncertain (aggregated counts from visual rollouts).

(*@\textbf{\sffamily --- MATCH HISTORY (COMPARED EPISODE ROLLOUTS: 3) ---}@*)
[Round 1] [Match 1] (*@\textit{\{k\_i\}}@*) vs (*@\textit{\{k\_j\}}@*) -> Winner: (*@\textit{\{k\_winner\}}@*) (Confidence: (*@\textit{\{0.0-1.0\}}@*))
    Evidence: (*@\textit{\{k\_i\}}@*)[S/F/U] vs (*@\textit{\{k\_j\}}@*)[S/F/U]
...
[Round (*@\textit{\{R\}}@*)] [Match (*@\textit{\{M\}}@*)] ...

(*@\textbf{\sffamily --- ADDITIONAL TOURNAMENT METRICS ---}@*)
1. Scoring Rules: Win=1.0, Draw=0.5, Loss=0.0. (Draws triggered by explicit VLM verdict or parsing failure).
2. Buchholz Score: Sum of opponents' scores (B(k) = Sum(Score(o)) for o in Opponents(k)).

3. Per-Candidate Opponent Log (for verification):
   (*@\textit{\{k\_i\}}@*)    : vs [(*@\textit{\{Opponent\_1\}}@*) W/L, (*@\textit{\{Opponent\_2\}}@*) W/L, ...]
   ...

(*@\textbf{\sffamily --- CANDIDATE NOTATIONS ---}@*)
(*@\textit{\{k\_i\}}@*): (*@\textit{\{Full\_Class\_Name\_i\}}@*)
...
\end{lstlisting}

\begin{lstlisting}[
  basicstyle=\ttfamily\scriptsize,
  frame=single,
  columns=fullflexible,
  breaklines=true,
  keepspaces=true,
  showstringspaces=false,
  aboveskip=0.6em,
  belowskip=0.2em,
  escapeinside={(*@}{@*)},
  caption={Generated Artefact: Evolutionary Recommendations for Actor},
  label={lst:evo_recommendation},
  captionpos=b
]
(*@\textbf{\underline{\sffamily Evolutionary Recommendations for Actor}}@*)

A prior assessment from an evolutionary tournament judge has determined the standing of the most promising candidates in solving this task. If you wish to escalate (ESCALATE-OPTION-A/B), the evolutionary recommendation is to consult and analyse the code of the top-performing candidates and inherit their reward by copying their parameters and adapting these to the current candidate's code.

If you do wish to copy code or parameters from the top-performing candidates, you must declare them anew in the current candidate's code - cross-referencing is not permitted. Moreover, given the concept of the survival of the fittest, the evolutionary recommendation is to terminate (ABORT) the weakest candidate if it continues to struggle over the duration of the learning process to save computational resources. 

(*@\textbf{\sffamily Important}@*): You must base this primarily on the tournament summaries over the learning process (also appended); use confidence scores only as supporting evidence not as a sole decision criterion. For context, the learning progress is currently at (*@\textit{\{\{LEARNING\_PROCESS\_PERCENTAGE\}\}}@*)%, given that we are now at N = {n} (*@\textit{\{\{EPISODES\_OR\_STEPS\}\}}@*), until the maximum of at N = {n_max} (*@\textit{\{\{EPISODES\_OR\_STEPS\}\}}@*). Calibrate your decision on whether to inherit from the top-performing candidates and whether to abort the current candidate primarily based on the learning progress (currently at (*@\textit{\{\{LEARNING\_PROCESS\_PERCENTAGE\}\}}@*)%) and the history in the tournament summary(-ies); use confidence scores only as supporting evidence.
\end{lstlisting}

\begin{lstlisting}[
  basicstyle=\ttfamily\scriptsize,
  frame=single,
  columns=fixed,
  breaklines=true,
  keepspaces=true,
  showstringspaces=false,
  aboveskip=0.6em,
  belowskip=0.2em,
  escapeinside={(*@}{@*)},
  caption={Generated Artefact: Assessment Actor Decision History (Indicative Example)},
  label={lst:assessment_history},
  captionpos=b
]
(*@\textbf{\underline{\sffamily Assessment History Digest: Longitudinal Decision Log}}@*)

(*@\textbf{\sffamily Current Checkpoint:}@*) 3 | (*@\textbf{\sffamily Total Steps:}@*) 20,000,000

(*@\textbf{\sffamily --- DECISION TRACE PER CANDIDATE (K1...K10) ---}@*)
Format: N={steps} -> {Decision} | ...

K1: N=1M -> (*@\textbf{\sffamily Continue}@*)    | N=10M -> (*@\textbf{\sffamily Escalate(A)}@*) | N=20M -> (*@\textbf{\sffamily Stop}@*) (*@\textit{(Solved)}@*)
K2: N=1M -> (*@\textbf{\sffamily Escalate(B)}@*)  | N=10M -> (*@\textbf{\sffamily Continue}@*)   | N=20M -> (*@\textbf{\sffamily Escalate(A)}@*)
K3: N=1M -> (*@\textbf{\sffamily Continue}@*)    | N=10M -> (*@\textbf{\sffamily Abort}@*)     | (*@\textit{(Terminated and Killed)}@*)
K4: N=1M -> (*@\textbf{\sffamily Escalate(A)}@*)  | N=10M -> (*@\textbf{\sffamily Escalate(A)}@*) | N=20M -> (*@\textbf{\sffamily Continue}@*)
K5: N=1M -> (*@\textbf{\sffamily Escalate(B)}@*)  | N=10M -> (*@\textbf{\sffamily Escalate(B)}@*) | N=20M -> (*@\textbf{\sffamily Stop}@*) (*@\textit{(Solved)}@*)
...
K10: N=1M -> (*@\textbf{\sffamily Continue}@*)   | N=10M -> (*@\textbf{\sffamily Continue}@*)   | N=20M -> (*@\textbf{\sffamily Abort}@*)

(*@\textbf{\sffamily --- LEGEND ---}@*)
(*@\textbf{\sffamily Continue}@*)   : No code changes; policy training proceeds.
(*@\textbf{\sffamily Escalate}@*)   : Reward with potential evolutionary mutation (A=Tuning, B=Exploration).
(*@\textbf{\sffamily Stop}@*)       : Candidate solved task; frozen.
(*@\textbf{\sffamily Abort}@*)      : Candidate pruned; resources freed.
\end{lstlisting}

\begin{lstlisting}[
  basicstyle=\ttfamily\tiny,
  frame=single,
  columns=fixed,
  breaklines=true,
  keepspaces=true,
  showstringspaces=false,
  aboveskip=0.6em,
  belowskip=0.2em,
  escapeinside={(*@}{@*)},
  caption={Generated Artefact: Rejection History Digest (Indicative Example)},
  label={lst:rejection_history},
  captionpos=b
]
(*@\textbf{\underline{\sffamily Rejection History Digest}}@*)

(*@\textbf{\sffamily Stage:}@*) I & III (Aggregated) | (*@\textbf{\sffamily Total Candidates Processed:}@*) 10

(*@\textbf{\sffamily --- DECISION LOG PER CANDIDATE (K1...K10) ---}@*)
Format: Candidate_ID : {STATUS} (Reason if Declined)

K1 : (*@\textbf{\sffamily ACCEPT}@*)
K2 : (*@\textbf{\sffamily ACCEPT}@*)
K3 : (*@\textbf{\sffamily ACCEPT}@*)
K4 : (*@\textbf{\sffamily ACCEPT}@*)
K5 : (*@\textbf{\sffamily ACCEPT}@*)
...
K10: (*@\textbf{\sffamily DECLINE}@*) (Reason: Omits RLTemplate members 'stepsSoFar' and 'CollectObservationsShared', violating "methods/functions retained" requirement and potentially breaking compilation.)
\end{lstlisting}

\section{WATERFALL's Code Input}
This section details the code provided as input for the initial generation (Stage~I) of WATERFALL's candidates. The listings below show the POMDP variants; note that hidden-state semantics are not revealed beforehand. Each template contains only above-boundary code, exposing the policy-visible observation schema and action interface while omitting hidden environment dynamics and ground-truth task logic. Naturally, since WATERFALL and other reward-synthesis baselines require a coding interface, some variable names still pose a risk of semantic leakage. We minimise this risk as follows. First, all four environments use a generic template named \texttt{RLEnvironment}, rather than a descriptive name such as \texttt{MiniGrid\_Labyrinth}. Second, unrevealed objects are stored in a hidden-state buffer whose entries are withheld until revealed; in Env.~III, for example, the buffer emits $-1$ until the door is unlocked (\autoref{lst:env3_medusas_revenge_po_initial_template}). Third, the size of this buffer, \texttt{HIDDEN\_STATE\_SIZE}, is not disclosed. Finally, WATERFALL is not given the episode's termination step and must instead infer it from the supplied code and image.

\label{app:code_input}
\begin{lstlisting}[
  basicstyle=\ttfamily\scriptsize,
  frame=single,
  columns=fullflexible,
  breaklines=true,
  keepspaces=true,
  showstringspaces=false,
  aboveskip=0.6em,
  belowskip=0.2em,
  caption={\textbf{Initial code-generation environment template for Environment~I (\emph{MiniGrid's Locked Room Variant}, PO).} This above-boundary Unity ML-Agents template is provided to WATERFALL during initial reward-code generation. It exposes the local egocentric observation schema and action interface while omitting hidden environment dynamics and ground-truth task logic.},
  label={lst:env1_minigrid_lockedroom_po_initial_template},
  captionpos=b
]
#define WATERFALL
using System.Collections.Generic;
using UnityEngine;
using Unity.MLAgents;
using Unity.MLAgents.Actuators;
using Unity.MLAgents.Sensors;

public class RLEnvironment : Agent
{
    private int gridWidth = 13;
    private int gridHeight = 13;
    private int[,] grid;
    private int agentOrientation; // 0=North, 1=East, 2=South, 3=West
    
    // 7x7x3 = 147 ray values + 1 dir + 1 inv slot + 1 inv color = 150 total
    private readonly List<float> _cachedObs = new List<float>(150);

    public override void Initialize()
    {
        InitializeEnv();
    }

    public override void OnEpisodeBegin()
    {
        ResetEnvironment();
    }

    private void CollectObservationsInternal(List<float> outputList, bool isLoggingPass)
    {
        outputList.Clear();
        
        // 7 columns x 7 depth x 3 channels = 147 ray observations
        // Each ray cell returns 3 scalars: (type_id, color_id, state_id).
        float[] perception = PerformRayPerception(isLoggingPass);
        outputList.AddRange(perception);
        outputList.Add((float)agentOrientation / 3.0f);
        float invVal = GetInventoryObservation();
        outputList.Add(invVal);
        outputList.Add(GetInventoryColorId());

        LoggingDynamics(invVal, isLoggingPass);
    }

    public override void CollectObservations(VectorSensor sensor)
    {
        CollectObservationsInternal(_cachedObs, false);
        foreach (var val in _cachedObs) sensor.AddObservation(val);
    }

    public override void OnActionReceived(ActionBuffers actions)
    {
        int act = actions.DiscreteActions[0];
        string actName = "Stay";
        
        switch (act)
        {
            case 0: actName = "TurnLeft"; break;
            case 1: actName = "TurnRight"; break;
            case 2: actName = "Forward"; break;
            case 3: actName = "Pickup"; break;
            case 4: actName = "Drop"; break;
            case 5: actName = "Toggle"; break;
            case 6: actName = "Stay"; break;
        }
        
        AgentMove(act, actName);
    }
}
\end{lstlisting}

\begin{lstlisting}[
  basicstyle=\ttfamily\scriptsize,
  frame=single,
  columns=fullflexible,
  breaklines=true,
  keepspaces=true,
  showstringspaces=false,
  aboveskip=0.6em,
  belowskip=0.2em,
  caption={\textbf{Initial code-generation environment template for Environment~II (\emph{Dark Corridor}, PO).} This above-boundary Unity ML-Agents template is provided to WATERFALL during initial reward-code generation. It exposes the T-shaped local ray observation schema and action interface while omitting hidden environment dynamics and ground-truth task logic.},
  label={lst:env2_darkcorridor_po_initial_template},
  captionpos=b
]
#define WATERFALL

using System;
using System.Collections.Generic;
using UnityEngine;
using Unity.MLAgents;
using Unity.MLAgents.Actuators;
using Unity.MLAgents.Sensors;

public class RLEnvironment : Agent
{
    private int gridWidth = 9;
    private int gridHeight = 9;
    private int[,] grid;
    private int agentOrientation; // 0=North, 1=East, 2=South, 3=West
    
    // (1 Left + 1 Right + 4 Fwd) x 3 = 18 ray values + 1 dir + 1 inv slot + 1 inv type = 21 total
    private readonly List<float> _cachedObs = new List<float>(21);

    public override void Initialize()
    {
        InitializeEnv();
    }

    public override void OnEpisodeBegin()
    {
        ResetEnvironment();
    }

    private void CollectObservationsInternal(List<float> outputList, bool isLoggingPass)
    {
        outputList.Clear();
        
        // T-Shaped Raycast: (1 Left + 1 Right + 4 Fwd) x 3 channels = 18 scalars
        float[] perception = PerformRayPerception(isLoggingPass);
        outputList.AddRange(perception);
        outputList.Add((float)agentOrientation / 3.0f);
        
        float invVal = GetInventoryObservation();
        outputList.Add(invVal);
        outputList.Add(GetInventoryType());

        LoggingDynamics(invVal, isLoggingPass);
    }

    public override void CollectObservations(VectorSensor sensor)
    {
        CollectObservationsInternal(_cachedObs, false);
        foreach (var val in _cachedObs) sensor.AddObservation(val);
    }

    public override void OnActionReceived(ActionBuffers actions)
    {
        int act = actions.DiscreteActions[0];
        string actName = "Stay";
        
        switch (act)
        {
            case 0: actName = "TurnLeft"; break;
            case 1: actName = "TurnRight"; break;
            case 2: actName = "Forward"; break;
            case 3: actName = "Pickup"; break;
            case 4: actName = "Drop"; break;
            case 5: actName = "Toggle"; break;
            case 6: actName = "Stay"; break;
        }
        
        AgentMove(act, actName);
    }
}
\end{lstlisting}

\begin{lstlisting}[
  basicstyle=\ttfamily\scriptsize,
  frame=single,
  columns=fullflexible,
  breaklines=true,
  keepspaces=true,
  showstringspaces=false,
  aboveskip=0.6em,
  belowskip=0.2em,
  caption={\textbf{Initial code-generation environment template for Environment~III (\emph{Medusa's Revenge}, PO).} This above-boundary Unity ML-Agents template is provided to WATERFALL during initial reward-code generation. It exposes the partially observable state interface, reveal-gated hidden-state buffer, and action interface while omitting hidden environment dynamics and ground-truth task logic.},
  label={lst:env3_medusas_revenge_po_initial_template},
  captionpos=b
]
#define WATERFALL
using System;
using System.Collections.Generic;
using UnityEngine;
using Unity.MLAgents;
using Unity.MLAgents.Actuators;
using Unity.MLAgents.Sensors;
using Unity.MLAgents.Policies;

public class RLEnvironment : Agent
{
    public GameObject agentPrefab;
    public int gridS = 17;
    public GameObject closedChest;
    public GameObject key;
    public GameObject door;

    private Vector2Int closedChestPos;
    private Vector2Int agentPos;
    private int actualGridSize;
    private int[,] grid;
    private int stepsSoFar;
    private bool hasKey = false;
    private bool doorUnlocked = false;
    private Vector2Int doorPos;
    private Vector2Int keyPos;
    private float[] hiddenStateBuffer = new float[HIDDEN_STATE_SIZE];

    void Start()
    {
        InitialiseLoggers();
    }

    public override void OnEpisodeBegin()
    {
        ApplyLoggers();
        ResetEnvironment();
    }

    private void CollectObservationsShared(System.Action<float> addObs)
    {
        if (!SyncHiddenState(addObs)) return;
        addObs(agentPos.x / (float)actualGridSize);
        addObs(agentPos.y / (float)actualGridSize);
        RayPerceptor(addObs);
        addObs(hasKey ? 1 : 0);
        addObs(doorPos.x / (float)actualGridSize);
        addObs(doorPos.y / (float)actualGridSize);
        addObs(keyPos.x / (float)actualGridSize);
        addObs(keyPos.y / (float)actualGridSize);
        addObs(closedChestPos.x / (float)actualGridSize);
        addObs(closedChestPos.y / (float)actualGridSize);
        if (doorUnlocked)
            for (int i = 0; i < HIDDEN_STATE_SIZE; i++) addObs(hiddenStateBuffer[i]);
        else
            for (int i = 0; i < HIDDEN_STATE_SIZE; i++) addObs(-1.0f);
    }

    public override void CollectObservations(VectorSensor sensor)
    {
        CollectObservationsShared(val => sensor.AddObservation(val));
    }

    public override void OnActionReceived(ActionBuffers actions)
    {
        int act = actions.DiscreteActions[0];
        Vector2Int delta = Vector2Int.zero;
        switch (act)
        {
            case 0: delta = new Vector2Int(0, 1); break;
            case 1: delta = new Vector2Int(0, -1); break;
            case 2: delta = new Vector2Int(-1, 0); break;
            case 3: delta = new Vector2Int(1, 0); break;
            case 4: delta = Vector2Int.zero; break;
        }

        AttemptMove(delta, act);
    }
}
\end{lstlisting}

\begin{lstlisting}[
  basicstyle=\ttfamily\scriptsize,
  frame=single,
  columns=fullflexible,
  breaklines=true,
  keepspaces=true,
  showstringspaces=false,
  aboveskip=0.6em,
  belowskip=0.2em,
  caption={\textbf{Initial code-generation environment template for Environment~IV (Reveal-Gated Robotic Button Task, PO).} This above-boundary Unity ML-Agents template is provided to WATERFALL during initial reward-code generation. It exposes the policy-visible proprioceptive and reveal-gated observation schema and continuous Cartesian action interface while omitting hidden environment dynamics and ground-truth task logic.},
  label={lst:env4_po_template},
  captionpos=b
]
#define WATERFALL

using System;
using System.Collections;
using System.Collections.Generic;
using UnityEngine;
using Unity.MLAgents;
using Unity.MLAgents.Actuators;
using Unity.MLAgents.Policies;
using Unity.MLAgents.Sensors;

public class RLEnvironment : Agent
{
    private const int PROPRIOCEPTIVE_INFORMATION_SIZE = 13;
    private const int VISIBLE_EXTEROCEPTIVE_INFORMATION_SIZE = 3;
    private const int LIGHT_STATE_INFORMATION_SIZE = 3;
    private const int OBSERVATION_SIZE =
        PROPRIOCEPTIVE_INFORMATION_SIZE +
        VISIBLE_EXTEROCEPTIVE_INFORMATION_SIZE +
        LIGHT_STATE_INFORMATION_SIZE +
        HIDDEN_STATE_SIZE;

    private Vector3 visibleButtonPositionRelativeToOrigin;
    private float lightStateA;
    private float lightStateB;
    private float lightStateC;
    private readonly float[] hiddenStateBuffer =
        new float[HIDDEN_STATE_SIZE];

    public override void Initialize()
    {
        InitializeEnvironment();
    }

    public override void OnEpisodeBegin()
    {
        ApplyEpisodeLoggers();
        ResetEnvironment();
    }

    private void CollectObservationsShared(Action<float> addObservation)
    {
        if (!SyncObservationState(addObservation)) return;

        Vector3 linearVelocityRelativeToOrigin =
            originReference.InverseTransformVector(eeAB.velocity);
        Vector3 angularVelocityRelativeToOrigin =
            originReference.InverseTransformVector(eeAB.angularVelocity);
        Vector3 endEffectorPositionRelativeToOrigin =
            PosRelativeTo(originReference, robotEndEffector.position);
        Quaternion endEffectorRotationRelativeToOrigin =
            RotRelativeTo(originReference, robotEndEffector.rotation);

        AddVector3(addObservation,
            SafeMath.ClampFinite(linearVelocityRelativeToOrigin));
        AddVector3(addObservation,
            SafeMath.ClampFinite(angularVelocityRelativeToOrigin));
        AddVector3(addObservation,
            SafeMath.ClampFinite(endEffectorPositionRelativeToOrigin));
        AddQuaternion(addObservation,
            SafeMath.ClampFinite(endEffectorRotationRelativeToOrigin));

        AddVector3(addObservation, visibleButtonPositionRelativeToOrigin);

        addObservation(lightStateA);
        addObservation(lightStateB);
        addObservation(lightStateC);

        for (int i = 0; i < HIDDEN_STATE_SIZE; i++)
            addObservation(hiddenStateBuffer[i]);
    }

    public override void CollectObservations(VectorSensor sensor)
    {
        CollectObservationsShared(value => sensor.AddObservation(value));
        InternalCheck(sensor);
    }

    public override void OnActionReceived(ActionBuffers actions)
    {
        if (!ikCapturedAndInitialised || episodeEnded || resetRecoveryActive)
            return;

        var continuousActions = actions.ContinuousActions;

        stepCountTaken = StepCount;

        Vector3 cartesianActionRelativeToOrigin =
            Vector3.ClampMagnitude(
                new Vector3(
                    continuousActions[0],
                    continuousActions[1],
                    continuousActions[2]),
                1f);

        _lastActionLog = FormatAction(cartesianActionRelativeToOrigin);

        ApplyCartesianAction(cartesianActionRelativeToOrigin);

        if (!episodeEnded)
            LogAblationStepMaybe(_lastActionLog, false);
    }
}
\end{lstlisting}

\section{Visual Inputs and Rollout Evidence}
\label{app:visual_inputs}
\newcommand{\visionplaceholder}{%
    \fbox{\rule{0pt}{0.92\linewidth}\rule{0.92\linewidth}{0pt}}%
}

This section provides representative examples of the visual evidence supplied to WATERFALL across its three stages. Stage~I receives an initial environment-context image (\autoref{fig:initial_visual_context}) alongside the permitted code interface (Appendix~\ref{app:code_input}) and a persona-conditioned prompt (Appendix~\ref{app:personas}). Stage~II evaluates candidate behaviour from sampled visual rollout sequences, whereas Stage~III receives only the corresponding episode-boundary frames, $\{o^v_0,o^v_T\}$, together with longitudinal assessment and tournament context. Rollout interiors are sampled independently of task-specific events under $\pi_{\mathrm{sample}}(t)$, while the initial and terminal frames are always retained.

\subsection{Initial Visual Context}
\label{app:visual_inputs_initial}
\begin{figure}[ht]
    \centering

    \begin{minipage}[t]{0.23\linewidth}
        \centering
       \includegraphics[width=\linewidth]{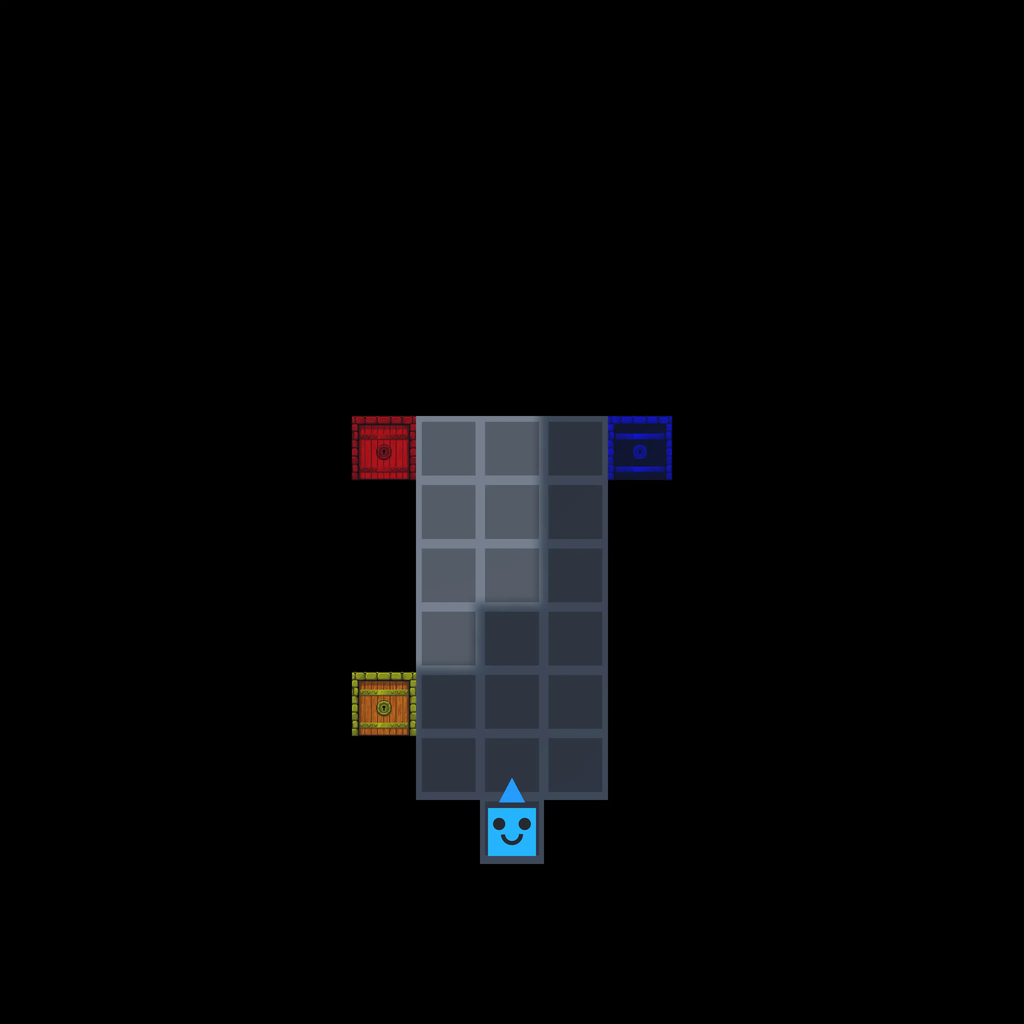}
        {\scriptsize Env.~I (PO)}
    \end{minipage}\hfill
    \begin{minipage}[t]{0.23\linewidth}
        \centering
         \includegraphics[width=\linewidth]{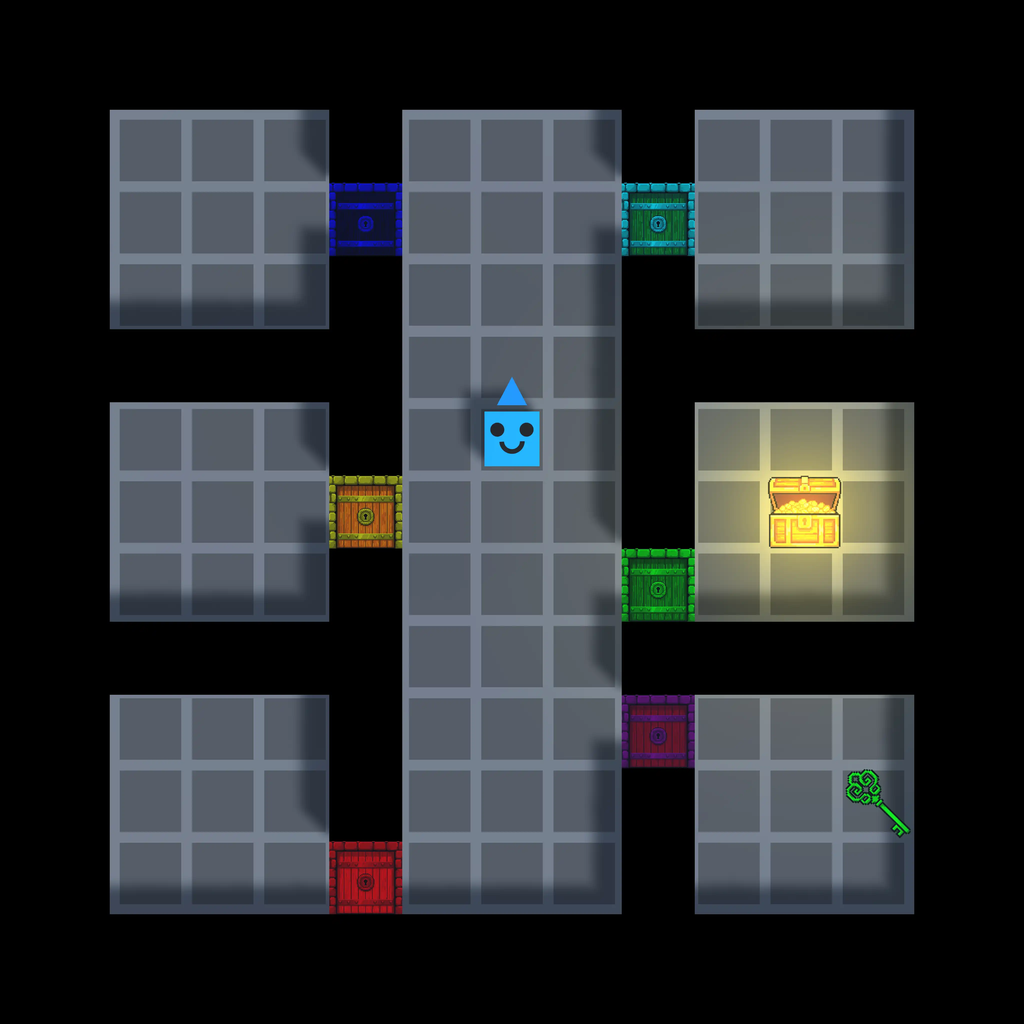}
        {\scriptsize Env.~I (FO)}
    \end{minipage}\hfill
    \begin{minipage}[t]{0.23\linewidth}
        \centering
         \includegraphics[width=\linewidth]{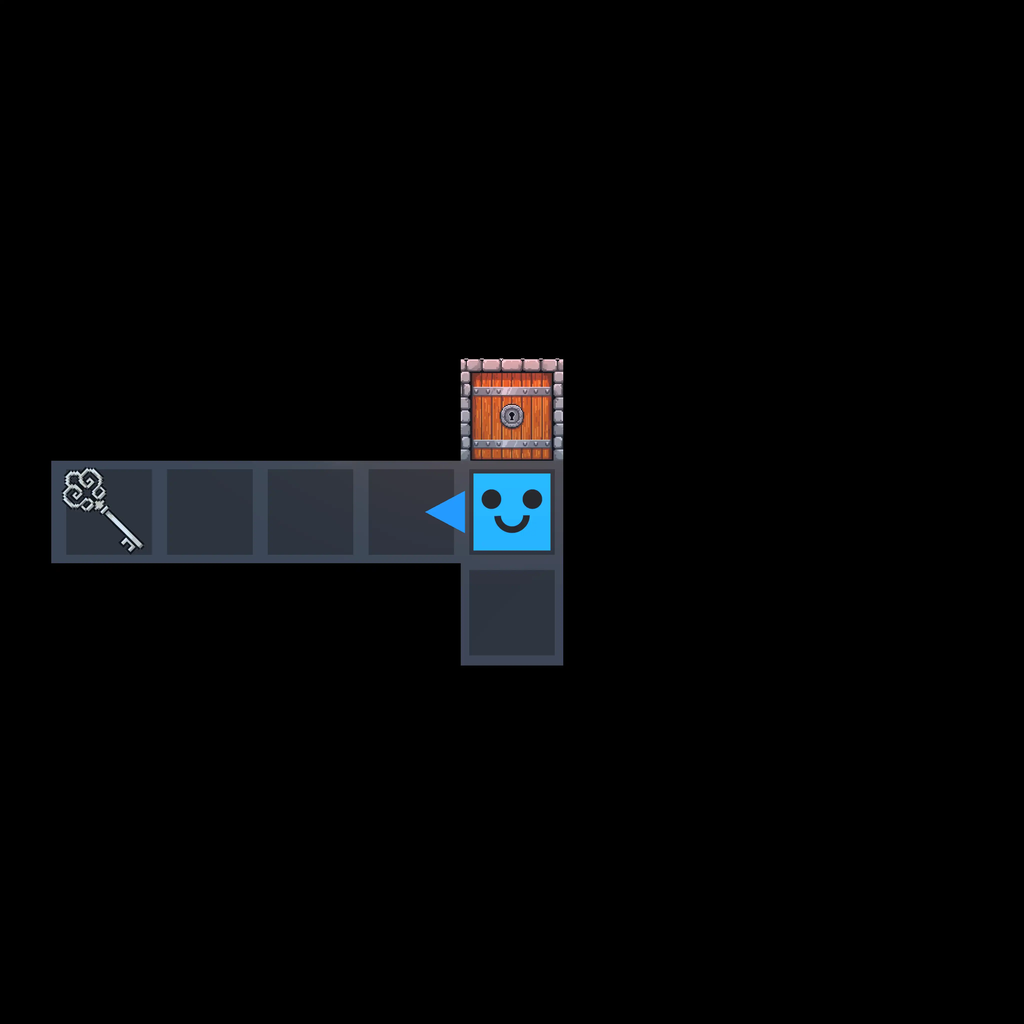}
        {\scriptsize Env.~II (PO)}
    \end{minipage}\hfill
    \begin{minipage}[t]{0.23\linewidth}
        \centering
         \includegraphics[width=\linewidth]{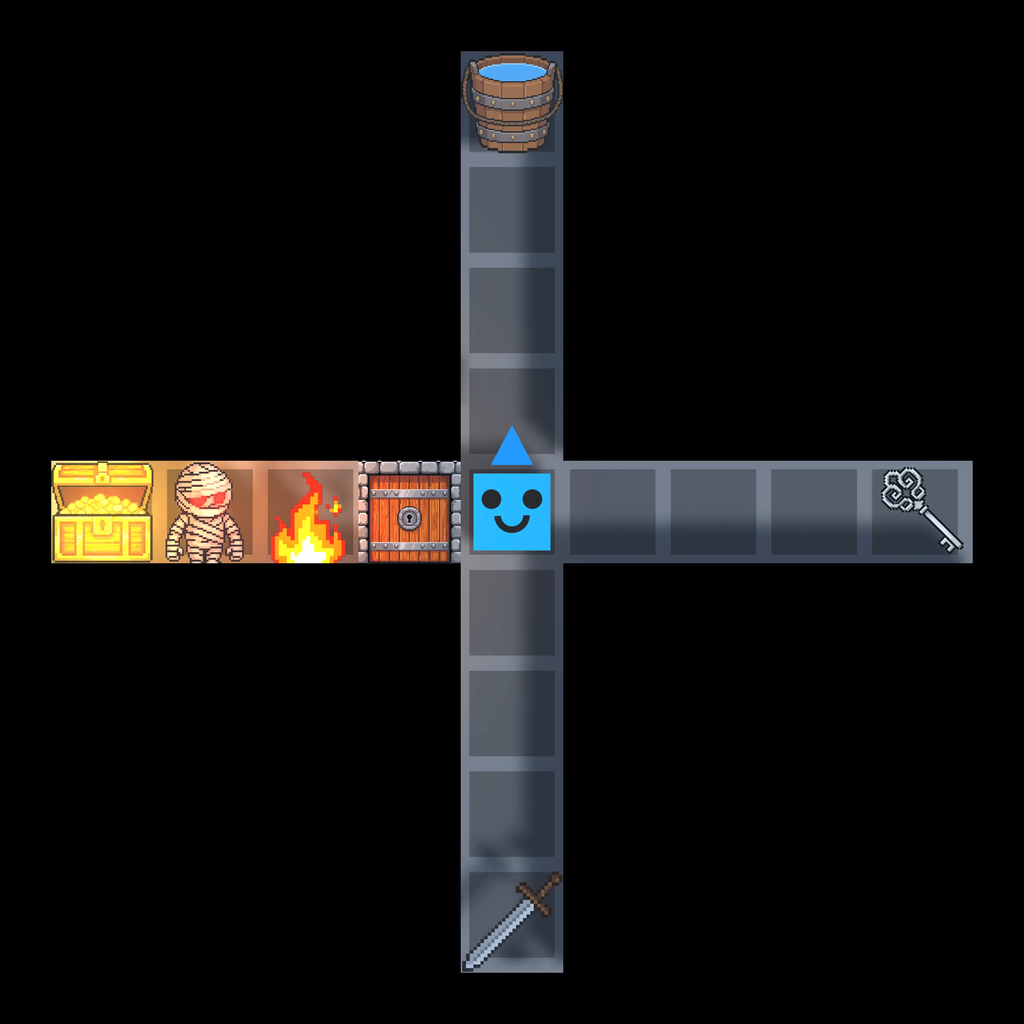}
        {\scriptsize Env.~II (FO)}
    \end{minipage}

    \vspace{0.5cm}

    \makebox[\linewidth][c]{%
        \begin{minipage}[t]{0.23\linewidth}
            \centering
            \includegraphics[width=\linewidth]{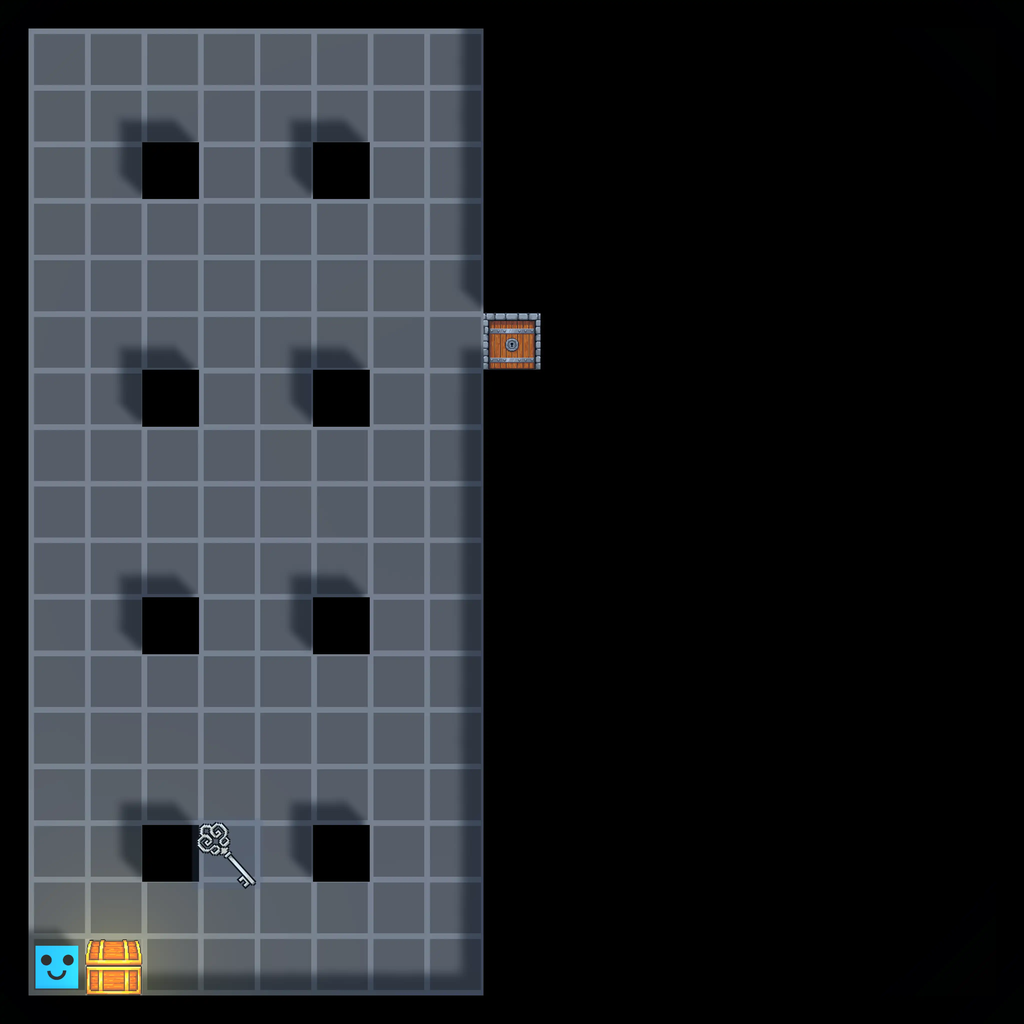}
            {\scriptsize Env.~III (PO)}
        \end{minipage}\hspace{0.02\linewidth}
        \begin{minipage}[t]{0.23\linewidth}
            \centering
            \includegraphics[width=\linewidth]{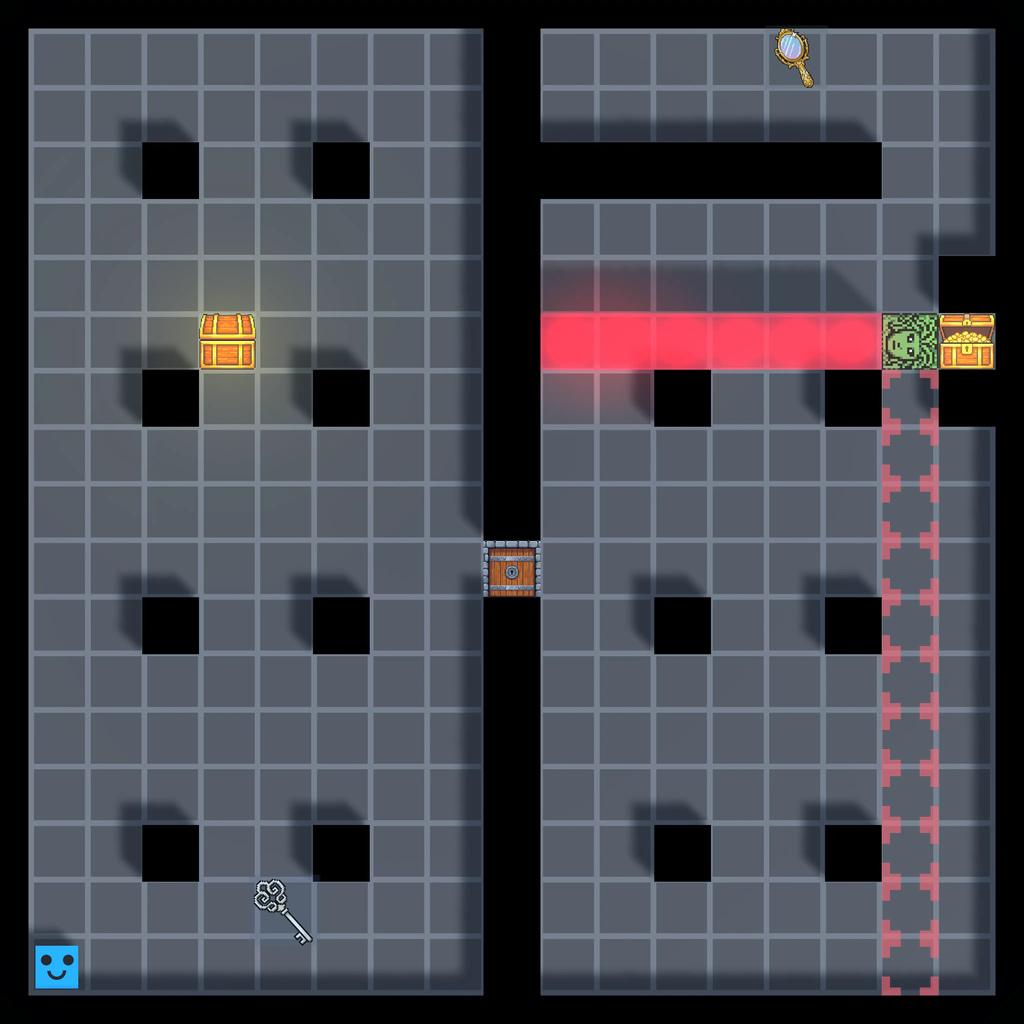}
            {\scriptsize Env.~III (FO)}
        \end{minipage}\hspace{0.02\linewidth}
        \begin{minipage}[t]{0.23\linewidth}
            \centering
           \includegraphics[width=\linewidth]{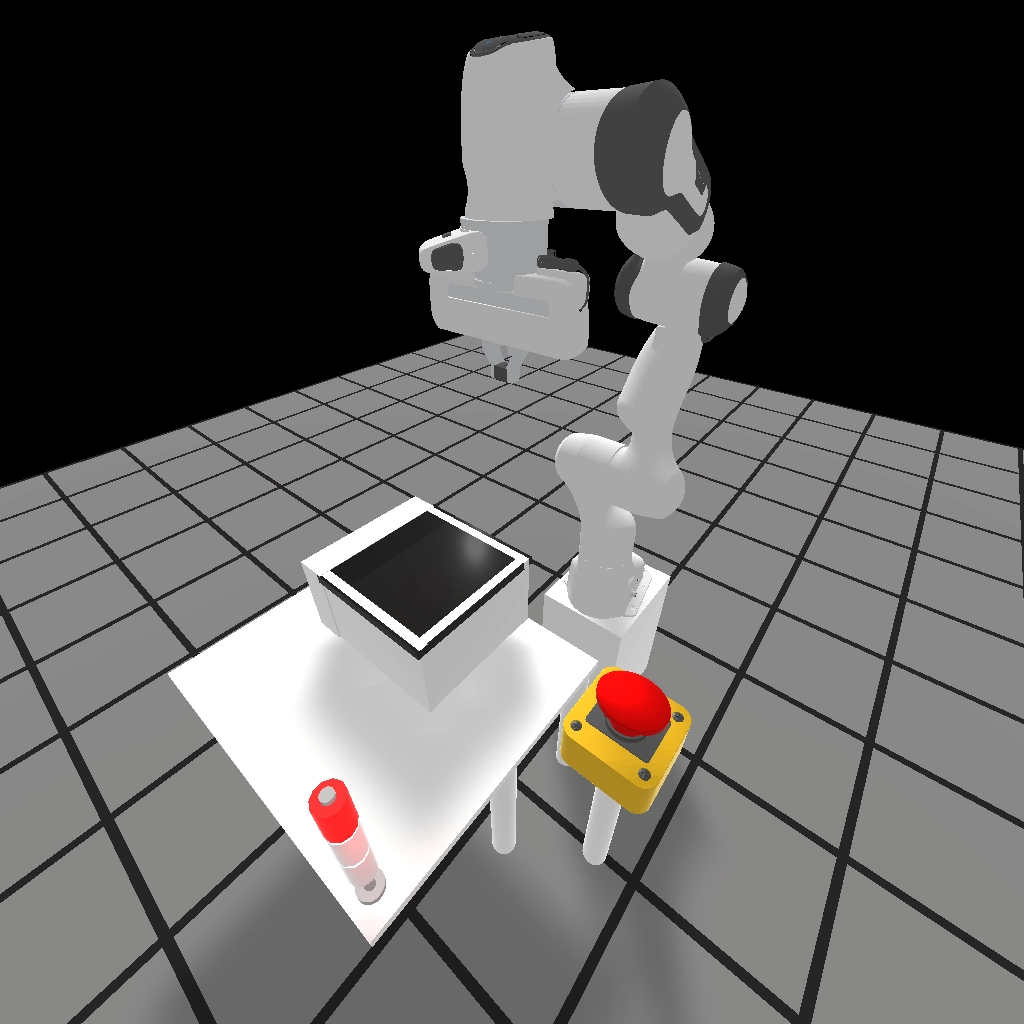}
            {\scriptsize Env.~IV (PO)}
        \end{minipage}%
    }

    \caption{\textbf{Initial visual context supplied to Stage~I across all evaluated settings.}}
    \label{fig:initial_visual_context}
\end{figure}

Stage~I receives a single environment-context image for each observability setting: a policy-visible render with no task description, goal annotation, or reward signal. \autoref{fig:initial_visual_context} groups the seven settings used in the main experiments: PO and FO variants for Env.~I--III, followed by the PO-only Env.~IV. These images complement the permitted observation--action interface supplied to the reward-synthesis actor and provide the initial visual grounding from which candidate reward programs are generated.

\subsection{Representative Successful Rollouts}
\label{app:visual_inputs_successful}

Representative successful rollouts from candidates associated with Stage~III \texttt{STOP} calls are shown for Env.~I--III in \autoref{fig:success_rollouts_env1}, \autoref{fig:success_rollouts_env2}, and \autoref{fig:success_rollouts_env3}, with PO episodes in the first three rows and FO episodes in the following three. Env.~IV PO is shown separately in \autoref{fig:success_rollouts_env4}. Each row corresponds to a distinct inference episode, with three episodes shown per setting, matching the default rollout count used for assessment for the same candidate and model. The first and final columns are the actual episode-boundary frames, while the intermediate columns are representative samples from the recorded trajectory, selected only to compactly visualise environment dynamics and distinct interaction phases. These figures therefore summarise the evidence rather than reproducing the complete Stage~II payload.

\newcommand{\successRolloutFirstImage}[2]{%
    \begin{minipage}[t]{0.242\linewidth}
        \centering
        \setlength{\unitlength}{\linewidth}
        \begin{picture}(1,1)
            \put(0,0){\includegraphics[width=\linewidth]{#1}}
            \put(0.04,0.88){%
                \outlinedwhite{Ep.~#2}%
            }
            \put(0.96,0.09){%
                \makebox(0,0)[r]{%
                    \outlinedwhite{$t_0$}%
                }%
            }
        \end{picture}
    \end{minipage}%
}

\newcommand{\successRolloutPlainImage}[2]{%
    \begin{minipage}[t]{0.242\linewidth}
        \centering
        \setlength{\unitlength}{\linewidth}
        \begin{picture}(1,1)
            \put(0,0){\includegraphics[width=\linewidth]{#1}}
            \put(0.96,0.09){%
                \makebox(0,0)[r]{%
                    \outlinedwhite{$#2$}%
                }%
            }
        \end{picture}
    \end{minipage}%
}

\newcommand{\successRolloutLastImage}[1]{%
    \begin{minipage}[t]{0.242\linewidth}
        \centering
        \setlength{\unitlength}{\linewidth}
        \begin{picture}(1,1)
            \put(0,0){\includegraphics[width=\linewidth]{#1}}
            \put(0.96,0.90){%
                \makebox(0,0)[r]{%
                    \outlinedcolor{successgreen}{Success}%
                }%
            }
            \put(0.96,0.09){%
                \makebox(0,0)[r]{%
                    \outlinedwhite{$t_3$}%
                }%
            }
        \end{picture}
    \end{minipage}%
}

\newcommand{\successRolloutEpisodeRow}[3]{%
    \makebox[\linewidth][c]{%
        \successRolloutFirstImage
            {figures/appendix/success_rollouts_representative/env_#1_#2_episode_#3_t_0.png}
            {#3}%
        \hspace{0.008\linewidth}%
        \successRolloutPlainImage
            {figures/appendix/success_rollouts_representative/env_#1_#2_episode_#3_t_1.png}
            {t_1}%
        \hspace{0.008\linewidth}%
        \successRolloutPlainImage
            {figures/appendix/success_rollouts_representative/env_#1_#2_episode_#3_t_2.png}
            {t_2}%
        \hspace{0.008\linewidth}%
        \successRolloutLastImage
            {figures/appendix/success_rollouts_representative/env_#1_#2_episode_#3_t_3.png}%
    }%
}

\newcommand{\successRolloutSettingBlock}[3]{%
    \begin{subfigure}[t]{\linewidth}
        \centering

        \successRolloutEpisodeRow{#1}{#2}{1}

        \vspace{0.5mm}

        \successRolloutEpisodeRow{#1}{#2}{2}

        \vspace{0.5mm}

        \successRolloutEpisodeRow{#1}{#2}{3}

        \caption{#3}
    \end{subfigure}%
}

\begin{figure}[p]
    \centering

    \successRolloutSettingBlock{1}{po}{POMDP Variant}

    \vspace{1mm}

    \successRolloutSettingBlock{1}{fo}{FOMDP Variant}

    \caption{\textbf{Representative Successful Rollouts of Env.~I for POMDP (a) and FOMDP (b).}}
    \label{fig:success_rollouts_env1}
\end{figure}

\begin{figure}[p]
    \centering

    \successRolloutSettingBlock{2}{po}{POMDP Variant}

    \vspace{1mm}

    \successRolloutSettingBlock{2}{fo}{FOMDP Variant}

    \caption{\textbf{Representative Successful Rollouts of Env.~II for POMDP (a) and FOMDP (b).}}
    \label{fig:success_rollouts_env2}
\end{figure}

\begin{figure}[p]
    \centering

    \successRolloutSettingBlock{3}{po}{POMDP Variant}

    \vspace{1mm}

    \successRolloutSettingBlock{3}{fo}{FOMDP Variant}

    \caption{\textbf{Representative Successful Rollouts of Env.~III for POMDP (a) and FOMDP (b).}}
    \label{fig:success_rollouts_env3}
\end{figure}

\clearpage

\begin{figure}[ht]
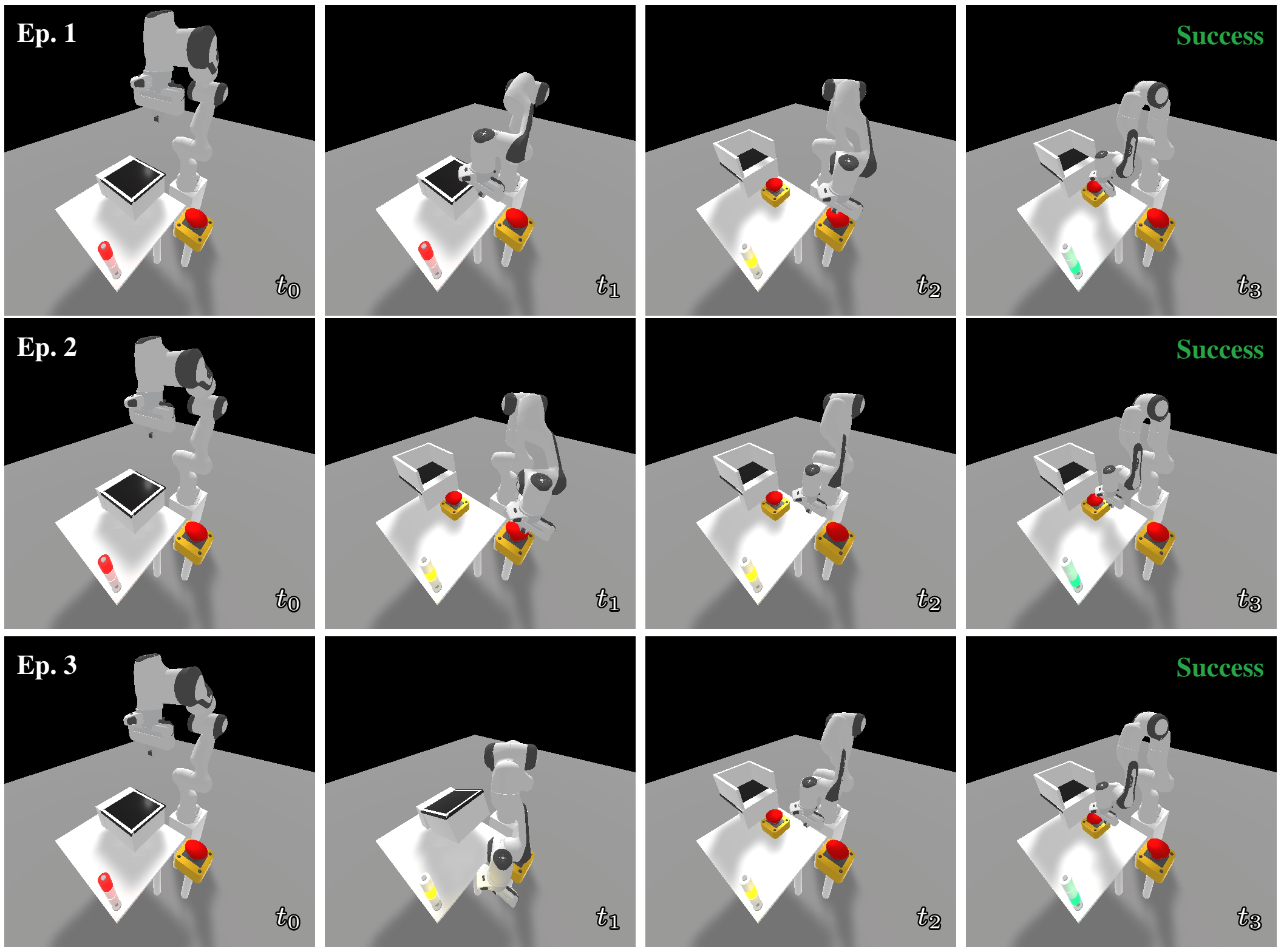

    \centering

    \successRolloutEpisodeRow{4}{po}{1}  \vspace{0.5mm}  \successRolloutEpisodeRow{4}{po}{2}  \vspace{0.5mm}  \successRolloutEpisodeRow{4}{po}{3}

    \caption{\textbf{Representative Successful Rollouts of Env.~IV's POMDP.}}
    \label{fig:success_rollouts_env4}
\end{figure}

\subsection{Candidate Lifecycle Across Checkpoints}
\label{app:visual_inputs_lifecycle}
\autoref{fig:lifecycle_env4_po_500k_2m} and \autoref{fig:lifecycle_env4_po_3m_5m} follow the candidate that received the Stage~III \texttt{STOP} in Env.~IV PO across successive assessment checkpoints from 500k to 5M environment steps. At each checkpoint, three inference episodes are shown, with six representative frames ordered chronologically from left to right. The episode-level outcome is overlaid on the terminal frame in each row, making the behavioural progression across checkpoints directly visible. At 500k and 1M, all three displayed episodes fail. At 2M, two of three episodes succeed; at 3M, one of three succeeds; at 4M, two of three succeed; and by 5M all three displayed episodes succeed. Together, the two figures provide a qualitative view of how the behaviour associated with the same candidate evolves longitudinally during WATERFALL's training and assessment process.

\newcommand{\lifecycleFirstImage}[2]{%
    \begin{minipage}[t]{0.162\linewidth}
        \centering
        \setlength{\unitlength}{\linewidth}
        \begin{picture}(1,1)
            \put(0,0){\includegraphics[width=\linewidth]{#1}}
            \put(0.04,0.88){%
                \outlinedwhite{Ep.~#2}%
            }
            \put(0.96,0.09){%
                \makebox(0,0)[r]{%
                    \outlinedwhite{$t_0$}%
                }%
            }
        \end{picture}
    \end{minipage}%
}

\newcommand{\lifecyclePlainImage}[2]{%
    \begin{minipage}[t]{0.162\linewidth}
        \centering
        \setlength{\unitlength}{\linewidth}
        \begin{picture}(1,1)
            \put(0,0){\includegraphics[width=\linewidth]{#1}}
            \put(0.96,0.09){%
                \makebox(0,0)[r]{%
                    \outlinedwhite{$#2$}%
                }%
            }
        \end{picture}
    \end{minipage}%
}

\newcommand{\lifecycleLastImage}[3]{%
    \begin{minipage}[t]{0.162\linewidth}
        \centering
        \setlength{\unitlength}{\linewidth}
        \begin{picture}(1,1)
            \put(0,0){\includegraphics[width=\linewidth]{#1}}
            \put(0.96,0.90){%
                \makebox(0,0)[r]{%
                    \outlinedcolor{#2}{#3}%
                }%
            }
            \put(0.96,0.09){%
                \makebox(0,0)[r]{%
                    \outlinedwhite{$t_5$}%
                }%
            }
        \end{picture}
    \end{minipage}%
}

\newcommand{\lifecycleEpisodeRow}[4]{%
    \makebox[\linewidth][c]{%
        \lifecycleFirstImage
            {figures/appendix/lifecycle_rollouts/#1_lifecycle_env_4_po_episode_#2_t_0.png}
            {#2}%
        \hspace{0.003\linewidth}%
        \lifecyclePlainImage
            {figures/appendix/lifecycle_rollouts/#1_lifecycle_env_4_po_episode_#2_t_1.png}
            {t_1}%
        \hspace{0.003\linewidth}%
        \lifecyclePlainImage
            {figures/appendix/lifecycle_rollouts/#1_lifecycle_env_4_po_episode_#2_t_2.png}
            {t_2}%
        \hspace{0.003\linewidth}%
        \lifecyclePlainImage
            {figures/appendix/lifecycle_rollouts/#1_lifecycle_env_4_po_episode_#2_t_3.png}
            {t_3}%
        \hspace{0.003\linewidth}%
        \lifecyclePlainImage
            {figures/appendix/lifecycle_rollouts/#1_lifecycle_env_4_po_episode_#2_t_4.png}
            {t_4}%
        \hspace{0.003\linewidth}%
        \lifecycleLastImage
            {figures/appendix/lifecycle_rollouts/#1_lifecycle_env_4_po_episode_#2_t_5.png}
            {#3}
            {#4}%
    }%
}

\newcommand{\lifecycleCheckpointBlock}[8]{%
    \begin{subfigure}[t]{\linewidth}
        \centering

        \lifecycleEpisodeRow{#1}{1}{#3}{#4}

        \vspace{0.5mm}

        \lifecycleEpisodeRow{#1}{2}{#5}{#6}

        \vspace{0.5mm}

        \lifecycleEpisodeRow{#1}{3}{#7}{#8}

        \caption{#2}
    \end{subfigure}%
}

\begin{figure}[p]
    \centering

    \lifecycleCheckpointBlock
        {500K}
        {500k Environment Steps}
        {failred}{Fail}
        {failred}{Fail}
        {failred}{Fail}

    \lifecycleCheckpointBlock
        {1M}
        {1M Environment Steps}
        {failred}{Fail}
        {failred}{Fail}
        {failred}{Fail}

    \lifecycleCheckpointBlock
        {2M}
        {2M Environment Steps}
        {failred}{Fail}
        {successgreen}{Success}
        {successgreen}{Success}

    \caption{\textbf{Candidate Lifecycle in Env.~IV PO from 500k to 2M Environment Steps.}}
    \label{fig:lifecycle_env4_po_500k_2m}
\end{figure}

\begin{figure}[p]
    \centering

    \lifecycleCheckpointBlock
        {3M}
        {3M Environment Steps}
        {successgreen}{Success}
        {failred}{Fail}
        {failred}{Fail}

    \lifecycleCheckpointBlock
        {4M}
        {4M Environment Steps}
        {failred}{Fail}
        {successgreen}{Success}
        {successgreen}{Success}

    \lifecycleCheckpointBlock
        {5M}
        {5M Environment Steps}
        {successgreen}{Success}
        {successgreen}{Success}
        {successgreen}{Success}

    \caption{\textbf{Candidate Lifecycle in Env.~IV PO from 3M to 5M Environment Steps.}}
    \label{fig:lifecycle_env4_po_3m_5m}
\end{figure}

\clearpage

\subsection{Raw Rollout Evidence Supplied to the Judge}
\label{app:visual_inputs_raw}
\begin{figure}[ht]
    \centering

    \begin{subfigure}[t]{0.32\linewidth}
        \centering
        \includegraphics[width=\linewidth]{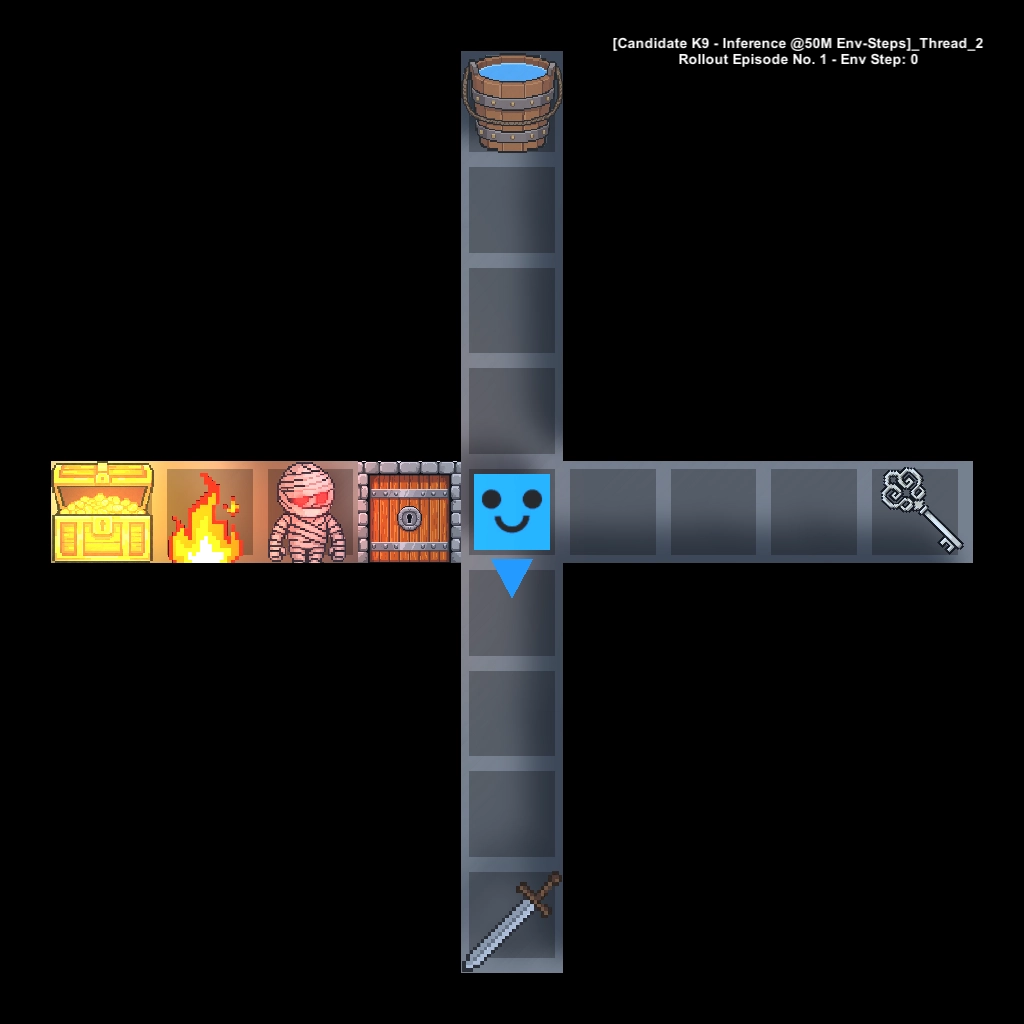}
        \caption*{\scriptsize Env.~Step $=0$ (start)}
    \end{subfigure}\hfill
    \begin{subfigure}[t]{0.32\linewidth}
        \centering
        \includegraphics[width=\linewidth]{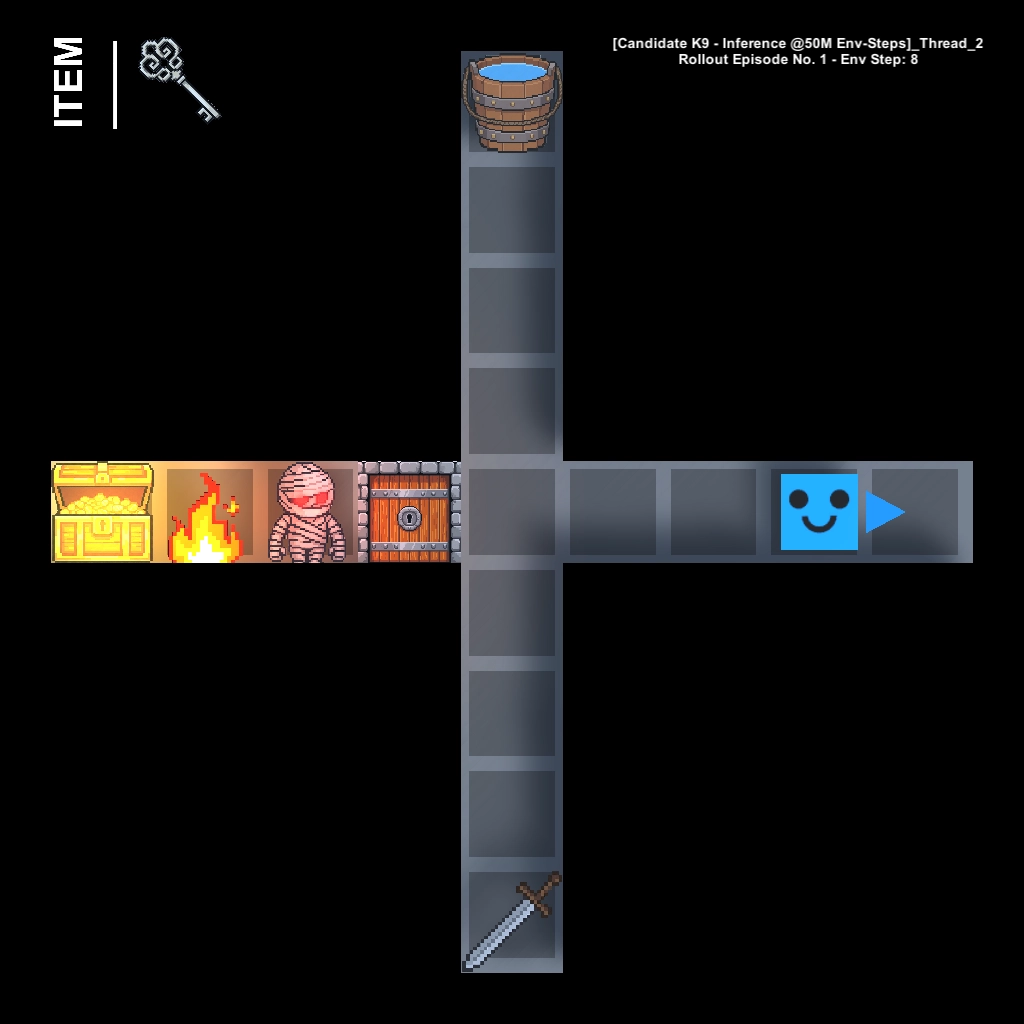}
        \caption*{\scriptsize Env.~Step $=8$}
    \end{subfigure}\hfill
    \begin{subfigure}[t]{0.32\linewidth}
        \centering
        \includegraphics[width=\linewidth]{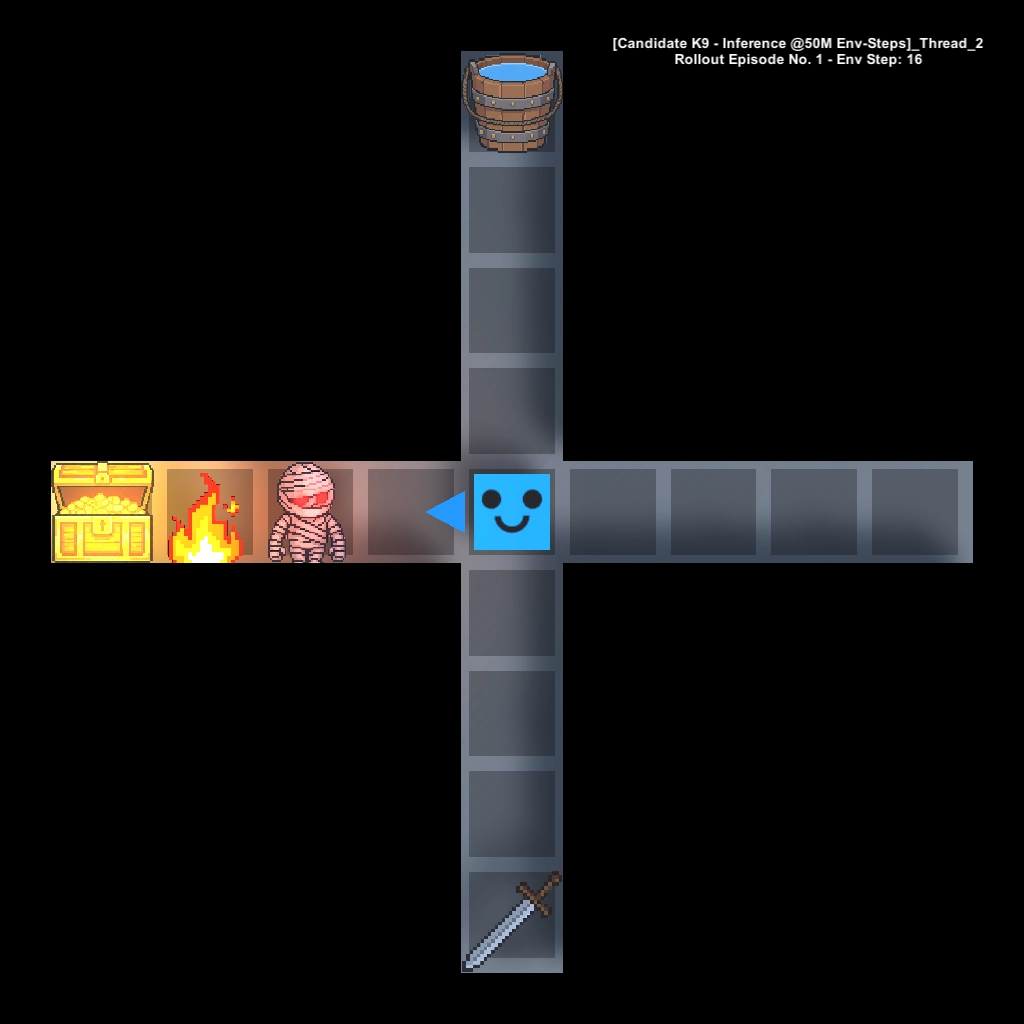}
        \caption*{\scriptsize Env.~Step $=16$}
    \end{subfigure}

    \begin{subfigure}[t]{0.32\linewidth}
        \centering
        \includegraphics[width=\linewidth]{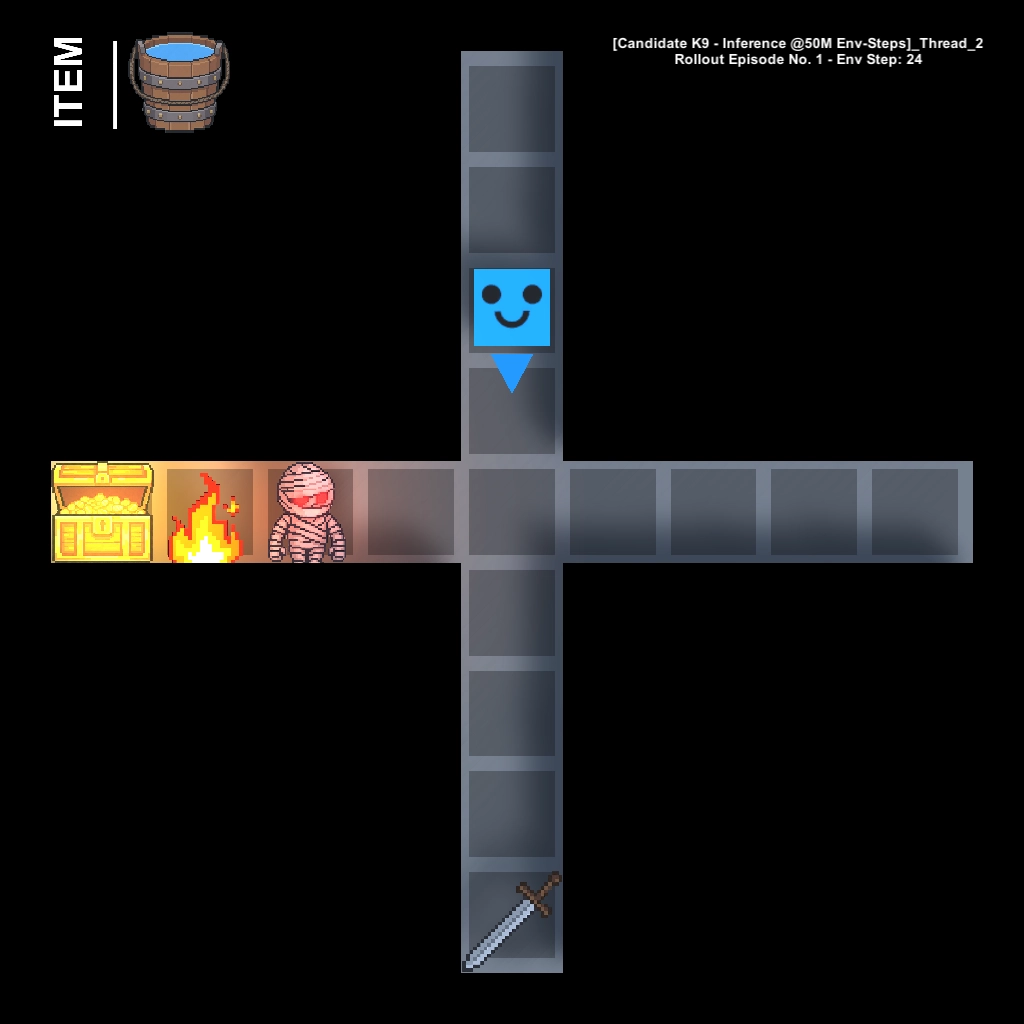}
        \caption*{\scriptsize Env.~Step $=24$}
    \end{subfigure}\hfill
    \begin{subfigure}[t]{0.32\linewidth}
        \centering
        \includegraphics[width=\linewidth]{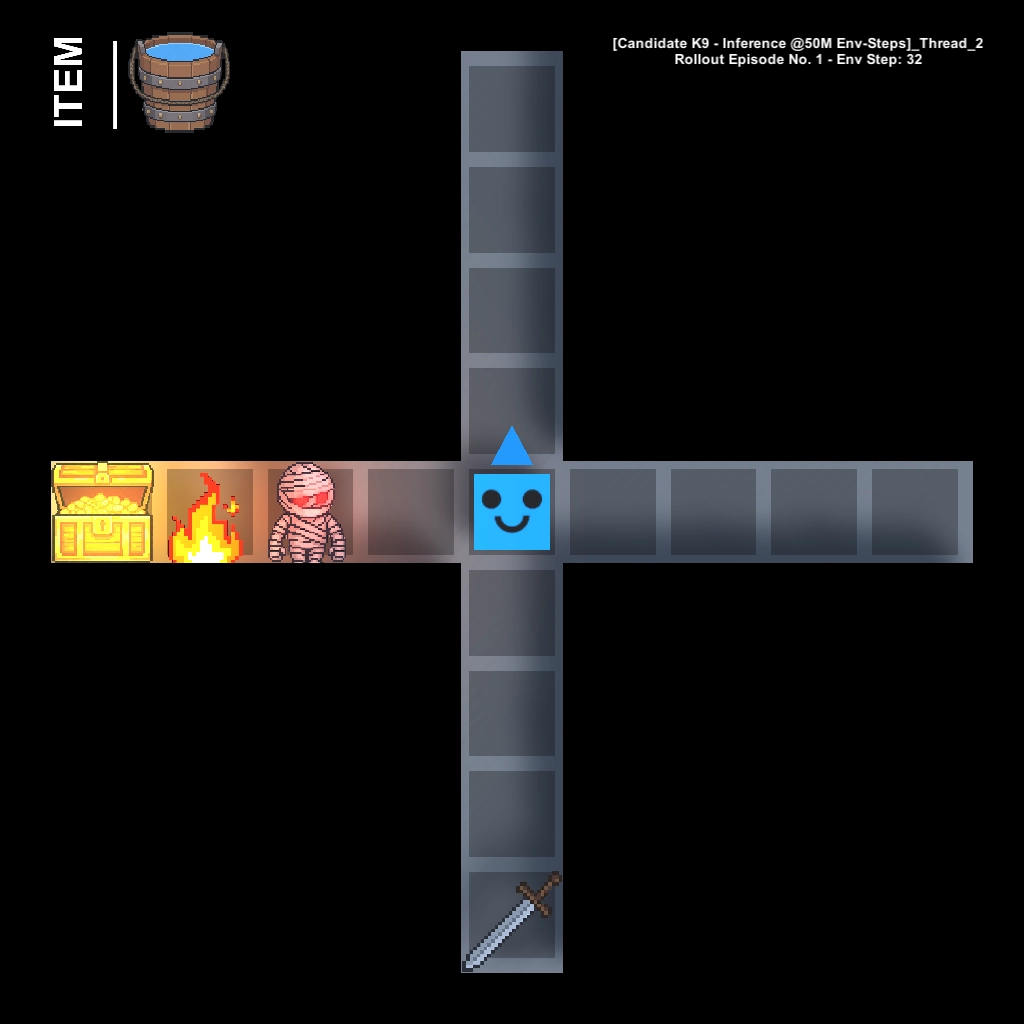}
        \caption*{\scriptsize Env.~Step $=32$}
    \end{subfigure}\hfill
    \begin{subfigure}[t]{0.32\linewidth}
        \centering
        \includegraphics[width=\linewidth]{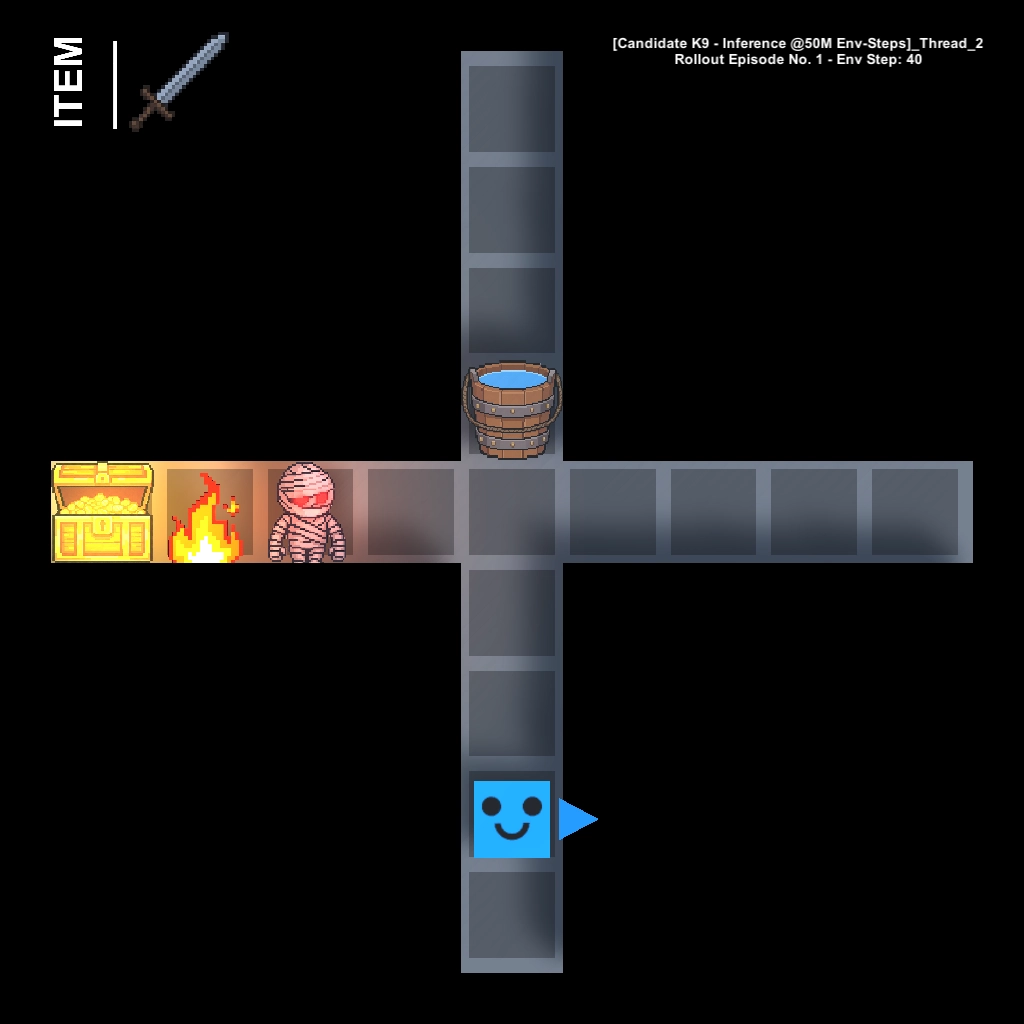}
        \caption*{\scriptsize Env.~Step $=40$}
    \end{subfigure}

    \begin{subfigure}[t]{0.32\linewidth}
        \centering
        \includegraphics[width=\linewidth]{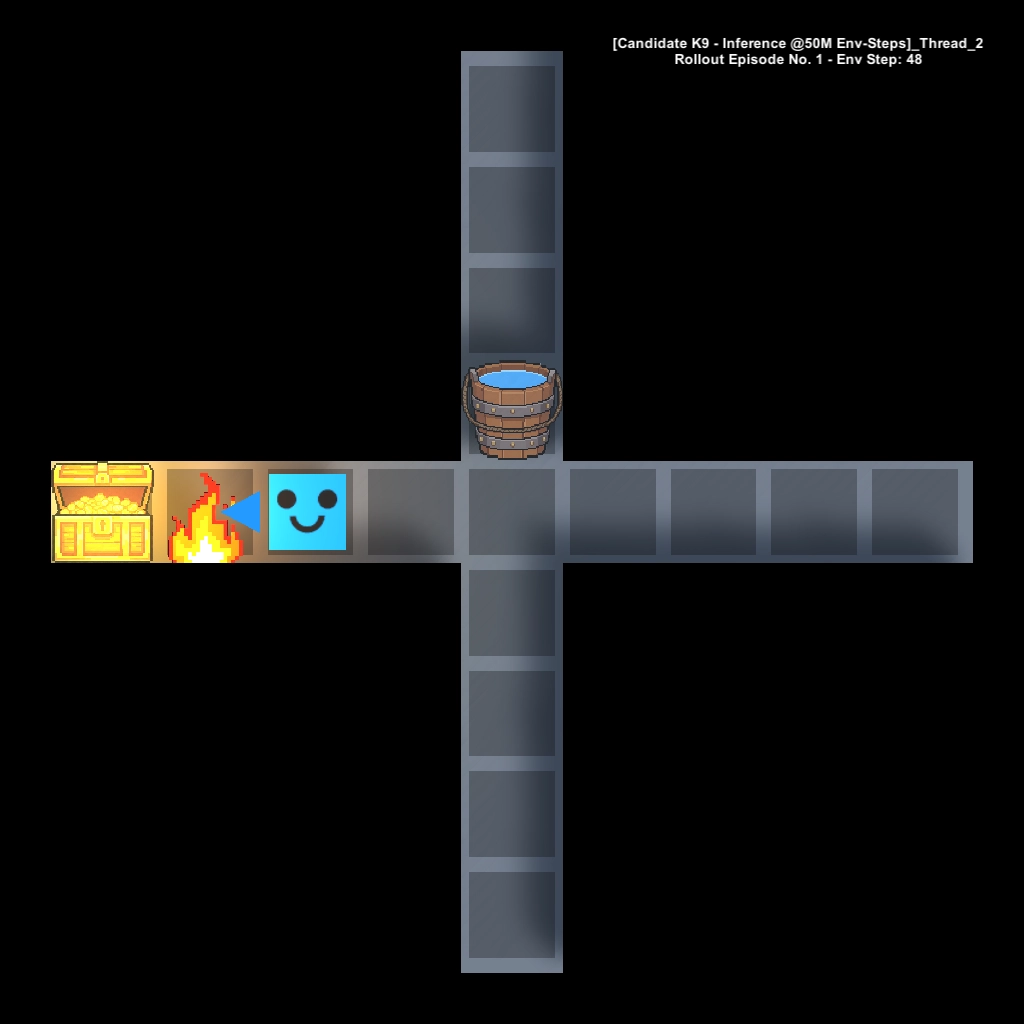}
        \caption*{\scriptsize Env.~Step $=48$}
    \end{subfigure}\hfill
    \begin{subfigure}[t]{0.32\linewidth}
        \centering
        \includegraphics[width=\linewidth]{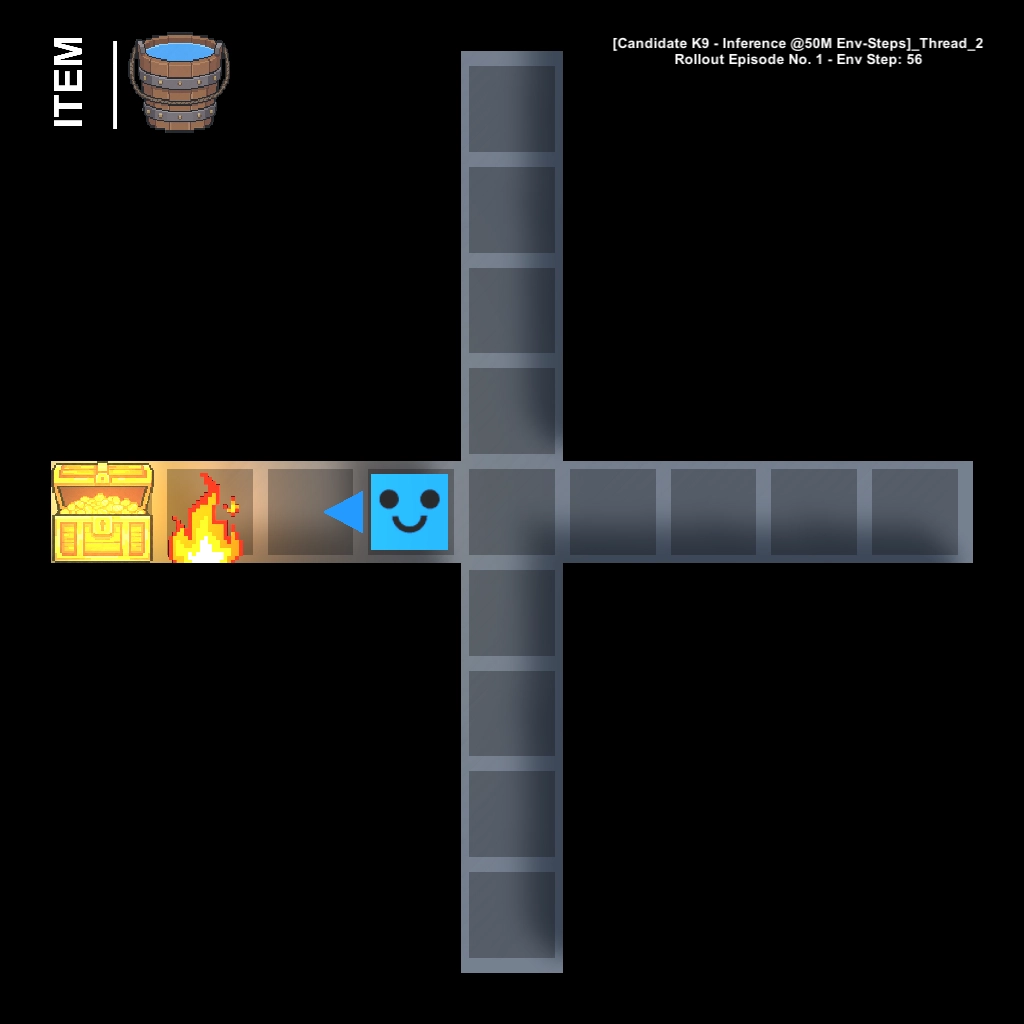}
        \caption*{\scriptsize Env.~Step $=56$}
    \end{subfigure}\hfill
    \begin{subfigure}[t]{0.32\linewidth}
        \centering
        \includegraphics[width=\linewidth]{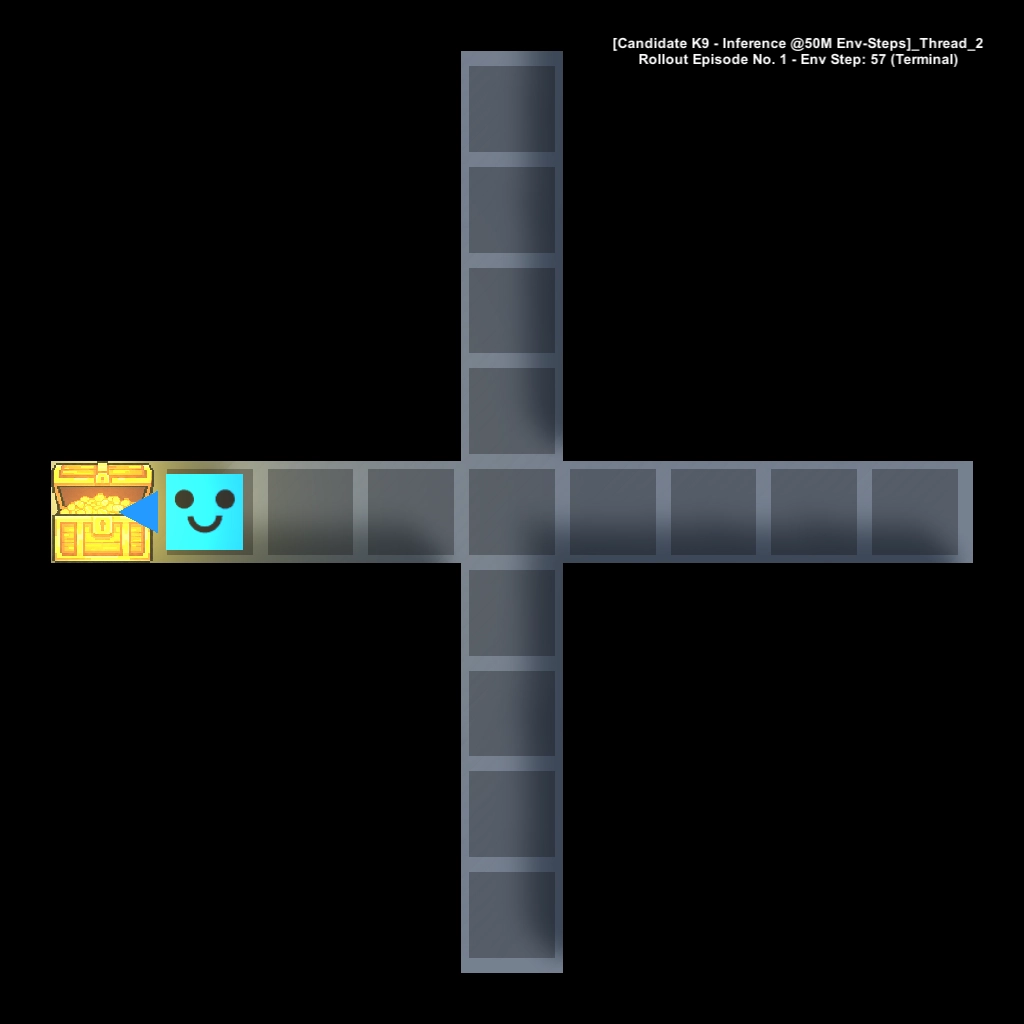}
        \caption*{\scriptsize Env.~Step $=57$ (terminal)}
    \end{subfigure}

    \caption{\textbf{Raw rollout evidence supplied to the Stage~II VLM judge for Env.~II FO.}
    The source images shown here are identical to those supplied to the judge and are presented in their original chronological order. Stage~II received the complete sampled rollout sequence, whereas Stage~III received only the byte-identical episode-boundary images at Env.~Step $0$ and Env.~Step $57$.}
    \label{fig:raw_rollout_judge}
\end{figure}

The compact filmstrips from \autoref{app:visual_inputs_successful} intentionally select only a few representative frames for readability. \autoref{fig:raw_rollout_judge} shows one complete episode as supplied to the Stage~II VLM judge: Env.~II FO, candidate \texttt{k9} at the 50M checkpoint, recorded every eight environment steps plus the mandatory start and terminal frames. Frames are ordered chronologically from left to right and then top to bottom, including the mandatory start and terminal observations. In the actual judge input, this visual sequence is accompanied by the burn-in annotation identifying the episode, environment step, candidate, and assessment checkpoint; no ground-truth return, success metric, or task-specific event annotation is supplied. Stage~III receives only the first and last frames of this sequence.

\let\visionplaceholder\relax

\clearpage
\section*{NeurIPS Paper Checklist}
\begin{enumerate}
\item {\bf Claims}
    \item[] Question: Do the main claims made in the abstract and introduction accurately reflect the paper's contributions and scope?
\item[] Answer: \answerYes{}
\item[] Justification: The abstract and introduction state WATERFALL's intended scope: automated reward discovery under strict observability parity in FO and reveal-gated PO environments. These claims are supported by the method, baseline comparisons, ablations, and robustness analyses in the main results and appendix, while limitations are discussed separately.

\item {\bf Limitations}
    \item[] Question: Does the paper discuss the limitations of the work performed by the authors?
\item[] Answer: \answerYes{}
\item[] Justification: Limitations are discussed in the Limitations and Future Work section, including reliance on foundation-model evaluators, hyperparameter tuning, VLM judge noise, compute overhead, and the need for broader validation on additional domains.

\item {\bf Theory assumptions and proofs}
    \item[] Question: For each theoretical result, does the paper provide the full set of assumptions and a complete (and correct) proof?
\item[] Answer: \answerNA{}
\item[] Justification: The paper does not present formal theoretical results requiring proofs. The mathematical expressions define the problem setting, observability protocol, PPO objective, and tournament aggregation rules.

    \item {\bf Experimental result reproducibility}
    \item[] Question: Does the paper fully disclose all the information needed to reproduce the main experimental results of the paper to the extent that it affects the main claims and/or conclusions of the paper (regardless of whether the code and data are provided or not)?
\item[] Answer: \answerYes{}
\item[] Justification: The main paper and appendix specify the algorithmic stages, prompt contracts, observability-parity rules, environment definitions, checkpoint schedule, randomness sources, evaluation protocol, and logging structure needed to reproduce the experimental protocol. Executable code is publicly available at \url{https://github.com/etriantafyllidis/waterfall}. The appendix further provides the implementation-level details required to independently reconstruct the experiments.

\item {\bf Open access to data and code}
    \item[] Question: Does the paper provide open access to the data and code, with sufficient instructions to faithfully reproduce the main experimental results, as described in supplemental material?
\item[] Answer: \answerYes{}
\item[] Justification: Publicly available code to run WATERFALL is provided at \url{https://github.com/etriantafyllidis/waterfall}. This includes simulator source files, environment wrappers and prompts. The accompanying technical appendix provides the full methodological details, prompt contracts and environment descriptions.

\item {\bf Experimental setting/details}
    \item[] Question: Does the paper specify all the training and test details (e.g., data splits, hyperparameters, how they were chosen, type of optimizer) necessary to understand the results?
\item[] Answer: \answerYes{}
\item[] Justification: The paper specifies the environment variants, observability protocol, candidate count, seed protocol, checkpoint schedule, PPO training setup, VLM evaluation protocol, baseline interfaces, and hyperparameters in Sections~\ref{sec:methods} and~\ref{sec:evaluation} and Appendices~\ref{app:full_methods}, \ref{app:evaluation_protocol} and~\ref{app:hyperparameters}.

\item {\bf Experiment statistical significance}
    \item[] Question: Does the paper report error bars suitably and correctly defined or other appropriate information about the statistical significance of the experiments?
\item[] Answer: \answerYes{}
\item[] Justification: Learning curves report mean performance over independent training seeds with shaded regions denoting \(\pm 1\) standard deviation. Additional evaluator-sensitivity experiments report mean \(\pm 1\sigma\) over repeated frozen-tournament evaluations, as described in the Results section and appendix.

\item {\bf Experiments compute resources}
    \item[] Question: For each experiment, does the paper provide sufficient information on the computer resources (type of compute workers, memory, time of execution) needed to reproduce the experiments?
\item[] Answer: \answerYes{}
\item[] Justification: The appendix reports the hardware, software stack, simulator version, GPU memory, training setup, wall-clock/runtime heterogeneity, and foundation-model API usage/costs required to reproduce the experiments.
    
\item {\bf Code of ethics}
    \item[] Question: Does the research conducted in the paper conform, in every respect, with the NeurIPS Code of Ethics \url{https://neurips.cc/public/EthicsGuidelines}?
\item[] Answer: \answerYes{}
\item[] Justification: The research uses synthetic environments and does not involve human subjects, private data, scraped datasets, or deployment on real users. We discuss limitations, artefact availability, and responsible release considerations.

\item {\bf Broader impacts}
    \item[] Question: Does the paper discuss both potential positive societal impacts and negative societal impacts of the work performed?
\item[] Answer: \answerYes{}
\item[] Justification: The Broader Impacts paragraph at the start of the Technical Appendix discusses both potential benefits, such as reducing manual reward engineering, and risks, such as reward misalignment, noisy VLM judgements, and compute/API costs.
    
\item {\bf Safeguards}
    \item[] Question: Does the paper describe safeguards that have been put in place for responsible release of data or models that have a high risk for misuse (e.g., pre-trained language models, image generators, or scraped datasets)?
\item[] Answer: \answerNA{}
\item[] Justification: The paper does not release high-risk pretrained models, image generators, scraped datasets, or human-subject datasets. The proposed method is evaluated only in synthetic RL environments.

\item {\bf Licenses for existing assets}
    \item[] Question: Are the creators or original owners of assets (e.g., code, data, models), used in the paper, properly credited and are the license and terms of use explicitly mentioned and properly respected?
\item[] Answer: \answerYes{}
\item[] Justification: The paper cites the main prior environments and software stack, with their licences or terms of use in the Existing Assets and Licences paragraph of the Technical Appendix; no third-party code or datasets are redistributed.

\item {\bf New assets}
    \item[] Question: Are new assets introduced in the paper well documented and is the documentation provided alongside the assets?
\item[] Answer: \answerYes{}
\item[] Justification: The paper introduces custom reveal-gated POMDP environment variants and generated evaluation artefacts. The environments are released with documentation at \url{https://github.com/etriantafyllidis/waterfall}, and the appendix documents their design, including environment descriptions, observability rules, prompts, and artefact structure.

\item {\bf Crowdsourcing and research with human subjects}
    \item[] Question: For crowdsourcing experiments and research with human subjects, does the paper include the full text of instructions given to participants and screenshots, if applicable, as well as details about compensation (if any)? 
    \item[] Answer: \answerNA{}
    \item[] Justification: No human subjects participated in this work.

\item {\bf Institutional review board (IRB) approvals or equivalent for research with human subjects}
    \item[] Question: Does the paper describe potential risks incurred by study participants, whether such risks were disclosed to the subjects, and whether Institutional Review Board (IRB) approvals (or an equivalent approval/review based on the requirements of your country or institution) were obtained?
    \item[] Answer: \answerNA{}
    \item[] Justification: No human subjects participated in this work.

\item {\bf Declaration of LLM usage}
    \item[] Question: Does the paper describe the usage of LLMs if it is an important, original, or non-standard component of the core methods in this research? Note that if the LLM is used only for writing, editing, or formatting purposes and does \emph{not} impact the core methodology, scientific rigor, or originality of the research, declaration is not required.
\item[] Answer: \answerYes{}
\item[] Justification: LLMs/VLMs are central to the method. WATERFALL uses foundation models for persona-conditioned reward synthesis, vision-only pairwise tournament judging, qualitative digest generation, and longitudinal assessment/mutation; these uses and prompt contracts are described in the Methods section and appendix.

\end{enumerate}
\end{document}